# Fuzzy, Neutrosophic, and Uncertain GRAPH THEORY:

## Properties and Applications

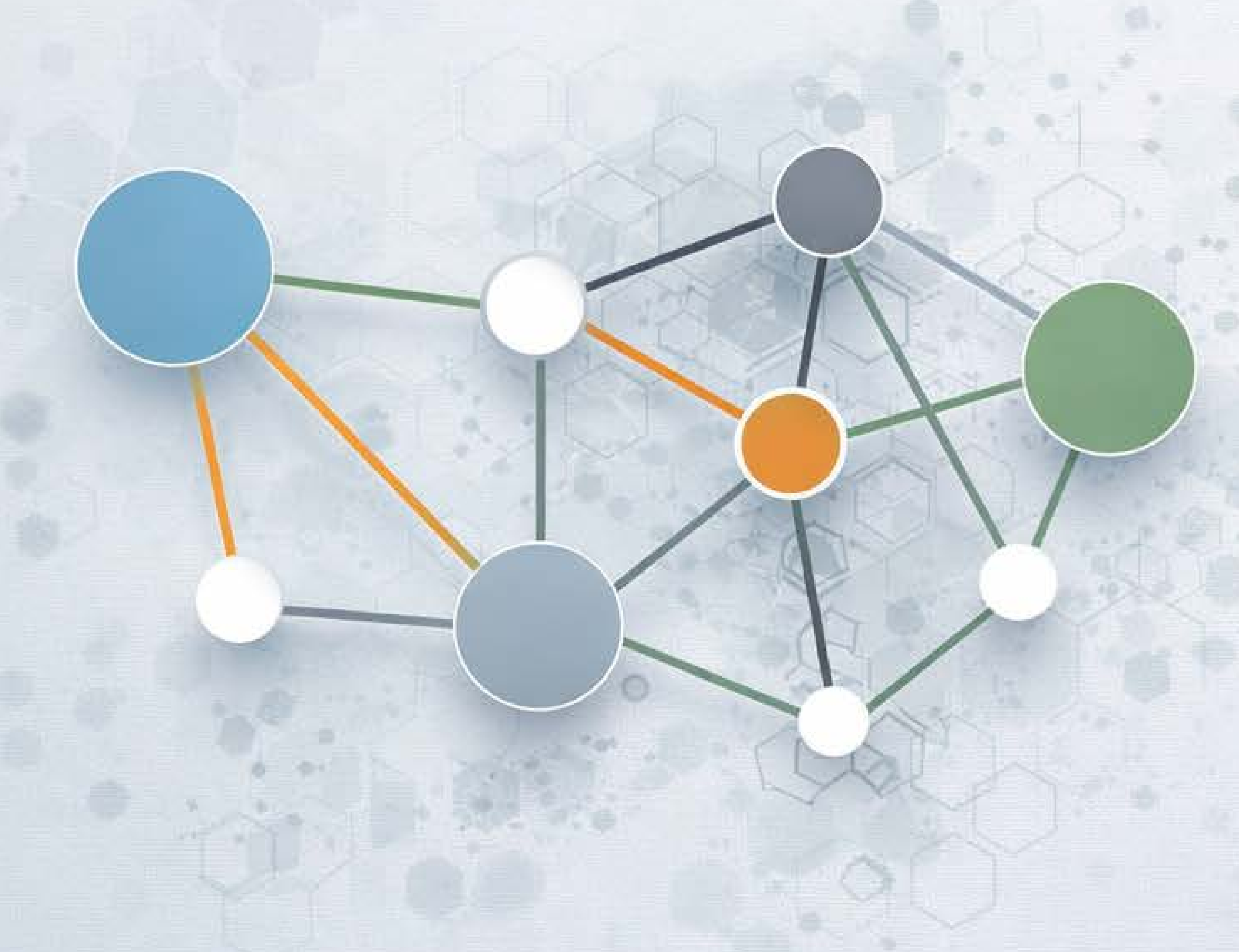

Takaaki Fujita
Florentin Smarandache

**Takaaki Fujita, Florentin Smarandache**

# Fuzzy, Neutrosophic, and Uncertain Graph Theory: Properties and Applications

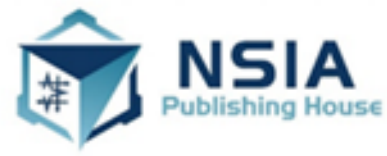

*Editor:*

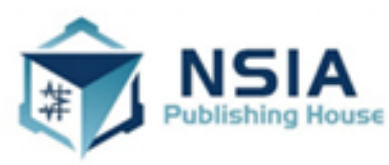

***Peer-Reviewers:***

**John Frederick D. Tapia**
Chemical Engineering Department, De La Salle University - Manila, 2401 Taft Avenue, Malate, Manila, Philippines
Email: *john.frederick.tapia@dlsu.edu.ph*

**Darren Chong**
Independent researcher, Singapore
Email: *darrenchong2001@yahoo.com.sg*

**Umit Cali**
Norwegian University of Science and Technology, NO-7491 Trondheim, Norway
Email: *umit.cali@ntnu.no*

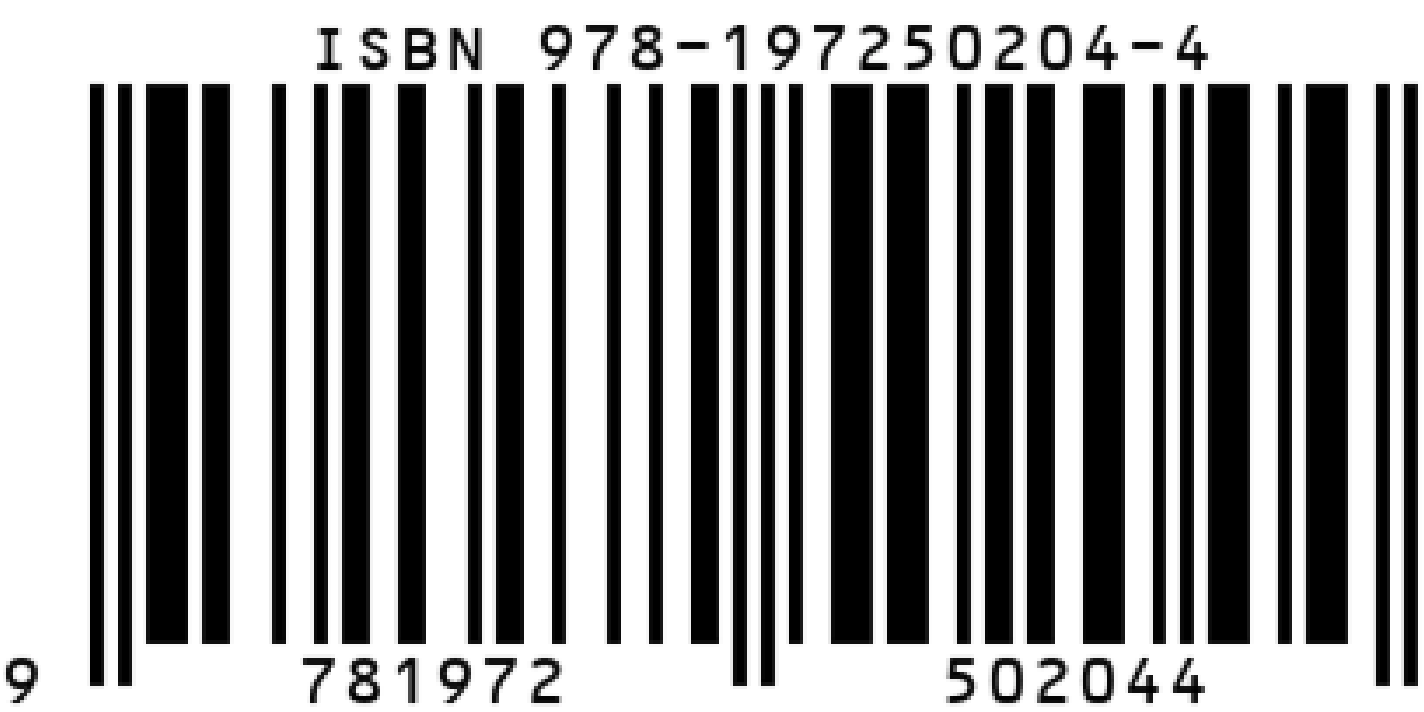

# Contents in this book

The remainder of this book is organized as follows.

# Chapter 1

# Introduction

## 1.1 Graph Theory

Graph theory is a fundamental branch of mathematics concerned with structures formed by vertices and edges. It provides a rigorous language for representing connectivity, interaction, and organization, and has long served as an essential framework in both pure and applied mathematics [1]. Over the years, graph-theoretic methods have been used successfully in a wide range of disciplines, including computer science, biology, social network analysis, communication systems, and chemistry [2–4]. More recently, graph-based models have also become increasingly important in artificial intelligence, particularly through graph neural networks, hypergraph learning, and related data-driven paradigms [5–9].

The development of graph theory has led to many important graph classes, structural notions, and algorithmic methodologies. Representative directions include the study of tree-like structures, path-based properties, and graph classes related to linear layouts or other structural restrictions [10–14]. A recurring theme in this area is that restricting attention to well-structured graph classes often yields stronger theoretical results and substantially more efficient algorithms than those available for arbitrary graphs [15]. For this reason, graph theory remains both a rich mathematical discipline and a practical foundation for modeling complex systems.

## 1.2 Uncertain Set

Many real-world phenomena involve vagueness, incompleteness, partial truth, inconsistency, or hesitation. To represent such uncertainty in a mathematically meaningful way, numerous generalized set-theoretic frameworks have been introduced. Among the most influential are Fuzzy Sets [16], Intuitionistic Fuzzy Sets [17], Neutrosophic Sets [18, 19], Vague Sets [20], Hesitant Fuzzy Sets [21], Picture Fuzzy Sets [22], Quadripartitioned Neutrosophic Sets [23], PentaPartitioned Neutrosophic Sets [24], Plithogenic Sets [25], HyperFuzzy Sets [26], and HyperNeutrosophic Sets [27]. Such frameworks have been applied in diverse areas including decision science, chemistry, control, and machine learning, where the ability to represent nonclassical information is essential [28].

In a classical fuzzy set, each element $x \in X$ is assigned a single membership degree

$$\mu(x) \in [0, 1],$$

which indicates the extent to which $x$ belongs to the set under consideration [16]. An intuitionistic fuzzy set enriches this description by associating with each element a membership degree $\mu(x)$ and a non-membership degree $\nu(x)$, subject to

$$0 \leq \mu(x) + \nu(x) \leq 1,$$

so that the remaining quantity $1 - \mu(x) - \nu(x)$ expresses hesitation [17, 29].

A neutrosophic set further extends this viewpoint by assigning to each element a triple

$$(T(x), I(x), F(x)),$$

where $T(x)$, $I(x)$, and $F(x)$ represent the degrees of truth, indeterminacy, and falsity, respectively. Unlike the intuitionistic fuzzy setting, these three components are not constrained to sum to 1, which makes it possible to model incomplete, inconsistent, or redundant information in a more flexible manner [29,30]. This additional expressive power has made neutrosophic frameworks important in a broad range of uncertainty-aware theories, including neutrosophic logic, probability, statistics, measure, integral, and related analytical formalisms [28, 31].

Plithogenic sets provide a further refinement by describing each element through attribute values together with corresponding degrees of appurtenance, while also incorporating a contradiction or dissimilarity function between distinct attribute values [25, 32, 33]. This additional structure enables context-sensitive aggregation of heterogeneous and potentially conflicting evaluations, thereby generalizing and refining classical fuzzy, intuitionistic fuzzy, and neutrosophic models [31, 34].

For convenience, Table 1.1 summarizes the canonical information associated with each element in several representative set extensions.

Table 1.1: Representative set extensions and the canonical information stored per element.

| **Set Type** | **Canonical data attached to each element** |
|---|---|
| Fuzzy Set | Membership mapping $\mu : X \to [0, 1]$. |
| Intuitionistic Fuzzy Set | Membership $\mu$ and non-membership $\nu$ with $\mu(x) + \nu(x) \leq 1$; the gap $1 - \mu(x) - \nu(x)$ represents hesitation. |
| Neutrosophic Set | Triple $(T, I, F)$ with $T, I, F \in [0, 1]$, representing truth, indeterminacy, and falsity as mutually independent coordinates. |
| Plithogenic Set | Tuple $(P, v, Pv, \mathrm{pdf}, \mathrm{pCF})$ where $\mathrm{pdf} : P \times Pv \to [0, 1]^s$ encodes $s$-dimensional appurtenance and $\mathrm{pCF} : Pv \times Pv \to [0, 1]^t$ is a symmetric contradiction map in $[0, 1]^t$. |

## 1.3 Fuzzy, Neutrosophic, Quadripartitioned Neutrosophic, and Plithogenic Graphs

Since many practical systems involve uncertainty not only in attributes but also in relations, several graph-theoretic frameworks have been developed to incorporate uncertainty directly into vertices, edges, and higher-level structural information. Among these, fuzzy graphs, neutrosophic graphs, quadripartitioned neutrosophic graphs, and plithogenic graphs form an important family of uncertainty-aware network models.

A fuzzy graph assigns to each vertex and each edge a membership degree in $[0, 1]$, thereby expressing the extent to which the corresponding object belongs to the modeled structure [35, 36]. In this sense, a fuzzy graph may be viewed as a graph-theoretic realization of fuzzy-set-based uncertainty [37,38]. Because many real-world relationships are inherently imprecise, fuzzy graphs have been applied to problems in social networks, decision-making, transportation systems, and related areas [35, 36]. This broad applicability has led to the development of many variants and refinements, including Intuitionistic Fuzzy Graphs [39], Bipolar Fuzzy Graphs [40], Fuzzy Planar Graphs [41], Irregular Bipolar Fuzzy Graphs [42], General Fuzzy Graphs [43, 44], and Complex Hesitant Fuzzy Graphs [45].

More generally, a wide variety of graph models have been proposed to capture uncertainty and enriched relational information. These include fuzzy graphs [35, 36], vague graphs [46–48], plithogenic graphs [32, 49–51], probabilistic graphs [52–54], vague hypergraphs [55], $N$-graphs [56], $N$-hypergraphs [57], Markov graphs [58], soft graphs [59, 60], hypersoft graphs [61, 62], and rough graphs [63, 64]. Together, these frameworks illustrate the breadth of approaches that have been developed to represent uncertainty, ambiguity, and enriched semantic structure in graph-based models.

In recent years, neutrosophic graphs [65, 66] and neutrosophic hypergraphs [67, 68] have attracted increasing attention within the broader development of neutrosophic set theory [69, 70]. The term *neutrosophic* refers to a framework in which truth, indeterminacy, and falsity are treated as distinct components. From a graph-theoretic perspective, this makes it possible to represent ambiguous or inconsistent relational information more flexibly than in ordinary fuzzy graphs. Accordingly, many related classes have been introduced, including Bipolar Neutrosophic Graphs [68, 71–73], Neutrosophic Incidence Graphs [74–77], single-valued neutrosophic signed graphs [78], Strong Neutrosophic Graphs [79], $m$-polar neutrosophic graphs [80–82], Complex Neutrosophic Hypergraphs [67], and Bipolar Neutrosophic Hypergraphs [68].

Plithogenic graphs extend uncertainty-aware graph models even further by describing vertices and edges through attribute values together with corresponding degrees of appurtenance, while also introducing a contradiction function that quantifies incompatibility between distinct attribute values [31,83–85]. They may therefore be regarded as graph-theoretic counterparts of plithogenic sets [25, 32, 33]. This richer structure supports context-dependent aggregation of heterogeneous and potentially conflicting information on networks, thereby refining classical fuzzy, intuitionistic fuzzy, and neutrosophic graph models [31, 34, 86–88].

For convenience, Table 1.2 summarizes the canonical information attached to vertices and edges in several representative graph extensions.

Table 1.2: Representative graph extensions and the canonical information stored on vertices and/or edges.

| **Graph Type** | **Canonical data attached to vertices/edges** |
|---|---|
| Fuzzy Graph | Vertex membership $\sigma : V \to [0,1]$ and edge membership $\mu : E \to [0,1]$ (typically with $\mu(uv) \le \sigma(u) \wedge \sigma(v)$). |
| Intuitionistic Fuzzy Graph | Vertex degrees $(\mu_A, \nu_A) : V \to [0,1]^2$ and edge degrees $(\mu_B, \nu_B) : E \to [0,1]^2$ with $\mu + \nu \le 1$; the residual represents hesitation. |
| Neutrosophic Graph | Vertex triple $(T_A, I_A, F_A) : V \to [0,1]^3$ and edge triple $(T_B, I_B, F_B) : E \to [0,1]^3$ (truth, indeterminacy, falsity). |
| Quadripartitioned Neutrosophic Graph | Vertex quadruple $(T, C, U, F) : V \to [0,1]^4$ and edge quadruple $(T, C, U, F) : E \to [0,1]^4$, typically encoding truth, contradiction, unknown, and falsity. |
| Pentapartitioned Neutrosophic Graph | Vertex quintuple $(T, C, U, F, S) : V \to [0,1]^5$ and edge quintuple $(T, C, U, F, S) : E \to [0,1]^5$, that is, a five-component refinement of neutrosophic information. |
| Plithogenic Graph | Vertex structure $PM = (M, \ell, M_\ell, \mathrm{adf}, \mathrm{aCf})$ and edge structure $PN = (N, m, N_m, \mathrm{bdf}, \mathrm{bCf})$, where $\mathrm{adf} : M \times M_\ell \to [0,1]^s$ and $\mathrm{bdf} : N \times N_m \to [0,1]^s$ encode $s$-dimensional appurtenance, while aCf and bCf are symmetric contradiction maps in $[0,1]^t$. |

## 1.4 Our Contributions

Numerous graph classes have been introduced within frameworks such as fuzzy graphs, neutrosophic graphs, and related uncertainty-aware graph models. The notion of an Uncertain Graph may be regarded as a general framework that enables these concepts to be treated in a more unified manner. In this book, we survey representative graph classes that are well known in frameworks such as fuzzy graphs, neutrosophic graphs, and plithogenic graphs, and we organize them from the viewpoint of a common uncertainty-based structure. In particular, we discuss basic graph classes, structural properties, graph parameters, and several application-oriented extensions, with the aim of providing a clearer overview of how these graph-theoretic notions can be interpreted under different uncertainty-aware settings.

# Fuzzy, Neutrosophic, and Uncertain Graph Theory: Properties and Applications

**Takaaki Fujita** [1] * **and Florentin Smarandache**[2]
[1] Independent Researcher, Tokyo, Japan.
Email: Takaaki.fujita060@gmail.com
[2] University of New Mexico, Gallup Campus, NM 87301, USA.
Email: fsmarandache@gmail.com

## Abstract

Since many practical systems involve uncertainty not only in attributes but also in relations, several graph-theoretic frameworks have been developed to incorporate uncertainty directly into vertices, edges, and higher-level structural information. Among these, fuzzy graphs, neutrosophic graphs, and plithogenic graphs form an important family of uncertainty-aware network models. Numerous graph classes have been introduced within frameworks such as fuzzy graphs and neutrosophic graphs. The notion of an Uncertain Graph may be regarded as a new framework that enables these concepts to be considered in a unified manner. In this book, we survey graph classes that are well known in frameworks such as fuzzy graphs, neutrosophic graphs, and plithogenic graphs.

# Chapter 2

# Preliminaries

This chapter collects the basic notation and background used throughout the book. Except when stated otherwise, all sets are assumed to be finite.

## 2.1 Fuzzy Graph

A fuzzy set assigns each element a membership degree between 0 and 1, modeling partial belonging and uncertainty in classification [16,89]. A fuzzy graph combines fuzzy vertex and edge membership functions, representing relationships with uncertainty and graded connectivity among nodes [35,90].

**Definition 2.1.1** (Fuzzy set)**.** [16] Let $Y$ be a non-empty universe. A *fuzzy set* $\tau$ on $Y$ is a function

$$\tau : Y \longrightarrow [0,1],$$

assigning to each $y \in Y$ a membership value $\tau(y)$. A *fuzzy relation* on $Y$ is a fuzzy subset $\delta$ of $Y \times Y$. Given a fuzzy set $\tau$ on $Y$, the relation $\delta$ is said to be a *fuzzy relation on* $\tau$ whenever

$$\delta(y,z) \;\le\; \min\{\tau(y),\tau(z)\}, \quad \forall\, y,z \in Y.$$

**Definition 2.1.2** (Fuzzy graph)**.** [35] A *fuzzy graph* on a vertex set $V$ is a pair $G = (\sigma, \mu)$ consisting of:

- A vertex membership function $\sigma : V \to [0,1]$, where $\sigma(x)$ gives the degree to which $x \in V$ belongs to the graph.
- An edge membership function $\mu : V \times V \to [0,1]$, which is a fuzzy relation on $\sigma$, satisfying

  $$\mu(x,y) \;\le\; \sigma(x) \wedge \sigma(y), \quad \forall\, x,y \in V,$$

  where $\wedge$ denotes the minimum operator.

The associated *crisp graph* $G^* = (\sigma^*, \mu^*)$ is determined by

$$\sigma^* = \{\, x \in V \mid \sigma(x) > 0 \,\}, \qquad \mu^* = \{\, (x,y) \in V \times V \mid \mu(x,y) > 0 \,\}.$$

A *fuzzy subgraph* $H = (\sigma', \mu')$ of $G$ is obtained by choosing a subset $X \subseteq V$ and defining

- a restricted vertex membership $\sigma' : X \to [0,1]$,

- an edge membership $\mu' : X \times X \to [0,1]$ such that

$$\mu'(x,y) \;\le\; \sigma'(x) \wedge \sigma'(y), \quad \forall\, x,y \in X.$$

**Example 2.1.3** (A fuzzy graph and one of its fuzzy subgraphs)**.** Let

$$V = \{v_1, v_2, v_3, v_4\}.$$

Define a vertex-membership function $\sigma : V \to [0,1]$ by

$$\sigma(v_1) = 0.9, \qquad \sigma(v_2) = 0.7, \qquad \sigma(v_3) = 0.5, \qquad \sigma(v_4) = 0.6.$$

Next, define an edge-membership function $\mu : V \times V \to [0,1]$ by

$$\mu(v_1,v_2) = 0.6, \qquad \mu(v_2,v_3) = 0.4, \qquad \mu(v_3,v_4) = 0.3, \qquad \mu(v_1,v_4) = 0.5,$$

and let

$$\mu(v_i,v_j) = 0$$

for all other unordered pairs $\{v_i,v_j\} \subseteq V$, with

$$\mu(v_i,v_j) = \mu(v_j,v_i)$$

for all $i,j$.

Then $G = (\sigma,\mu)$ is a fuzzy graph, because for every edge with positive membership we have

$$\mu(v_1,v_2) = 0.6 \le \min\{0.9, 0.7\} = 0.7,$$
$$\mu(v_2,v_3) = 0.4 \le \min\{0.7, 0.5\} = 0.5,$$
$$\mu(v_3,v_4) = 0.3 \le \min\{0.5, 0.6\} = 0.5,$$

and

$$\mu(v_1,v_4) = 0.5 \le \min\{0.9, 0.6\} = 0.6.$$

Hence the associated crisp graph is

$$G^* = (V^*, E^*),$$

where

$$V^* = \{v_1, v_2, v_3, v_4\},$$

since all vertex-memberships are positive, and

$$E^* = \big\{\{v_1,v_2\}, \{v_2,v_3\}, \{v_3,v_4\}, \{v_1,v_4\}\big\},$$

since these are exactly the pairs having positive edge-membership.

Now choose

$$X = \{v_1, v_2, v_4\} \subseteq V.$$

Define a fuzzy subgraph $H = (\sigma',\mu')$ on $X$ by

$$\sigma'(v_1) = 0.9, \qquad \sigma'(v_2) = 0.7, \qquad \sigma'(v_4) = 0.4,$$

and

$$\mu'(v_1,v_2) = 0.5, \qquad \mu'(v_1,v_4) = 0.3, \qquad \mu'(v_2,v_4) = 0.2,$$

with

$$\mu'(x,y) = 0$$

for all other pairs in $X \times X$, and $\mu'(x,y) = \mu'(y,x)$.

Again, $H$ is a fuzzy graph, since

$$\mu'(v_1,v_2) = 0.5 \le \min\{0.9, 0.7\} = 0.7,$$
$$\mu'(v_1,v_4) = 0.3 \le \min\{0.9, 0.4\} = 0.4,$$

and

$$\mu'(v_2,v_4) = 0.2 \le \min\{0.7, 0.4\} = 0.4.$$

Therefore $H$ is a fuzzy subgraph of $G$. For reference, the illustrative figure is shown in Figure 2.1.

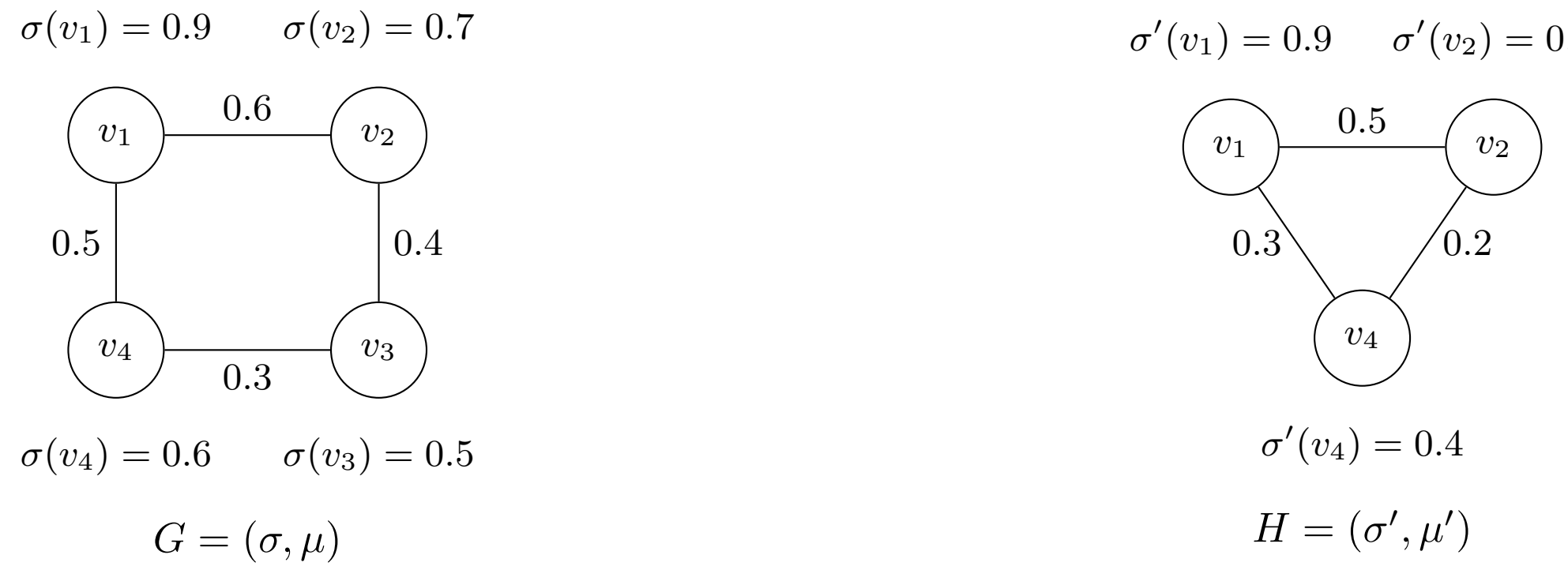

Figure 2.1: A fuzzy graph $G$ and a fuzzy subgraph $H$. Vertex labels indicate the elements of $V$, numbers near vertices represent vertex-memberships, and numbers on edges represent edge-memberships.

## 2.2 Intuitionistic Fuzzy Graph

An intuitionistic fuzzy set assigns each element membership and nonmembership degrees, with their sum at most one, explicitly representing hesitation and incomplete information in contexts [91,92]. An intuitionistic fuzzy graph extends a graph by assigning membership and nonmembership degrees to vertices and edges, thereby modeling uncertain relations, partial connectivity, and hesitation [93].

**Definition 2.2.1** (Intuitionistic Fuzzy Graph)**.** Let $V$ be a nonempty vertex set. An *intuitionistic fuzzy graph* on $V$ is a pair

$$G = (A, B),$$

where

$$A = \{(v, \mu_A(v), \nu_A(v)) : v \in V\}$$

is an intuitionistic fuzzy set on $V$, and

$$B = \{((u, v), \mu_B(u, v), \nu_B(u, v)) : u, v \in V\}$$

is an intuitionistic fuzzy relation on $V$, satisfying

$$0 \leq \mu_A(v) + \nu_A(v) \leq 1 \qquad \text{for all } v \in V,$$

and

$$\mu_B(u, v) \leq \min\{\mu_A(u), \mu_A(v)\}, \qquad \nu_B(u, v) \geq \max\{\nu_A(u), \nu_A(v)\}$$

for all $u, v \in V$, with

$$0 \leq \mu_B(u, v) + \nu_B(u, v) \leq 1.$$

Here, $\mu_A$ and $\mu_B$ denote the membership degrees, while $\nu_A$ and $\nu_B$ denote the non-membership degrees of vertices and edges, respectively.

**Example 2.2.2** (An intuitionistic fuzzy graph)**.** Let

$$V = \{v_1, v_2, v_3, v_4\}.$$

Define an intuitionistic fuzzy set

$$A = \{(v, \mu_A(v), \nu_A(v)) : v \in V\}$$

on $V$ by

$$A = \{(v_1, 0.8, 0.1),\ (v_2, 0.7, 0.2),\ (v_3, 0.6, 0.2),\ (v_4, 0.5, 0.3)\}.$$

Clearly,

$$0 \leq \mu_A(v_i) + \nu_A(v_i) \leq 1 \qquad (i = 1, 2, 3, 4),$$

since

$$0.8 + 0.1 = 0.9, \qquad 0.7 + 0.2 = 0.9, \qquad 0.6 + 0.2 = 0.8, \qquad 0.5 + 0.3 = 0.8.$$

Next, define an intuitionistic fuzzy relation

$$B = \{((u, v), \mu_B(u, v), \nu_B(u, v)) : u, v \in V\}$$

as follows:

$$\mu_B(v_1, v_2) = 0.6, \qquad \nu_B(v_1, v_2) = 0.2,$$

$$\mu_B(v_2, v_3) = 0.5, \qquad \nu_B(v_2, v_3) = 0.3,$$

$$\mu_B(v_3, v_4) = 0.4, \qquad \nu_B(v_3, v_4) = 0.4,$$

$$\mu_B(v_1, v_4) = 0.4, \qquad \nu_B(v_1, v_4) = 0.3,$$

and for all remaining unordered pairs,

$$\mu_B(u, v) = 0, \qquad \nu_B(u, v) = 1.$$

Assume also that $B$ is symmetric, that is,

$$\mu_B(u, v) = \mu_B(v, u), \qquad \nu_B(u, v) = \nu_B(v, u)$$

for all $u, v \in V$.

Now we verify the defining conditions. For example,

$$\mu_B(v_1, v_2) = 0.6 \leq \min\{0.8, 0.7\} = 0.7, \qquad \nu_B(v_1, v_2) = 0.2 \geq \max\{0.1, 0.2\} = 0.2,$$

and

$$0 \leq \mu_B(v_1, v_2) + \nu_B(v_1, v_2) = 0.8 \leq 1.$$

Similarly,

$$\mu_B(v_2, v_3) = 0.5 \leq \min\{0.7, 0.6\} = 0.6, \qquad \nu_B(v_2, v_3) = 0.3 \geq \max\{0.2, 0.2\} = 0.2,$$

$$0 \leq \mu_B(v_2, v_3) + \nu_B(v_2, v_3) = 0.8 \leq 1,$$

$$\mu_B(v_3, v_4) = 0.4 \leq \min\{0.6, 0.5\} = 0.5, \qquad \nu_B(v_3, v_4) = 0.4 \geq \max\{0.2, 0.3\} = 0.3,$$

$$0 \leq \mu_B(v_3, v_4) + \nu_B(v_3, v_4) = 0.8 \leq 1,$$

and

$$\mu_B(v_1, v_4) = 0.4 \leq \min\{0.8, 0.5\} = 0.5, \qquad \nu_B(v_1, v_4) = 0.3 \geq \max\{0.1, 0.3\} = 0.3,$$

$$0 \leq \mu_B(v_1, v_4) + \nu_B(v_1, v_4) = 0.7 \leq 1.$$

Hence,

$$G = (A, B)$$

is an intuitionistic fuzzy graph on $V$. For reference, the illustrative diagram is shown in Figure 2.2.

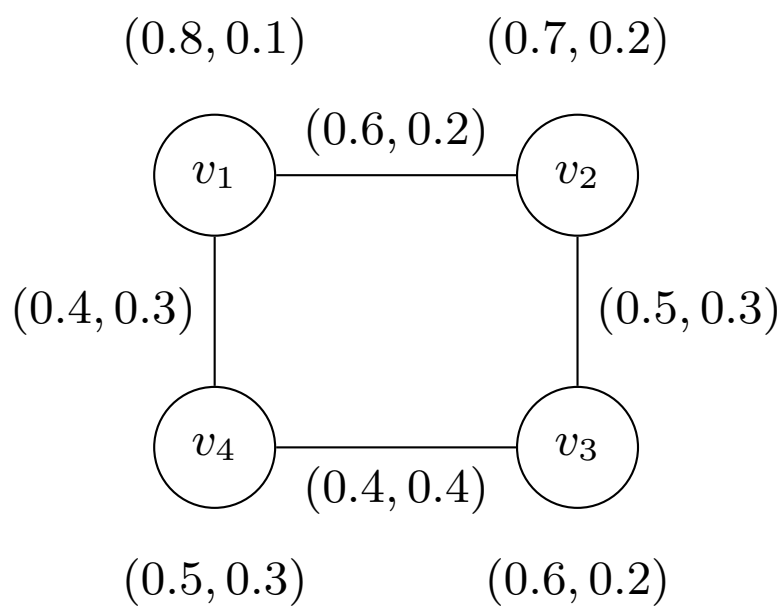

Figure 2.2: An intuitionistic fuzzy graph. The label near each vertex is $(\mu_A, \nu_A)$, and the label on each edge is $(\mu_B, \nu_B)$.

## 2.3 Neutrosophic Graph

A Single-Valued Neutrosophic Graph assigns truth, indeterminacy, and falsity degrees to vertices and edges, extending classical graphs [19, 76, 94, 95].

**Definition 2.3.1** (Single-Valued Neutrosophic Graph)**.** [19] Let $G^* = (V, E)$ be a crisp (classical) graph, where $V$ is the vertex set and $E \subseteq V \times V$ the edge set. A *single-valued neutrosophic graph* (SVNG) on $G^*$ is defined as a pair

$$G = (A, B),$$

where

- $A = \{\langle v, T_A(v), I_A(v), F_A(v)\rangle : v \in V\}$ is the *single-valued neutrosophic vertex set*, with

  $$T_A, I_A, F_A : V \to [0, 1],$$

  denoting respectively the *truth-membership*, *indeterminacy-membership*, and *falsity-membership* functions of vertices, such that for every $v \in V$,

  $$0 \le T_A(v) + I_A(v) + F_A(v) \le 3.$$

- $B = \{\langle uv, T_B(uv), I_B(uv), F_B(uv)\rangle : uv \in E\}$ is the *single-valued neutrosophic edge set*, with

  $$T_B, I_B, F_B : E \to [0, 1],$$

  satisfying for all $u, v \in V$ with $uv \in E$:

  $$T_B(uv) \le \min\{T_A(u), T_A(v)\}, \qquad I_B(uv) \le \min\{I_A(u), I_A(v)\}, \qquad F_B(uv) \ge \max\{F_A(u), F_A(v)\}.$$

If $B$ is symmetric, $G = (A, B)$ is called an *undirected SVNG*; otherwise, it is a *directed SVNG*.

**Example 2.3.2** (A single-valued neutrosophic graph)**.** Let the underlying crisp graph be

$$G^* = (V, E),$$

where

$$V = \{v_1, v_2, v_3, v_4\}$$

and

$$E = \{v_1v_2,\ v_2v_3,\ v_3v_4,\ v_1v_4\}.$$

Define the single-valued neutrosophic vertex set

$$A = \{\langle v, T_A(v), I_A(v), F_A(v)\rangle : v \in V\}$$

by

$$A = \{\langle v_1, 0.8, 0.2, 0.1\rangle,\ \langle v_2, 0.7, 0.3, 0.2\rangle,\ \langle v_3, 0.6, 0.2, 0.3\rangle,\ \langle v_4, 0.5, 0.4, 0.2\rangle\}.$$

Then, for each vertex $v_i \in V$,

$$0 \le T_A(v_i) + I_A(v_i) + F_A(v_i) \le 3.$$

Indeed,

$$0.8 + 0.2 + 0.1 = 1.1, \qquad 0.7 + 0.3 + 0.2 = 1.2,$$
$$0.6 + 0.2 + 0.3 = 1.1, \qquad 0.5 + 0.4 + 0.2 = 1.1.$$

Next, define the single-valued neutrosophic edge set

$$B = \{\langle uv, T_B(uv), I_B(uv), F_B(uv)\rangle : uv \in E\}$$

by

$$B = \{\langle v_1v_2, 0.6, 0.2, 0.2\rangle,\ \langle v_2v_3, 0.5, 0.2, 0.3\rangle,\ \langle v_3v_4, 0.4, 0.2, 0.3\rangle,\ \langle v_1v_4, 0.4, 0.2, 0.2\rangle\}.$$

We now verify the defining conditions.

For the edge $v_1v_2$,

$$T_B(v_1v_2) = 0.6 \le \min\{0.8, 0.7\} = 0.7,$$

$$I_B(v_1v_2) = 0.2 \le \min\{0.2, 0.3\} = 0.2,$$

$$F_B(v_1v_2) = 0.2 \ge \max\{0.1, 0.2\} = 0.2.$$

For the edge $v_2v_3$,

$$T_B(v_2v_3) = 0.5 \le \min\{0.7, 0.6\} = 0.6,$$

$$I_B(v_2v_3) = 0.2 \le \min\{0.3, 0.2\} = 0.2,$$

$$F_B(v_2v_3) = 0.3 \ge \max\{0.2, 0.3\} = 0.3.$$

For the edge $v_3v_4$,

$$T_B(v_3v_4) = 0.4 \le \min\{0.6, 0.5\} = 0.5,$$

$$I_B(v_3v_4) = 0.2 \le \min\{0.2, 0.4\} = 0.2,$$

$$F_B(v_3v_4) = 0.3 \ge \max\{0.3, 0.2\} = 0.3.$$

For the edge $v_1v_4$,

$$T_B(v_1v_4) = 0.4 \le \min\{0.8, 0.5\} = 0.5,$$

$$I_B(v_1v_4) = 0.2 \le \min\{0.2, 0.4\} = 0.2,$$

$$F_B(v_1v_4) = 0.2 \ge \max\{0.1, 0.2\} = 0.2.$$

Hence $G = (A, B)$ is a single-valued neutrosophic graph. Since the edge assignments are symmetric, $G$ is an undirected SVNG.

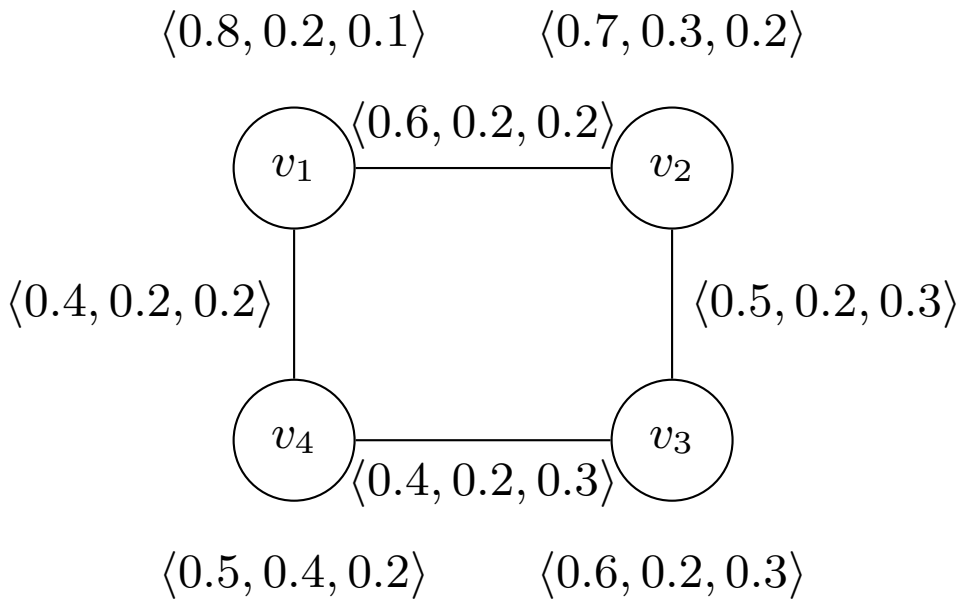

Figure 2.3: A single-valued neutrosophic graph. The label near each vertex is $\langle T_A, I_A, F_A\rangle$, and the label on each edge is $\langle T_B, I_B, F_B\rangle$.

## 2.4 Plithogenic Graph

A plithogenic set models elements through attribute values, appurtenance degrees, and contradiction degrees, capturing multi-valued, attribute-dependent uncertainty, diversity, inconsistency, and context in complex systems formally [32, 49]. A plithogenic graph extends graphs using attribute-based appurtenance and contradiction degrees on vertices and edges, representing heterogeneous, context-sensitive, multi-valued relationships under uncertainty and inconsistency formally [83].

**Definition 2.4.1** (Plithogenic Set)**.** [32, 49] Let $S$ be a universal set and $P \subseteq S$ a nonempty subset. A *Plithogenic Set* is a quintuple

$$PS = (P, v, Pv, pdf, pCF),$$

where

- $v$ is an attribute,
- $Pv$ is the set of possible values of the attribute $v$,
- $pdf : P \times Pv \to [0,1]^s$ is the *Degree of Appurtenance Function (DAF)*,[1]
- $pCF : Pv \times Pv \to [0,1]^t$ is the *Degree of Contradiction Function (DCF)*.

The DCF satisfies, for all $a, b \in Pv$,

$$\text{Reflexivity: } pCF(a,a) = 0, \qquad \text{Symmetry: } pCF(a,b) = pCF(b,a).$$

Here $s \in \mathbb{N}$ is the appurtenance dimension and $t \in \mathbb{N}$ the contradiction dimension.

**Definition 2.4.2** (Plithogenic Graph)**.** (cf. [33, 83]) Let $G = (V, E)$ be a crisp (simple, undirected) graph with $E \subseteq \{\{x, y\} : x, y \in V, \ x \neq y\}$. A *plithogenic graph* is a pair

$$PG = (PM, PN),$$

where the vertex and edge components are specified as follows.

**Vertex component.**

$$PM = (M, \ \ell, \ ML, \ \text{adf}, \ \text{aCf}),$$

with

- $M \subseteq V$ a chosen vertex subset;
- $\ell$ an attribute attached to vertices;
- $ML$ the set of possible values of $\ell$;
- $\text{adf} : M \times ML \to [0,1]^s$ the vertex DAF;
- $\text{aCf} : ML \times ML \to [0,1]^t$ the vertex DCF.

**Edge component.**

$$PN = (N, \ m, \ ML', \ \text{bdf}, \ \text{bCf}),$$

with

[1] In the literature, DAF is defined in slightly different ways: some variants use powerset–valued constructions, others the simple cube $[0,1]^s$. We adopt the latter (classical) form here; cf. [96].

- $N \subseteq E$ a chosen edge subset;
- $m$ an attribute attached to edges;
- $ML'$ the set of possible values of $m$;
- $\mathrm{bdf} : N \times ML' \to [0,1]^s$ the edge DAF;
- $\mathrm{bCf} : ML' \times ML' \to [0,1]^t$ the edge DCF.

All inequalities in $[0,1]^k$ are interpreted *componentwise*. Fix $s, t \in \mathbb{N}$. The following axioms are required.

(A1) **Edge–vertex compatibility (appurtenance bound).** For all $\{x, y\} \in N$ and $a, b \in ML$,

$$\mathrm{bdf}\big(\{x, y\}, (a, b)\big) \ \le \ \min\big\{\mathrm{adf}(x, a),\ \mathrm{adf}(y, b)\big\}. \tag{2.1}$$

(A2) **Contradiction consistency (edge vs. vertices).** For all $(a, b), (c, d) \in ML'$,

$$\mathrm{bCf}\big((a, b), (c, d)\big) \ \le \ \min\big\{\mathrm{aCf}(a, c),\ \mathrm{aCf}(b, d)\big\}. \tag{2.2}$$

(A3) **Reflexivity and symmetry of DCF.**

$$\mathrm{aCf}(u, u) = 0, \quad \mathrm{aCf}(u, v) = \mathrm{aCf}(v, u) \qquad (\forall\, u, v \in ML),$$

$$\mathrm{bCf}(u, u) = 0, \quad \mathrm{bCf}(u, v) = \mathrm{bCf}(v, u) \qquad (\forall\, u, v \in ML').$$

When $s = t = 1$, all maps are scalar-valued in $[0, 1]$ and (2.1)–(2.2) are scalar inequalities.

**Example 2.4.3** (A plithogenic graph)**.** We construct a simple scalar-valued plithogenic graph, so we take

$$s = t = 1.$$

Let the underlying crisp graph be

$$G = (V, E), \qquad V = \{v_1, v_2, v_3\}, \qquad E = \big\{\{v_1, v_2\}, \{v_2, v_3\}\big\}.$$

We use the vertex attribute

$$\ell = \text{reliability level},$$

with possible values

$$ML = \{H, L\},$$

where $H$ means *high* and $L$ means *low*.

We also use the edge attribute

$$m = \text{interaction type},$$

and we take

$$ML' = ML \times ML = \{(H, H), (H, L), (L, H), (L, L)\}.$$

Define the vertex component

$$PM = (M, \ell, ML, \mathrm{adf}, \mathrm{aCf}),$$

where

$$M = V.$$

Let the vertex degree-of-appurtenance function

$$\mathrm{adf} : M \times ML \to [0, 1]$$

be given by

$$\mathrm{adf}(v_1, H) = 0.9, \qquad \mathrm{adf}(v_1, L) = 0.2,$$
$$\mathrm{adf}(v_2, H) = 0.8, \qquad \mathrm{adf}(v_2, L) = 0.3,$$
$$\mathrm{adf}(v_3, H) = 0.4, \qquad \mathrm{adf}(v_3, L) = 0.7.$$

Define the vertex contradiction function

$$\mathrm{aCf} : ML \times ML \to [0,1]$$

by

$$\mathrm{aCf}(H, H) = 0, \qquad \mathrm{aCf}(L, L) = 0,$$
$$\mathrm{aCf}(H, L) = \mathrm{aCf}(L, H) = 0.6.$$

Thus aCf is reflexive and symmetric.

Next, define the edge component

$$PN = (N, m, ML', \mathrm{bdf}, \mathrm{bCf}),$$

where

$$N = E.$$

Let the edge degree-of-appurtenance function

$$\mathrm{bdf} : N \times ML' \to [0,1]$$

be defined as follows.

For the edge $\{v_1, v_2\}$,

$$\mathrm{bdf}(\{v_1, v_2\}, (H, H)) = 0.7, \qquad \mathrm{bdf}(\{v_1, v_2\}, (H, L)) = 0.2,$$
$$\mathrm{bdf}(\{v_1, v_2\}, (L, H)) = 0.2, \qquad \mathrm{bdf}(\{v_1, v_2\}, (L, L)) = 0.1.$$

For the edge $\{v_2, v_3\}$,

$$\mathrm{bdf}(\{v_2, v_3\}, (H, H)) = 0.4, \qquad \mathrm{bdf}(\{v_2, v_3\}, (H, L)) = 0.5,$$
$$\mathrm{bdf}(\{v_2, v_3\}, (L, H)) = 0.2, \qquad \mathrm{bdf}(\{v_2, v_3\}, (L, L)) = 0.3.$$

Now define the edge contradiction function

$$\mathrm{bCf} : ML' \times ML' \to [0,1]$$

by

$$\mathrm{bCf}\big((a, b), (c, d)\big) = \min\big\{\mathrm{aCf}(a, c),\ \mathrm{aCf}(b, d)\big\}$$

for all $(a, b), (c, d) \in ML'$.

In particular,

$$\mathrm{bCf}\big((H, H), (H, H)\big) = 0, \qquad \mathrm{bCf}\big((L, L), (L, L)\big) = 0,$$
$$\mathrm{bCf}\big((H, H), (H, L)\big) = \min\{0, 0.6\} = 0,$$
$$\mathrm{bCf}\big((H, H), (L, L)\big) = \min\{0.6, 0.6\} = 0.6.$$

Hence bCf is also reflexive and symmetric.

We verify the axioms.

**(A1) Edge–vertex compatibility.** For the edge $\{v_1, v_2\}$, we have

$$\mathrm{bdf}(\{v_1, v_2\}, (H, H)) = 0.7 \le \min\{\mathrm{adf}(v_1, H), \mathrm{adf}(v_2, H)\} = \min\{0.9, 0.8\} = 0.8,$$

$$\mathrm{bdf}(\{v_1, v_2\}, (H, L)) = 0.2 \le \min\{\mathrm{adf}(v_1, H), \mathrm{adf}(v_2, L)\} = \min\{0.9, 0.3\} = 0.3,$$
$$\mathrm{bdf}(\{v_1, v_2\}, (L, H)) = 0.2 \le \min\{\mathrm{adf}(v_1, L), \mathrm{adf}(v_2, H)\} = \min\{0.2, 0.8\} = 0.2,$$
$$\mathrm{bdf}(\{v_1, v_2\}, (L, L)) = 0.1 \le \min\{\mathrm{adf}(v_1, L), \mathrm{adf}(v_2, L)\} = \min\{0.2, 0.3\} = 0.2.$$

Similarly, for the edge $\{v_2, v_3\}$,

$$\mathrm{bdf}(\{v_2, v_3\}, (H, H)) = 0.4 \le \min\{0.8, 0.4\} = 0.4,$$

$$\mathrm{bdf}(\{v_2, v_3\}, (H, L)) = 0.5 \le \min\{0.8, 0.7\} = 0.7,$$
$$\mathrm{bdf}(\{v_2, v_3\}, (L, H)) = 0.2 \le \min\{0.3, 0.4\} = 0.3,$$
$$\mathrm{bdf}(\{v_2, v_3\}, (L, L)) = 0.3 \le \min\{0.3, 0.7\} = 0.3.$$

**(A2) Contradiction consistency.** Because

$$\mathrm{bCf}\big((a, b), (c, d)\big) = \min\big\{\mathrm{aCf}(a, c), \mathrm{aCf}(b, d)\big\},$$

we automatically have

$$\mathrm{bCf}\big((a, b), (c, d)\big) \le \min\big\{\mathrm{aCf}(a, c), \mathrm{aCf}(b, d)\big\}$$

for all $(a, b), (c, d) \in ML'$.

**(A3) Reflexivity and symmetry.** These hold by construction for both aCf and bCf.

Therefore,

$$PG = (PM, PN)$$

is a plithogenic graph.

## 2.5 Uncertain Graph

An Uncertain Set assigns to each element a degree from an uncertainty model, unifying fuzzy, intuitionistic, neutrosophic and plithogenic frameworks [97]. An Uncertain Graph is a graph where vertices or edges carry degrees in an uncertainty model, subsuming fuzzy, intuitionistic, neutrosophic. We first recall the notion of an Uncertain Model, which provides the membership–degree domain.

**Definition 2.5.1** (Uncertain Model)**.** [97] Let $U$ denote the class of all *uncertain models*. Each $M \in U$ is specified by

- a nonempty set $\mathrm{Dom}(M) \subseteq [0, 1]^k$ of *admissible degree tuples* for some fixed integer $k \ge 1$;
- model–specific algebraic or geometric constraints on elements of $\mathrm{Dom}(M)$ (for example, $\mu + \nu \le 1$ in the intuitionistic fuzzy case, or $T + I + F \le 3$ in the neutrosophic case).

Typical examples include:

- Fuzzy model: $\mathrm{Dom}(M) = [0, 1]$;

- Intuitionistic fuzzy model: $\mathrm{Dom}(M) = \{(\mu, \nu) \in [0, 1]^2 \mid \mu + \nu \leq 1\}$;
- Neutrosophic model: $\mathrm{Dom}(M) = \{(T, I, F) \in [0, 1]^3 \mid 0 \leq T + I + F \leq 3\}$;
- Plithogenic model, and many other extensions.

**Definition 2.5.2** (Uncertain Set (U-Set))**.** [97] Let $X$ be a nonempty universe, and let $M$ be a fixed uncertain model with degree–domain $\mathrm{Dom}(M) \subseteq [0, 1]^k$. An *Uncertain Set of type M* (or *U-Set* for short) on $X$ is a pair

$$\mathcal{U} = (X, \mu_M),$$

where

$$\mu_M : X \longrightarrow \mathrm{Dom}(M)$$

is called the *uncertainty–degree function* (or membership map) of $\mathcal{U}$.

For $x \in X$, the value $\mu_M(x) \in \mathrm{Dom}(M)$ encodes the degree(s) to which $x$ belongs to the uncertain set, according to the model $M$.

**Remark 2.5.3.** Special cases:

- If $M$ is the fuzzy model and $\mathrm{Dom}(M) = [0, 1]$, then $\mu_M : X \to [0, 1]$ is a usual fuzzy membership function and $\mathcal{U}$ is a fuzzy set.
- If $M$ is neutrosophic, then $\mu_M(x) = (T(x), I(x), F(x))$ gives a neutrosophic set.
- Other choices of $M$ recover intuitionistic fuzzy sets, picture fuzzy sets, plithogenic sets, and so on.

As noted in the remark, various generalizations are possible. For reference, Table 2.1 presents a catalogue of uncertainty-set families (U-Sets) organized by the dimension $k$ of the degree-domain $\mathrm{Dom}(M) \subseteq [0, 1]^k$ (cf. [98]).

Table 2.1: A catalogue of uncertainty-set families (U-Sets) by the dimension $k$ of the degree-domain $\mathrm{Dom}(M) \subseteq [0, 1]^k$ [98].

| $k$ | note | Representative U-Set model(s) whose degree-domain is a subset of $[0, 1]^k$ |
|---|---|---|
| 1 | | Fuzzy Set [16,36]; N-Fuzzy Set [99–101] Shadowed Set [102–104] |
| 2 | | Intuitionistic Fuzzy Set [17,92]; Vague Set [20,55]; Bipolar Fuzzy Set (two-component description) [105]; Variable Fuzzy Set [106–108]; Paraconsistent Fuzzy Set [109,110]; Bifuzzy Set [111,112] |
| 3 | | Single-Valued Neutrosophic Set [113,114]; Picture Fuzzy Set [22,115]; Spherical Fuzzy Set [116,117]; Tripolar Fuzzy Set (three-component formalisms) [118–120]; Neutrosophic Vague Set [69,121] |
| 4 | | Quadripartitioned Neutrosophic Set [23,122]; Double-Valued Neutrosophic Set [123,124]; Dual Hesitant Fuzzy Set [125,126]; Ambiguous Set [127–129]; Turiyam Neutrosophic Set [130–133] |
| 5 | | Pentapartitioned Neutrosophic Set [134–136]; Triple-Valued Neutrosophic Set [137–139] |
| 6 | | Hexapartitioned Neutrosophic Set; Quadruple-Valued Neutrosophic Set [138,140] |
| 7 | | Heptapartitioned Neutrosophic Set; Quintuple-Valued Neutrosophic Set [138,141,142] |
| 8 | | Octapartitioned Neutrosophic Set [143,144] |
| 9 | | Nonapartitioned Neutrosophic Set [143,144] |
| $n$ | $(n \geq 1)$ | Multi-valued (Fuzzy) Sets [145]; MultiFuzzy Set [146]; $n$-Refined Fuzzy Set [147,148] |
| $2n$ | $(n \geq 1)$ | $n$-Refined Intuitionistic Fuzzy Set [148]; Multi-Intuitionistic Fuzzy Set [146] |
| $3n$ | $(n \geq 1)$ | $n$-Refined Neutrosophic Set [148]; Multi-Neutrosophic Set [146,149] |

**Reading guide.** In the U-Set scheme [97], each model $M$ is specified by a degree-domain $\mathrm{Dom}(M) \subseteq [0, 1]^k$ and a membership map $\mu_M : X \to \mathrm{Dom}(M)$. The table groups representative families by the ambient dimension $k$ (i.e., how many numerical components are stored per element).
(a) A widely cited viewpoint is that neutrosophic sets provide a unifying umbrella covering several earlier multi-component fuzzy models (and their generalizations); see [150].
(b) Ambiguous sets are commonly presented as subclasses of certain four-component neutrosophic families; see [23,122,129].
(c) Turiyam neutrosophic sets are reported as subclasses of quadripartitioned neutrosophic sets; see [151].

The definitions and related concepts of Uncertain Graphs are presented below.

**Definition 2.5.4** (Uncertain Graph)**.** Let $G = (V, E)$ be a (finite, undirected, loopless) graph and let $M$ be an uncertain model with degree–domain $\mathrm{Dom}(M)$. An *Uncertain Graph of type $M$* is a triple

$$\mathcal{G}_M = (V, E, \mu_M),$$

where

$$\mu_M : V \cup E \longrightarrow \mathrm{Dom}(M)$$

assigns to each vertex $v \in V$ and each edge $e \in E$ an uncertainty degree $\mu_M(v)$ or $\mu_M(e)$ in $\mathrm{Dom}(M)$.

Optionally, one may impose model–specific consistency conditions between vertex and edge degrees (for instance, $\mu_M(e)$ bounded in terms of $\mu_M(u)$ and $\mu_M(v)$ for $e = \{u, v\}$ in fuzzy or intuitionistic fuzzy graph models), but these constraints are encoded in the choice of $M$ and are not fixed at the level of this general definition.

**Remark 2.5.5.** Again, particular choices of $M$ recover well–known graph models:

- Fuzzy graph (when $M$ is fuzzy and $\mu_M : V \cup E \to [0, 1]$);
- Intuitionistic fuzzy graph, neutrosophic graph, plithogenic graph, etc., for the corresponding models $M$.

As a reference, Table 2.2 presents a catalogue of uncertainty-graph families (Uncertain Graphs) organised by the dimension $k$ of the degree-domain $\mathrm{Dom}(M) \subseteq [0, 1]^k$.

Table 2.2: A catalogue of uncertainty-graph families (Uncertain Graphs) by the dimension $k$ of the degree-domain $\mathrm{Dom}(M) \subseteq [0, 1]^k$.

| $k$ | Representative uncertainty-graph type(s) $\mathcal{G}_M = (V, E, \mu_M)$ with $\mu_M : V \cup E \to \mathrm{Dom}(M) \subseteq [0, 1]^k$ |
|---|---|
| 1 | Fuzzy graph; $N$-graph; shadowed-graph variants |
| 2 | Intuitionistic fuzzy graph [152]; vague graph [153]; bipolar fuzzy graph [40]; intuitionistic evidence graph; variable fuzzy graph; paraconsistent fuzzy graph; bifuzzy graph [154, 155] |
| 3 | Neutrosophic graph [19][(a)]; hesitant fuzzy graph [156]; tripolar fuzzy graph; three-way fuzzy graph; picture fuzzy graph [157, 158]; spherical fuzzy graph [116]; inconsistent intuitionistic fuzzy graph; ternary fuzzy / neutrosophic-fuzzy graph; neutrosophic vague graph |
| 4 | Quadripartitioned neutrosophic graph [159, 160]; double-valued neutrosophic graph [123]; dual hesitant fuzzy graph [161]; ambiguous graph[(b)]; local-neutrosophic graph; support-neutrosophic graph; turiyam neutrosophic graph [162][(c)] |
| 5 | Pentapartitioned neutrosophic graph [163]; triple-valued neutrosophic graph |
| 6 | Hexapartitioned neutrosophic graph; quadruple-valued neutrosophic graph |
| 7 | Heptapartitioned neutrosophic graph [164]; quintuple-valued neutrosophic graph |
| 8 | Octapartitioned neutrosophic graph |
| 9 | Nonapartitioned neutrosophic graph |
| $n$ | $n$-refined fuzzy graph; multi-valued (fuzzy) graphs; multi-fuzzy graphs [165] |
| $2n$ | $n$-refined intuitionistic fuzzy graph; multi-intuitionistic fuzzy graphs |
| $3n$ | $n$-refined neutrosophic graph; multi-neutrosophic graphs |

(a) Neutrosophic graph models are often treated as broad frameworks that can specialize to many degree-based graph formalisms under suitable constraints.
(b) Ambiguous-graph models are commonly presented as subclasses of certain quadripartitioned and also double-valued neutrosophic graph models.
(c) Turiyam neutrosophic graphs are reported as subclasses of certain quadripartitioned neutrosophic graph models.

## 2.6 Soft Graph

Soft graph is a parameterized graph structure assigning to each parameter a subgraph, enabling flexible modeling of systems whose relations vary across contexts and scenarios.

**Definition 2.6.1** (Soft Graph)**.** Let $G^* = (V, E)$ be a simple graph, and let $A$ be a nonempty set of parameters. A *soft graph* over $G^*$ is a quadruple

$$G = (G^*, F, K, A),$$

where

$$F : A \to \mathcal{P}(V), \qquad K : A \to \mathcal{P}(E),$$

such that, for every $a \in A$, the pair

$$H(a) = \bigl(F(a), K(a)\bigr)$$

is a subgraph of $G^*$.

## 2.7 Rough Graph

A rough graph represents a graph through lower and upper approximation graphs under an equivalence relation, thereby modeling indiscernibility, vagueness, and boundary uncertainty structurally formally [166, 167].

**Definition 2.7.1** (Rough Graph)**.** Let $U = (V, E)$ be a universe graph, and let $R$ be an equivalence relation on $E$, inducing edge equivalence classes $[e]_R$ for $e \in E$. For a graph $T = (W, X)$ with $W \subseteq V$ and $X \subseteq E$, define

$$\underline{R}(X) = \{\, e \in E : [e]_R \subseteq X \,\}, \qquad \overline{R}(X) = \{\, e \in E : [e]_R \cap X \neq \varnothing \,\}.$$

Then the pair

$$\big(\underline{R}(T), \overline{R}(T)\big) = \big((W, \underline{R}(X)), (W, \overline{R}(X))\big)$$

is called the *rough graph* associated with $T$. If $X$ is not a union of $R$-equivalence classes, then $T$ is said to be an *$R$-rough graph*.

# Chapter 3

# Basic Concepts in Uncertain Graph

In this chapter, we discuss the basic concepts in uncertain graph theory.

## 3.1 Uncertain Path

A fuzzy path is a sequence of distinct vertices connected by positive-membership edges, whose overall strength equals the minimum membership value among its constituent edges.

**Definition 3.1.1** (Fuzzy Path)**.** Let

$$G = (V, \sigma, \mu)$$

be a fuzzy graph, where

$$\sigma : V \to [0, 1], \qquad \mu : V \times V \to [0, 1], \qquad \mu(u, v) \leq \min\{\sigma(u), \sigma(v)\} \quad (\forall\, u, v \in V).$$

A *fuzzy path* of length $n$ in $G$ is a sequence of distinct vertices

$$P : u_0, u_1, \ldots, u_n$$

such that

$$\mu(u_{i-1}, u_i) > 0 \qquad (i = 1, 2, \ldots, n).$$

The *strength* of the fuzzy path $P$ is defined by

$$s(P) := \min_{1 \leq i \leq n} \mu(u_{i-1}, u_i).$$

That is, the strength of $P$ is the membership value of its weakest edge.

**Example 3.1.2** (A fuzzy path and its strength)**.** Let

$$V = \{v_1, v_2, v_3, v_4, v_5\}.$$

Define a vertex-membership function

$$\sigma : V \to [0, 1]$$

by

$$\sigma(v_1) = 0.9, \qquad \sigma(v_2) = 0.8, \qquad \sigma(v_3) = 0.7, \qquad \sigma(v_4) = 0.6, \qquad \sigma(v_5) = 0.5.$$

Next, define an edge-membership function

$$\mu : V \times V \to [0,1]$$

by

$$\mu(v_1, v_2) = 0.6, \qquad \mu(v_2, v_3) = 0.5, \qquad \mu(v_3, v_4) = 0.4,$$
$$\mu(v_2, v_5) = 0.3, \qquad \mu(v_4, v_5) = 0.2,$$

and let

$$\mu(u, v) = 0$$

for all other unordered pairs $\{u, v\} \subseteq V$, with

$$\mu(u, v) = \mu(v, u)$$

for all $u, v \in V$.

Then $G = (V, \sigma, \mu)$ is a fuzzy graph, since each positive edge-membership satisfies the required bound. For example,

$$\mu(v_1, v_2) = 0.6 \leq \min\{0.9, 0.8\} = 0.8,$$

$$\mu(v_2, v_3) = 0.5 \leq \min\{0.8, 0.7\} = 0.7,$$

$$\mu(v_3, v_4) = 0.4 \leq \min\{0.7, 0.6\} = 0.6.$$

Now consider the sequence of distinct vertices

$$P : v_1, v_2, v_3, v_4.$$

Since

$$\mu(v_1, v_2) > 0, \qquad \mu(v_2, v_3) > 0, \qquad \mu(v_3, v_4) > 0,$$

the sequence $P$ is a fuzzy path of length 3.

Its strength is

$$s(P) = \min\bigl\{\mu(v_1, v_2), \mu(v_2, v_3), \mu(v_3, v_4)\bigr\} = \min\{0.6, 0.5, 0.4\} = 0.4.$$

Hence the strength of the fuzzy path $P$ is 0.4, which is the membership value of its weakest edge. An illustrative diagram is shown in Figure 3.1.

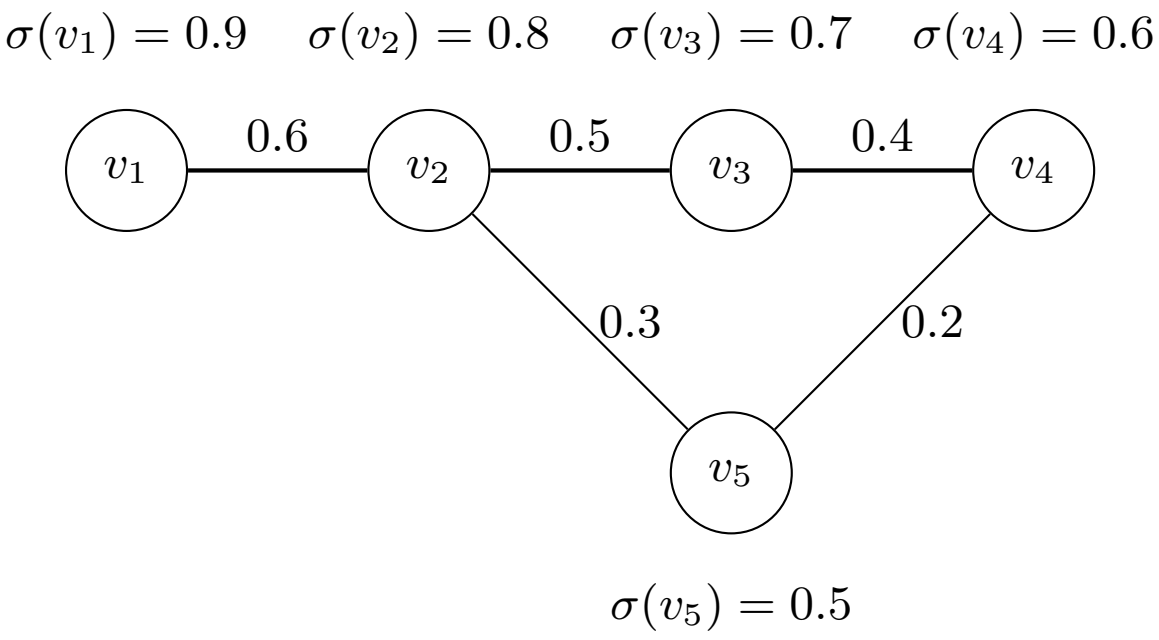

Figure 3.1: A fuzzy graph containing the fuzzy path $P : v_1, v_2, v_3, v_4$. The numbers on vertices indicate vertex-memberships, and the numbers on edges indicate edge-memberships.

An uncertain path is a sequence of distinct vertices joined by support edges in an uncertain graph, together with a model-dependent path strength.

**Definition 3.1.3** (Path-Evaluable Uncertain Model)**.** Let $M$ be an uncertain model with degree-domain

$$\mathrm{Dom}(M) \subseteq [0,1]^k.$$

We say that $M$ is *path-evaluable* if it is equipped with:

1. a distinguished element
$$0_M \in \mathrm{Dom}(M),$$
called the *zero degree*;

2. for each integer $n \geq 1$, a map
$$\Psi_M^{(n)} : \mathrm{Dom}(M)^n \longrightarrow \mathrm{Dom}(M),$$
called the *path-strength operator of length n*.

**Definition 3.1.4** (Uncertain Path)**.** Let $M$ be a path-evaluable uncertain model with degree-domain

$$\mathrm{Dom}(M) \subseteq [0,1]^k,$$

zero degree

$$0_M \in \mathrm{Dom}(M),$$

and path-strength operators

$$\Psi_M^{(n)} : \mathrm{Dom}(M)^n \to \mathrm{Dom}(M) \qquad (n \geq 1).$$

Let

$$\mathcal{G}_M = (V, E, \sigma_M, \eta_M)$$

be an Uncertain Graph of type $M$, where

$$\sigma_M : V \to \mathrm{Dom}(M), \qquad \eta_M : E \to \mathrm{Dom}(M).$$

Define the support edge set by

$$E^*(\mathcal{G}_M) := \{\, e \in E \mid \eta_M(e) \neq 0_M \,\}.$$

An *Uncertain Path* of length $n$ in $\mathcal{G}_M$ is a sequence of distinct vertices

$$P : u_0, u_1, \ldots, u_n$$

such that

$$\{u_{i-1}, u_i\} \in E^*(\mathcal{G}_M) \qquad (i = 1, 2, \ldots, n).$$

The *strength* of the uncertain path $P$ is defined by

$$s_M(P) := \Psi_M^{(n)}\big(\eta_M(\{u_0, u_1\}), \eta_M(\{u_1, u_2\}), \ldots, \eta_M(\{u_{n-1}, u_n\})\big).$$

**Theorem 3.1.5** (Well-definedness of Uncertain Path)**.** *Let $M$ be a path-evaluable uncertain model with degree-domain*

$$\mathrm{Dom}(M) \subseteq [0,1]^k,$$

*zero degree*

$$0_M \in \mathrm{Dom}(M),$$

*and path-strength operators*

$$\Psi_M^{(n)} : \mathrm{Dom}(M)^n \to \mathrm{Dom}(M) \qquad (n \geq 1).$$

*Let*

$$\mathcal{G}_M = (V, E, \sigma_M, \eta_M)$$

*be an Uncertain Graph of type $M$.*

*Then:*

1. *the support edge set*
$$E^*(\mathcal{G}_M) = \{\, e \in E \mid \eta_M(e) \neq 0_M \,\}$$
*is well-defined;*

2. *the statement*
$$\text{“}P : u_0, u_1, \ldots, u_n \text{ is an Uncertain Path of length } n\text{”}$$
*is well-defined;*

3. *for every Uncertain Path*
$$P : u_0, u_1, \ldots, u_n,$$
*the quantity*
$$s_M(P) = \Psi_M^{(n)}\big(\eta_M(\{u_0, u_1\}), \eta_M(\{u_1, u_2\}), \ldots, \eta_M(\{u_{n-1}, u_n\})\big)$$
*is well-defined.*

*Hence the notions of uncertain path and uncertain path strength are well-defined.*

*Proof.* Since $M$ is an uncertain model, its degree-domain
$$\mathrm{Dom}(M)$$
is fixed. Since $M$ is path-evaluable, the element
$$0_M \in \mathrm{Dom}(M)$$
and the maps
$$\Psi_M^{(n)} : \mathrm{Dom}(M)^n \to \mathrm{Dom}(M) \qquad (n \geq 1)$$
are also fixed.

Because
$$\eta_M : E \to \mathrm{Dom}(M)$$
is a function, for each edge $e \in E$ the value $\eta_M(e)$ is uniquely determined in $\mathrm{Dom}(M)$. Therefore the condition
$$\eta_M(e) \neq 0_M$$
is meaningful for every $e \in E$. Hence
$$E^*(\mathcal{G}_M) = \{\, e \in E \mid \eta_M(e) \neq 0_M \,\}$$
is a well-defined subset of $E$.

Now let
$$P : u_0, u_1, \ldots, u_n$$
be a sequence of distinct vertices in $V$. For each $i = 1, \ldots, n$, the unordered pair
$$\{u_{i-1}, u_i\}$$
is uniquely determined. Since $E^*(\mathcal{G}_M) \subseteq E$ is well-defined, the condition
$$\{u_{i-1}, u_i\} \in E^*(\mathcal{G}_M)$$
is meaningful. Therefore the predicate
$$\text{“}P : u_0, u_1, \ldots, u_n \text{ is an Uncertain Path of length } n\text{”}$$
is well-defined.

Assume now that $P : u_0, u_1, \dots, u_n$ is an Uncertain Path. Then for each $i = 1, \dots, n$,

$$\{u_{i-1}, u_i\} \in E^*(\mathcal{G}_M),$$

so in particular

$$\eta_M(\{u_{i-1}, u_i\}) \in \mathrm{Dom}(M).$$

Thus the $n$-tuple

$$\big(\eta_M(\{u_0, u_1\}), \eta_M(\{u_1, u_2\}), \dots, \eta_M(\{u_{n-1}, u_n\})\big)$$

belongs to $\mathrm{Dom}(M)^n$. Since

$$\Psi_M^{(n)} : \mathrm{Dom}(M)^n \to \mathrm{Dom}(M)$$

is a well-defined map, the value

$$\Psi_M^{(n)}\big(\eta_M(\{u_0, u_1\}), \eta_M(\{u_1, u_2\}), \dots, \eta_M(\{u_{n-1}, u_n\})\big)$$

is uniquely determined in $\mathrm{Dom}(M)$. Therefore $s_M(P)$ is well-defined.

Hence both the notion of an uncertain path and its path strength are well-defined. □

## 3.2 Uncertain Cycle

A fuzzy cycle is a cycle in a fuzzy graph having at least two weakest edges, so it forms a non-tree circular uncertain connection structure [168–170].

**Definition 3.2.1** (Fuzzy Cycle)**.** Let

$$G = (V, \sigma, \mu)$$

be a fuzzy graph.

A *cycle* in $G$ is a sequence

$$C : u_0, u_1, \dots, u_{n-1}, u_n$$

such that

$$u_0 = u_n, \qquad n \geq 3,$$

the vertices

$$u_0, u_1, \dots, u_{n-1}$$

are distinct, and

$$\mu(u_{i-1}, u_i) > 0 \qquad (i = 1, 2, \dots, n).$$

Such a cycle $C$ is called a *fuzzy cycle* if it contains more than one weakest edge; equivalently, if the minimum value among

$$\mu(u_0, u_1), \mu(u_1, u_2), \dots, \mu(u_{n-1}, u_n)$$

is attained by at least two edges of $C$.

An uncertain cycle extends this idea from ordinary fuzzy membership values to general uncertainty degrees belonging to an arbitrary uncertain model.

**Definition 3.2.2** (Cycle-Comparable Uncertain Model)**.** Let $M$ be an uncertain model with degree-domain

$$\mathrm{Dom}(M) \subseteq [0, 1]^k.$$

We say that $M$ is *cycle-comparable* if it is equipped with:

1. a distinguished element
$$0_M \in \mathrm{Dom}(M),$$
called the *zero degree*;
2. a total order
$$\preceq_M \subseteq \mathrm{Dom}(M) \times \mathrm{Dom}(M),$$
called the *cycle order*.

The strict part of $\preceq_M$ is denoted by $\prec_M$.

An uncertain cycle is a cycle in the support graph of an uncertain graph such that the minimum edge-degree, with respect to the cycle order, is attained by at least two cycle edges.

**Definition 3.2.3** (Uncertain Cycle)**.** Let $M$ be a cycle-comparable uncertain model with degree-domain
$$\mathrm{Dom}(M) \subseteq [0,1]^k,$$
zero degree
$$0_M \in \mathrm{Dom}(M),$$
and cycle order
$$\preceq_M .$$

Let
$$\mathcal{G}_M = (V, E, \sigma_M, \eta_M)$$
be an Uncertain Graph of type $M$, where
$$\sigma_M : V \to \mathrm{Dom}(M), \qquad \eta_M : E \to \mathrm{Dom}(M)$$
are uncertainty-degree functions on the vertex set and edge set, respectively. Equivalently,
$$(V, \sigma_M)$$
and
$$(E, \eta_M)$$
are Uncertain Sets of type $M$.

Define the support vertex set by
$$V^*(\mathcal{G}_M) := \{\, v \in V \mid \sigma_M(v) \neq 0_M \,\},$$
the support edge set by
$$E^*(\mathcal{G}_M) := \big\{\{u,v\} \in E \mid u, v \in V^*(\mathcal{G}_M),\ \eta_M(\{u,v\}) \neq 0_M\big\},$$
and the support graph by
$$G^*_{\mathrm{supp}}(\mathcal{G}_M) := \big(V^*(\mathcal{G}_M), E^*(\mathcal{G}_M)\big).$$

A *cycle* in $\mathcal{G}_M$ is a sequence
$$C : u_0, u_1, \ldots, u_{n-1}, u_n$$
such that
$$u_0 = u_n, \qquad n \geq 3,$$
the vertices
$$u_0, u_1, \ldots, u_{n-1}$$
are distinct, and
$$\{u_{i-1}, u_i\} \in E^*(\mathcal{G}_M) \qquad (i = 1, 2, \ldots, n).$$

For such a cycle $C$, define its edge set by

$$E(C) := \{\{u_{i-1}, u_i\} \mid i = 1, 2, \ldots, n\}.$$

An edge

$$e \in E(C)$$

is called a *weakest edge* of $C$ if

$$\eta_M(e) \preceq_M \eta_M(f) \qquad \text{for all } f \in E(C).$$

Define the set of weakest edges of $C$ by

$$W_M(C) := \{\, e \in E(C) \mid \eta_M(e) \preceq_M \eta_M(f) \text{ for all } f \in E(C)\}.$$

Then $C$ is called an *Uncertain Cycle* if

$$|W_M(C)| \geq 2.$$

Equivalently, $C$ is an uncertain cycle if the minimum value among

$$\eta_M(\{u_0, u_1\}), \eta_M(\{u_1, u_2\}), \ldots, \eta_M(\{u_{n-1}, u_n\})$$

with respect to $\preceq_M$ is attained by at least two edges of $C$.

**Theorem 3.2.4** (Well-definedness of Uncertain Cycle)**.** *Let $M$ be a cycle-comparable uncertain model with degree-domain*

$$\mathrm{Dom}(M) \subseteq [0, 1]^k,$$

*zero degree*

$$0_M \in \mathrm{Dom}(M),$$

*and cycle order*

$$\preceq_M .$$

*Let*

$$\mathcal{G}_M = (V, E, \sigma_M, \eta_M)$$

*be an Uncertain Graph of type $M$.*

*Then:*

1. *the support sets*

   $$V^*(\mathcal{G}_M) = \{\, v \in V \mid \sigma_M(v) \neq 0_M \,\}$$

   *and*

   $$E^*(\mathcal{G}_M) = \{\{u, v\} \in E \mid u, v \in V^*(\mathcal{G}_M),\ \eta_M(\{u, v\}) \neq 0_M\}$$

   *are well-defined;*

2. *the support graph*

   $$G^*_{\mathrm{supp}}(\mathcal{G}_M) = \big(V^*(\mathcal{G}_M), E^*(\mathcal{G}_M)\big)$$

   *is well-defined;*

3. *for every cycle*
$$C : u_0, u_1, \ldots, u_{n-1}, u_n$$
*in* $G^*_{\mathrm{supp}}(\mathcal{G}_M)$*, the set*
$$E(C)$$
*of its cycle edges and the set*
$$W_M(C)$$
*of its weakest edges are well-defined;*

4. *if*
$$C'$$
*is obtained from* $C$ *by a cyclic permutation of the vertices or by reversing the direction of traversal, then*
$$E(C') = E(C) \qquad \textit{and} \qquad W_M(C') = W_M(C).$$

*Consequently, the statement*

*"C is an Uncertain Cycle"*

*is well-defined and depends only on the underlying cycle in the support graph, not on the chosen starting vertex or orientation.*

*Hence the notion of an uncertain cycle is well-defined.*

*Proof.* Since $M$ is an uncertain model, its degree-domain
$$\mathrm{Dom}(M)$$
is fixed. Since $M$ is cycle-comparable, the element
$$0_M \in \mathrm{Dom}(M)$$
and the total order
$$\preceq_M$$
on $\mathrm{Dom}(M)$ are fixed as well.

Because
$$\sigma_M : V \to \mathrm{Dom}(M)$$
is a function, for each vertex $v \in V$ the value
$$\sigma_M(v) \in \mathrm{Dom}(M)$$
is uniquely determined. Therefore the predicate
$$\sigma_M(v) \neq 0_M$$
is meaningful for every $v \in V$, and hence
$$V^*(\mathcal{G}_M) = \{\, v \in V \mid \sigma_M(v) \neq 0_M \,\}$$
is a well-defined subset of $V$.

Likewise, because
$$\eta_M : E \to \mathrm{Dom}(M)$$
is a function, for each edge $e \in E$ the value
$$\eta_M(e) \in \mathrm{Dom}(M)$$

is uniquely determined. Hence the condition

$$\eta_M(e) \neq 0_M$$

is meaningful for every $e \in E$. Therefore

$$E^*(\mathcal{G}_M) = \big\{\{u,v\} \in E \mid u,v \in V^*(\mathcal{G}_M),\ \eta_M(\{u,v\}) \neq 0_M\big\}$$

is a well-defined subset of $E$. Consequently, the support graph

$$G^*_{\text{supp}}(\mathcal{G}_M) = \big(V^*(\mathcal{G}_M), E^*(\mathcal{G}_M)\big)$$

is well-defined. This proves (1) and (2).

Now let

$$C : u_0, u_1, \ldots, u_{n-1}, u_n$$

be a cycle in $G^*_{\text{supp}}(\mathcal{G}_M)$. For each

$$i = 1, 2, \ldots, n,$$

the unordered pair

$$\{u_{i-1}, u_i\}$$

is uniquely determined and belongs to

$$E^*(\mathcal{G}_M) \subseteq E.$$

Therefore

$$E(C) = \big\{\{u_{i-1}, u_i\} \mid i = 1, 2, \ldots, n\big\}$$

is a well-defined finite nonempty subset of $E$.

Since

$$\eta_M : E \to \mathrm{Dom}(M),$$

the set of cycle-edge degrees

$$\eta_M(E(C)) := \{\, \eta_M(e) \mid e \in E(C)\,\}$$

is a well-defined finite nonempty subset of $\mathrm{Dom}(M)$. Because $\preceq_M$ is a total order on $\mathrm{Dom}(M)$, every finite nonempty subset of $\mathrm{Dom}(M)$ has a unique minimum with respect to $\preceq_M$. Denote this minimum by

$$m_C.$$

Then

$$W_M(C) = \{\, e \in E(C) \mid \eta_M(e) = m_C\,\}$$

is a well-defined subset of $E(C)$. Hence (3) holds.

It remains to prove (4). If $C'$ is obtained from $C$ by a cyclic permutation of the vertices, then $C'$ traverses exactly the same consecutive unordered pairs as $C$, only with a different starting point. Therefore

$$E(C') = E(C).$$

If $C'$ is obtained from $C$ by reversing the direction of traversal, then each edge

$$\{u_{i-1}, u_i\}$$

is replaced by

$$\{u_i, u_{i-1}\},$$

which is the same unordered pair. Hence again

$$E(C') = E(C).$$

Since the edge set is unchanged and $\eta_M$ is a function on $E$, the multiset of cycle-edge degrees is unchanged as well. Therefore the minimum degree $m_C$ and the set of weakest edges are unchanged:

$$W_M(C') = W_M(C).$$

Thus the property

$$|W_M(C)| \geq 2$$

does not depend on the choice of representative of the same geometric cycle. Consequently, the statement

"$C$ is an Uncertain Cycle"

is well-defined and depends only on the underlying cycle in the support graph.

Hence the notion of an uncertain cycle is well-defined. □

## 3.3 Uncertain Tree

A fuzzy tree is a connected fuzzy graph having a spanning tree-like fuzzy subgraph, where every non-tree edge is weaker than the corresponding connecting path [171, 172].

**Definition 3.3.1** (Fuzzy Tree)**.** A connected fuzzy graph

$$G = (V, \sigma, \mu)$$

is called a *fuzzy tree* if there exists a fuzzy spanning subgraph

$$F = (V, \sigma, \nu)$$

whose underlying crisp graph is a tree, and for every edge $(x, y)$ not in $F$, there exists an $x$-$y$ path in $F$ having strength greater than $\mu(x, y)$.

An uncertain tree is a connected uncertain graph having a spanning uncertain subgraph whose support graph is a tree, and every non-tree support edge is strictly weaker than the corresponding support path in that spanning subgraph.

**Definition 3.3.2** (Tree-Evaluable Uncertain Model)**.** Let $M$ be an uncertain model with degree-domain

$$\mathrm{Dom}(M) \subseteq [0, 1]^k.$$

We say that $M$ is *tree-evaluable* if it is equipped with:

1. a distinguished element
$$0_M \in \mathrm{Dom}(M),$$
called the *zero degree*;
2. a strict binary relation
$$\prec_M \subseteq \mathrm{Dom}(M) \times \mathrm{Dom}(M),$$
called the *strength order*;
3. for each integer $n \geq 1$, a map
$$\Psi_M^{(n)} : \mathrm{Dom}(M)^n \to \mathrm{Dom}(M),$$
called the *path-strength operator of length $n$*.

An uncertain tree is a connected uncertain graph having a spanning uncertain subgraph whose support graph is a tree, and every non-tree support edge is strictly weaker than the corresponding support path in that spanning subgraph.

**Definition 3.3.3** (Tree-Evaluable Uncertain Model)**.** Let $M$ be an uncertain model with degree-domain

$$\mathrm{Dom}(M) \subseteq [0, 1]^k.$$

We say that $M$ is *tree-evaluable* if it is equipped with:

1. a distinguished element
$$0_M \in \mathrm{Dom}(M),$$
called the *zero degree*;
2. a strict binary relation
$$\prec_M \subseteq \mathrm{Dom}(M) \times \mathrm{Dom}(M),$$
called the *strength order*;
3. for each integer $n \geq 1$, a map
$$\Psi_M^{(n)} : \mathrm{Dom}(M)^n \to \mathrm{Dom}(M),$$
called the *path-strength operator of length n*.

**Definition 3.3.4** (Uncertain Tree)**.** Let $M$ be a tree-evaluable uncertain model with degree-domain
$$\mathrm{Dom}(M) \subseteq [0,1]^k,$$
zero degree
$$0_M \in \mathrm{Dom}(M),$$
strength order
$$\prec_M,$$
and path-strength operators
$$\Psi_M^{(n)} : \mathrm{Dom}(M)^n \to \mathrm{Dom}(M) \qquad (n \geq 1).$$

Let
$$\mathcal{G}_M = (V, E, \sigma_M, \eta_M)$$
be an Uncertain Graph of type $M$.

Define the support edge set of $\mathcal{G}_M$ by
$$E^*(\mathcal{G}_M) := \{\, e \in E \mid \eta_M(e) \neq 0_M \,\},$$
and its support graph by
$$G^*_{\mathrm{supp}}(\mathcal{G}_M) := \big(V, E^*(\mathcal{G}_M)\big).$$

The graph $\mathcal{G}_M$ is called an *Uncertain Tree* if:

1. the support graph $G^*_{\mathrm{supp}}(\mathcal{G}_M)$ is connected;
2. there exists an uncertain spanning subgraph
$$\mathcal{F}_M = (V, E_F, \sigma_M, \eta_F)$$
of $\mathcal{G}_M$ such that the support graph
$$G^*_{\mathrm{supp}}(\mathcal{F}_M) := \big(V, E^*(\mathcal{F}_M)\big), \qquad E^*(\mathcal{F}_M) := \{\, e \in E_F \mid \eta_F(e) \neq 0_M \,\},$$
is a crisp tree;
3. for every support edge
$$e = \{x, y\} \in E^*(\mathcal{G}_M) \setminus E^*(\mathcal{F}_M),$$
there exists an $x$-$y$ path
$$P : x = u_0, u_1, \ldots, u_n = y$$
in $G^*_{\mathrm{supp}}(\mathcal{F}_M)$ such that
$$\eta_M(e) \prec_M s^{\mathcal{F}}_M(P),$$
where
$$s^{\mathcal{F}}_M(P) := \Psi_M^{(n)}\big(\eta_F(\{u_0,u_1\}), \eta_F(\{u_1,u_2\}), \ldots, \eta_F(\{u_{n-1},u_n\})\big).$$

**Theorem 3.3.5** (Well-definedness of Uncertain Tree)**.** *Let $M$ be a tree-evaluable uncertain model with degree-domain*

$$\mathrm{Dom}(M) \subseteq [0,1]^k,$$

*zero degree*

$$0_M \in \mathrm{Dom}(M),$$

*strength order*

$$\prec_M,$$

*and path-strength operators*

$$\Psi_M^{(n)} : \mathrm{Dom}(M)^n \to \mathrm{Dom}(M) \qquad (n \geq 1).$$

*Let*

$$\mathcal{G}_M = (V, E, \sigma_M, \eta_M)$$

*be an Uncertain Graph of type $M$.*

*Then:*

1. *the support edge set*
$$E^*(\mathcal{G}_M) = \{\, e \in E \mid \eta_M(e) \neq 0_M \,\}$$
*and the support graph*
$$G^*_{\mathrm{supp}}(\mathcal{G}_M) = \big(V, E^*(\mathcal{G}_M)\big)$$
*are well-defined;*
2. *for every uncertain spanning subgraph*
$$\mathcal{F}_M = (V, E_F, \sigma_M, \eta_F)$$
*of $\mathcal{G}_M$, the support graph*
$$G^*_{\mathrm{supp}}(\mathcal{F}_M) = \big(V, E^*(\mathcal{F}_M)\big)$$
*is well-defined;*
3. *for every support edge*
$$e = \{x, y\} \in E^*(\mathcal{G}_M) \setminus E^*(\mathcal{F}_M),$$
*the statement*
"*there exists an $x$-$y$ path $P$ in $G^*_{\mathrm{supp}}(\mathcal{F}_M)$ such that $\eta_M(e) \prec_M s_M^{\mathcal{F}}(P)$*"
*is well-defined.*

*Consequently, the statement*

"*$\mathcal{G}_M$ is an Uncertain Tree*"

*is well-defined. Hence the notion of an uncertain tree is well-defined.*

*Proof.* Since $M$ is an uncertain model, its degree-domain

$$\mathrm{Dom}(M)$$

is fixed. Since $M$ is tree-evaluable, the element

$$0_M \in \mathrm{Dom}(M),$$

the strict relation

$$\prec_M \subseteq \mathrm{Dom}(M) \times \mathrm{Dom}(M),$$

and the maps

$$\Psi_M^{(n)} : \mathrm{Dom}(M)^n \to \mathrm{Dom}(M)$$

are all fixed.

Because

$$\eta_M : E \to \mathrm{Dom}(M)$$

is a function, for each edge $e \in E$ the value $\eta_M(e)$ is uniquely determined in $\mathrm{Dom}(M)$. Therefore the condition

$$\eta_M(e) \neq 0_M$$

is meaningful for every $e \in E$, and so

$$E^*(\mathcal{G}_M) = \{\, e \in E \mid \eta_M(e) \neq 0_M \,\}$$

is a well-defined subset of $E$. Hence

$$G^*_{\mathrm{supp}}(\mathcal{G}_M) = \big(V, E^*(\mathcal{G}_M)\big)$$

is a well-defined graph.

Now let

$$\mathcal{F}_M = (V, E_F, \sigma_M, \eta_F)$$

be an uncertain spanning subgraph of $\mathcal{G}_M$. By definition,

$$E_F \subseteq E \qquad \text{and} \qquad \eta_F = \eta_M|_{E_F}.$$

Hence $\eta_F$ is a well-defined function on $E_F$. Therefore

$$E^*(\mathcal{F}_M) = \{\, e \in E_F \mid \eta_F(e) \neq 0_M \,\}$$

is well-defined, and so

$$G^*_{\mathrm{supp}}(\mathcal{F}_M) = \big(V, E^*(\mathcal{F}_M)\big)$$

is also well-defined.

Next, let

$$e = \{x, y\} \in E^*(\mathcal{G}_M) \setminus E^*(\mathcal{F}_M).$$

Because $G^*_{\mathrm{supp}}(\mathcal{F}_M)$ is a well-defined crisp graph, the statement

"$P : x = u_0, u_1, \ldots, u_n = y$ is an $x$-$y$ path in $G^*_{\mathrm{supp}}(\mathcal{F}_M)$"

is well-defined in the ordinary graph-theoretic sense.

For such a path $P$, each edge

$$\{u_{i-1}, u_i\} \in E^*(\mathcal{F}_M)$$

has a uniquely determined uncertainty degree

$$\eta_F(\{u_{i-1}, u_i\}) \in \mathrm{Dom}(M).$$

Hence the tuple

$$\big(\eta_F(\{u_0, u_1\}), \eta_F(\{u_1, u_2\}), \ldots, \eta_F(\{u_{n-1}, u_n\})\big)$$

belongs to $\mathrm{Dom}(M)^n$, and therefore

$$s^{\mathcal{F}}_M(P) = \Psi^{(n)}_M\big(\eta_F(\{u_0, u_1\}), \ldots, \eta_F(\{u_{n-1}, u_n\})\big)$$

is a well-defined element of $\mathrm{Dom}(M)$.

Since also

$$\eta_M(e) \in \mathrm{Dom}(M),$$

the comparison

$$\eta_M(e) \prec_M s^{\mathcal{F}}_M(P)$$

is a meaningful statement in the model $M$. Therefore the entire predicate

"there exists an $x$-$y$ path $P$ in $G^*_{\mathrm{supp}}(\mathcal{F}_M)$ such that $\eta_M(e) \prec_M s^{\mathcal{F}}_M(P)$"

is well-defined.

Finally, the definition of uncertain tree requires:

- that $G^*_{\mathrm{supp}}(\mathcal{G}_M)$ be connected,
- that there exist an uncertain spanning subgraph $\mathcal{F}_M$,
- that $G^*_{\mathrm{supp}}(\mathcal{F}_M)$ be a crisp tree,
- and that the above comparison condition hold for every support edge outside $E^*(\mathcal{F}_M)$.

Each of these is meaningful because all objects involved have already been shown to be well-defined.

Therefore the statement

"$\mathcal{G}_M$ is an Uncertain Tree"

is well-defined. Hence the notion of an uncertain tree is well-defined. □

Table 3.1: Related tree concepts under fuzzy and uncertainty-aware frameworks

| **Concept** | **Reference(s)** |
|---|---|
| Fuzzy Tree | — |
| Intuitionistic Fuzzy Tree | cf. [173] |
| Neutrosophic Tree | [174, 175] |

## 3.4 Uncertain Degree, Order, and Size

The degree of a vertex in a fuzzy graph is the sum of memberships of all incident edges, measuring its membership-weighted local connectivity strength therein [176,177]. The order of a fuzzy graph is the sum of all vertex membership values, representing the total weighted size of its vertices taken together [176,177]. The size of a fuzzy graph is the sum of memberships of all positive edges, representing the total weighted extent of adjacency in the graph [176,177].

**Definition 3.4.1** (Degree, Order, and Size of a Fuzzy Graph)**.** Let

$$G = (V, \sigma, \mu)$$

be a finite fuzzy graph, where

$$\sigma : V \to [0,1], \qquad \mu : V \times V \to [0,1], \qquad \mu(u,v) \le \min\{\sigma(u), \sigma(v)\} \quad (\forall\, u, v \in V),$$

and assume that $\mu$ is symmetric and $G$ has no loops.

The *degree* of a vertex $v \in V$ is defined by

$$d_G(v) := \sum_{\substack{u \in V \\ u \ne v}} \mu(v,u).$$

The *order* of the fuzzy graph $G$ is defined by

$$O(G) := \sum_{v \in V} \sigma(v).$$

Define the support edge set by

$$E^*(G) := \big\{\{u,v\} \subseteq V : u \ne v,\ \mu(u,v) > 0\big\}.$$

Then the *size* of the fuzzy graph $G$ is defined by

$$S(G) := \sum_{\{u,v\} \in E^*(G)} \mu(u,v).$$

**Example 3.4.2** (Degree, order, and size of a fuzzy graph)**.** Let

$$V = \{v_1, v_2, v_3, v_4\}.$$

Define a fuzzy graph

$$G = (V, \sigma, \mu)$$

by the vertex-membership function

$$\sigma(v_1) = 0.9, \qquad \sigma(v_2) = 0.7, \qquad \sigma(v_3) = 0.8, \qquad \sigma(v_4) = 0.6,$$

and the symmetric edge-membership function $\mu : V \times V \to [0, 1]$ given by

$$\mu(v_1, v_2) = 0.5, \qquad \mu(v_2, v_3) = 0.4, \qquad \mu(v_3, v_4) = 0.3, \qquad \mu(v_4, v_1) = 0.4, \qquad \mu(v_1, v_3) = 0.2,$$

and

$$\mu(u, v) = 0$$

for all other unordered pairs $\{u, v\} \subseteq V$, with

$$\mu(u, v) = \mu(v, u)$$

for all $u, v \in V$, and

$$\mu(v_i, v_i) = 0 \qquad (i = 1, 2, 3, 4).$$

First, we verify that $G$ is a fuzzy graph. Indeed,

$$\mu(v_1, v_2) = 0.5 \le \min\{0.9, 0.7\} = 0.7,$$

$$\mu(v_2, v_3) = 0.4 \le \min\{0.7, 0.8\} = 0.7,$$
$$\mu(v_3, v_4) = 0.3 \le \min\{0.8, 0.6\} = 0.6,$$
$$\mu(v_4, v_1) = 0.4 \le \min\{0.6, 0.9\} = 0.6,$$

and

$$\mu(v_1, v_3) = 0.2 \le \min\{0.9, 0.8\} = 0.8.$$

Hence all edge-membership values satisfy the required condition

$$\mu(u, v) \le \min\{\sigma(u), \sigma(v)\}.$$

The support edge set is

$$E^*(G) = \big\{\{v_1, v_2\}, \{v_2, v_3\}, \{v_3, v_4\}, \{v_4, v_1\}, \{v_1, v_3\}\big\}.$$

The degree of each vertex is computed as follows:

$$d_G(v_1) = \mu(v_1, v_2) + \mu(v_1, v_4) + \mu(v_1, v_3) = 0.5 + 0.4 + 0.2 = 1.1,$$

$$d_G(v_2) = \mu(v_2, v_1) + \mu(v_2, v_3) = 0.5 + 0.4 = 0.9,$$
$$d_G(v_3) = \mu(v_3, v_2) + \mu(v_3, v_4) + \mu(v_3, v_1) = 0.4 + 0.3 + 0.2 = 0.9,$$
$$d_G(v_4) = \mu(v_4, v_3) + \mu(v_4, v_1) = 0.3 + 0.4 = 0.7.$$

The order of $G$ is

$$O(G) = \sum_{v \in V} \sigma(v) = \sigma(v_1) + \sigma(v_2) + \sigma(v_3) + \sigma(v_4) = 0.9 + 0.7 + 0.8 + 0.6 = 3.0.$$

The size of $G$ is

$$S(G) = \sum_{\{u,v\} \in E^*(G)} \mu(u, v) = 0.5 + 0.4 + 0.3 + 0.4 + 0.2 = 1.8.$$

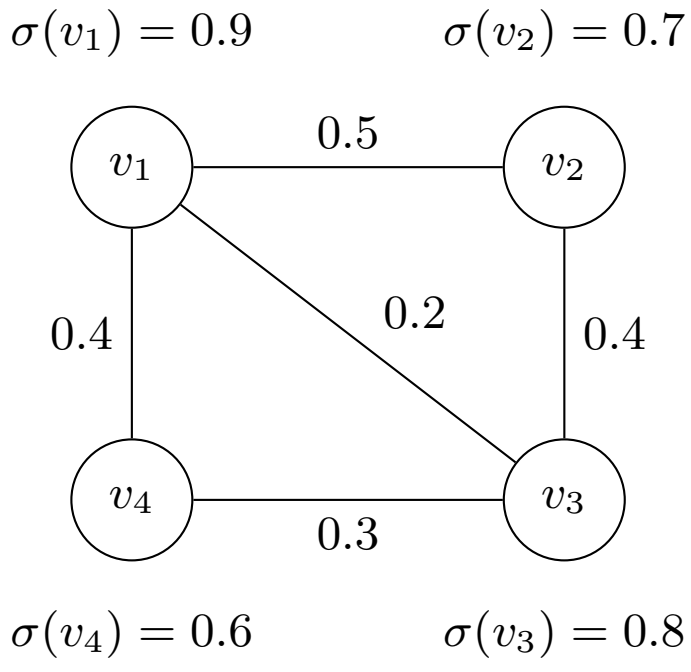

Figure 3.2: A fuzzy graph illustrating degree, order, and size

Therefore, for this fuzzy graph,

$$d_G(v_1) = 1.1, \qquad d_G(v_2) = 0.9, \qquad d_G(v_3) = 0.9, \qquad d_G(v_4) = 0.7,$$

$$O(G) = 3.0, \qquad S(G) = 1.8.$$

A schematic illustration of this fuzzy graph is shown in Figure 3.2.

**Definition 3.4.3** (Measure-Evaluable Uncertain Model)**.** Let $M$ be an uncertain model with degree-domain

$$\mathrm{Dom}(M) \subseteq [0,1]^k.$$

We say that $M$ is *measure-evaluable* if it is equipped with a map

$$\Delta_M : \mathrm{Dom}(M) \longrightarrow [0,\infty),$$

called the *evaluation map* of $M$.

**Definition 3.4.4** (Degree, Order, and Size of an Uncertain Graph)**.** Let

$$G^* = (V, E)$$

be a finite undirected loopless graph, and let $M$ be a measure-evaluable uncertain model with degree-domain $\mathrm{Dom}(M)$ and evaluation map

$$\Delta_M : \mathrm{Dom}(M) \to [0,\infty).$$

An *Uncertain Graph of type $M$* on $G^*$ is a quadruple

$$\mathcal{G}_M = (V, E, \sigma_M, \eta_M),$$

where

$$\sigma_M : V \to \mathrm{Dom}(M), \qquad \eta_M : E \to \mathrm{Dom}(M)$$

are uncertainty-degree functions on the vertex set and edge set, respectively.

Then the following quantities are defined.

(i) The *degree* of a vertex $v \in V$ is

$$d_{\mathcal{G}_M}(v) := \sum_{\substack{e \in E \\ v \in e}} \Delta_M\big(\eta_M(e)\big).$$

(ii) The *order* of $\mathcal{G}_M$ is

$$O(\mathcal{G}_M) := \sum_{v \in V} \Delta_M\big(\sigma_M(v)\big).$$

(iii) The *size* of $\mathcal{G}_M$ is

$$S(\mathcal{G}_M) := \sum_{e\in E} \Delta_M\big(\eta_M(e)\big).$$

**Definition 3.4.5** (Measure-Evaluable Uncertain Model)**.** Let $M$ be an uncertain model with degree-domain

$$\mathrm{Dom}(M) \subseteq [0,1]^k.$$

We say that $M$ is *measure-evaluable* if it is equipped with a map

$$\Delta_M : \mathrm{Dom}(M) \longrightarrow [0,\infty),$$

called the *evaluation map* of $M$.

**Definition 3.4.6** (Degree, Order, and Size of an Uncertain Graph)**.** Let

$$G^* = (V,E)$$

be a finite undirected loopless graph, and let $M$ be a measure-evaluable uncertain model with degree-domain $\mathrm{Dom}(M)$ and evaluation map

$$\Delta_M : \mathrm{Dom}(M) \to [0,\infty).$$

An *Uncertain Graph of type $M$* on $G^*$ is a quadruple

$$\mathcal{G}_M = (V, E, \sigma_M, \eta_M),$$

where

$$\sigma_M : V \to \mathrm{Dom}(M), \qquad \eta_M : E \to \mathrm{Dom}(M)$$

are uncertainty-degree functions on the vertex set and edge set, respectively.

Then the following quantities are defined.

(i) The *degree* of a vertex $v \in V$ is

$$d_{\mathcal{G}_M}(v) := \sum_{\substack{e\in E\\ v\in e}} \Delta_M\big(\eta_M(e)\big).$$

(ii) The *order* of $\mathcal{G}_M$ is

$$O(\mathcal{G}_M) := \sum_{v\in V} \Delta_M\big(\sigma_M(v)\big).$$

(iii) The *size* of $\mathcal{G}_M$ is

$$S(\mathcal{G}_M) := \sum_{e\in E} \Delta_M\big(\eta_M(e)\big).$$

## 3.5 Uncertain Distance

Fuzzy distance in a fuzzy graph is the minimum path length between two vertices, computed from edge memberships, quantifying separation under uncertain connectivity and relations.

**Definition 3.5.1** (Fuzzy Distance in a Fuzzy Graph)**.** Let

$$G = (V, \sigma, \mu)$$

be a finite connected fuzzy graph, where

$$\sigma : V \to [0,1], \qquad \mu : V\times V \to [0,1], \qquad \mu(u,v) \le \min\{\sigma(u), \sigma(v)\} \quad (\forall\, u,v \in V),$$

and assume that $\mu$ is symmetric.

A path
$$P : u_0, u_1, \ldots, u_n$$
from $u_0$ to $u_n$ is a sequence of vertices such that
$$\mu(u_{i-1}, u_i) > 0 \qquad (i = 1, 2, \ldots, n).$$

The *$\mu$-length* of the path $P$ is defined by
$$\ell_\mu(P) := \sum_{i=1}^{n} \frac{1}{\mu(u_{i-1}, u_i)}.$$

For two vertices $u, v \in V$, the *fuzzy distance* (or *$\mu$-distance*) between $u$ and $v$ is defined by
$$d_\mu(u, v) := \min\{\ell_\mu(P) : P \text{ is a path from } u \text{ to } v\}.$$
Also,
$$d_\mu(u, u) := 0 \qquad (\forall\, u \in V).$$

**Example 3.5.2** (Fuzzy distance in a fuzzy graph)**.** Let
$$V = \{v_1, v_2, v_3, v_4\}.$$
Define a fuzzy graph
$$G = (V, \sigma, \mu)$$
by
$$\sigma(v_1) = 0.9, \qquad \sigma(v_2) = 0.8, \qquad \sigma(v_3) = 0.85, \qquad \sigma(v_4) = 0.6,$$
and let the symmetric edge-membership function $\mu : V \times V \to [0, 1]$ be given by
$$\mu(v_1, v_2) = 0.8, \qquad \mu(v_2, v_3) = 0.7, \qquad \mu(v_1, v_4) = 0.5, \qquad \mu(v_4, v_3) = 0.5, \qquad \mu(v_1, v_3) = 0.3,$$
and
$$\mu(u, v) = 0$$
for all other unordered pairs $\{u, v\} \subseteq V$, with
$$\mu(u, v) = \mu(v, u) \qquad (\forall\, u, v \in V).$$

First, we verify that $G$ is a fuzzy graph. Indeed,
$$\mu(v_1, v_2) = 0.8 \le \min\{0.9, 0.8\} = 0.8,$$
$$\mu(v_2, v_3) = 0.7 \le \min\{0.8, 0.85\} = 0.8,$$
$$\mu(v_1, v_4) = 0.5 \le \min\{0.9, 0.6\} = 0.6,$$
$$\mu(v_4, v_3) = 0.5 \le \min\{0.6, 0.85\} = 0.6,$$
and
$$\mu(v_1, v_3) = 0.3 \le \min\{0.9, 0.85\} = 0.85.$$
Hence $G$ satisfies the condition
$$\mu(u, v) \le \min\{\sigma(u), \sigma(v)\} \qquad (\forall\, u, v \in V).$$

Moreover, the support graph is connected, since the positive edges
$$\{v_1, v_2\},\ \{v_2, v_3\},\ \{v_1, v_4\},\ \{v_4, v_3\},\ \{v_1, v_3\}$$
connect all vertices.

Now consider the fuzzy distance between $v_1$ and $v_3$. Possible paths from $v_1$ to $v_3$ include:

$$P_1 : v_1, v_3,$$

$$P_2 : v_1, v_2, v_3,$$

and

$$P_3 : v_1, v_4, v_3.$$

Their $\mu$-lengths are:

$$\ell_\mu(P_1) = \frac{1}{\mu(v_1, v_3)} = \frac{1}{0.3} = \frac{10}{3},$$
$$\ell_\mu(P_2) = \frac{1}{\mu(v_1, v_2)} + \frac{1}{\mu(v_2, v_3)} = \frac{1}{0.8} + \frac{1}{0.7} = \frac{5}{4} + \frac{10}{7} = \frac{75}{28},$$
$$\ell_\mu(P_3) = \frac{1}{\mu(v_1, v_4)} + \frac{1}{\mu(v_4, v_3)} = \frac{1}{0.5} + \frac{1}{0.5} = 2 + 2 = 4.$$

Therefore,

$$d_\mu(v_1, v_3) = \min\left\{\frac{10}{3}, \frac{75}{28}, 4\right\} = \frac{75}{28}.$$

Thus, although there is a direct edge between $v_1$ and $v_3$, the shortest fuzzy route is

$$v_1 \to v_2 \to v_3,$$

because its edge memberships are stronger and hence its reciprocal-sum length is smaller.

As another example, consider the distance between $v_2$ and $v_4$. Two natural paths are

$$Q_1 : v_2, v_1, v_4, \qquad Q_2 : v_2, v_3, v_4.$$

Their $\mu$-lengths are

$$\ell_\mu(Q_1) = \frac{1}{0.8} + \frac{1}{0.5} = \frac{5}{4} + 2 = \frac{13}{4},$$

and

$$\ell_\mu(Q_2) = \frac{1}{0.7} + \frac{1}{0.5} = \frac{10}{7} + 2 = \frac{24}{7}.$$

Hence

$$d_\mu(v_2, v_4) = \min\left\{\frac{13}{4}, \frac{24}{7}\right\} = \frac{13}{4}.$$

Also, by definition,

$$d_\mu(v_i, v_i) = 0 \qquad (i = 1, 2, 3, 4).$$

A schematic illustration of this fuzzy graph is shown in Figure 3.3.

Next, we present the extensions based on Uncertain Sets below.

**Definition 3.5.3** (Distance-Evaluable Uncertain Model)**.** Let $M$ be an uncertain model with degree-domain

$$\mathrm{Dom}(M) \subseteq [0, 1]^k.$$

We say that $M$ is *distance-evaluable* if it is equipped with:

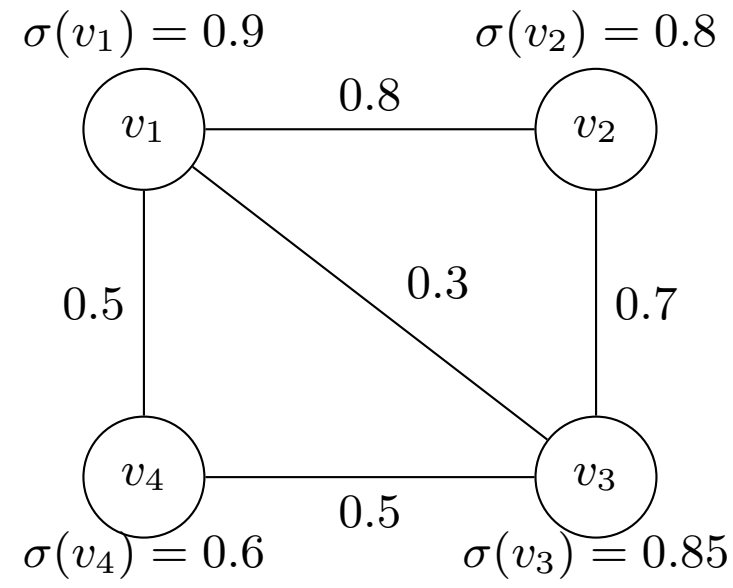

Figure 3.3: A fuzzy graph illustrating fuzzy distance

1. a distinguished element
$$0_M \in \mathrm{Dom}(M),$$
called the *zero degree*;
2. a map
$$\Lambda_M : \mathrm{Dom}(M) \setminus \{0_M\} \longrightarrow (0, \infty),$$
called the *edge-length evaluation map*.

**Definition 3.5.4** (Uncertain Distance)**.** Let
$$G^* = (V, E)$$
be a finite simple graph, and let $M$ be a distance-evaluable uncertain model.

Let
$$\mathcal{G}_M = (V, E, \sigma_M, \eta_M)$$
be an Uncertain Graph of type $M$, where
$$\sigma_M : V \to \mathrm{Dom}(M), \qquad \eta_M : E \to \mathrm{Dom}(M).$$

Define the *support edge set* of $\mathcal{G}_M$ by
$$E^*(\mathcal{G}_M) := \{\, e \in E \mid \eta_M(e) \neq 0_M \,\},$$
and let
$$G^*_{\mathrm{supp}}(\mathcal{G}_M) := (V, E^*(\mathcal{G}_M))$$
be the corresponding support graph.

Assume that $G^*_{\mathrm{supp}}(\mathcal{G}_M)$ is connected.

A *path* from $u$ to $v$ in $\mathcal{G}_M$ is a path
$$P : u_0, u_1, \dots, u_n$$
in the support graph $G^*_{\mathrm{supp}}(\mathcal{G}_M)$, where
$$u_0 = u, \qquad u_n = v,$$
and
$$\{u_{i-1}, u_i\} \in E^*(\mathcal{G}_M) \qquad (i = 1, \dots, n).$$

The *uncertain length* of such a path $P$ is defined by
$$\ell_M(P) := \sum_{i=1}^{n} \Lambda_M\big(\eta_M(\{u_{i-1}, u_i\})\big).$$

For two vertices $u, v \in V$, the *uncertain distance* between $u$ and $v$ is defined by
$$d_M(u, v) := \min\big\{\ell_M(P) \mid P \text{ is a path from } u \text{ to } v \text{ in } \mathcal{G}_M\big\}.$$
Also,
$$d_M(u, u) := 0 \qquad (\forall\, u \in V).$$

**Theorem 3.5.5** (Well-definedness of Uncertain Distance)**.** *Let*

$$\mathcal{G}_M = (V, E, \sigma_M, \eta_M)$$

*be an Uncertain Graph of type $M$, where $M$ is a distance-evaluable uncertain model with zero degree*

$$0_M \in \mathrm{Dom}(M)$$

*and edge-length evaluation map*

$$\Lambda_M : \mathrm{Dom}(M) \setminus \{0_M\} \to (0, \infty).$$

*Assume that the support graph*

$$G^*_{\mathrm{supp}}(\mathcal{G}_M) = (V, E^*(\mathcal{G}_M))$$

*is connected.*

*Then, for every pair of vertices $u, v \in V$, the quantity*

$$d_M(u, v) = \min\big\{\ell_M(P) \mid P \text{ is a path from } u \text{ to } v\big\}$$

*is well-defined as a nonnegative real number.*

*Hence the notion of uncertain distance is well-defined.*

*Proof.* Fix $u, v \in V$.

Since the support graph

$$G^*_{\mathrm{supp}}(\mathcal{G}_M)$$

is connected, there exists at least one path from $u$ to $v$. Thus the set

$$\mathcal{P}(u, v) := \{\, P \mid P \text{ is a path from } u \text{ to } v \text{ in } G^*_{\mathrm{supp}}(\mathcal{G}_M) \,\}$$

is nonempty.

Because $V$ is finite, there are only finitely many simple $u$-$v$ paths in the finite graph

$$G^*_{\mathrm{supp}}(\mathcal{G}_M).$$

Hence $\mathcal{P}(u, v)$ is finite.

Now let

$$P : u_0, u_1, \ldots, u_n$$

be a path in $\mathcal{P}(u, v)$. For each $i = 1, \ldots, n$, we have

$$\{u_{i-1}, u_i\} \in E^*(\mathcal{G}_M),$$

so by definition of the support edge set,

$$\eta_M(\{u_{i-1}, u_i\}) \neq 0_M.$$

Therefore

$$\eta_M(\{u_{i-1}, u_i\}) \in \mathrm{Dom}(M) \setminus \{0_M\},$$

and so

$$\Lambda_M\big(\eta_M(\{u_{i-1}, u_i\})\big) \in (0, \infty)$$

is well-defined.

Thus the path length

$$\ell_M(P) = \sum_{i=1}^{n} \Lambda_M\big(\eta_M(\{u_{i-1}, u_i\})\big)$$

is a finite sum of positive real numbers, hence is a well-defined element of $(0, \infty)$. In the case $u = v$, we define $d_M(u, u) = 0$, which is also well-defined.

Therefore the set

$$\{\ell_M(P) \mid P \in \mathcal{P}(u, v)\}$$

is a finite nonempty subset of $[0, \infty)$. Every finite nonempty subset of $\mathbb{R}$ has a minimum. Hence

$$d_M(u, v) := \min\{\ell_M(P) \mid P \in \mathcal{P}(u, v)\}$$

exists and is uniquely determined.

Consequently, $d_M(u, v)$ is well-defined for all $u, v \in V$. Hence the notion of uncertain distance is well-defined. □

## 3.6 Uncertain Clique

A clique in a fuzzy graph is a vertex subset whose induced fuzzy subgraph is complete, so every pair attains maximal admissible edge membership [178–181].

**Definition 3.6.1** (Clique in a Fuzzy Graph)**.** Let

$$G = (V, \sigma, \mu)$$

be a finite fuzzy graph, where

$$\sigma : V \to [0, 1], \qquad \mu : V \times V \to [0, 1], \qquad \mu(u, v) \leq \min\{\sigma(u), \sigma(v)\} \quad (\forall\, u, v \in V),$$

and assume that $\mu$ is symmetric.

For a nonempty subset

$$C \subseteq V,$$

the fuzzy subgraph of $G$ induced by $C$ is

$$G[C] = \big(C, \sigma|_C, \mu|_{C\times C}\big).$$

Then $C$ is called a *clique* of the fuzzy graph $G$ if the induced fuzzy subgraph $G[C]$ is complete; that is, for every two distinct vertices $u, v \in C$,

$$\mu(u, v) = \min\{\sigma(u), \sigma(v)\}.$$

Equivalently, every pair of distinct vertices of $C$ is adjacent with the maximum possible membership value allowed by their vertex memberships.

**Example 3.6.2** (Clique in a fuzzy graph)**.** Let

$$V = \{v_1, v_2, v_3, v_4\}.$$

Define a fuzzy graph

$$G = (V, \sigma, \mu)$$

by the vertex-membership function

$$\sigma(v_1) = 0.9, \qquad \sigma(v_2) = 0.7, \qquad \sigma(v_3) = 0.8, \qquad \sigma(v_4) = 0.6,$$

and the symmetric edge-membership function $\mu : V \times V \to [0,1]$ given by

$$\mu(v_1, v_2) = 0.7, \qquad \mu(v_1, v_3) = 0.8, \qquad \mu(v_2, v_3) = 0.7,$$

$$\mu(v_1, v_4) = 0.5, \qquad \mu(v_2, v_4) = 0.4, \qquad \mu(v_3, v_4) = 0.3,$$

and

$$\mu(v_i, v_i) = 0 \qquad (i = 1, 2, 3, 4),$$

with

$$\mu(u, v) = \mu(v, u) \qquad (\forall\, u, v \in V).$$

First, we verify that $G$ is a fuzzy graph. Indeed,

$$\mu(v_1, v_2) = 0.7 = \min\{0.9, 0.7\},$$

$$\mu(v_1, v_3) = 0.8 = \min\{0.9, 0.8\},$$

$$\mu(v_2, v_3) = 0.7 = \min\{0.7, 0.8\},$$

$$\mu(v_1, v_4) = 0.5 \leq \min\{0.9, 0.6\} = 0.6,$$

$$\mu(v_2, v_4) = 0.4 \leq \min\{0.7, 0.6\} = 0.6,$$

and

$$\mu(v_3, v_4) = 0.3 \leq \min\{0.8, 0.6\} = 0.6.$$

Hence,

$$\mu(u, v) \leq \min\{\sigma(u), \sigma(v)\} \qquad (\forall\, u, v \in V).$$

Now consider the subset

$$C = \{v_1, v_2, v_3\} \subseteq V.$$

Then the induced fuzzy subgraph is

$$G[C] = \big(C, \sigma|_C, \mu|_{C\times C}\big).$$

We check the three distinct pairs of vertices in $C$:

$$\mu(v_1, v_2) = 0.7 = \min\{\sigma(v_1), \sigma(v_2)\},$$

$$\mu(v_1, v_3) = 0.8 = \min\{\sigma(v_1), \sigma(v_3)\},$$

$$\mu(v_2, v_3) = 0.7 = \min\{\sigma(v_2), \sigma(v_3)\}.$$

Therefore, every pair of distinct vertices in $C$ is adjacent with the maximum possible membership value allowed by their vertex memberships. Hence $G[C]$ is complete, and so

$$C = \{v_1, v_2, v_3\}$$

is a clique of the fuzzy graph $G$.

On the other hand, the whole vertex set

$$V = \{v_1, v_2, v_3, v_4\}$$

is not a clique, because for example,

$$\mu(v_1, v_4) = 0.5 < \min\{0.9, 0.6\} = 0.6.$$

Thus, this example shows that a subset of vertices may form a clique even when the entire fuzzy graph is not complete.

A schematic illustration is given in Figure 3.4.

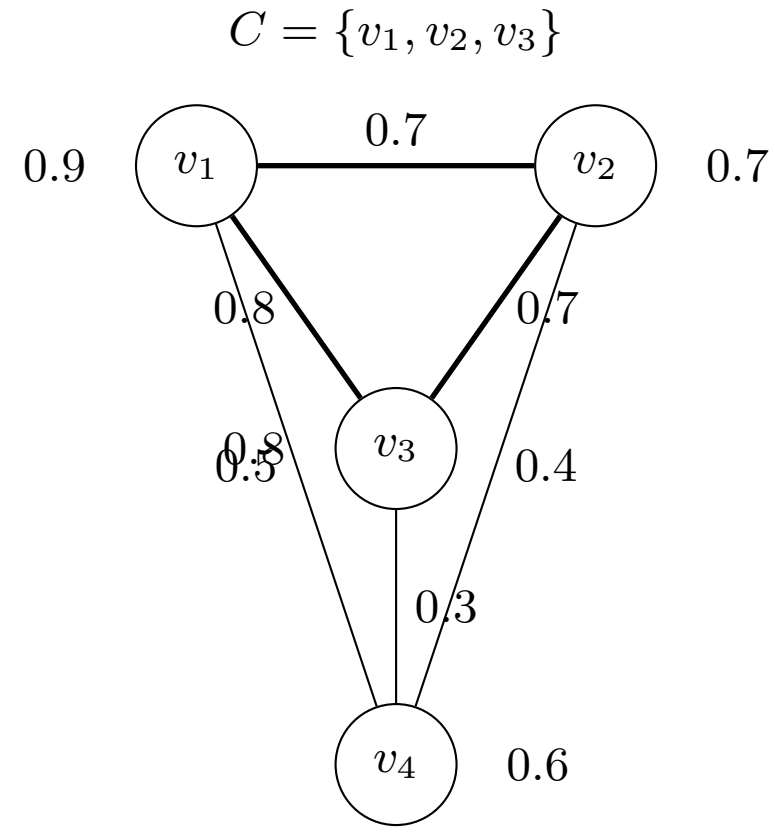

Figure 3.4: A fuzzy graph containing the clique $C = \{v_1, v_2, v_3\}$

An uncertain clique is a vertex subset whose induced uncertain subgraph is complete, so every pair of distinct vertices is joined by the model-dependent complete edge.

**Definition 3.6.3** (Uncertain Clique)**.** Let $M$ be a complete-edge-evaluable uncertain model with degree-domain

$$\mathrm{Dom}(M) \subseteq [0, 1]^k$$

and complete-edge operator

$$\Gamma_M : \mathrm{Dom}(M) \times \mathrm{Dom}(M) \to \mathrm{Dom}(M).$$

Let

$$\mathcal{G}_M = (V, E, \sigma_M, \eta_M)$$

be an Uncertain Graph of type $M$, where

$$\sigma_M : V \to \mathrm{Dom}(M), \qquad \eta_M : E \to \mathrm{Dom}(M).$$

For a nonempty subset

$$C \subseteq V,$$

define the induced edge set

$$E[C] := E \cap \big\{\{u, v\} \subseteq C \mid u \neq v\big\}.$$

The *induced uncertain subgraph* of $\mathcal{G}_M$ on $C$ is

$$\mathcal{G}_M[C] := \big(C, E[C], \sigma_M|_C, \eta_M|_{E[C]}\big).$$

Then $C$ is called an *Uncertain Clique* of $\mathcal{G}_M$ if the induced uncertain subgraph

$$\mathcal{G}_M[C]$$

is a Complete Uncertain Graph of type $M$. Equivalently, $C$ is an uncertain clique if

$$E[C] = \big\{\{u, v\} \subseteq C \mid u \neq v\big\},$$

and

$$\eta_M(\{u, v\}) = \Gamma_M\big(\sigma_M(u), \sigma_M(v)\big) \qquad \text{for all distinct } u, v \in C.$$

In other words, every pair of distinct vertices in $C$ is joined by the model-dependent complete edge determined by their uncertainty degrees.

**Theorem 3.6.4** (Well-definedness of Uncertain Clique)**.** *Let $M$ be a complete-edge-evaluable uncertain model with degree-domain*

$$\mathrm{Dom}(M) \subseteq [0,1]^k$$

*and symmetric complete-edge operator*

$$\Gamma_M : \mathrm{Dom}(M) \times \mathrm{Dom}(M) \to \mathrm{Dom}(M).$$

*Let*

$$\mathcal{G}_M = (V, E, \sigma_M, \eta_M)$$

*be an Uncertain Graph of type $M$, and let*

$$C \subseteq V$$

*be a nonempty subset.*

*Then the induced object*

$$\mathcal{G}_M[C] = \bigl(C, E[C], \sigma_M|_C, \eta_M|_{E[C]}\bigr)$$

*is a well-defined Uncertain Graph of type $M$.*

*Consequently, the statement*

$$\text{"}C \text{ is an Uncertain Clique of } \mathcal{G}_M\text{"}$$

*is well-defined. Hence the notion of an uncertain clique is well-defined.*

*Proof.* Let

$$C \subseteq V$$

be nonempty. Define

$$E[C] := E \cap \bigl\{\{u,v\} \subseteq C \mid u \neq v\bigr\}.$$

Since both $E$ and

$$\bigl\{\{u,v\} \subseteq C \mid u \neq v\bigr\}$$

are well-defined sets, the intersection $E[C]$ is well-defined.

Because

$$\sigma_M : V \to \mathrm{Dom}(M)$$

is a function, its restriction

$$\sigma_M|_C : C \to \mathrm{Dom}(M)$$

is also a well-defined function. Likewise, since

$$\eta_M : E \to \mathrm{Dom}(M)$$

is a function and $E[C] \subseteq E$, the restriction

$$\eta_M|_{E[C]} : E[C] \to \mathrm{Dom}(M)$$

is well-defined.

Therefore

$$\mathcal{G}_M[C] = \bigl(C, E[C], \sigma_M|_C, \eta_M|_{E[C]}\bigr)$$

is a well-defined Uncertain Graph of type $M$.

Now consider the condition that $\mathcal{G}_M[C]$ be complete. The requirement

$$E[C] = \bigl\{\{u,v\} \subseteq C \mid u \neq v\bigr\}$$

is meaningful because both sides are well-defined sets of unordered pairs of vertices.

Further, for each distinct $u, v \in C$, the values

$$\sigma_M(u), \sigma_M(v) \in \mathrm{Dom}(M)$$

are well-defined, and since

$$\Gamma_M : \mathrm{Dom}(M) \times \mathrm{Dom}(M) \to \mathrm{Dom}(M)$$

is a well-defined symmetric map, the value

$$\Gamma_M\big(\sigma_M(u), \sigma_M(v)\big)$$

is well-defined and independent of the order of $u$ and $v$. Thus the equality

$$\eta_M(\{u, v\}) = \Gamma_M\big(\sigma_M(u), \sigma_M(v)\big)$$

is a meaningful statement for every distinct $u, v \in C$.

Hence the predicate

$$\mathcal{G}_M[C] \text{ is a Complete Uncertain Graph}$$

is well-defined. By definition, this is exactly the predicate

$$C \text{ is an Uncertain Clique of } \mathcal{G}_M.$$

Therefore the notion of uncertain clique is well-defined. □

For convenience, Table 3.2 summarizes representative clique-related concepts according to the dimension $k$ of the information associated with vertices and/or edges.

Table 3.2: Representative clique-related concepts under uncertainty-aware graph frameworks, classified by the dimension $k$ of the information attached to vertices and/or edges.

| $k$ | **Clique-related concept** | **Typical coordinate form** | **Canonical information attached to vertices/edges** |
|---|---|---|---|
| 1 | Fuzzy Clique | $\mu$ | A clique is studied in a fuzzy graph, where each vertex and edge is associated with a single membership degree in $[0, 1]$. |
| 2 | Intuitionistic Fuzzy Clique [182] | $(\mu, \nu)$ | A clique is defined in an intuitionistic fuzzy graph, where each vertex and edge carries a membership degree and a non-membership degree, usually satisfying $\mu + \nu \leq 1$. |
| 3 | Neutrosophic Clique | $(T, I, F)$ | A clique is defined in a neutrosophic graph, where each vertex and edge is described by truth, indeterminacy, and falsity degrees. |

Related concepts such as biclique [183, 184], hyperclique [185, 186], quasi-clique [187, 188], k-core [189], k-plex [190], and k-club [191] are also well known.

## 3.7 Uncertain Star

A fuzzy star is a fuzzy graph whose support has one central vertex adjacent to all leaves, while no positive edges exist between leaves pairwise [192–194].

**Definition 3.7.1** (Fuzzy Star)**.** [192, 193] Let

$$G = (V, \sigma, \mu)$$

be a finite fuzzy graph, where

$$\sigma : V \to [0,1], \qquad \mu : V \times V \to [0,1], \qquad \mu(u,v) \le \min\{\sigma(u), \sigma(v)\} \quad (\forall\, u, v \in V),$$

and assume that $\mu$ is symmetric and $G$ has no loops.

Define the support vertex set and support edge set by

$$V^* := \{v \in V : \sigma(v) > 0\}, \qquad E^* := \big\{\{u,v\} \subseteq V^* : u \neq v,\ \mu(u,v) > 0\big\}.$$

Then $G$ is called a *fuzzy star* if there exist a vertex

$$c \in V^*$$

and distinct vertices

$$u_1, u_2, \ldots, u_n \in V^* \qquad (n \ge 1)$$

such that

$$V^* = \{c, u_1, u_2, \ldots, u_n\},$$

and

$$E^* = \big\{\{c, u_i\} : 1 \le i \le n\big\}.$$

Equivalently,

$$\mu(c, u_i) > 0 \qquad (1 \le i \le n),$$

and

$$\mu(u_i, u_j) = 0 \qquad (1 \le i < j \le n).$$

In this case, $c$ is called the *center* of the fuzzy star, the vertices $u_1, \ldots, u_n$ are called its *leaves*, and the fuzzy star is denoted by

$$S_{1,n}.$$

**Example 3.7.2** (Fuzzy star)**.** Let

$$V = \{c, u_1, u_2, u_3, u_4\}.$$

Define a fuzzy graph

$$G = (V, \sigma, \mu)$$

by the vertex-membership function

$$\sigma(c) = 0.9, \qquad \sigma(u_1) = 0.7, \qquad \sigma(u_2) = 0.6, \qquad \sigma(u_3) = 0.8, \qquad \sigma(u_4) = 0.5,$$

and the symmetric edge-membership function $\mu : V \times V \to [0,1]$ given by

$$\mu(c, u_1) = 0.6, \qquad \mu(c, u_2) = 0.5, \qquad \mu(c, u_3) = 0.7, \qquad \mu(c, u_4) = 0.4,$$

$$\mu(u_i, u_j) = 0 \qquad (1 \le i < j \le 4),$$
$$\mu(v, v) = 0 \qquad (\forall\, v \in V),$$

and

$$\mu(u, v) = \mu(v, u) \qquad (\forall\, u, v \in V).$$

First, we verify that $G$ is a fuzzy graph. Indeed,

$$\mu(c, u_1) = 0.6 \le \min\{\sigma(c), \sigma(u_1)\} = \min\{0.9, 0.7\} = 0.7,$$

$$\mu(c, u_2) = 0.5 \le \min\{\sigma(c), \sigma(u_2)\} = \min\{0.9, 0.6\} = 0.6,$$

$$\mu(c, u_3) = 0.7 \le \min\{\sigma(c), \sigma(u_3)\} = \min\{0.9, 0.8\} = 0.8,$$

and

$$\mu(c, u_4) = 0.4 \le \min\{\sigma(c), \sigma(u_4)\} = \min\{0.9, 0.5\} = 0.5.$$

All remaining pairs of distinct leaves have edge-membership value 0, so the condition

$$\mu(u, v) \le \min\{\sigma(u), \sigma(v)\} \qquad (\forall\, u, v \in V)$$

is satisfied.

Since all vertex-membership values are positive, the support vertex set is

$$V^* = \{c, u_1, u_2, u_3, u_4\}.$$

Moreover, the only pairs with positive edge-membership are

$$\{c, u_1\},\ \{c, u_2\},\ \{c, u_3\},\ \{c, u_4\}.$$

Hence the support edge set is

$$E^* = \big\{\{c, u_1\}, \{c, u_2\}, \{c, u_3\}, \{c, u_4\}\big\}.$$

Therefore,

$$V^* = \{c, u_1, u_2, u_3, u_4\},$$

and

$$E^* = \big\{\{c, u_i\} : 1 \le i \le 4\big\}.$$

Thus $G$ is a fuzzy star. Its center is

$$c,$$

its leaves are

$$u_1, u_2, u_3, u_4,$$

and the support graph is the star

$$S_{1,4}.$$

A schematic illustration of this fuzzy star is shown in Figure 3.5.

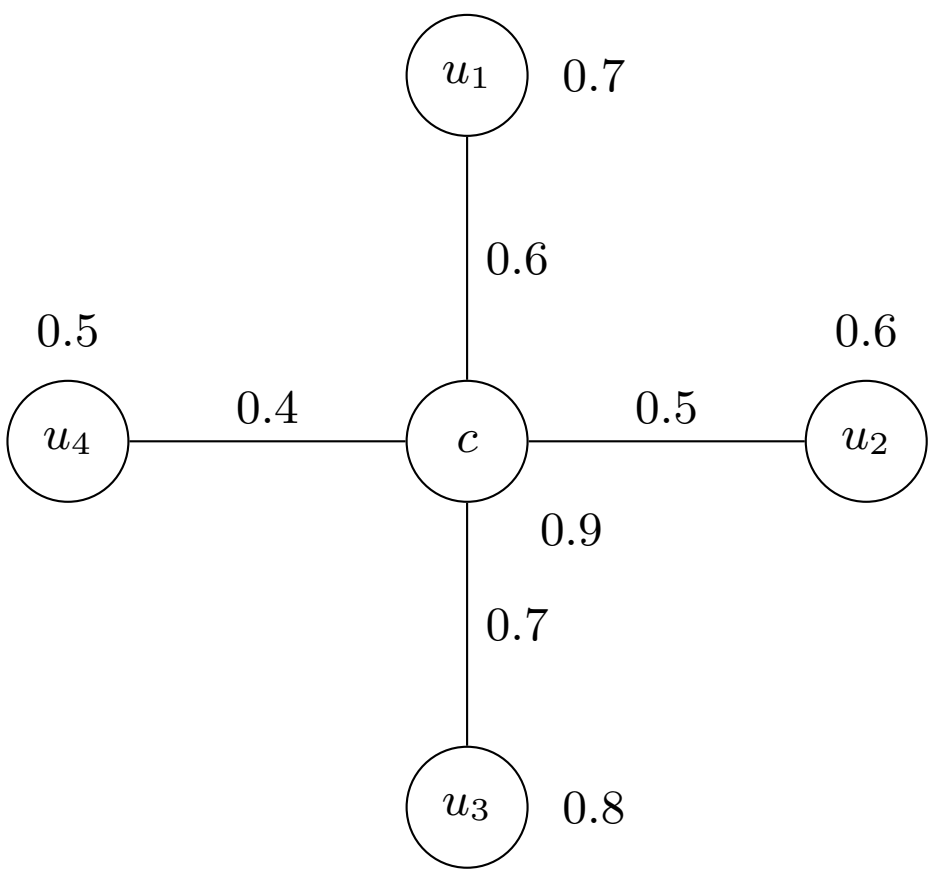

Figure 3.5: A fuzzy star with center $c$ and leaves $u_1, u_2, u_3, u_4$

An uncertain star is an uncertain graph whose support graph has one central vertex adjacent to all support leaves, while no support edge joins two distinct leaves.

**Definition 3.7.3** (Support-Evaluable Uncertain Model)**.** Let $M$ be an uncertain model with degree-domain

$$\mathrm{Dom}(M) \subseteq [0,1]^k.$$

We say that $M$ is *support-evaluable* if it is equipped with a distinguished element

$$0_M \in \mathrm{Dom}(M),$$

called the *zero degree.*

**Definition 3.7.4** (Uncertain Star)**.** Let $M$ be a support-evaluable uncertain model with degree-domain

$$\mathrm{Dom}(M) \subseteq [0,1]^k$$

and zero degree

$$0_M \in \mathrm{Dom}(M).$$

Let

$$\mathcal{G}_M = (V, E, \sigma_M, \eta_M)$$

be an Uncertain Graph of type $M$, where

$$\sigma_M : V \to \mathrm{Dom}(M), \qquad \eta_M : E \to \mathrm{Dom}(M).$$

Define the support vertex set and support edge set by

$$V^*(\mathcal{G}_M) := \{\, v \in V \mid \sigma_M(v) \neq 0_M \,\},$$

and

$$E^*(\mathcal{G}_M) := \{\, e \in E \mid \eta_M(e) \neq 0_M \,\}.$$

The *support graph* of $\mathcal{G}_M$ is

$$G^*_{\mathrm{supp}}(\mathcal{G}_M) := \big(V^*(\mathcal{G}_M),\, E^*(\mathcal{G}_M)\big).$$

Then $\mathcal{G}_M$ is called an *Uncertain Star* if there exist a vertex

$$c \in V^*(\mathcal{G}_M)$$

and distinct vertices

$$u_1, u_2, \ldots, u_n \in V^*(\mathcal{G}_M) \qquad (n \geq 1)$$

such that

$$V^*(\mathcal{G}_M) = \{c, u_1, u_2, \ldots, u_n\},$$

and

$$E^*(\mathcal{G}_M) = \big\{\{c, u_i\} \mid 1 \leq i \leq n\big\}.$$

Equivalently, the support graph $G^*_{\mathrm{supp}}(\mathcal{G}_M)$ is isomorphic to the star graph

$$K_{1,n}.$$

In this case, $c$ is called a *center* of the uncertain star and

$$u_1, \ldots, u_n$$

are called its *leaves.*

**Theorem 3.7.5** (Well-definedness of Uncertain Star)**.** *Let $M$ be a support-evaluable uncertain model with degree-domain*

$$\mathrm{Dom}(M) \subseteq [0,1]^k$$

*and zero degree*

$$0_M \in \mathrm{Dom}(M).$$

*Let*

$$\mathcal{G}_M = (V, E, \sigma_M, \eta_M)$$

*be an Uncertain Graph of type $M$.*

*Then the support sets*

$$V^*(\mathcal{G}_M) = \{\, v \in V \mid \sigma_M(v) \neq 0_M \,\}$$

*and*

$$E^*(\mathcal{G}_M) = \{\, e \in E \mid \eta_M(e) \neq 0_M \,\}$$

*are well-defined.*

*Consequently, the support graph*

$$G^*_{\mathrm{supp}}(\mathcal{G}_M) = \big(V^*(\mathcal{G}_M), E^*(\mathcal{G}_M)\big)$$

*is well-defined, and the statement*

*"$\mathcal{G}_M$ is an Uncertain Star"*

*is well-defined.*

*Hence the notion of an uncertain star is well-defined.*

*Proof.* Since $M$ is an uncertain model, its degree-domain

$$\mathrm{Dom}(M)$$

is fixed. Since $M$ is support-evaluable, the element

$$0_M \in \mathrm{Dom}(M)$$

is also fixed.

Because

$$\sigma_M : V \to \mathrm{Dom}(M)$$

is a function, for each $v \in V$ the value $\sigma_M(v)$ is uniquely determined in $\mathrm{Dom}(M)$. Therefore the condition

$$\sigma_M(v) \neq 0_M$$

is meaningful for every $v \in V$. Hence

$$V^*(\mathcal{G}_M) = \{\, v \in V \mid \sigma_M(v) \neq 0_M \,\}$$

is a well-defined subset of $V$.

Likewise, since

$$\eta_M : E \to \mathrm{Dom}(M)$$

is a function, for each $e \in E$ the value $\eta_M(e)$ is uniquely determined in $\mathrm{Dom}(M)$. Therefore the condition

$$\eta_M(e) \neq 0_M$$

is meaningful for every $e \in E$. Hence

$$E^*(\mathcal{G}_M) = \{\, e \in E \mid \eta_M(e) \neq 0_M \,\}$$

is a well-defined subset of $E$.

Thus the pair

$$G^*_{\text{supp}}(\mathcal{G}_M) = \big(V^*(\mathcal{G}_M), E^*(\mathcal{G}_M)\big)$$

is a well-defined graph.

Now the condition that $\mathcal{G}_M$ be an uncertain star asserts the existence of a vertex

$$c \in V^*(\mathcal{G}_M)$$

and distinct vertices

$$u_1, \ldots, u_n \in V^*(\mathcal{G}_M)$$

such that

$$V^*(\mathcal{G}_M) = \{c, u_1, \ldots, u_n\}$$

and

$$E^*(\mathcal{G}_M) = \big\{\{c, u_i\} \mid 1 \leq i \leq n\big\}.$$

Since both $V^*(\mathcal{G}_M)$ and $E^*(\mathcal{G}_M)$ are already well-defined, these set equalities are meaningful statements. Equivalently, the condition that

$$G^*_{\text{supp}}(\mathcal{G}_M) \cong K_{1,n}$$

is also meaningful, because both graphs are well-defined crisp graphs.

Therefore the predicate

$$\text{“}\mathcal{G}_M \text{ is an Uncertain Star”}$$

is well-defined.

Hence the notion of an uncertain star is well-defined. □

For convenience, Table 3.3 summarizes representative star-related concepts according to the dimension $k$ of the information associated with vertices and/or edges.

Related concepts other than the uncertain star are also known, such as the double star [198–200], subdivided star [201, 202], spider graph [203], and directed star [204, 205].

## 3.8 Uncertain Radius and Diameter

The radius of a fuzzy graph is the minimum eccentricity among its vertices, measuring how close the most central vertex is to all others overall (cf. [206]). The diameter of a fuzzy graph is the maximum eccentricity among its vertices, measuring the greatest distance from a vertex to any other vertex overall(cf. [206]).

Table 3.3: Representative star-related concepts under uncertainty-aware graph frameworks, classified by the dimension $k$ of the information attached to vertices and/or edges.

| $k$ | **Star-related concept** | **Typical coordinate form** | **Canonical information attached to vertices/edges** |
|---|---|---|---|
| 1 | Fuzzy Star | $\mu$ | A star is studied in a fuzzy graph, where each vertex and edge is associated with a single membership degree in $[0, 1]$. |
| 2 | Intuitionistic Fuzzy Star [195] | $(\mu, \nu)$ | A star is defined in an intuitionistic fuzzy graph, where each vertex and edge carries a membership degree and a non-membership degree, usually satisfying $\mu + \nu \leq 1$. |
| 3 | Neutrosophic Star [196] | $(T, I, F)$ | A star is defined in a neutrosophic graph, where each vertex and edge is described by truth, indeterminacy, and falsity degrees. |
| $s + t$ | Plithogenic Star [197] | $(\mathbf{a}, \mathbf{c}) \in [0, 1]^s \times [0, 1]^t$ | A star is defined in a plithogenic graph, where each vertex and edge is described by attribute-based information together with an $s$-dimensional appurtenance vector and a $t$-dimensional contradiction vector. |

**Definition 3.8.1** (Radius and Diameter in a Fuzzy Graph)**.** Let

$$G = (V, \sigma, \mu)$$

be a finite connected fuzzy graph, and let

$$d_\mu : V \times V \to [0, \infty)$$

be the fuzzy distance on $G$, defined by

$$d_\mu(u, v) := \min\big\{\ell_\mu(P) : P \text{ is a path from } u \text{ to } v\big\},$$

where, for a path

$$P : u_0, u_1, \ldots, u_n,$$

its $\mu$-length is

$$\ell_\mu(P) := \sum_{i=1}^{n} \frac{1}{\mu(u_{i-1}, u_i)}.$$

For each vertex $v \in V$, the *eccentricity* of $v$ is defined by

$$e_\mu(v) := \max_{u \in V} d_\mu(v, u).$$

The *radius* of the fuzzy graph $G$ is defined by

$$r_\mu(G) := \min_{v \in V} e_\mu(v),$$

and the *diameter* of the fuzzy graph $G$ is defined by

$$d_\mu(G) := \max_{v \in V} e_\mu(v).$$

**Example 3.8.2** (Radius and diameter in a fuzzy graph)**.** Let

$$V = \{v_1, v_2, v_3, v_4\}.$$

Define a fuzzy graph

$$G = (V, \sigma, \mu)$$

by the vertex-membership function

$$\sigma(v_1) = 0.9, \qquad \sigma(v_2) = 0.8, \qquad \sigma(v_3) = 0.8, \qquad \sigma(v_4) = 0.9,$$

and the symmetric edge-membership function $\mu : V \times V \to [0, 1]$ given by

$$\mu(v_1, v_2) = 0.8, \qquad \mu(v_2, v_3) = 0.5, \qquad \mu(v_3, v_4) = 0.8,$$

and

$$\mu(u, v) = 0$$

for all other unordered pairs $\{u, v\} \subseteq V$, with

$$\mu(u, v) = \mu(v, u) \qquad (\forall\, u, v \in V).$$

First, we verify that $G$ is a fuzzy graph. Indeed,

$$\mu(v_1, v_2) = 0.8 \le \min\{0.9, 0.8\} = 0.8,$$

$$\mu(v_2, v_3) = 0.5 \le \min\{0.8, 0.8\} = 0.8,$$

and

$$\mu(v_3, v_4) = 0.8 \le \min\{0.8, 0.9\} = 0.8.$$

Hence

$$\mu(u, v) \le \min\{\sigma(u), \sigma(v)\} \qquad (\forall\, u, v \in V).$$

Since the positive edges are exactly

$$\{v_1, v_2\}, \qquad \{v_2, v_3\}, \qquad \{v_3, v_4\},$$

the support graph is the path

$$v_1 - v_2 - v_3 - v_4,$$

which is connected. Therefore the fuzzy distance $d_\mu$ is well defined for every pair of vertices.

Because the support graph is a path, each pair of vertices is joined by a unique path. Thus the fuzzy distances are obtained by summing reciprocals of the corresponding edge-membership values.

We have

$$d_\mu(v_1, v_2) = \frac{1}{0.8} = \frac{5}{4}, \qquad d_\mu(v_2, v_3) = \frac{1}{0.5} = 2, \qquad d_\mu(v_3, v_4) = \frac{1}{0.8} = \frac{5}{4}.$$

Also,

$$d_\mu(v_1, v_3) = \frac{1}{0.8} + \frac{1}{0.5} = \frac{5}{4} + 2 = \frac{13}{4},$$

$$d_\mu(v_2, v_4) = \frac{1}{0.5} + \frac{1}{0.8} = 2 + \frac{5}{4} = \frac{13}{4},$$

and

$$d_\mu(v_1, v_4) = \frac{1}{0.8} + \frac{1}{0.5} + \frac{1}{0.8} = \frac{5}{4} + 2 + \frac{5}{4} = \frac{9}{2}.$$

Of course,

$$d_\mu(v_i, v_i) = 0 \qquad (i = 1, 2, 3, 4).$$

Hence the eccentricity of each vertex is

$$e_\mu(v_1) = \max\left\{0, \frac{5}{4}, \frac{13}{4}, \frac{9}{2}\right\} = \frac{9}{2},$$

$$e_\mu(v_2) = \max\left\{\frac{5}{4}, 0, 2, \frac{13}{4}\right\} = \frac{13}{4},$$

$$e_\mu(v_3) = \max\left\{\frac{13}{4}, 2, 0, \frac{5}{4}\right\} = \frac{13}{4},$$

$$e_\mu(v_4) = \max\left\{\frac{9}{2}, \frac{13}{4}, \frac{5}{4}, 0\right\} = \frac{9}{2}.$$

Therefore the radius of $G$ is

$$r_\mu(G) = \min_{v\in V} e_\mu(v) = \min\left\{\frac{9}{2}, \frac{13}{4}, \frac{13}{4}, \frac{9}{2}\right\} = \frac{13}{4},$$

and the diameter of $G$ is

$$d_\mu(G) = \max_{v\in V} e_\mu(v) = \max\left\{\frac{9}{2}, \frac{13}{4}, \frac{13}{4}, \frac{9}{2}\right\} = \frac{9}{2}.$$

Thus the vertices $v_2$ and $v_3$ are the central vertices of this fuzzy graph, while the largest fuzzy distance occurs between the end vertices $v_1$ and $v_4$.

A schematic illustration of this fuzzy graph is shown in Figure 3.6.

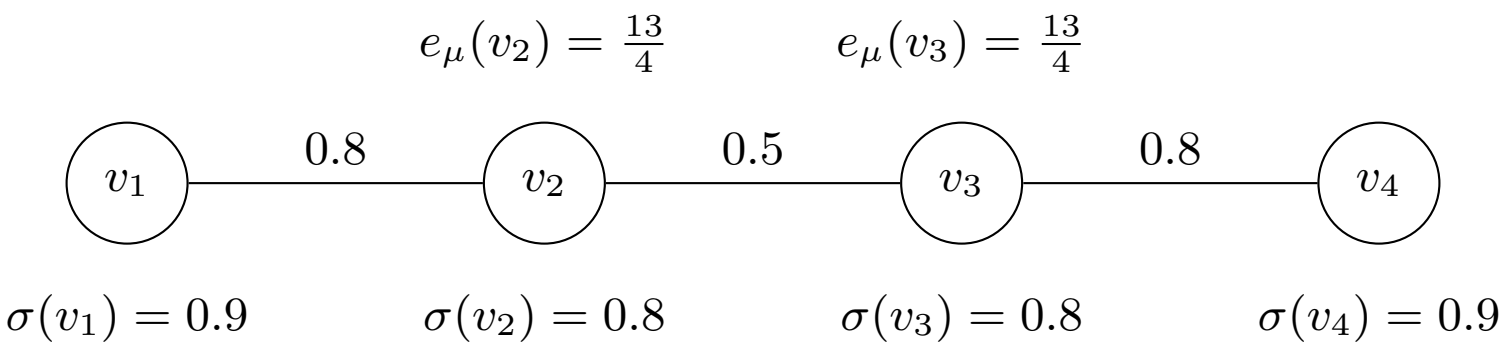

Figure 3.6: A fuzzy graph illustrating radius and diameter

The radius of an uncertain graph is the minimum eccentricity among its vertices, while the diameter is the maximum eccentricity among its vertices.

**Definition 3.8.3** (Uncertain Radius and Diameter)**.** Let

$$\mathcal{G}_M = (V, E, \sigma_M, \eta_M)$$

be a finite connected Uncertain Graph of type $M$, and let

$$d_M : V \times V \to [0, \infty)$$

be the uncertain distance on $\mathcal{G}_M$.

For each vertex $v \in V$, the *uncertain eccentricity* of $v$ is defined by

$$e_M(v) := \max_{u\in V} d_M(v, u).$$

The *uncertain radius* of $\mathcal{G}_M$ is defined by

$$r_M(\mathcal{G}_M) := \min_{v\in V} e_M(v),$$

and the *uncertain diameter* of $\mathcal{G}_M$ is defined by

$$D_M(\mathcal{G}_M) := \max_{v\in V} e_M(v).$$

**Theorem 3.8.4** (Well-definedness of Uncertain Radius and Diameter). *Let*

$$\mathcal{G}_M = (V, E, \sigma_M, \eta_M)$$

*be a finite connected Uncertain Graph of type $M$, and suppose that the uncertain distance*

$$d_M : V \times V \to [0, \infty)$$

*is well-defined.*

*Then:*

1. *for every vertex $v \in V$, the uncertain eccentricity*

$$e_M(v) = \max_{u \in V} d_M(v, u)$$

*is well-defined;*

2. *the uncertain radius*

$$r_M(\mathcal{G}_M) = \min_{v \in V} e_M(v)$$

*is well-defined;*

3. *the uncertain diameter*

$$D_M(\mathcal{G}_M) = \max_{v \in V} e_M(v)$$

*is well-defined.*

*Hence the notions of uncertain eccentricity, uncertain radius, and uncertain diameter are well-defined.*

*Proof.* Since $\mathcal{G}_M$ is finite, the vertex set $V$ is a finite nonempty set.

Fix a vertex $v \in V$. Because the uncertain distance

$$d_M : V \times V \to [0, \infty)$$

is well-defined, the value

$$d_M(v, u) \in [0, \infty)$$

is well-defined for every $u \in V$. Therefore the set

$$\{ d_M(v, u) \mid u \in V \}$$

is a finite nonempty subset of $[0, \infty)$. Every finite nonempty subset of $\mathbb{R}$ has a maximum, so

$$e_M(v) := \max_{u \in V} d_M(v, u)$$

exists and is uniquely determined. Hence the uncertain eccentricity of $v$ is well-defined.

Now consider the set of all vertex eccentricities:

$$\{ e_M(v) \mid v \in V \}.$$

Since $V$ is finite and nonempty, this is again a finite nonempty subset of $[0, \infty)$. Therefore it has both a minimum and a maximum. Consequently,

$$r_M(\mathcal{G}_M) := \min_{v \in V} e_M(v)$$

and

$$D_M(\mathcal{G}_M) := \max_{v \in V} e_M(v)$$

exist and are uniquely determined.

Thus uncertain eccentricity, uncertain radius, and uncertain diameter are all well-defined. □

## 3.9 Uncertain Wheel

A fuzzy wheel is a fuzzy graph whose support is a wheel, with one central hub joined to all vertices of an outer cycle [170, 207, 208].

**Definition 3.9.1** (Fuzzy Wheel)**.** Let

$$G = (V, \sigma, \mu)$$

be a finite fuzzy graph, where

$$\sigma : V \to [0,1], \qquad \mu : V \times V \to [0,1], \qquad \mu(u,v) \le \min\{\sigma(u), \sigma(v)\} \quad (\forall\, u, v \in V),$$

and assume that $\mu$ is symmetric and $G$ has no loops.

Define the support vertex set and support edge set by

$$V^* := \{v \in V : \sigma(v) > 0\}, \qquad E^* := \big\{\{u,v\} \subseteq V^* : u \ne v,\ \mu(u,v) > 0\big\}.$$

Then $G$ is called a *fuzzy wheel* if there exist a vertex

$$c \in V^*$$

and distinct vertices

$$v_1, v_2, \ldots, v_n \in V^* \qquad (n \ge 3)$$

such that

$$V^* = \{c, v_1, v_2, \ldots, v_n\},$$

and

$$E^* = \big\{\{v_i, v_{i+1}\} : 1 \le i \le n-1\big\} \cup \big\{\{v_n, v_1\}\big\} \cup \big\{\{c, v_i\} : 1 \le i \le n\big\}.$$

Equivalently, the support graph $G^* = (V^*, E^*)$ is isomorphic to the wheel graph

$$W_{n+1} = K_1 + C_n.$$

In this case, $c$ is called the *hub* (or *center*) of the fuzzy wheel, and

$$v_1, v_2, \ldots, v_n$$

form its outer fuzzy cycle.

**Example 3.9.2** (Fuzzy wheel)**.** Let

$$V = \{c, v_1, v_2, v_3, v_4, v_5\}.$$

Define a fuzzy graph

$$G = (V, \sigma, \mu)$$

by the vertex-membership function

$$\sigma(c) = 0.9, \qquad \sigma(v_1) = 0.8, \qquad \sigma(v_2) = 0.7, \qquad \sigma(v_3) = 0.8, \qquad \sigma(v_4) = 0.6, \qquad \sigma(v_5) = 0.7,$$

and the symmetric edge-membership function $\mu : V \times V \to [0,1]$ given by

$$\mu(v_1, v_2) = 0.6, \qquad \mu(v_2, v_3) = 0.6, \qquad \mu(v_3, v_4) = 0.5, \qquad \mu(v_4, v_5) = 0.5, \qquad \mu(v_5, v_1) = 0.6,$$

$$\mu(c, v_1) = 0.7, \qquad \mu(c, v_2) = 0.6, \qquad \mu(c, v_3) = 0.7, \qquad \mu(c, v_4) = 0.5, \qquad \mu(c, v_5) = 0.6,$$

$$\mu(v,v) = 0 \qquad (\forall\, v \in V),$$

and

$$\mu(u,v) = 0$$

for all other distinct pairs $\{u, v\} \subseteq V$, with

$$\mu(u, v) = \mu(v, u) \qquad (\forall\, u, v \in V).$$

First, we verify that $G$ is a fuzzy graph. Indeed,

$$\mu(v_1, v_2) = 0.6 \le \min\{0.8, 0.7\} = 0.7,$$

$$\mu(v_2, v_3) = 0.6 \le \min\{0.7, 0.8\} = 0.7,$$
$$\mu(v_3, v_4) = 0.5 \le \min\{0.8, 0.6\} = 0.6,$$
$$\mu(v_4, v_5) = 0.5 \le \min\{0.6, 0.7\} = 0.6,$$
$$\mu(v_5, v_1) = 0.6 \le \min\{0.7, 0.8\} = 0.7,$$

and

$$\mu(c, v_1) = 0.7 \le \min\{0.9, 0.8\} = 0.8,$$
$$\mu(c, v_2) = 0.6 \le \min\{0.9, 0.7\} = 0.7,$$
$$\mu(c, v_3) = 0.7 \le \min\{0.9, 0.8\} = 0.8,$$
$$\mu(c, v_4) = 0.5 \le \min\{0.9, 0.6\} = 0.6,$$
$$\mu(c, v_5) = 0.6 \le \min\{0.9, 0.7\} = 0.7.$$

Hence

$$\mu(u, v) \le \min\{\sigma(u), \sigma(v)\} \qquad (\forall\, u, v \in V).$$

Since all vertex-membership values are positive, the support vertex set is

$$V^* = \{c, v_1, v_2, v_3, v_4, v_5\}.$$

Moreover, the positive edges are exactly

$$\{v_1, v_2\},\ \{v_2, v_3\},\ \{v_3, v_4\},\ \{v_4, v_5\},\ \{v_5, v_1\},$$

together with

$$\{c, v_1\},\ \{c, v_2\},\ \{c, v_3\},\ \{c, v_4\},\ \{c, v_5\}.$$

Therefore,

$$E^* = \big\{\{v_i, v_{i+1}\} : 1 \le i \le 4\big\} \cup \big\{\{v_5, v_1\}\big\} \cup \big\{\{c, v_i\} : 1 \le i \le 5\big\}.$$

Thus the support graph is a wheel:

$$G^* \cong W_6 = K_1 + C_5.$$

Hence $G$ is a fuzzy wheel with hub $c$, and

$$v_1, v_2, v_3, v_4, v_5$$

form its outer fuzzy cycle.

A schematic illustration of this fuzzy wheel is shown in Figure 3.7.

An uncertain wheel is an uncertain graph whose support graph is a wheel, consisting of one hub adjacent to every vertex of an outer cycle.

**Definition 3.9.3** (Support-Evaluable Uncertain Model)**.** Let $M$ be an uncertain model with degree-domain

$$\mathrm{Dom}(M) \subseteq [0, 1]^k.$$

We say that $M$ is *support-evaluable* if it is equipped with a distinguished element

$$0_M \in \mathrm{Dom}(M),$$

called the *zero degree.*

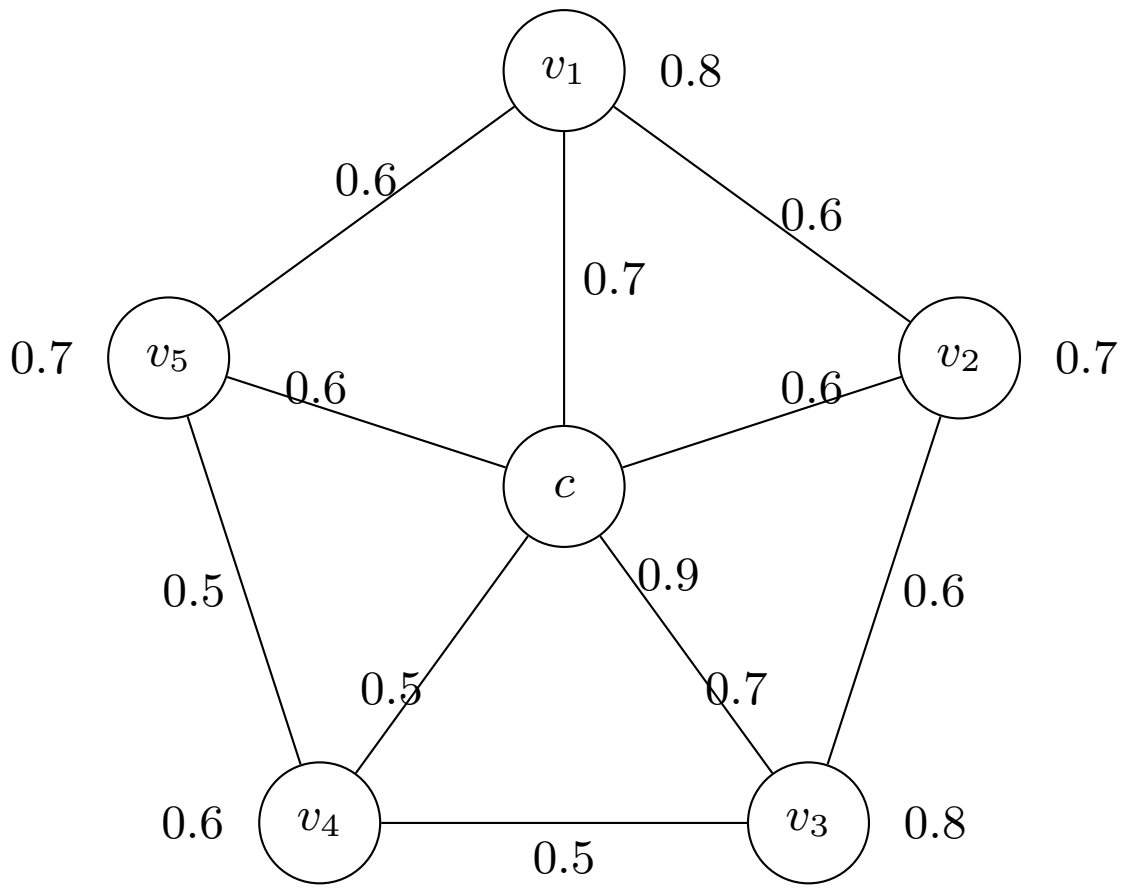

Figure 3.7: A fuzzy wheel with hub $c$ and outer fuzzy cycle $v_1v_2v_3v_4v_5v_1$

**Definition 3.9.4** (Uncertain Wheel)**.** Let $M$ be a support-evaluable uncertain model with degree-domain

$$\mathrm{Dom}(M) \subseteq [0,1]^k$$

and zero degree

$$0_M \in \mathrm{Dom}(M).$$

Let

$$\mathcal{G}_M = (V, E, \sigma_M, \eta_M)$$

be an Uncertain Graph of type $M$, where

$$\sigma_M : V \to \mathrm{Dom}(M), \qquad \eta_M : E \to \mathrm{Dom}(M).$$

Define the support vertex set by

$$V^*(\mathcal{G}_M) := \{\, v \in V \mid \sigma_M(v) \neq 0_M \,\},$$

and define the support edge set by

$$E^*(\mathcal{G}_M) := \big\{\{u,v\} \in E \mid u,v \in V^*(\mathcal{G}_M),\ \eta_M(\{u,v\}) \neq 0_M\big\}.$$

The *support graph* of $\mathcal{G}_M$ is

$$G^*_{\mathrm{supp}}(\mathcal{G}_M) := \big(V^*(\mathcal{G}_M),\, E^*(\mathcal{G}_M)\big).$$

Then $\mathcal{G}_M$ is called an *Uncertain Wheel* if there exist a vertex

$$c \in V^*(\mathcal{G}_M)$$

and distinct vertices

$$v_1, v_2, \ldots, v_n \in V^*(\mathcal{G}_M) \qquad (n \geq 3)$$

such that

$$V^*(\mathcal{G}_M) = \{c, v_1, v_2, \ldots, v_n\},$$

and

$$E^*(\mathcal{G}_M) = \big\{\{v_i, v_{i+1}\} : 1 \leq i \leq n-1\big\} \cup \big\{\{v_n, v_1\}\big\} \cup \big\{\{c, v_i\} : 1 \leq i \leq n\big\}.$$

Equivalently, $\mathcal{G}_M$ is an uncertain wheel if its support graph is isomorphic to the wheel graph

$$W_{n+1} = K_1 + C_n.$$

In this case, $c$ is called the *hub* (or *center*) of the uncertain wheel, and

$$v_1, v_2, \ldots, v_n$$

form its outer cycle.

**Theorem 3.9.5** (Well-definedness of Uncertain Wheel)**.** *Let $M$ be a support-evaluable uncertain model with degree-domain*

$$\mathrm{Dom}(M) \subseteq [0,1]^k$$

*and zero degree*

$$0_M \in \mathrm{Dom}(M).$$

*Let*

$$\mathcal{G}_M = (V, E, \sigma_M, \eta_M)$$

*be an Uncertain Graph of type $M$.*

*Then the support sets*

$$V^*(\mathcal{G}_M) = \{\, v \in V \mid \sigma_M(v) \neq 0_M \,\}$$

*and*

$$E^*(\mathcal{G}_M) = \big\{ \{u,v\} \in E \mid u,v \in V^*(\mathcal{G}_M),\ \ \eta_M(\{u,v\}) \neq 0_M \big\}$$

*are well-defined.*

*Consequently, the support graph*

$$G^*_{\mathrm{supp}}(\mathcal{G}_M) = \big(V^*(\mathcal{G}_M), E^*(\mathcal{G}_M)\big)$$

*is well-defined, and the statement*

*"$\mathcal{G}_M$ is an Uncertain Wheel"*

*is well-defined.*

*Hence the notion of an uncertain wheel is well-defined.*

*Proof.* Since $M$ is an uncertain model, its degree-domain

$$\mathrm{Dom}(M)$$

is fixed. Since $M$ is support-evaluable, the element

$$0_M \in \mathrm{Dom}(M)$$

is fixed as well.

Because

$$\sigma_M : V \to \mathrm{Dom}(M)$$

is a function, for each $v \in V$ the value $\sigma_M(v)$ is uniquely determined in $\mathrm{Dom}(M)$. Hence the condition

$$\sigma_M(v) \neq 0_M$$

is meaningful for every $v \in V$. Therefore

$$V^*(\mathcal{G}_M) = \{\, v \in V \mid \sigma_M(v) \neq 0_M \,\}$$

is a well-defined subset of $V$.

Likewise, because

$$\eta_M : E \to \mathrm{Dom}(M)$$

is a function, for each edge $e \in E$ the value $\eta_M(e)$ is uniquely determined in $\mathrm{Dom}(M)$. Hence the condition

$$\eta_M(e) \neq 0_M$$

is meaningful for every $e \in E$.

Now define

$$E^*(\mathcal{G}_M) = \big\{\{u,v\} \in E \mid u,v \in V^*(\mathcal{G}_M),\ \eta_M(\{u,v\}) \neq 0_M\big\}.$$

Since $E$ and $V^*(\mathcal{G}_M)$ are well-defined, and since the predicate

$$u,v \in V^*(\mathcal{G}_M),\ \eta_M(\{u,v\}) \neq 0_M$$

is meaningful, the set $E^*(\mathcal{G}_M)$ is a well-defined subset of $E$.

Therefore the pair

$$G^*_{\text{supp}}(\mathcal{G}_M) = \big(V^*(\mathcal{G}_M), E^*(\mathcal{G}_M)\big)$$

is a well-defined graph.

The statement that $\mathcal{G}_M$ is an uncertain wheel asserts that there exist a vertex

$$c \in V^*(\mathcal{G}_M)$$

and distinct vertices

$$v_1, \ldots, v_n \in V^*(\mathcal{G}_M) \qquad (n \geq 3)$$

such that

$$V^*(\mathcal{G}_M) = \{c, v_1, \ldots, v_n\},$$

and

$$E^*(\mathcal{G}_M) = \big\{\{v_i, v_{i+1}\} : 1 \leq i \leq n-1\big\} \cup \big\{\{v_n, v_1\}\big\} \cup \big\{\{c, v_i\} : 1 \leq i \leq n\big\}.$$

Since both $V^*(\mathcal{G}_M)$ and $E^*(\mathcal{G}_M)$ are well-defined sets, these equalities are meaningful. Equivalently, the statement

$$G^*_{\text{supp}}(\mathcal{G}_M) \cong W_{n+1}$$

is meaningful because both graphs are well-defined crisp graphs.

Hence the predicate

$$\text{“}\mathcal{G}_M \text{ is an Uncertain Wheel”}$$

is well-defined.

Therefore the notion of an uncertain wheel is well-defined. □

Representative wheel-related concepts under uncertainty-aware graph frameworks are listed in Table 3.4.

Related concepts such as the gear graph [212, 213], helm graph [214], fan graph [215], friendship graph [216], multi-wheel graph [217, 218], and double wheel graph [219] are also well known.

Table 3.4: Representative wheel-related concepts under uncertainty-aware graph frameworks, classified by the dimension $k$ of the information attached to vertices and/or edges.

| $k$ | **Wheel-related concept** | **Typical coordinate form** | **Canonical information attached to vertices/edges** |
|---|---|---|---|
| 1 | Fuzzy Wheel | $\mu$ | A wheel is studied in a fuzzy graph, where each vertex and edge is associated with a single membership degree in $[0, 1]$. |
| 2 | Intuitionistic Fuzzy Wheel [209, 210] | $(\mu, \nu)$ | A wheel is defined in an intuitionistic fuzzy graph, where each vertex and edge carries a membership degree and a non-membership degree, usually satisfying $\mu + \nu \leq 1$. |
| 3 | Neutrosophic Wheel [211] | $(T, I, F)$ | A wheel is defined in a neutrosophic graph, where each vertex and edge is described by truth, indeterminacy, and falsity degrees. |

# Chapter 4

# Graph Classes

In this chapter, graph classes based on fuzzy graphs and uncertain graphs are introduced and investigated.

## 4.1 Uncertain Digraph

A fuzzy directed graph assigns membership values to vertices and directed edges [220–223].

**Definition 4.1.1** (Fuzzy Directed Graph)**.** [224–226] A *fuzzy directed graph* is a quadruple

$$G = (V, E, \sigma, \mu),$$

where

- $V$ is a nonempty set of vertices,
- $E \subseteq V \times V$ is a set of directed edges,
- $\sigma : V \to [0, 1]$ assigns a membership degree to each vertex,
- $\mu : E \to [0, 1]$ assigns a membership degree to each directed edge,

such that, for every $(u, v) \in E$,

$$\mu(u, v) \le \min\{\sigma(u), \sigma(v)\}.$$

**Example 4.1.2** (Fuzzy directed graph)**.** Let

$$V = \{v_1, v_2, v_3, v_4\},$$

and define the directed edge set by

$$E = \{(v_1, v_2), (v_2, v_3), (v_3, v_1), (v_1, v_4), (v_4, v_3)\}.$$

Define the vertex-membership function

$$\sigma : V \to [0, 1]$$

by

$$\sigma(v_1) = 0.9, \qquad \sigma(v_2) = 0.7, \qquad \sigma(v_3) = 0.8, \qquad \sigma(v_4) = 0.6,$$

and define the directed edge-membership function

$$\mu : E \to [0,1]$$

by

$$\mu(v_1, v_2) = 0.6, \qquad \mu(v_2, v_3) = 0.5, \qquad \mu(v_3, v_1) = 0.7, \qquad \mu(v_1, v_4) = 0.4, \qquad \mu(v_4, v_3) = 0.5.$$

Then

$$G = (V, E, \sigma, \mu)$$

is a fuzzy directed graph, because for every directed edge $(u, v) \in E$,

$$\mu(u, v) \leq \min\{\sigma(u), \sigma(v)\}.$$

Indeed,

$$\mu(v_1, v_2) = 0.6 \leq \min\{0.9, 0.7\} = 0.7,$$
$$\mu(v_2, v_3) = 0.5 \leq \min\{0.7, 0.8\} = 0.7,$$
$$\mu(v_3, v_1) = 0.7 \leq \min\{0.8, 0.9\} = 0.8,$$
$$\mu(v_1, v_4) = 0.4 \leq \min\{0.9, 0.6\} = 0.6,$$

and

$$\mu(v_4, v_3) = 0.5 \leq \min\{0.6, 0.8\} = 0.6.$$

Hence all required conditions are satisfied.

Observe that this graph is genuinely directed. For example,

$$(v_1, v_2) \in E, \qquad \text{but} \qquad (v_2, v_1) \notin E,$$

so the edge relation is not symmetric.

A schematic illustration of this fuzzy directed graph is shown in Figure 4.1.

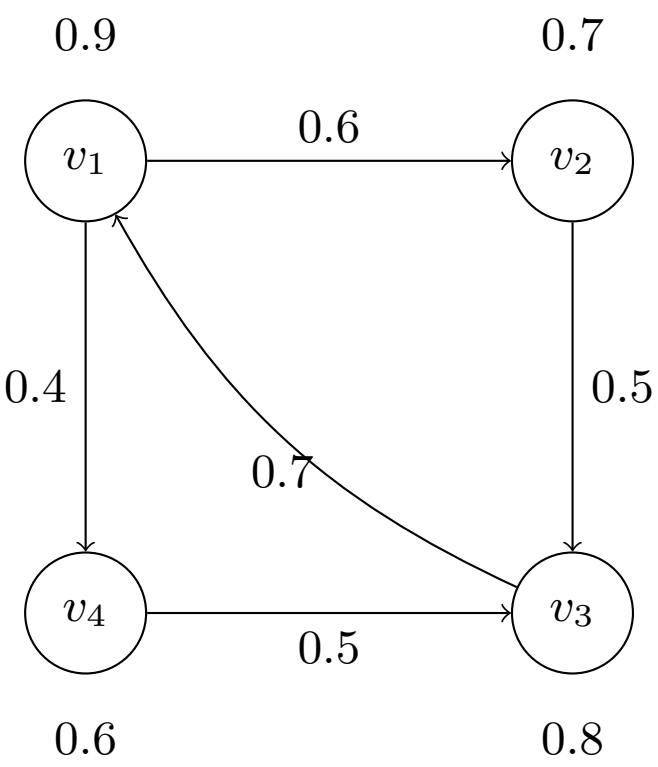

Figure 4.1: A fuzzy directed graph

An uncertain directed graph assigns uncertainty degrees to vertices and directed edges, thereby representing asymmetric uncertain relations among vertices.

**Definition 4.1.3** (Uncertain Directed Graph)**.** Let

$$D^* = (V, A)$$

be a finite directed graph, where

$$A \subseteq V \times V$$

is the set of directed edges (arcs). Assume that $D^*$ is loopless, that is,

$$(u,u) \notin A \qquad (\forall\, u \in V).$$

Let $M$ be a fixed uncertain model with degree-domain

$$\mathrm{Dom}(M) \subseteq [0,1]^k.$$

An *Uncertain Directed Graph of type $M$* is a quadruple

$$\mathcal{D}_M = (V, A, \sigma_M, \alpha_M),$$

where

$$\sigma_M : V \longrightarrow \mathrm{Dom}(M)$$

and

$$\alpha_M : A \longrightarrow \mathrm{Dom}(M)$$

are uncertainty-degree functions on the vertex set and the arc set, respectively.

Equivalently,

$$(V, \sigma_M)$$

is an Uncertain Set of type $M$ on $V$, and

$$(A, \alpha_M)$$

is an Uncertain Set of type $M$ on $A$.

For each vertex $v \in V$, the value

$$\sigma_M(v) \in \mathrm{Dom}(M)$$

represents the uncertainty degree of $v$, and for each arc

$$(u,v) \in A,$$

the value

$$\alpha_M((u,v)) \in \mathrm{Dom}(M)$$

represents the uncertainty degree of the directed edge from $u$ to $v$.

If desired, one may additionally impose model-specific compatibility conditions involving

$$\sigma_M(u), \qquad \sigma_M(v), \qquad \alpha_M((u,v)),$$

but such conditions depend on the chosen uncertain model $M$ and are not fixed at the level of this general definition.

**Theorem 4.1.4** (Well-definedness of Uncertain Directed Graph)**.** *Let*

$$D^* = (V, A)$$

*be a finite loopless directed graph, let $M$ be an uncertain model with degree-domain*

$$\mathrm{Dom}(M) \subseteq [0,1]^k,$$

*and let*

$$\sigma_M : V \to \mathrm{Dom}(M), \qquad \alpha_M : A \to \mathrm{Dom}(M)$$

*be functions.*

*Then*

$$\mathcal{D}_M = (V, A, \sigma_M, \alpha_M)$$

*is a well-defined Uncertain Directed Graph of type $M$.*

*Moreover,*

$$(V, \sigma_M)$$

*is an Uncertain Set of type $M$ on $V$, and*

$$(A, \alpha_M)$$

*is an Uncertain Set of type $M$ on $A$.*

*Proof.* Since $M$ is an uncertain model, its degree-domain

$$\mathrm{Dom}(M)$$

is fixed. Hence the maps

$$\sigma_M : V \to \mathrm{Dom}(M) \qquad \text{and} \qquad \alpha_M : A \to \mathrm{Dom}(M)$$

are ordinary set-theoretic functions with well-specified codomain $\mathrm{Dom}(M)$.

Therefore the pairs

$$(V, \sigma_M) \quad \text{and} \quad (A, \alpha_M)$$

are Uncertain Sets of type $M$ on $V$ and $A$, respectively.

Next, because

$$A \subseteq V \times V,$$

every arc $a \in A$ is a uniquely determined ordered pair

$$a = (u, v)$$

with $u, v \in V$. Hence each directed edge has a uniquely determined source $u$ and a uniquely determined target $v$.

Accordingly, for every arc $(u, v) \in A$, the value

$$\alpha_M((u, v)) \in \mathrm{Dom}(M)$$

is unambiguously assigned to that ordered pair. Since order matters in $V \times V$, the arcs $(u, v)$ and $(v, u)$ are distinct whenever $u \neq v$, so no ambiguity arises between opposite directions.

Thus all components of

$$\mathcal{D}_M = (V, A, \sigma_M, \alpha_M)$$

are simultaneously and uniquely specified:

- $V$ is the vertex set;
- $A$ is the arc set;
- $\sigma_M$ assigns a unique uncertainty degree to each vertex;
- $\alpha_M$ assigns a unique uncertainty degree to each directed edge.

Hence

$$\mathcal{D}_M = (V, A, \sigma_M, \alpha_M)$$

defines a unique mathematical object, namely an Uncertain Directed Graph of type $M$. Therefore the definition is well-defined. □

Representative directed-graph concepts under uncertainty-aware graph frameworks are listed in Table 4.1.

Table 4.1: Representative directed-graph concepts under uncertainty-aware graph frameworks, classified by the dimension $k$ of the information attached to vertices and/or edges.

| $k$ | **Directed-graph concept** | **Typical coordinate form** | **Canonical information attached to vertices/edges** |
|---|---|---|---|
| 1 | Fuzzy Directed Graph [227–229] | $\mu$ | A directed graph in which each vertex and arc is associated with a single membership degree in $[0, 1]$. |
| 2 | Vague Directed Graph [230, 231] | $(t, f)$ | A directed graph in which each vertex and arc is described by a truth-membership degree and a falsity-membership degree, typically with $t+f \leq 1$. |
| 2 | Intuitionistic Fuzzy Directed Graph [232–234] | $(\mu, \nu)$ | A directed graph in which each vertex and arc carries a membership degree and a non-membership degree, usually satisfying $\mu + \nu \leq 1$. |
| 3 | Spherical Fuzzy Directed Graph [235] | $(\mu, \eta, \nu)$ | A directed graph in which each vertex and arc is assigned positive, neutral, and negative membership degrees, usually satisfying $\mu^2 + \eta^2 + \nu^2 \leq 1$. |
| 3 | Picture Fuzzy Digraph [236, 237] | $(\mu, \eta, \nu)$ | A directed graph in which each vertex and arc is described by positive, neutral, and negative membership degrees, usually satisfying $\mu+\eta+\nu \leq 1$. |
| 3 | Neutrosophic Directed Graph [131, 238, 239] | $(T, I, F)$ | A directed graph in which each vertex and arc is described by truth, indeterminacy, and falsity degrees. |
| $s + t$ | Plithogenic Directed Graph [240] | $(\mathbf{a}, \mathbf{c}) \in [0, 1]^s \times [0, 1]^t$ | A directed graph in which each vertex and arc is described by attribute-based information together with an $s$-dimensional appurtenance vector and a $t$-dimensional contradiction vector. |

## 4.2 Uncertain Bidirected Graph

A fuzzy bidirected graph assigns membership values to vertices and edges, while each endpoint of an edge has its own local orientation [241].

**Definition 4.2.1** (Fuzzy Bidirected Graph)**.** [241] A *fuzzy bidirected graph* is a quintuple

$$G = (V, E, \sigma, \mu, \tau),$$

where

- $V$ is a nonempty set of vertices,
- $E \subseteq \big\{\{u, v\} \mid u, v \in V,\ u \neq v\big\}$ is a set of bidirected edges,
- $\sigma : V \to [0, 1]$ assigns a membership degree to each vertex,
- $\mu : E \to [0, 1]$ assigns a membership degree to each edge,
- $\tau : V \times E \to \{-1, 0, 1\}$ is a bidirection function,

such that, for each edge $e = \{u, v\} \in E$,

$$\tau(u, e), \tau(v, e) \in \{-1, 1\}, \qquad \tau(w, e) = 0 \text{ for all } w \in V \setminus \{u, v\},$$

and

$$\mu(e) \leq \min\{\sigma(u), \sigma(v)\}.$$

Thus each endpoint of a fuzzy edge carries its own local orientation.

**Example 4.2.2** (Fuzzy bidirected graph)**.** Let

$$V = \{v_1, v_2, v_3, v_4\},$$

and let

$$E = \{e_{12}, e_{23}, e_{13}, e_{34}\},$$

where

$$e_{12} = \{v_1, v_2\}, \qquad e_{23} = \{v_2, v_3\}, \qquad e_{13} = \{v_1, v_3\}, \qquad e_{34} = \{v_3, v_4\}.$$

Define the vertex-membership function

$$\sigma : V \to [0, 1]$$

by

$$\sigma(v_1) = 0.9, \qquad \sigma(v_2) = 0.8, \qquad \sigma(v_3) = 0.7, \qquad \sigma(v_4) = 0.6.$$

Define the edge-membership function

$$\mu : E \to [0, 1]$$

by

$$\mu(e_{12}) = 0.6, \qquad \mu(e_{23}) = 0.5, \qquad \mu(e_{13}) = 0.7, \qquad \mu(e_{34}) = 0.4.$$

Next, define the bidirection function

$$\tau : V \times E \to \{-1, 0, 1\}$$

as follows:

$$\begin{aligned} \tau(v_1, e_{12}) = 1, &\qquad \tau(v_2, e_{12}) = -1, \\ \tau(v_2, e_{23}) = 1, &\qquad \tau(v_3, e_{23}) = 1, \\ \tau(v_1, e_{13}) = -1, &\qquad \tau(v_3, e_{13}) = 1, \\ \tau(v_3, e_{34}) = -1, &\qquad \tau(v_4, e_{34}) = -1, \end{aligned}$$

and

$$\tau(w, e) = 0$$

for every pair $(w, e)$ not listed above.

Then

$$G = (V, E, \sigma, \mu, \tau)$$

is a fuzzy bidirected graph.

Indeed, for each edge $e = \{u, v\} \in E$, the values at its endpoints belong to

$$\{-1, 1\},$$

while every non-incident vertex receives value 0. For example, for the edge

$$e_{12} = \{v_1, v_2\},$$

we have

$$\tau(v_1, e_{12}) = 1, \qquad \tau(v_2, e_{12}) = -1,$$

and

$$\tau(v_3, e_{12}) = \tau(v_4, e_{12}) = 0.$$

Similarly, the same property holds for $e_{23}, e_{13}$, and $e_{34}$.

Moreover, the membership condition is satisfied:

$$\mu(e_{12}) = 0.6 \leq \min\{\sigma(v_1), \sigma(v_2)\} = \min\{0.9, 0.8\} = 0.8,$$

$$\mu(e_{23}) = 0.5 \le \min\{\sigma(v_2), \sigma(v_3)\} = \min\{0.8, 0.7\} = 0.7,$$
$$\mu(e_{13}) = 0.7 \le \min\{\sigma(v_1), \sigma(v_3)\} = \min\{0.9, 0.7\} = 0.7,$$

and

$$\mu(e_{34}) = 0.4 \le \min\{\sigma(v_3), \sigma(v_4)\} = \min\{0.7, 0.6\} = 0.6.$$

Hence all requirements in the definition are fulfilled. Therefore,

$$G = (V, E, \sigma, \mu, \tau)$$

is a fuzzy bidirected graph.

A schematic illustration is shown in Figure 4.2. In the figure, the number written near the midpoint of each edge is $\mu(e)$, while the small signs near the endpoints indicate the corresponding values of $\tau(\cdot, e)$.

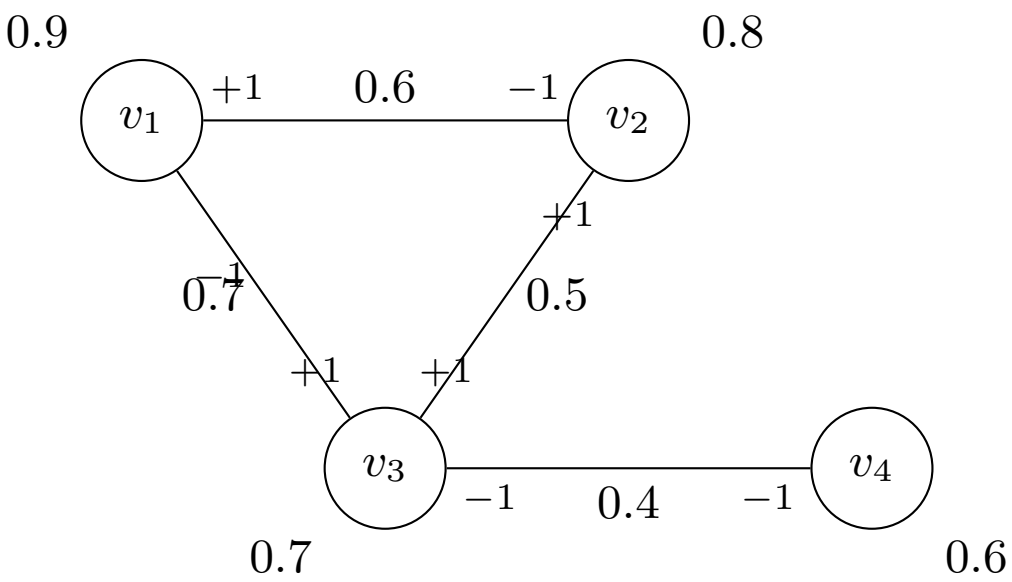

Figure 4.2: A fuzzy bidirected graph

The extensions based on Uncertain Sets are presented below.

**Definition 4.2.3** (Uncertain Bidirected Graph)**.** Let

$$B^* = (V, E, \tau)$$

be a finite bidirected graph, where

$$E \subseteq \big\{\{u, v\} \mid u, v \in V,\ u \ne v\big\}$$

is a set of undirected edges, and

$$\tau : V \times E \to \{-1, 0, 1\}$$

is a bidirection function satisfying the following condition: for each edge

$$e = \{u, v\} \in E,$$

we have

$$\tau(u, e), \tau(v, e) \in \{-1, 1\}, \qquad \tau(w, e) = 0 \quad \text{for all } w \in V \setminus \{u, v\}.$$

Let $M$ be a fixed uncertain model with degree-domain

$$\mathrm{Dom}(M) \subseteq [0, 1]^k.$$

An *Uncertain Bidirected Graph of type $M$* is a quintuple

$$\mathcal{B}_M = (V, E, \tau, \sigma_M, \eta_M),$$

where

$$\sigma_M : V \longrightarrow \mathrm{Dom}(M)$$

and

$$\eta_M : E \longrightarrow \mathrm{Dom}(M)$$

are uncertainty-degree functions on the vertex set and the edge set, respectively.

Equivalently, $(V, \sigma_M)$ is an Uncertain Set of type $M$ on $V$, and $(E, \eta_M)$ is an Uncertain Set of type $M$ on $E$.

For each vertex $v \in V$, the value

$$\sigma_M(v) \in \mathrm{Dom}(M)$$

represents the uncertainty degree of $v$, and for each edge $e \in E$, the value

$$\eta_M(e) \in \mathrm{Dom}(M)$$

represents the uncertainty degree of $e$, while the function $\tau$ assigns an independent local orientation at each endpoint of every edge.

If desired, one may additionally impose model-specific compatibility conditions between $\eta_M(e)$ and the endpoint degrees $\sigma_M(u), \sigma_M(v)$ for $e = \{u, v\}$, but such conditions depend on the chosen uncertain model $M$ and are not fixed at the level of this general definition.

**Theorem 4.2.4** (Well-definedness of Uncertain Bidirected Graph)**.** *Let*

$$B^* = (V, E, \tau)$$

*be a finite bidirected graph, where*

$$E \subseteq \big\{\{u, v\} \mid u, v \in V,\ u \neq v\big\}$$

*and*

$$\tau : V \times E \to \{-1, 0, 1\}$$

*satisfies the bidirected-incidence condition: for every edge*

$$e = \{u, v\} \in E,$$

*there exist exactly two distinct vertices $u, v \in V$ such that*

$$\tau(u, e), \tau(v, e) \in \{-1, 1\}, \qquad \tau(w, e) = 0 \quad \text{for all } w \in V \setminus \{u, v\}.$$

*Let $M$ be an uncertain model with degree-domain*

$$\mathrm{Dom}(M) \subseteq [0, 1]^k,$$

*and let*

$$\sigma_M : V \to \mathrm{Dom}(M), \qquad \eta_M : E \to \mathrm{Dom}(M)$$

*be functions.*

*Then the quintuple*

$$\mathcal{B}_M = (V, E, \tau, \sigma_M, \eta_M)$$

*is a well-defined Uncertain Bidirected Graph of type $M$.*

*Moreover,*

$$(V, \sigma_M)$$

*is an Uncertain Set of type $M$ on the vertex set $V$, and*

$$(E, \eta_M)$$

*is an Uncertain Set of type $M$ on the edge set $E$.*

*Proof.* Since $M$ is an uncertain model, its degree-domain $\mathrm{Dom}(M)$ is a fixed admissible set of uncertainty degrees. Hence the maps

$$\sigma_M : V \to \mathrm{Dom}(M) \qquad \text{and} \qquad \eta_M : E \to \mathrm{Dom}(M)$$

are ordinary set-theoretic functions with well-specified codomain $\mathrm{Dom}(M)$. Therefore the pairs

$$(V, \sigma_M) \quad \text{and} \quad (E, \eta_M)$$

are Uncertain Sets of type $M$ on $V$ and $E$, respectively.

Next, because $B^* = (V, E, \tau)$ is a bidirected graph, every edge $e \in E$ is an unordered two-element subset of $V$, say

$$e = \{u, v\}, \qquad u \neq v,$$

and the bidirection function $\tau$ assigns local orientations only to the two incidences

$$(u, e) \quad \text{and} \quad (v, e),$$

while all nonincident pairs $(w, e)$ have value 0. Thus the local orientation data attached to each edge is unambiguous.

Therefore all components of

$$\mathcal{B}_M = (V, E, \tau, \sigma_M, \eta_M)$$

are simultaneously well specified:

- $V$ is the vertex set;
- $E$ is the set of bidirected edges;
- $\tau$ is the bidirection function on incidences;
- $\sigma_M$ assigns to each vertex a unique uncertainty degree in $\mathrm{Dom}(M)$;
- $\eta_M$ assigns to each edge a unique uncertainty degree in $\mathrm{Dom}(M)$.

Hence the quintuple

$$\mathcal{B}_M = (V, E, \tau, \sigma_M, \eta_M)$$

defines a unique mathematical object, namely an Uncertain Bidirected Graph of type $M$. Therefore the definition is well-defined. □

## 4.3 Uncertain MutliDirected Graph

A fuzzy multidirected graph assigns membership values to vertices and directed edges and allows multiple parallel directed edges between the same ordered pair [241].

**Definition 4.3.1** (Fuzzy MultiDirected Graph)**.** A *fuzzy multidirected graph* is a quadruple

$$G = (V, E, \sigma, \mu),$$

where

- $V$ is a nonempty set of vertices,
- $E$ is a multiset of directed edges, each edge $e \in E$ being an ordered pair

  $$e = (u, v) \in V \times V,$$

  so that multiple parallel directed edges between the same ordered pair are allowed,

- $\sigma : V \to [0, 1]$ assigns a membership degree to each vertex,
- $\mu : E \to [0, 1]$ assigns a membership degree to each directed edge,

such that, for every edge $e = (u, v) \in E$,

$$\mu(e) \le \min\{\sigma(u), \sigma(v)\}.$$

Equivalently, if $s, t : E \to V$ denote the source and target maps, then

$$\mu(e) \le \min\{\sigma(s(e)), \sigma(t(e))\} \qquad (\forall e \in E).$$

**Example 4.3.2** (Fuzzy multidirected graph)**.** Let

$$V = \{v_1, v_2, v_3\},$$

and let $E$ be the multiset of directed edges

$$E = \{e_1, e_2, e_3, e_4, e_5\},$$

where

$$e_1 = (v_1, v_2), \qquad e_2 = (v_1, v_2), \qquad e_3 = (v_2, v_3), \qquad e_4 = (v_3, v_1), \qquad e_5 = (v_2, v_1).$$

Thus $e_1$ and $e_2$ are two distinct parallel directed edges from $v_1$ to $v_2$.

Define the vertex-membership function

$$\sigma : V \to [0, 1]$$

by

$$\sigma(v_1) = 0.9, \qquad \sigma(v_2) = 0.8, \qquad \sigma(v_3) = 0.7.$$

Define the edge-membership function

$$\mu : E \to [0, 1]$$

by

$$\mu(e_1) = 0.5, \qquad \mu(e_2) = 0.7, \qquad \mu(e_3) = 0.6, \qquad \mu(e_4) = 0.6, \qquad \mu(e_5) = 0.4.$$

Then

$$G = (V, E, \sigma, \mu)$$

is a fuzzy multidirected graph.

Indeed, for each edge $e = (u, v) \in E$, we verify that

$$\mu(e) \le \min\{\sigma(u), \sigma(v)\}.$$

For the two parallel edges from $v_1$ to $v_2$, we have

$$\mu(e_1) = 0.5 \le \min\{\sigma(v_1), \sigma(v_2)\} = \min\{0.9, 0.8\} = 0.8,$$

and

$$\mu(e_2) = 0.7 \le \min\{\sigma(v_1), \sigma(v_2)\} = \min\{0.9, 0.8\} = 0.8.$$

Also,

$$\mu(e_3) = 0.6 \le \min\{\sigma(v_2), \sigma(v_3)\} = \min\{0.8, 0.7\} = 0.7,$$
$$\mu(e_4) = 0.6 \le \min\{\sigma(v_3), \sigma(v_1)\} = \min\{0.7, 0.9\} = 0.7,$$

and

$$\mu(e_5) = 0.4 \le \min\{\sigma(v_2), \sigma(v_1)\} = \min\{0.8, 0.9\} = 0.8.$$

Hence all conditions in the definition are satisfied.

Therefore,

$$G = (V, E, \sigma, \mu)$$

is a fuzzy multidirected graph. This example illustrates that multiple distinct directed edges with the same source and target may coexist, each carrying its own membership value.

An uncertain multidirected graph assigns uncertainty degrees to vertices and directed edges, while allowing multiple distinct parallel directed edges between the same ordered pair of vertices.

**Definition 4.3.3** (Uncertain MultiDirected Graph)**.** Let

$$D^* = (V, A, s, t)$$

be a finite multidirected graph, where

- $V$ is a nonempty set of vertices,
- $A$ is a finite set of directed edge identifiers (arcs),
- $s : A \to V$ is the source map,
- $t : A \to V$ is the target map.

Thus each arc $a \in A$ is directed from $s(a)$ to $t(a)$, and multiple parallel directed edges are allowed in the sense that there may exist distinct arcs $a, b \in A$ such that

$$s(a) = s(b), \qquad t(a) = t(b).$$

Let $M$ be a fixed uncertain model with degree-domain

$$\mathrm{Dom}(M) \subseteq [0,1]^k.$$

An *Uncertain MultiDirected Graph of type* $M$ is a quadruple

$$\mathcal{D}_M = (V, A, \sigma_M, \alpha_M),$$

or, when the source and target maps are to be displayed explicitly,

$$\mathcal{D}_M = (V, A, s, t, \sigma_M, \alpha_M),$$

where

$$\sigma_M : V \longrightarrow \mathrm{Dom}(M)$$

and

$$\alpha_M : A \longrightarrow \mathrm{Dom}(M)$$

are uncertainty-degree functions on the vertex set and the arc set, respectively.

Equivalently,

$$(V, \sigma_M)$$

is an Uncertain Set of type $M$ on $V$, and

$$(A, \alpha_M)$$

is an Uncertain Set of type $M$ on $A$.

For each vertex $v \in V$, the value

$$\sigma_M(v) \in \mathrm{Dom}(M)$$

represents the uncertainty degree of $v$, and for each arc $a \in A$, the value

$$\alpha_M(a) \in \mathrm{Dom}(M)$$

represents the uncertainty degree of the directed edge from $s(a)$ to $t(a)$.

If desired, one may additionally impose model-specific compatibility conditions involving

$$\alpha_M(a), \quad \sigma_M(s(a)), \quad \sigma_M(t(a)),$$

but such conditions depend on the chosen uncertain model $M$ and are not fixed at the level of this general definition.

**Theorem 4.3.4** (Well-definedness of Uncertain MultiDirected Graph)**.** *Let*

$$D^* = (V, A, s, t)$$

*be a finite multidirected graph, where $V$ is a nonempty set, $A$ is a finite set of arc identifiers, and*

$$s, t : A \to V$$

*are the source and target maps.*

*Let $M$ be an uncertain model with degree-domain*

$$\mathrm{Dom}(M) \subseteq [0, 1]^k,$$

*and let*

$$\sigma_M : V \to \mathrm{Dom}(M), \qquad \alpha_M : A \to \mathrm{Dom}(M)$$

*be functions.*

*Then*

$$\mathcal{D}_M = (V, A, s, t, \sigma_M, \alpha_M)$$

*is a well-defined Uncertain MultiDirected Graph of type $M$.*

*Moreover,*

$$(V, \sigma_M)$$

*is an Uncertain Set of type $M$ on $V$, and*

$$(A, \alpha_M)$$

*is an Uncertain Set of type $M$ on $A$.*

*Proof.* Since $M$ is an uncertain model, its degree-domain

$$\mathrm{Dom}(M)$$

is a fixed admissible set of uncertainty degrees. Hence the maps

$$\sigma_M : V \to \mathrm{Dom}(M) \qquad \text{and} \qquad \alpha_M : A \to \mathrm{Dom}(M)$$

are ordinary set-theoretic functions with well-specified codomain $\mathrm{Dom}(M)$.

Therefore the pairs

$$(V, \sigma_M) \quad \text{and} \quad (A, \alpha_M)$$

are Uncertain Sets of type $M$ on the sets $V$ and $A$, respectively.

Next, because

$$D^* = (V, A, s, t)$$

is a multidirected graph, every arc $a \in A$ has a uniquely determined source

$$s(a) \in V$$

and a uniquely determined target

$$t(a) \in V.$$

Thus each arc identifier $a$ determines a unique directed edge from $s(a)$ to $t(a)$.

Even when parallel arcs are present, that is, when distinct arcs $a, b \in A$ satisfy

$$s(a) = s(b), \qquad t(a) = t(b),$$

the two arcs remain distinct as elements of the identifier set $A$. Hence the uncertainty-degree assignment

$$\alpha_M : A \to \mathrm{Dom}(M)$$

is unambiguous, because it is attached to arc identifiers rather than merely to ordered pairs of vertices.

Therefore all components of

$$\mathcal{D}_M = (V, A, s, t, \sigma_M, \alpha_M)$$

are simultaneously and uniquely specified:

- $V$ is the vertex set,
- $A$ is the arc-identifier set,
- $s$ and $t$ determine the direction of each arc,
- $\sigma_M$ assigns a unique uncertainty degree to each vertex,
- $\alpha_M$ assigns a unique uncertainty degree to each arc.

Hence

$$\mathcal{D}_M = (V, A, s, t, \sigma_M, \alpha_M)$$

defines a unique mathematical object, namely an Uncertain MultiDirected Graph of type $M$. Therefore the definition is well-defined. □

## 4.4 Uncertain Mixed Graph

A fuzzy mixed graph combines undirected and directed edges with membership values, representing symmetric and asymmetric relationships among vertices simultaneously [242, 243].

**Definition 4.4.1** (Fuzzy Mixed Graph). [242, 243] Let

$$M = (V, E, A)$$

be a finite mixed graph, where

$$E \subseteq \big\{\{u, v\} \mid u, v \in V,\ u \neq v\big\}$$

is the set of undirected edges and

$$A \subseteq \big\{(u, v) \in V \times V \mid u \neq v\big\}$$

is the set of directed edges (arcs).

A *fuzzy mixed graph* on $M$ is a sextuple

$$G = (V, E, A, \sigma, \mu_E, \mu_A),$$

where

$$\sigma : V \to [0, 1]$$

is a fuzzy subset of vertices,

$$\mu_E : E \to [0, 1]$$

is the membership function of undirected edges, and

$$\mu_A : A \to [0, 1]$$

is the membership function of directed edges, such that

$$\mu_E(\{u, v\}) \leq \min\{\sigma(u), \sigma(v)\} \qquad \text{for all } \{u, v\} \in E,$$

and

$$\mu_A((u, v)) \leq \min\{\sigma(u), \sigma(v)\} \qquad \text{for all } (u, v) \in A.$$

**Example 4.4.2** (Fuzzy mixed graph)**.** Let

$$V = \{v_1, v_2, v_3, v_4\},$$

and define the undirected edge set and the directed edge set by

$$E = \big\{\{v_1, v_2\}, \{v_2, v_3\}\big\}, \qquad A = \big\{(v_1, v_3), (v_3, v_4), (v_4, v_2)\big\}.$$

Define the vertex-membership function

$$\sigma : V \to [0, 1]$$

by

$$\sigma(v_1) = 0.9, \qquad \sigma(v_2) = 0.8, \qquad \sigma(v_3) = 0.7, \qquad \sigma(v_4) = 0.6.$$

Define the membership function of undirected edges

$$\mu_E : E \to [0, 1]$$

by

$$\mu_E(\{v_1, v_2\}) = 0.7, \qquad \mu_E(\{v_2, v_3\}) = 0.5,$$

and define the membership function of directed edges

$$\mu_A : A \to [0, 1]$$

by

$$\mu_A((v_1, v_3)) = 0.6, \qquad \mu_A((v_3, v_4)) = 0.5, \qquad \mu_A((v_4, v_2)) = 0.4.$$

Then

$$G = (V, E, A, \sigma, \mu_E, \mu_A)$$

is a fuzzy mixed graph.

Indeed, for each undirected edge, we have

$$\mu_E(\{v_1, v_2\}) = 0.7 \le \min\{\sigma(v_1), \sigma(v_2)\} = \min\{0.9, 0.8\} = 0.8,$$

and

$$\mu_E(\{v_2, v_3\}) = 0.5 \le \min\{\sigma(v_2), \sigma(v_3)\} = \min\{0.8, 0.7\} = 0.7.$$

Similarly, for each directed edge, we obtain

$$\mu_A((v_1, v_3)) = 0.6 \le \min\{\sigma(v_1), \sigma(v_3)\} = \min\{0.9, 0.7\} = 0.7,$$

$$\mu_A((v_3, v_4)) = 0.5 \le \min\{\sigma(v_3), \sigma(v_4)\} = \min\{0.7, 0.6\} = 0.6,$$

and

$$\mu_A((v_4, v_2)) = 0.4 \le \min\{\sigma(v_4), \sigma(v_2)\} = \min\{0.6, 0.8\} = 0.6.$$

Hence all defining conditions are satisfied. Therefore,

$$G = (V, E, A, \sigma, \mu_E, \mu_A)$$

is a fuzzy mixed graph.

In this example, the pairs

$$\{v_1, v_2\} \quad \text{and} \quad \{v_2, v_3\}$$

are connected by undirected fuzzy edges, whereas

$$(v_1, v_3), \ (v_3, v_4), \ (v_4, v_2)$$

are fuzzy directed edges. Thus the graph contains both undirected and directed fuzzy relations.

A schematic illustration of this fuzzy mixed graph is shown in Figure 4.3.

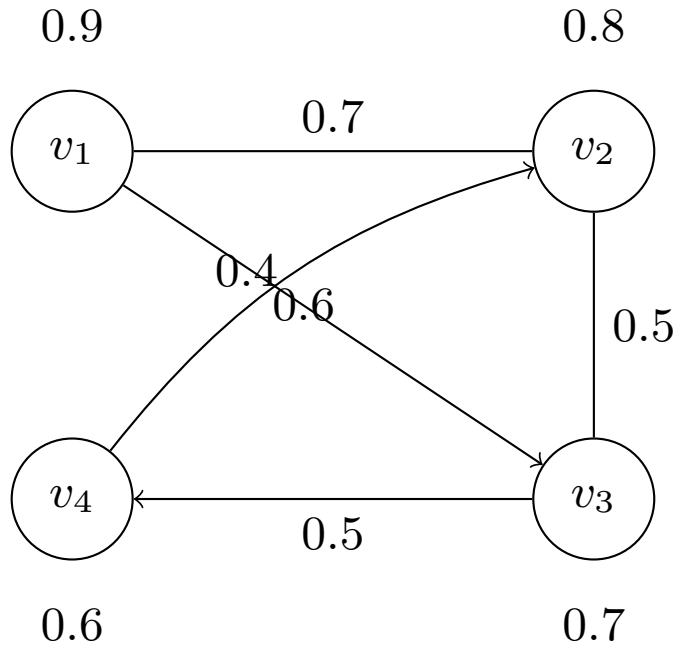

Figure 4.3: A fuzzy mixed graph

An uncertain mixed graph combines uncertain undirected edges and uncertain directed edges in a single structure, allowing symmetric and asymmetric uncertain relations among vertices simultaneously.

**Definition 4.4.3** (Uncertain Mixed Graph)**.** Let

$$M^* = (V, E, A)$$

be a finite mixed graph, where

$$E \subseteq \big\{\{u, v\} \mid u, v \in V,\ u \neq v\big\}$$

is the set of undirected edges and

$$A \subseteq \big\{(u, v) \in V \times V \mid u \neq v\big\}$$

is the set of directed edges (arcs).

Let $M$ be a fixed uncertain model with degree-domain

$$\mathrm{Dom}(M) \subseteq [0, 1]^k.$$

An *Uncertain Mixed Graph of type* $M$ is a sextuple

$$\mathcal{M}_M = (V, E, A, \sigma_M, \eta_M, \alpha_M),$$

where

$$\sigma_M : V \longrightarrow \mathrm{Dom}(M), \qquad \eta_M : E \longrightarrow \mathrm{Dom}(M), \qquad \alpha_M : A \longrightarrow \mathrm{Dom}(M)$$

are uncertainty-degree functions on the vertex set, the undirected-edge set, and the directed-edge set, respectively.

Equivalently,

$$(V, \sigma_M), \qquad (E, \eta_M), \qquad (A, \alpha_M)$$

are Uncertain Sets of type $M$ on $V$, $E$, and $A$, respectively.

For each vertex $v \in V$, the value

$$\sigma_M(v) \in \mathrm{Dom}(M)$$

represents the uncertainty degree of $v$. For each undirected edge

$$e = \{u, v\} \in E,$$

the value

$$\eta_M(e) \in \mathrm{Dom}(M)$$

represents the uncertainty degree of $e$. For each directed edge

$$a = (u, v) \in A,$$

the value

$$\alpha_M(a) \in \mathrm{Dom}(M)$$

represents the uncertainty degree of the arc from $u$ to $v$.

If desired, one may additionally impose model-specific compatibility conditions involving

$$\eta_M(\{u,v\}), \qquad \alpha_M((u,v)), \qquad \sigma_M(u), \qquad \sigma_M(v),$$

but such conditions depend on the chosen uncertain model $M$ and are not fixed at the level of this general definition.

**Theorem 4.4.4** (Well-definedness of Uncertain Mixed Graph)**.** *Let*

$$M^* = (V, E, A)$$

*be a finite mixed graph, where*

$$E \subseteq \big\{\{u,v\} \mid u,v \in V,\ u \neq v\big\}$$

*and*

$$A \subseteq \big\{(u,v) \in V \times V \mid u \neq v\big\}.$$

*Let $M$ be an uncertain model with degree-domain*

$$\mathrm{Dom}(M) \subseteq [0,1]^k,$$

*and let*

$$\sigma_M : V \to \mathrm{Dom}(M), \qquad \eta_M : E \to \mathrm{Dom}(M), \qquad \alpha_M : A \to \mathrm{Dom}(M)$$

*be functions.*

*Then the sextuple*

$$\mathcal{M}_M = (V, E, A, \sigma_M, \eta_M, \alpha_M)$$

*is a well-defined Uncertain Mixed Graph of type $M$.*

*Moreover,*

$$(V, \sigma_M), \qquad (E, \eta_M), \qquad (A, \alpha_M)$$

*are Uncertain Sets of type $M$ on $V$, $E$, and $A$, respectively.*

*Proof.* Since $M$ is an uncertain model, its degree-domain

$$\mathrm{Dom}(M)$$

is a fixed admissible set of uncertainty degrees. Hence the maps

$$\sigma_M : V \to \mathrm{Dom}(M), \qquad \eta_M : E \to \mathrm{Dom}(M), \qquad \alpha_M : A \to \mathrm{Dom}(M)$$

are ordinary set-theoretic functions with well-specified codomain $\mathrm{Dom}(M)$.

Therefore the pairs

$$(V, \sigma_M), \qquad (E, \eta_M), \qquad (A, \alpha_M)$$

are Uncertain Sets of type $M$ on $V$, $E$, and $A$, respectively.

Next, every element of $E$ is an unordered two-element subset

$$e = \{u,v\} \qquad (u \neq v),$$

so each undirected edge has uniquely determined endpoints $u$ and $v$, independent of order. Likewise, every element of $A$ is an ordered pair

$$a = (u,v) \qquad (u \neq v),$$

so each arc has uniquely determined source $u$ and target $v$.

Moreover, an undirected edge

$$\{u,v\} \in E$$

and a directed edge

$$(u,v) \in A$$

are objects of different types, belonging to different domains $E$ and $A$. Hence even if they involve the same vertices, there is no ambiguity between their uncertainty-degree assignments

$$\eta_M(\{u,v\}) \quad \text{and} \quad \alpha_M((u,v)).$$

Thus all components of

$$\mathcal{M}_M = (V, E, A, \sigma_M, \eta_M, \alpha_M)$$

are simultaneously and uniquely specified:

- $V$ is the vertex set,
- $E$ is the undirected-edge set,
- $A$ is the directed-edge set,
- $\sigma_M$ assigns a unique uncertainty degree to each vertex,
- $\eta_M$ assigns a unique uncertainty degree to each undirected edge,
- $\alpha_M$ assigns a unique uncertainty degree to each directed edge.

Hence

$$\mathcal{M}_M = (V, E, A, \sigma_M, \eta_M, \alpha_M)$$

determines a unique mathematical object, namely an Uncertain Mixed Graph of type $M$. Therefore the definition is well-defined. □

Related mixed graph concepts under fuzzy and uncertainty-aware frameworks are listed in Table 4.2.

Table 4.2: Related mixed graph concepts under fuzzy and uncertainty-aware frameworks

| **Concept** | **Reference(s)** |
|---|---|
| Fuzzy Mixed Graph | — |
| Intuitionistic Fuzzy Mixed Graph | [88] |
| Neutrosophic Mixed Graph | [88] |
| Picture Fuzzy Mixed Graph | — |
| Plithogenic Mixed Graph | [88] |

## 4.5 Uncertain Regular Graph

A regular fuzzy graph is a fuzzy graph in which every vertex has the same degree, yielding uniform membership-weighted structure [244–247].

**Definition 4.5.1** (Regular Fuzzy Graph)**.** [244, 245] Let

$$G = (V, \sigma, \mu)$$

be a finite fuzzy graph, where

$$\sigma : V \to [0,1], \qquad \mu : V \times V \to [0,1], \qquad \mu(u,v) \le \min\{\sigma(u), \sigma(v)\} \quad (\forall\, u, v \in V),$$

and assume that $\mu$ is symmetric and $G$ has no loops.

The degree of a vertex $v \in V$ is defined by

$$d_G(v) := \sum_{\substack{u \in V \\ u \neq v}} \mu(v,u).$$

The fuzzy graph $G$ is called *regular* if there exists a constant

$$r \ge 0$$

such that

$$d_G(v) = r \qquad \text{for all } v \in V.$$

In this case, $G$ is also called an *$r$-regular fuzzy graph.*

The following is defined for the extensions based on Uncertain Sets.

**Definition 4.5.2** (Degree-Evaluable Uncertain Model)**.** Let $M$ be an uncertain model with degree-domain

$$\mathrm{Dom}(M) \subseteq [0,1]^k.$$

We say that $M$ is *degree-evaluable* if it is equipped with a map

$$\Delta_M : \mathrm{Dom}(M) \longrightarrow [0, \infty),$$

called the *degree-evaluation map.*

**Definition 4.5.3** (Regular Uncertain Graph)**.** Let

$$G^* = (V, E)$$

be a finite undirected loopless graph, and let $M$ be a degree-evaluable uncertain model with degree-domain $\mathrm{Dom}(M)$ and degree-evaluation map

$$\Delta_M : \mathrm{Dom}(M) \to [0, \infty).$$

An *Uncertain Graph of type $M$* on $G^*$ is a quadruple

$$\mathcal{G}_M = (V, E, \sigma_M, \eta_M),$$

where

$$\sigma_M : V \to \mathrm{Dom}(M), \qquad \eta_M : E \to \mathrm{Dom}(M)$$

are uncertainty-degree functions on the vertex set and edge set, respectively.

For a vertex $v \in V$, define its *degree* in $\mathcal{G}_M$ by

$$d_{\mathcal{G}_M}(v) := \sum_{\substack{e \in E \\ v \in e}} \Delta_M\big(\eta_M(e)\big).$$

The uncertain graph $\mathcal{G}_M$ is called *regular* if there exists a constant

$$r \in [0, \infty)$$

such that

$$d_{\mathcal{G}_M}(v) = r \qquad \text{for all } v \in V.$$

In this case, $\mathcal{G}_M$ is called an *$r$-regular uncertain graph.*

**Theorem 4.5.4** (Well-definedness of Regular Uncertain Graph)**.** *Let*

$$G^* = (V, E)$$

*be a finite undirected loopless graph, let $M$ be a degree-evaluable uncertain model with degree-domain* $\mathrm{Dom}(M) \subseteq [0,1]^k$*, and let*

$$\mathcal{G}_M = (V, E, \sigma_M, \eta_M)$$

*be an uncertain graph of type $M$.*

*Then, for every vertex $v \in V$, the quantity*

$$d_{\mathcal{G}_M}(v) = \sum_{\substack{e \in E \\ v \in e}} \Delta_M\big(\eta_M(e)\big)$$

*is a well-defined element of* $[0, \infty)$*.*

*Consequently, the statement*

$$\exists\, r \in [0,\infty) \text{ such that } d_{\mathcal{G}_M}(v) = r \text{ for all } v \in V$$

*is well-defined. Hence the notion of a regular uncertain graph is well-defined.*

*Moreover, if $\mathcal{G}_M$ is regular, then the constant $r$ is unique.*

*Proof.* Fix a vertex $v \in V$. Since $G^* = (V, E)$ is finite, the set

$$E(v) := \{\, e \in E : v \in e \,\}$$

is finite.

For each edge $e \in E(v)$, we have

$$\eta_M(e) \in \mathrm{Dom}(M),$$

because $\eta_M : E \to \mathrm{Dom}(M)$ is a function. Since $M$ is degree-evaluable,

$$\Delta_M\big(\eta_M(e)\big) \in [0, \infty).$$

Therefore every summand in

$$\sum_{e \in E(v)} \Delta_M\big(\eta_M(e)\big)$$

is a well-defined nonnegative real number.

Because $E(v)$ is finite, this is a finite sum of real numbers, and hence it exists and is uniquely determined. Also, since addition in $[0, \infty) \subseteq \mathbb{R}$ is associative and commutative, the value of the sum does not depend on the order in which the incident edges are listed. Thus

$$d_{\mathcal{G}_M}(v) \in [0, \infty)$$

is well-defined.

Since this holds for every $v \in V$, the predicate

$$d_{\mathcal{G}_M}(v) = r \qquad (\forall\, v \in V)$$

is a meaningful statement for any $r \in [0, \infty)$. Therefore the regularity condition

$$\exists\, r \in [0,\infty) \text{ such that } d_{\mathcal{G}_M}(v) = r \text{ for all } v \in V$$

is well-defined, and so the notion of a regular uncertain graph is well-defined.

Finally, suppose that $\mathcal{G}_M$ is regular and that both $r, s \in [0, \infty)$ satisfy

$$d_{\mathcal{G}_M}(v) = r \qquad \text{and} \qquad d_{\mathcal{G}_M}(v) = s \qquad (\forall\, v \in V).$$

Since $V \neq \varnothing$, choose any $v_0 \in V$. Then

$$r = d_{\mathcal{G}_M}(v_0) = s.$$

Hence the regularity constant is unique. □

Representative regular-graph concepts under uncertainty-aware graph frameworks are listed in Table 4.3.

Table 4.3: Representative regular-graph concepts under uncertainty-aware graph frameworks, classified by the dimension $k$ of the information attached to vertices and/or edges.

| $k$ | **Regular-graph concept** | **Typical coordinate form** | **Canonical information attached to vertices/edges** |
|---|---|---|---|
| 1 | Regular Fuzzy Graph | $\mu$ | A regular graph studied in a fuzzy framework, where each vertex and edge is associated with a single membership degree in $[0, 1]$. |
| 2 | Regular Intuitionistic Fuzzy Graph [248–250] | $(\mu, \nu)$ | A regular graph defined in an intuitionistic fuzzy framework, where each vertex and edge carries a membership degree and a non-membership degree, usually satisfying $\mu + \nu \leq 1$. |
| 2 | Regular Bipolar Fuzzy Graph [251–253] | $(\mu^+, \mu^-)$ | A regular graph defined in a bipolar fuzzy framework, where each vertex and edge is described by a positive membership degree and a negative membership degree. |
| 3 | Regular Picture Fuzzy Graph [254, 255] | $(\mu, \eta, \nu)$ | A regular graph defined in a picture fuzzy framework, where each vertex and edge is described by positive, neutral, and negative membership degrees, usually satisfying $\mu + \eta + \nu \leq 1$. |
| 3 | Regular Spherical Fuzzy Graph [256–258] | $(\mu, \eta, \nu)$ | A regular graph defined in a spherical fuzzy framework, where each vertex and edge is described by positive, neutral, and negative membership degrees, usually satisfying $\mu^2 + \eta^2 + \nu^2 \leq 1$. |
| 3 | Regular Neutrosophic Graph [259–262] | $(T, I, F)$ | A regular graph defined in a neutrosophic framework, where each vertex and edge is described by truth, indeterminacy, and falsity degrees. |

Other related concepts include the strongly regular graph [263–265], amply regular graph [266, 267], distance-regular graph [268, 269], distance-transitive graph [270, 271], walk-regular graph [272, 273], and regular hypergraph [274, 275].

## 4.6 Uncertain Intersection Graph

A fuzzy intersection graph represents fuzzy sets as vertices, with edge memberships determined by the strength or height of pairwise fuzzy intersections [276–278].

**Definition 4.6.1** (Fuzzy Intersection Graph)**.** Let

$$\mathcal{F} = \{A_1, A_2, \dots, A_n\}$$

be a finite family of fuzzy sets on a nonempty set $X$, where each

$$A_i : X \to [0, 1].$$

For a fuzzy set $A$ on $X$, define its height by

$$h(A) := \sup_{x \in X} A(x).$$

For two fuzzy sets $A, B$ on $X$, define their pointwise intersection by

$$(A \wedge B)(x) := \min\{A(x), B(x)\} \qquad (x \in X).$$

The *fuzzy intersection graph* of $\mathcal{F}$ is the fuzzy graph

$$\mathrm{Int}(\mathcal{F}) = (V, \sigma, \mu),$$

where

$$V = \{v_1, v_2, \ldots, v_n\}$$

is a crisp vertex set in one-to-one correspondence with the family $\mathcal{F}$, and

$$\sigma(v_i) := h(A_i) \qquad (1 \le i \le n),$$

$$\mu(v_i, v_j) := \begin{cases} h(A_i \wedge A_j), & i \neq j, \\ 0, & i = j. \end{cases}$$

Equivalently,

$$\mu(v_i, v_j) = \begin{cases} \sup_{x \in X} \min\{A_i(x), A_j(x)\}, & i \neq j, \\ 0, & i = j. \end{cases}$$

Thus, two distinct vertices are adjacent with positive membership if and only if the corresponding fuzzy sets have nonzero fuzzy intersection.

An uncertain intersection graph represents uncertain sets as vertices, and assigns to each vertex and edge an uncertainty degree obtained from model-dependent intersection and height operators.

**Definition 4.6.2** (Intersection-Evaluable Uncertain Model)**.** Let $M$ be an uncertain model with degree-domain

$$\mathrm{Dom}(M) \subseteq [0,1]^k.$$

We say that $M$ is *intersection-evaluable* if, for every nonempty universe $X$, the following two operations are specified:

1. a binary operation
$$\wedge_M : \mathrm{Dom}(M) \times \mathrm{Dom}(M) \longrightarrow \mathrm{Dom}(M),$$
called the *model intersection operator*;
2. a map
$$h_M : \{\, \mu : X \to \mathrm{Dom}(M) \,\} \longrightarrow \mathrm{Dom}(M),$$
called the *model height operator*.

For two Uncertain Sets

$$\mathcal{U} = (X, \mu), \qquad \mathcal{W} = (X, \nu)$$

of type $M$ on the same universe $X$, their *pointwise intersection* is the uncertain set

$$\mathcal{U} \wedge_M \mathcal{W} := \big(X, \mu \wedge_M \nu\big),$$

where

$$(\mu \wedge_M \nu)(x) := \mu(x) \wedge_M \nu(x) \qquad (\forall x \in X).$$

**Theorem 4.6.3** (Well-definedness of Uncertain Intersection Graph)**.** *Let $X$ be a nonempty set, let $M$ be an intersection-evaluable uncertain model, and let*

$$\mathcal{F} = \{\mathcal{U}_1, \mathcal{U}_2, \ldots, \mathcal{U}_n\}$$

*be a finite family of Uncertain Sets of type $M$ on $X$, where*

$$\mathcal{U}_i = (X, \mu_i), \qquad \mu_i : X \to \mathrm{Dom}(M).$$

*Then the object*

$$\mathrm{UInt}_M(\mathcal{F}) = (V, E, \sigma_M, \eta_M)$$

*defined above is a well-defined Uncertain Graph of type $M$.*

*Moreover, if*

$$V = \{v_1, \ldots, v_n\},$$

*then*

$$(V, \sigma_M)$$

*is an Uncertain Set of type $M$ on $V$, and*

$$(E, \eta_M)$$

*is an Uncertain Set of type $M$ on $E$.*

*Proof.* Since each $\mathcal{U}_i = (X, \mu_i)$ is an Uncertain Set of type $M$, the map

$$\mu_i : X \to \mathrm{Dom}(M)$$

is well-defined for every $i = 1, \ldots, n$.

Because $M$ is intersection-evaluable, the model intersection operator

$$\wedge_M : \mathrm{Dom}(M) \times \mathrm{Dom}(M) \to \mathrm{Dom}(M)$$

is well-defined. Hence for every pair $i, j$, the pointwise formula

$$(\mu_i \wedge_M \mu_j)(x) := \mu_i(x) \wedge_M \mu_j(x) \qquad (\forall x \in X)$$

defines a unique map

$$\mu_i \wedge_M \mu_j : X \to \mathrm{Dom}(M).$$

Therefore

$$\mathcal{U}_i \wedge_M \mathcal{U}_j$$

is a well-defined Uncertain Set of type $M$ on $X$.

Again, since $M$ is intersection-evaluable, the height operator

$$h_M : \{\, \mu : X \to \mathrm{Dom}(M) \,\} \to \mathrm{Dom}(M)$$

is well-defined. Consequently,

$$h_M(\mu_i) \in \mathrm{Dom}(M) \qquad \text{and} \qquad h_M(\mu_i \wedge_M \mu_j) \in \mathrm{Dom}(M)$$

for all admissible $i, j$.

Now define

$$V = \{v_1, \ldots, v_n\}$$

and

$$E = \big\{\{v_i, v_j\} \mid 1 \le i < j \le n\big\}.$$

Since $\mathcal{F}$ is finite, both $V$ and $E$ are finite and uniquely determined.

Define

$$\sigma_M(v_i) := h_M(\mu_i) \qquad (1 \leq i \leq n).$$

This gives a unique map

$$\sigma_M : V \to \mathrm{Dom}(M).$$

Likewise, define

$$\eta_M(\{v_i, v_j\}) := h_M(\mu_i \wedge_M \mu_j) \qquad (1 \leq i < j \leq n).$$

Since each unordered pair $\{v_i, v_j\} \in E$ corresponds uniquely to the pair of uncertain sets $\mathcal{U}_i, \mathcal{U}_j$, this gives a unique map

$$\eta_M : E \to \mathrm{Dom}(M).$$

Therefore

$$(V, \sigma_M)$$

is an Uncertain Set of type $M$ on $V$, and

$$(E, \eta_M)$$

is an Uncertain Set of type $M$ on $E$. Hence

$$\mathrm{UInt}_M(\mathcal{F}) = (V, E, \sigma_M, \eta_M)$$

is a well-defined Uncertain Graph of type $M$.

Thus the definition is well-defined. □

Representative intersection-graph concepts under uncertainty-aware graph frameworks are listed in Table 4.4.

Table 4.4: Representative intersection-graph concepts under uncertainty-aware graph frameworks, classified by the dimension $k$ of the information attached to vertices and/or edges.

| $k$ | **Intersection-graph concept** | **Typical coordinate form** | **Canonical information attached to vertices/edges** |
|---|---|---|---|
| 1 | Fuzzy Intersection Graph | $\mu$ | An intersection graph studied in a fuzzy framework, where each vertex and edge is associated with a single membership degree in $[0, 1]$. |
| 2 | Intuitionistic Fuzzy Intersection Graph | $(\mu, \nu)$ | An intersection graph defined in an intuitionistic fuzzy framework, where each vertex and edge carries a membership degree and a non-membership degree, usually satisfying $\mu + \nu \leq 1$. |
| 3 | Neutrosophic Intersection Graph [279] | $(T, I, F)$ | An intersection graph defined in a neutrosophic framework, where each vertex and edge is described by truth, indeterminacy, and falsity degrees. |

Moreover, a wide variety of concepts are known as classical intersection graphs. For example, these include interval graphs [280], proper interval graphs [281, 282], circular-arc graphs [283, 284], permutation graphs [285, 286], trapezoid graphs [287, 288], unit disk graphs [289, 290], intersection hypergraphs [291, 292], and string graphs [293, 294]. More broadly, intersection graphs give rise to a very large number of graph classes depending on what family of objects is allowed to intersect.

## 4.7 Uncertain Labeling Graph

A fuzzy labeling graph assigns distinct membership values to all vertices and edges, with each edge valued below both incident vertices, ensuring a consistent labeling [295–298].

**Definition 4.7.1** (Fuzzy Labeling Graph)**.** [299–301] Let

$$G^* = (V, E)$$

be a finite simple graph.

A *fuzzy labeling* of $G^*$ is a pair of maps

$$\sigma : V \to [0,1], \qquad \mu : E \to [0,1],$$

satisfying the following conditions:

(L1) $\sigma$ is injective on $V$;

(L2) $\mu$ is injective on $E$;

(L3) the labels of vertices and edges are mutually distinct, that is,

$$\sigma(u) \neq \mu(e) \qquad \text{for all } u \in V,\ e \in E;$$

(L4) for every edge $e = uv \in E$,

$$\mu(uv) < \min\{\sigma(u), \sigma(v)\}.$$

The fuzzy graph

$$G = (\sigma, \mu)$$

induced by such a labeling is called a *fuzzy labeling graph*.

An uncertain labeling graph assigns pairwise distinct uncertainty labels to vertices and edges, together with a model-dependent strict comparison ensuring that each edge label is strictly below the labels of its incident vertices.

**Definition 4.7.2** (Label-Comparable Uncertain Model)**.** Let $M$ be an uncertain model with degree-domain

$$\mathrm{Dom}(M) \subseteq [0,1]^k.$$

We say that $M$ is *label-comparable* if it is equipped with a strict binary relation

$$\prec_M \subseteq \mathrm{Dom}(M) \times \mathrm{Dom}(M),$$

called the *strict label order* of $M$.

For $a, b \in \mathrm{Dom}(M)$, the expression

$$a \prec_M b$$

means that the label $a$ is strictly smaller than the label $b$ in the model $M$.

**Definition 4.7.3** (Uncertain Labeling Graph)**.** Let

$$G^* = (V, E)$$

be a finite simple graph, and let $M$ be a label-comparable uncertain model with degree-domain

$$\mathrm{Dom}(M) \subseteq [0,1]^k$$

and strict label order

$$\prec_M .$$

An *Uncertain Labeling* of $G^*$ of type $M$ is a pair of maps

$$\sigma_M : V \to \mathrm{Dom}(M), \qquad \eta_M : E \to \mathrm{Dom}(M),$$

satisfying the following conditions:

(UL1) $\sigma_M$ is injective on $V$;

(UL2) $\eta_M$ is injective on $E$;

(UL3) the vertex labels and edge labels are mutually distinct, that is,

$$\sigma_M(u) \neq \eta_M(e) \qquad \text{for all } u \in V,\ e \in E;$$

(UL4) for every edge

$$e = \{u, v\} \in E,$$

the edge label is strictly below both incident vertex labels:

$$\eta_M(e) \prec_M \sigma_M(u) \qquad \text{and} \qquad \eta_M(e) \prec_M \sigma_M(v).$$

The pair

$$\mathcal{G}_M = (G^*, \sigma_M, \eta_M)$$

induced by such a labeling is called an *Uncertain Labeling Graph of type $M$*.

**Theorem 4.7.4** (Well-definedness of Uncertain Labeling Graph)**.** *Let*

$$G^* = (V, E)$$

*be a finite simple graph, let $M$ be a label-comparable uncertain model with degree-domain*

$$\mathrm{Dom}(M) \subseteq [0, 1]^k$$

*and strict label order $\prec_M$, and let*

$$\sigma_M : V \to \mathrm{Dom}(M), \qquad \eta_M : E \to \mathrm{Dom}(M)$$

*be maps satisfying* (UL1)–(UL4)*.*

*Then*

$$\mathcal{G}_M = (G^*, \sigma_M, \eta_M)$$

*is a well-defined Uncertain Labeling Graph of type $M$.*

*Moreover,*

$$(V, \sigma_M)$$

*is an Uncertain Set of type $M$ on $V$, and*

$$(E, \eta_M)$$

*is an Uncertain Set of type $M$ on $E$.*

*Proof.* Since $M$ is an uncertain model, its degree-domain

$$\mathrm{Dom}(M)$$

is fixed. Hence the maps

$$\sigma_M : V \to \mathrm{Dom}(M) \qquad \text{and} \qquad \eta_M : E \to \mathrm{Dom}(M)$$

are ordinary set-theoretic functions with codomain $\mathrm{Dom}(M)$. Therefore the pairs

$$(V, \sigma_M) \quad \text{and} \quad (E, \eta_M)$$

are Uncertain Sets of type $M$ on $V$ and $E$, respectively.

Next, because

$$G^* = (V, E)$$

is a finite simple graph, every edge $e \in E$ is a uniquely determined unordered pair

$$e = \{u, v\} \qquad (u \neq v),$$

so its incident vertices are uniquely specified.

Condition (UL1) ensures that distinct vertices receive distinct uncertainty labels. Condition (UL2) ensures that distinct edges receive distinct uncertainty labels. Condition (UL3) guarantees that no vertex label coincides with any edge label. Hence every element of

$$V \cup E$$

receives a unique label in $\mathrm{Dom}(M)$, and no ambiguity can arise between labels assigned to vertices and labels assigned to edges.

Because $M$ is label-comparable, the strict relation

$$\prec_M \subseteq \mathrm{Dom}(M) \times \mathrm{Dom}(M)$$

is already specified on the whole degree-domain. Therefore, for each edge

$$e = \{u, v\} \in E,$$

the conditions

$$\eta_M(e) \prec_M \sigma_M(u) \qquad \text{and} \qquad \eta_M(e) \prec_M \sigma_M(v)$$

are meaningful statements in the model $M$. Thus condition (UL4) is well-formed.

Consequently, all data entering

$$\mathcal{G}_M = (G^*, \sigma_M, \eta_M)$$

are simultaneously and uniquely specified:

- $G^* = (V, E)$ is the underlying simple graph,
- $\sigma_M$ assigns a unique uncertainty label to each vertex,
- $\eta_M$ assigns a unique uncertainty label to each edge,
- the comparison rule $\prec_M$ determines the strict labeling constraints.

Hence

$$\mathcal{G}_M = (G^*, \sigma_M, \eta_M)$$

defines a unique mathematical object, namely an Uncertain Labeling Graph of type $M$. Therefore the definition is well-defined. □

Representative labeling-graph concepts under uncertainty-aware graph frameworks are listed in Table 4.5.

Besides uncertain labeling graphs, several related concepts are also known, including graceful labeling graphs [311–313], harmonious labeling graphs [314,315], cordial labeling graphs [316,317], magic labeling graphs [307,318,319], labeling hypergraphs [320, 321], cordial labeling graphs [322, 323], antimagic labeling graphs [319, 324, 325], radio labeling graphs [326,327], set-graceful labeling graphs [328], and lucky labeling graphs [329–331].

Table 4.5: Representative labeling-graph concepts under uncertainty-aware graph frameworks, classified by the dimension $k$ of the information attached to vertices and/or edges.

| $k$ | **Labeling-graph concept** | **Typical coordinate form** | **Canonical information attached to vertices/edges** |
|---|---|---|---|
| 1 | Fuzzy Labeling Graph | $\mu$ | A labeling graph studied in a fuzzy framework, where each vertex and edge is associated with a single membership degree in $[0,1]$. |
| 2 | Intuitionistic Fuzzy Labeling Graph [302, 303] | $(\mu,\nu)$ | A labeling graph defined in an intuitionistic fuzzy framework, where each vertex and edge carries a membership degree and a non-membership degree, usually satisfying $\mu+\nu \leq 1$. |
| 3 | Picture Fuzzy Labeling Graph [304, 305] | $(\mu,\eta,\nu)$ | A labeling graph defined in a picture fuzzy framework, where each vertex and edge is described by positive, neutral, and negative membership degrees, usually satisfying $\mu+\eta+\nu \leq 1$. |
| 3 | Spherical Fuzzy Labeling Graph [306] | $(\mu,\eta,\nu)$ | A labeling graph defined in a spherical fuzzy framework, where each vertex and edge is described by positive, neutral, and negative membership degrees, usually satisfying $\mu^2+\eta^2+\nu^2 \leq 1$. |
| 3 | Neutrosophic Labeling Graph [307–310] | $(T,I,F)$ | A labeling graph defined in a neutrosophic framework, where each vertex and edge is described by truth, indeterminacy, and falsity degrees. |

## 4.8 Complete Uncertain Graph

A complete fuzzy graph is a fuzzy graph where every edge attains the maximum possible membership allowed by the memberships of its incident vertices [332–334].

**Definition 4.8.1** (Complete Fuzzy Graph)**.** [334] Let

$$G = (V, \sigma, \mu)$$

be a finite fuzzy graph, where

$$\sigma : V \to [0,1], \qquad \mu : V \times V \to [0,1],$$

with

$$\mu(u,v) \leq \min\{\sigma(u), \sigma(v)\} \qquad (\forall\, u, v \in V),$$

and assume that $\mu$ is symmetric.

The fuzzy graph $G$ is called a *complete fuzzy graph* if

$$\mu(u,v) = \min\{\sigma(u), \sigma(v)\} \qquad \text{for all } u, v \in V.$$

**Example 4.8.2** (Complete fuzzy graph)**.** Let

$$V = \{v_1, v_2, v_3\}.$$

Define a fuzzy graph

$$G = (V, \sigma, \mu)$$

by the vertex-membership function

$$\sigma(v_1) = 0.9, \qquad \sigma(v_2) = 0.7, \qquad \sigma(v_3) = 0.5,$$

and define the edge-membership function

$$\mu : V \times V \to [0,1]$$

by

$$\mu(u,v) = \min\{\sigma(u), \sigma(v)\} \qquad (\forall\, u, v \in V).$$

Explicitly, this gives

$$\mu(v_1,v_1) = 0.9, \qquad \mu(v_2,v_2) = 0.7, \qquad \mu(v_3,v_3) = 0.5,$$
$$\mu(v_1,v_2) = \mu(v_2,v_1) = 0.7,$$
$$\mu(v_1,v_3) = \mu(v_3,v_1) = 0.5,$$
$$\mu(v_2,v_3) = \mu(v_3,v_2) = 0.5.$$

Then $G$ is a complete fuzzy graph, because for every $u, v \in V$,

$$\mu(u,v) = \min\{\sigma(u), \sigma(v)\}.$$

Indeed,

$$\mu(v_1,v_2) = 0.7 = \min\{0.9, 0.7\},$$
$$\mu(v_1,v_3) = 0.5 = \min\{0.9, 0.5\},$$
$$\mu(v_2,v_3) = 0.5 = \min\{0.7, 0.5\},$$

and similarly

$$\mu(v_i,v_i) = \sigma(v_i) = \min\{\sigma(v_i), \sigma(v_i)\} \qquad (i = 1, 2, 3).$$

Hence $G$ satisfies the definition of a complete fuzzy graph.

A schematic illustration is shown in Figure 4.4. For visual clarity, only the edges between distinct vertices are drawn; the diagonal values

$$\mu(v_i,v_i) = \sigma(v_i)$$

are not shown in the figure.

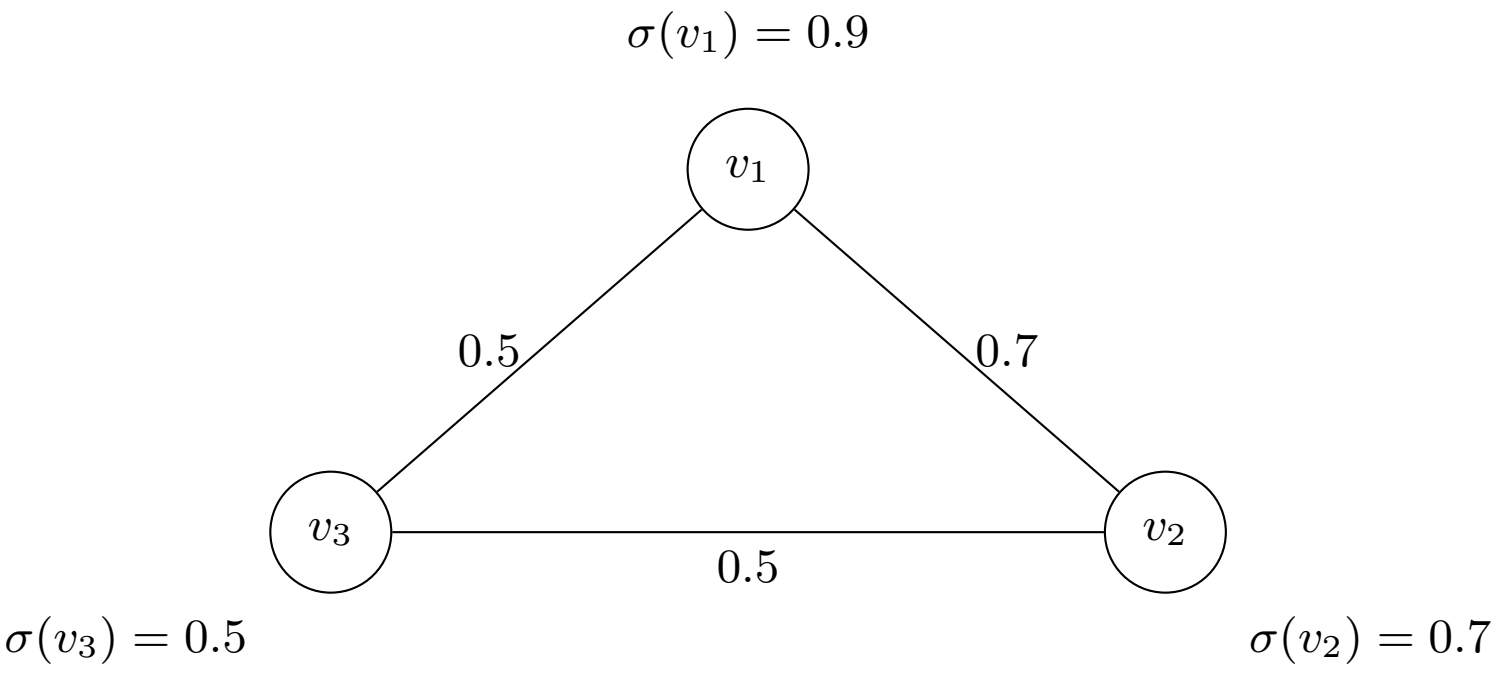

Figure 4.4: A complete fuzzy graph on three vertices

The extensions based on Uncertain Sets are described below.

**Definition 4.8.3** (Complete-Edge-Evaluable Uncertain Model)**.** Let $M$ be an uncertain model with degree-domain

$$\mathrm{Dom}(M) \subseteq [0,1]^k.$$

We say that $M$ is *complete-edge-evaluable* if it is equipped with a symmetric map

$$\Gamma_M : \mathrm{Dom}(M) \times \mathrm{Dom}(M) \longrightarrow \mathrm{Dom}(M),$$

called the *complete-edge operator*, such that

$$\Gamma_M(a,b) = \Gamma_M(b,a) \qquad (\forall\, a, b \in \mathrm{Dom}(M)).$$

For two vertex-degrees $a, b \in \mathrm{Dom}(M)$, the value

$$\Gamma_M(a,b)$$

is interpreted as the model-dependent edge degree assigned in the complete uncertain graph generated by the two endpoint degrees.

**Definition 4.8.4** (Complete-Edge-Evaluable Uncertain Model)**.** Let $M$ be an uncertain model with degree-domain

$$\mathrm{Dom}(M) \subseteq [0,1]^k.$$

We say that $M$ is *complete-edge-evaluable* if it is equipped with a symmetric map

$$\Gamma_M : \mathrm{Dom}(M) \times \mathrm{Dom}(M) \longrightarrow \mathrm{Dom}(M),$$

called the *complete-edge operator*, such that

$$\Gamma_M(a,b) = \Gamma_M(b,a) \qquad (\forall\, a,b \in \mathrm{Dom}(M)).$$

For two vertex-degrees $a, b \in \mathrm{Dom}(M)$, the value

$$\Gamma_M(a,b)$$

is interpreted as the model-dependent edge degree assigned in the complete uncertain graph generated by the two endpoint degrees.

**Theorem 4.8.5** (Well-definedness of Complete Uncertain Graph)**.** *Let $M$ be a complete-edge-evaluable uncertain model with degree-domain*

$$\mathrm{Dom}(M) \subseteq [0,1]^k$$

*and symmetric complete-edge operator*

$$\Gamma_M : \mathrm{Dom}(M) \times \mathrm{Dom}(M) \to \mathrm{Dom}(M).$$

*Let $V$ be a finite nonempty set, and let*

$$\sigma_M : V \to \mathrm{Dom}(M)$$

*be a function. Define*

$$E_V := \big\{\{u,v\} \subseteq V \mid u \neq v\big\},$$

*and define*

$$\eta_M : E_V \to \mathrm{Dom}(M)$$

*by*

$$\eta_M(\{u,v\}) := \Gamma_M\big(\sigma_M(u), \sigma_M(v)\big) \qquad (\forall\, \{u,v\} \in E_V).$$

*Then*

$$\mathcal{K}_M = (V, E_V, \sigma_M, \eta_M)$$

*is a well-defined Complete Uncertain Graph of type $M$.*

*Moreover, for the fixed data $V$, $\sigma_M$, and $\Gamma_M$, the edge uncertainty-degree function $\eta_M$ is uniquely determined.*

*Proof.* Since $M$ is an uncertain model, its degree-domain

$$\mathrm{Dom}(M)$$

is fixed. Hence the map

$$\sigma_M : V \to \mathrm{Dom}(M)$$

is an ordinary set-theoretic function with codomain $\mathrm{Dom}(M)$, so

$$(V, \sigma_M)$$

is an Uncertain Set of type $M$ on $V$.

Next, because $V$ is finite and nonempty, the set

$$E_V = \big\{\{u,v\} \subseteq V \mid u \neq v\big\}$$

is a well-defined finite set. It is precisely the edge set of the complete simple graph on $V$.

Now let

$$e = \{u, v\} \in E_V.$$

Since $e$ is an unordered two-element subset of $V$, the endpoints $u$ and $v$ are uniquely determined up to order. Because $\sigma_M(u), \sigma_M(v) \in \mathrm{Dom}(M)$ and

$$\Gamma_M : \mathrm{Dom}(M) \times \mathrm{Dom}(M) \to \mathrm{Dom}(M)$$

is well-defined, the value

$$\Gamma_M\big(\sigma_M(u), \sigma_M(v)\big)$$

belongs to $\mathrm{Dom}(M)$.

It remains to check that the definition of $\eta_M(e)$ does not depend on the order of the endpoints. Since $\Gamma_M$ is symmetric,

$$\Gamma_M\big(\sigma_M(u), \sigma_M(v)\big) = \Gamma_M\big(\sigma_M(v), \sigma_M(u)\big).$$

Therefore the rule

$$\eta_M(\{u, v\}) := \Gamma_M\big(\sigma_M(u), \sigma_M(v)\big)$$

is independent of the choice of ordering of $u$ and $v$. Hence $\eta_M$ is a well-defined function

$$\eta_M : E_V \to \mathrm{Dom}(M).$$

Consequently,

$$(E_V, \eta_M)$$

is an Uncertain Set of type $M$ on $E_V$, and so

$$\mathcal{K}_M = (V, E_V, \sigma_M, \eta_M)$$

is a well-defined Uncertain Graph of type $M$.

By construction, its underlying crisp graph is complete, and every edge degree is exactly the value prescribed by the complete-edge operator $\Gamma_M$. Therefore

$$\mathcal{K}_M$$

is a well-defined Complete Uncertain Graph of type $M$.

Finally, uniqueness of $\eta_M$ is immediate from the defining formula

$$\eta_M(\{u, v\}) = \Gamma_M\big(\sigma_M(u), \sigma_M(v)\big),$$

which determines the value of $\eta_M$ for every edge $\{u, v\} \in E_V$. Hence $\eta_M$ is uniquely determined by $V$, $\sigma_M$, and $\Gamma_M$. □

Representative complete-graph concepts under uncertainty-aware graph frameworks are listed in Table 4.6.

In addition to the uncertain complete graph, related concepts such as the complete bipartite graph [313,335], complete hypergraph [336,337], complete directed graph [204,338], and quasi-complete graph [339,340] are also known.

Table 4.6: Representative complete-graph concepts under uncertainty-aware graph frameworks, classified by the dimension $k$ of the information attached to vertices and/or edges.

| $k$ | Complete-graph concept | Typical coordinate form | Canonical information attached to vertices/edges |
|---|---|---|---|
| 1 | Complete Fuzzy Graph | $\mu$ | A complete graph studied in a fuzzy framework, where each vertex and edge is associated with a single membership degree in $[0, 1]$. |
| 2 | Complete Intuitionistic Fuzzy Graph | $(\mu, \nu)$ | A complete graph defined in an intuitionistic fuzzy framework, where each vertex and edge carries a membership degree and a non-membership degree, usually satisfying $\mu + \nu \leq 1$. |
| 3 | Complete Neutrosophic Graph [262] | $(T, I, F)$ | A complete graph defined in a neutrosophic framework, where each vertex and edge is described by truth, indeterminacy, and falsity degrees. |

## 4.9 Uncertain Zero-Divisor Graph

Zero-divisor graph represents the nonzero zero divisors of a ring as vertices, joining two distinct vertices whenever their product is zero in the ring structure [341, 342]. Fuzzy zero-divisor graph extends the zero-divisor graph by assigning membership degrees to vertices and edges, modeling uncertain algebraic relations among nonzero zero divisors of rings [343–347].

**Definition 4.9.1** (Fuzzy Zero-Divisor Graph)**.** [343, 344] Let $R$ be a commutative ring with identity, and let

$$Z(R) := \{a \in R : \exists\, b \in R \setminus \{0\} \text{ such that } ab = 0\}$$

be the set of zero divisors of $R$. Set

$$Z(R)^* := Z(R) \setminus \{0\}.$$

A *fuzzy zero-divisor graph* of $R$ is a pair

$$\Gamma_f(R) = (\sigma, \mu)$$

defined on the vertex set

$$V := Z(R)^*,$$

where

$$\sigma : V \to [0, 1]$$

is a vertex-membership function and

$$\mu : V \times V \to [0, 1]$$

is an edge-membership function satisfying the following conditions for all $x, y \in V$:

1. $$\mu(x, y) = \mu(y, x);$$
2. $$\mu(x, y) \leq \min\{\sigma(x), \sigma(y)\};$$
3. if
$$x = y \quad \text{or} \quad xy \neq 0,$$
then
$$\mu(x, y) = 0.$$

Equivalently, positive edge-membership is allowed only between distinct nonzero zero divisors whose product is zero.

If, moreover, one has

$$\mu(x,y) = \begin{cases} \min\{\sigma(x),\sigma(y)\}, & x \neq y \text{ and } xy = 0, \\ 0, & \text{otherwise,} \end{cases}$$

then $\Gamma_f(R)$ is called the *strong fuzzy zero-divisor graph* associated with $\sigma$.

**Example 4.9.2** (Fuzzy zero-divisor graph)**.** Consider the commutative ring

$$R = \mathbb{Z}_6 = \{0,1,2,3,4,5\}$$

with multiplication modulo 6.

First, determine the set of zero divisors of $R$. We have

$$2 \cdot 3 \equiv 0 \pmod 6, \qquad 3 \cdot 4 \equiv 0 \pmod 6,$$

so $2,3,4$ are nonzero zero divisors. Hence

$$Z(R)^* = \{2,3,4\}.$$

Define a vertex-membership function

$$\sigma : Z(R)^* \to [0,1]$$

by

$$\sigma(2) = 0.8, \qquad \sigma(3) = 0.9, \qquad \sigma(4) = 0.7.$$

Next, define the edge-membership function

$$\mu : Z(R)^* \times Z(R)^* \to [0,1]$$

by

$$\mu(2,3) = 0.8, \qquad \mu(3,4) = 0.7, \qquad \mu(2,4) = 0,$$

together with symmetry

$$\mu(x,y) = \mu(y,x) \qquad (\forall\, x,y \in Z(R)^*),$$

and

$$\mu(2,2) = \mu(3,3) = \mu(4,4) = 0.$$

We now verify that

$$\Gamma_f(R) = (\sigma,\mu)$$

is a fuzzy zero-divisor graph.

First, $\mu$ is symmetric by definition.

Second,

$$\mu(2,3) = 0.8 \leq \min\{\sigma(2),\sigma(3)\} = \min\{0.8,0.9\} = 0.8,$$

and

$$\mu(3,4) = 0.7 \leq \min\{\sigma(3),\sigma(4)\} = \min\{0.9,0.7\} = 0.7.$$

Also,

$$\mu(2,4) = 0 \leq \min\{0.8,0.7\} = 0.7.$$

Third, positive edge-membership occurs only between distinct nonzero zero divisors whose product is zero:

$$2 \cdot 3 \equiv 0 \pmod 6, \qquad 3 \cdot 4 \equiv 0 \pmod 6,$$

whereas

$$2 \cdot 4 \equiv 2 \not\equiv 0 \pmod 6.$$

Hence

$$\mu(2,4) = 0,$$

as required. Moreover,

$$\mu(x,x) = 0 \qquad (\forall\, x \in Z(R)^*).$$

Therefore,

$$\Gamma_f(R) = (\sigma, \mu)$$

is a fuzzy zero-divisor graph of $\mathbb{Z}_6$.

In fact, this example is a *strong fuzzy zero-divisor graph*, because

$$\mu(2,3) = \min\{\sigma(2), \sigma(3)\} = 0.8,$$

$$\mu(3,4) = \min\{\sigma(3), \sigma(4)\} = 0.7,$$

and all other pairs have edge-membership 0.

Thus the support graph is the path

$$2 - 3 - 4.$$

The uncertain version replaces these scalar memberships by general uncertainty degrees from a fixed uncertain model.

**Definition 4.9.3** (Uncertain Zero-Divisor Graph)**.** Let $R$ be a commutative ring with identity, and let

$$Z(R) := \{a \in R : \exists\, b \in R \setminus \{0\} \text{ such that } ab = 0\}$$

be the set of zero divisors of $R$. Define

$$V_R := Z(R) \setminus \{0\}.$$

Let

$$E_R := \big\{\{x,y\} \subseteq V_R : x \neq y,\ xy = 0\big\}.$$

Thus

$$\Gamma(R) := (V_R, E_R)$$

is the classical zero-divisor graph of $R$.

Let $M$ be a fixed uncertain model with degree-domain

$$\mathrm{Dom}(M) \subseteq [0,1]^k.$$

An *Uncertain Zero-Divisor Graph of type $M$* associated with $R$ is a quadruple

$$\Gamma_M(R) = (V_R, E_R, \sigma_M, \eta_M),$$

where

$$\sigma_M : V_R \longrightarrow \mathrm{Dom}(M)$$

and

$$\eta_M : E_R \longrightarrow \mathrm{Dom}(M)$$

are uncertainty-degree functions on the vertex set and edge set, respectively.

Equivalently,

$$(V_R, \sigma_M)$$

is an Uncertain Set of type $M$ on $V_R$, and

$$(E_R, \eta_M)$$

is an Uncertain Set of type $M$ on $E_R$.

For each

$$x \in V_R,$$

the value

$$\sigma_M(x) \in \mathrm{Dom}(M)$$

represents the uncertainty degree of the nonzero zero divisor $x$, and for each

$$e = \{x, y\} \in E_R,$$

the value

$$\eta_M(e) \in \mathrm{Dom}(M)$$

represents the uncertainty degree of the annihilating pair $\{x, y\}$.

If desired, one may additionally impose model-specific compatibility conditions between

$$\eta_M(\{x, y\}), \qquad \sigma_M(x), \qquad \sigma_M(y),$$

but such conditions depend on the chosen uncertain model $M$ and are not fixed at the level of this general definition.

**Definition 4.9.4** (Annihilation-Edge-Evaluable Uncertain Model)**.** Let $M$ be an uncertain model with degree-domain

$$\mathrm{Dom}(M) \subseteq [0, 1]^k.$$

We say that $M$ is *annihilation-edge-evaluable* if it is equipped with a symmetric map

$$\Gamma_M : \mathrm{Dom}(M) \times \mathrm{Dom}(M) \longrightarrow \mathrm{Dom}(M),$$

called the *annihilation-edge operator*, such that

$$\Gamma_M(a, b) = \Gamma_M(b, a) \qquad (\forall\, a, b \in \mathrm{Dom}(M)).$$

For two vertex-degrees $a, b \in \mathrm{Dom}(M)$, the value

$$\Gamma_M(a, b)$$

is interpreted as the model-dependent edge degree assigned to an annihilating pair in the strong uncertain zero-divisor graph generated by the two endpoint degrees.

**Definition 4.9.5** (Strong Uncertain Zero-Divisor Graph)**.** Let $R$ be a commutative ring with identity, and let

$$V_R := Z(R) \setminus \{0\}, \qquad E_R := \big\{\{x, y\} \subseteq V_R : x \neq y,\ xy = 0\big\}.$$

Let $M$ be an annihilation-edge-evaluable uncertain model with degree-domain

$$\mathrm{Dom}(M) \subseteq [0, 1]^k$$

and annihilation-edge operator

$$\Gamma_M : \mathrm{Dom}(M) \times \mathrm{Dom}(M) \to \mathrm{Dom}(M).$$

Given a function

$$\sigma_M : V_R \to \mathrm{Dom}(M),$$

define

$$\eta_M : E_R \to \mathrm{Dom}(M)$$

by

$$\eta_M(\{x, y\}) := \Gamma_M\big(\sigma_M(x), \sigma_M(y)\big) \qquad (\forall \{x, y\} \in E_R).$$

Then

$$\Gamma^{\mathrm{s}}_M(R) := (V_R, E_R, \sigma_M, \eta_M)$$

is called the *strong uncertain zero-divisor graph* of type $M$ associated with $\sigma_M$.

**Theorem 4.9.6** (Well-definedness of Uncertain Zero-Divisor Graph)**.** *Let $R$ be a commutative ring with identity, let $M$ be an uncertain model with degree-domain*

$$\mathrm{Dom}(M) \subseteq [0, 1]^k,$$

*and let*

$$\sigma_M : V_R \to \mathrm{Dom}(M), \qquad \eta_M : E_R \to \mathrm{Dom}(M)$$

*be functions, where*

$$V_R := Z(R) \setminus \{0\}, \qquad E_R := \big\{\{x, y\} \subseteq V_R : x \neq y,\ xy = 0\big\}.$$

*Then*

$$\Gamma_M(R) = (V_R, E_R, \sigma_M, \eta_M)$$

*is a well-defined Uncertain Zero-Divisor Graph of type $M$.*

*Moreover,*

$$(V_R, \sigma_M)$$

*and*

$$(E_R, \eta_M)$$

*are well-defined Uncertain Sets of type $M$.*

*Proof.* Since $R$ is a commutative ring with identity, its underlying set, multiplication, and zero element are fixed.

First, the set

$$Z(R) = \{a \in R : \exists\, b \in R \setminus \{0\} \text{ such that } ab = 0\}$$

is well-defined, because for each $a \in R$ the statement

$$\exists\, b \in R \setminus \{0\} \text{ such that } ab = 0$$

has a definite truth value in the ring $R$. Therefore

$$V_R = Z(R) \setminus \{0\}$$

is a well-defined subset of $R$. Its elements are precisely the nonzero zero divisors of $R$.

Next, define

$$E_R := \big\{\{x, y\} \subseteq V_R : x \neq y,\ xy = 0\big\}.$$

For any two elements $x, y \in V_R$ with $x \neq y$, the product $xy$ is uniquely determined in $R$, so the condition

$$xy = 0$$

is meaningful. Since $R$ is commutative,

$$xy = 0 \iff yx = 0,$$

and hence adjacency depends only on the unordered pair $\{x, y\}$, not on the order of the endpoints. Therefore $E_R$ is a well-defined subset of

$$\{\{x, y\} \subseteq V_R : x \neq y\},$$

and

$$\Gamma(R) = (V_R, E_R)$$

is a well-defined simple graph.

Now $M$ is a fixed uncertain model, so its degree-domain

$$\mathrm{Dom}(M)$$

is fixed. Because

$$\sigma_M : V_R \to \mathrm{Dom}(M)$$

is a function, each vertex $x \in V_R$ is assigned a unique uncertainty degree

$$\sigma_M(x) \in \mathrm{Dom}(M).$$

Hence

$$(V_R, \sigma_M)$$

is a well-defined Uncertain Set of type $M$ on $V_R$.

Similarly, because

$$\eta_M : E_R \to \mathrm{Dom}(M)$$

is a function, each edge $e \in E_R$ is assigned a unique uncertainty degree

$$\eta_M(e) \in \mathrm{Dom}(M).$$

Hence

$$(E_R, \eta_M)$$

is a well-defined Uncertain Set of type $M$ on $E_R$.

Consequently, the quadruple

$$\Gamma_M(R) = (V_R, E_R, \sigma_M, \eta_M)$$

is well-defined. By construction, its underlying crisp graph is the zero-divisor graph of $R$, its vertex uncertainty-degree function is $\sigma_M$, and its edge uncertainty-degree function is $\eta_M$. Therefore

$$\Gamma_M(R)$$

is a well-defined Uncertain Zero-Divisor Graph of type $M$. □

**Theorem 4.9.7** (Well-definedness of Strong Uncertain Zero-Divisor Graph)**.** *Let $R$ be a commutative ring with identity, and let*

$$V_R := Z(R) \setminus \{0\}, \qquad E_R := \{\{x, y\} \subseteq V_R : x \neq y,\ xy = 0\}.$$

*Let $M$ be an annihilation-edge-evaluable uncertain model with degree-domain*

$$\mathrm{Dom}(M) \subseteq [0, 1]^k$$

*and symmetric annihilation-edge operator*

$$\Gamma_M : \mathrm{Dom}(M) \times \mathrm{Dom}(M) \to \mathrm{Dom}(M).$$

*If*

$$\sigma_M : V_R \to \mathrm{Dom}(M)$$

*is any function and*

$$\eta_M : E_R \to \mathrm{Dom}(M)$$

*is defined by*

$$\eta_M(\{x, y\}) := \Gamma_M\big(\sigma_M(x), \sigma_M(y)\big) \qquad (\forall\, \{x, y\} \in E_R),$$

*then*

$$\Gamma^{\mathrm{s}}_M(R) = (V_R, E_R, \sigma_M, \eta_M)$$

*is a well-defined Strong Uncertain Zero-Divisor Graph of type $M$.*

*Moreover, for the fixed data*

$$R, \qquad M, \qquad \sigma_M,$$

*the edge uncertainty-degree function $\eta_M$ is uniquely determined.*

*Proof.* By the previous theorem, the sets

$$V_R \qquad \text{and} \qquad E_R$$

are well-defined.

Now let

$$e = \{x, y\} \in E_R.$$

Then $x, y \in V_R$, so

$$\sigma_M(x), \sigma_M(y) \in \mathrm{Dom}(M).$$

Because

$$\Gamma_M : \mathrm{Dom}(M) \times \mathrm{Dom}(M) \to \mathrm{Dom}(M)$$

is well-defined, the value

$$\Gamma_M\big(\sigma_M(x), \sigma_M(y)\big)$$

belongs to $\mathrm{Dom}(M)$.

It remains to show that the definition of $\eta_M(e)$ does not depend on the order of the endpoints. Since $e = \{x, y\}$ is an unordered pair and $\Gamma_M$ is symmetric,

$$\Gamma_M\big(\sigma_M(x), \sigma_M(y)\big) = \Gamma_M\big(\sigma_M(y), \sigma_M(x)\big).$$

Hence the rule

$$\eta_M(\{x, y\}) := \Gamma_M\big(\sigma_M(x), \sigma_M(y)\big)$$

is independent of the choice of ordering of $x$ and $y$. Therefore $\eta_M$ is a well-defined function

$$\eta_M : E_R \to \mathrm{Dom}(M).$$

Consequently,

$$(E_R, \eta_M)$$

is a well-defined Uncertain Set of type $M$, and thus

$$\Gamma^{\mathrm{s}}_M(R) = (V_R, E_R, \sigma_M, \eta_M)$$

is a well-defined Uncertain Zero-Divisor Graph of type $M$. By construction, every annihilating pair $\{x, y\} \in E_R$ receives exactly the degree prescribed by the annihilation-edge operator. Hence it is a well-defined Strong Uncertain Zero-Divisor Graph.

Finally, uniqueness is immediate: for each edge $\{x, y\} \in E_R$, the value

$$\eta_M(\{x, y\}) = \Gamma_M\big(\sigma_M(x), \sigma_M(y)\big)$$

is forced by the given data. Therefore no other edge uncertainty-degree function can satisfy the same defining rule. □

Representative zero-divisor-graph concepts under uncertainty-aware graph frameworks are listed in Table 4.7.

Besides uncertain zero-divisor graphs, several related concepts are also known, including ideal-based zero-divisor graphs [353–355], annihilating-ideal graphs [356–358], compressed zero-divisor graphs [359, 360], and zero-divisor hypergraphs [361, 362].

Table 4.7: Representative zero-divisor-graph concepts under uncertainty-aware graph frameworks, classified by the dimension $k$ of the information attached to vertices and/or edges.

| $k$ | **Zero-divisor-graph concept** | **Typical coordinate form** | **Canonical information attached to vertices/edges** |
|---|---|---|---|
| 1 | Fuzzy Zero-Divisor Graph [348, 349] | $\mu$ | A zero-divisor graph studied in a fuzzy framework, where each vertex and edge is associated with a single membership degree in $[0, 1]$. |
| 2 | Intuitionistic Fuzzy Zero-Divisor Graph [343] | $(\mu, \nu)$ | A zero-divisor graph defined in an intuitionistic fuzzy framework, where each vertex and edge carries a membership degree and a non-membership degree, usually satisfying $\mu + \nu \leq 1$. |
| 3 | Neutrosophic Zero-Divisor Graph [350–352] | $(T, I, F)$ | A zero-divisor graph defined in a neutrosophic framework, where each vertex and edge is described by truth, indeterminacy, and falsity degrees. |

## 4.10 Fuzzy tolerance graphs

Fuzzy tolerance graphs model uncertain interval overlap, where vertices represent fuzzy intervals and edges express graded adjacency whenever intersections satisfy corresponding fuzzy tolerances or supports [363, 364].

**Definition 4.10.1** (Fuzzy Interval)**.** A fuzzy set

$$I = (\mathbb{R}, m_I)$$

on the real line is called a *fuzzy interval* if it is normal and convex, that is,

$$\sup_{x\in\mathbb{R}} m_I(x) = 1,$$

and

$$m_I\big(\lambda x + (1-\lambda)y\big) \geq \min\{m_I(x), m_I(y)\} \qquad (\forall\, x, y \in \mathbb{R},\ \forall\, \lambda \in [0,1]).$$

The support and core of $I$ are defined by

$$\mathrm{Supp}(I) := \{x \in \mathbb{R} : m_I(x) > 0\}, \qquad \mathrm{Core}(I) := \{x \in \mathbb{R} : m_I(x) = 1\}.$$

Their lengths are denoted by

$$s(I) := \ell\big(\mathrm{Supp}(I)\big), \qquad c(I) := \ell\big(\mathrm{Core}(I)\big),$$

where $\ell(\cdot)$ denotes the ordinary length of an interval.

**Definition 4.10.2** (Fuzzy Tolerance)**.** A *fuzzy tolerance* is a fuzzy interval

$$T = (\mathbb{R}, m_T)$$

whose core length is positive, i.e.,

$$c(T) > 0.$$

The numbers

$$s(T) := \ell\big(\mathrm{Supp}(T)\big) \qquad \text{and} \qquad c(T) := \ell\big(\mathrm{Core}(T)\big)$$

are called the *support length* and *core length* of the fuzzy tolerance $T$, respectively.

**Definition 4.10.3** (Fuzzy Tolerance Graph)**.** Let

$$\mathcal{I} = \{I_1, I_2, \ldots, I_n\}$$

be a finite family of fuzzy intervals on $\mathbb{R}$, and let

$$\mathcal{T} = \{T_1, T_2, \ldots, T_n\}$$

be the corresponding family of fuzzy tolerances.

Associate with $\mathcal{I}$ the crisp vertex set

$$V = \{v_1, v_2, \ldots, v_n\},$$

where the vertex $v_i$ corresponds to the fuzzy interval $I_i$ and the fuzzy tolerance $T_i$.

The *fuzzy tolerance graph* determined by $(\mathcal{I}, \mathcal{T})$ is the fuzzy graph

$$G = (V, \sigma, \mu),$$

where

$$\sigma : V \to [0, 1]$$

and

$$\mu : V \times V \to [0, 1]$$

are defined as follows:

$$\sigma(v_i) := h(I_i) = 1 \qquad (i = 1, 2, \ldots, n),$$

since each fuzzy interval is normal, and for $i \neq j$,

$$\mu(v_i, v_j) := \begin{cases} 1, & c(I_i \cap I_j) \geq \min\{c(T_i), c(T_j)\}, \\ \dfrac{s(I_i \cap I_j)}{\min\{s(T_i), s(T_j)\}}, & c(I_i \cap I_j) < \min\{c(T_i), c(T_j)\} \text{ and } s(I_i \cap I_j) \geq \min\{s(T_i), s(T_j)\}, \\ 0, & \text{otherwise}, \end{cases}$$

and

$$\mu(v_i, v_i) := 0 \qquad (i = 1, 2, \ldots, n).$$

Here

$$I_i \cap I_j$$

denotes the fuzzy intersection of $I_i$ and $I_j$, defined by the minimum operator:

$$m_{I_i \cap I_j}(x) = \min\{m_{I_i}(x), m_{I_j}(x)\} \qquad (\forall\, x \in \mathbb{R}).$$

The pair

$$(\mathcal{I}, \mathcal{T})$$

is called a *fuzzy tolerance representation* of $G$.

The uncertain extension below separates the tolerance-incidence structure from the uncertainty carrier: the tolerance representation determines a crisp support graph, and Uncertain Sets assign model-dependent uncertainty degrees to its vertices and edges.

**Definition 4.10.4** (Support Tolerance Graph)**.** Let

$$(\mathcal{I}, \mathcal{T}) = \big(\{I_1, I_2, \ldots, I_n\}, \{T_1, T_2, \ldots, T_n\}\big)$$

be a fuzzy tolerance representation, and let

$$G_f = (V, \sigma_f, \mu_f)$$

be the fuzzy tolerance graph determined by $(\mathcal{I}, \mathcal{T})$, where

$$V = \{v_1, v_2, \ldots, v_n\}.$$

Define the support edge set by

$$E_{\mathrm{tol}} := \big\{\{v_i, v_j\} \subseteq V : \ i \neq j, \ \mu_f(v_i, v_j) > 0\big\}.$$

Then the graph

$$G^*_{\mathrm{tol}} := (V, E_{\mathrm{tol}})$$

is called the *support tolerance graph* associated with the fuzzy tolerance representation

$$(\mathcal{I}, \mathcal{T}).$$

**Definition 4.10.5** (Uncertain Tolerance Graph)**.** Let

$$(\mathcal{I},\mathcal{T}) = \big(\{I_1, I_2, \ldots, I_n\}, \{T_1, T_2, \ldots, T_n\}\big)$$

be a fuzzy tolerance representation, and let

$$G^*_{\text{tol}} = (V, E_{\text{tol}})$$

be its support tolerance graph.

Let $M$ be a fixed uncertain model with degree-domain

$$\mathrm{Dom}(M) \subseteq [0,1]^k.$$

An *Uncertain Tolerance Graph of type $M$* determined by

$$(\mathcal{I},\mathcal{T})$$

is a quadruple

$$\mathcal{G}^{\text{tol}}_M = (V, E_{\text{tol}}, \sigma_M, \eta_M),$$

where

$$\sigma_M : V \longrightarrow \mathrm{Dom}(M)$$

and

$$\eta_M : E_{\text{tol}} \longrightarrow \mathrm{Dom}(M)$$

are uncertainty-degree functions on the vertex set and edge set, respectively.

Equivalently,

$$(V, \sigma_M)$$

is an Uncertain Set of type $M$ on $V$, and

$$(E_{\text{tol}}, \eta_M)$$

is an Uncertain Set of type $M$ on $E_{\text{tol}}$.

For each vertex

$$v_i \in V,$$

the value

$$\sigma_M(v_i) \in \mathrm{Dom}(M)$$

represents the uncertainty degree attached to the tolerance object corresponding to the pair

$$(I_i, T_i),$$

and for each edge

$$e = \{v_i, v_j\} \in E_{\text{tol}},$$

the value

$$\eta_M(e) \in \mathrm{Dom}(M)$$

represents the uncertainty degree of the tolerance adjacency between

$$(I_i, T_i) \quad \text{and} \quad (I_j, T_j).$$

If desired, one may additionally impose model-specific compatibility conditions between

$$\eta_M(\{v_i, v_j\}) \quad \text{and} \quad \sigma_M(v_i),\ \sigma_M(v_j),$$

but such conditions depend on the chosen uncertain model $M$ and are not fixed at the level of this general definition.

**Theorem 4.10.6** (Well-definedness of Support Tolerance Graph)**.** *Let*

$$(\mathcal{I},\mathcal{T}) = \big(\{I_1, I_2, \ldots, I_n\}, \{T_1, T_2, \ldots, T_n\}\big)$$

*be a fuzzy tolerance representation, and let*

$$G_f = (V, \sigma_f, \mu_f)$$

*be the fuzzy tolerance graph determined by* $(\mathcal{I},\mathcal{T})$*, where*

$$V = \{v_1, v_2, \ldots, v_n\}.$$

*Define*

$$E_{\mathrm{tol}} := \big\{\{v_i, v_j\} \subseteq V : \ i \neq j, \ \mu_f(v_i, v_j) > 0\big\}.$$

*Then*

$$G^*_{\mathrm{tol}} = (V, E_{\mathrm{tol}})$$

*is a well-defined finite simple graph.*

*Proof.* Since

$$G_f = (V, \sigma_f, \mu_f)$$

is a fuzzy tolerance graph, the vertex set

$$V = \{v_1, v_2, \ldots, v_n\}$$

is fixed and finite, and the function

$$\mu_f : V \times V \to [0, 1]$$

is well-defined.

For any distinct vertices

$$v_i, v_j \in V,$$

the statement

$$\mu_f(v_i, v_j) > 0$$

has a definite truth value, because $\mu_f(v_i, v_j)$ is a uniquely determined real number in $[0, 1]$. Hence the condition defining membership in

$$E_{\mathrm{tol}}$$

is meaningful for every unordered pair of distinct vertices.

Moreover, since $G_f$ is a fuzzy graph, the edge-membership function $\mu_f$ is symmetric. Therefore,

$$\mu_f(v_i, v_j) = \mu_f(v_j, v_i) \qquad (\forall\, i, j),$$

so the property

$$\mu_f(v_i, v_j) > 0$$

depends only on the unordered pair

$$\{v_i, v_j\},$$

not on the ordering of its endpoints. Thus

$$E_{\mathrm{tol}}$$

is a well-defined subset of

$$\big\{\{u, v\} \subseteq V : u \neq v\big\}.$$

Also, by the defining property of the fuzzy tolerance graph,

$$\mu_f(v_i, v_i) = 0 \qquad (i = 1, 2, \ldots, n),$$

so no loop can belong to

$$E_{\text{tol}}.$$

Hence

$$G^*_{\text{tol}} = (V, E_{\text{tol}})$$

is loopless. Since its edges are unordered pairs of distinct vertices, it is simple.

Therefore

$$G^*_{\text{tol}} = (V, E_{\text{tol}})$$

is a well-defined finite simple graph. □

**Theorem 4.10.7** (Well-definedness of Uncertain Tolerance Graph)**.** *Let*

$$(\mathcal{I}, \mathcal{T}) = \big(\{I_1, I_2, \ldots, I_n\}, \{T_1, T_2, \ldots, T_n\}\big)$$

*be a fuzzy tolerance representation, let*

$$G^*_{\text{tol}} = (V, E_{\text{tol}})$$

*be its support tolerance graph, and let $M$ be an uncertain model with degree-domain*

$$\mathrm{Dom}(M) \subseteq [0, 1]^k.$$

*Suppose that*

$$\sigma_M : V \to \mathrm{Dom}(M) \qquad \text{and} \qquad \eta_M : E_{\text{tol}} \to \mathrm{Dom}(M)$$

*are functions.*

*Then*

$$\mathcal{G}^{\text{tol}}_M = (V, E_{\text{tol}}, \sigma_M, \eta_M)$$

*is a well-defined Uncertain Tolerance Graph of type $M$.*

*Moreover,*

$$(V, \sigma_M)$$

*and*

$$(E_{\text{tol}}, \eta_M)$$

*are well-defined Uncertain Sets of type $M$.*

*Proof.* By the previous theorem, the support tolerance graph

$$G^*_{\text{tol}} = (V, E_{\text{tol}})$$

is well-defined, where $V$ is a finite set and $E_{\text{tol}}$ is a well-defined set of unordered pairs of distinct vertices.

Since $M$ is an uncertain model, its degree-domain

$$\mathrm{Dom}(M)$$

is fixed.

Because

$$\sigma_M : V \to \mathrm{Dom}(M)$$

is a function, each vertex

$$v \in V$$

is assigned a unique uncertainty degree

$$\sigma_M(v) \in \mathrm{Dom}(M).$$

Hence

$$(V, \sigma_M)$$

is a well-defined Uncertain Set of type $M$ on $V$.

Similarly, because

$$\eta_M : E_{\mathrm{tol}} \to \mathrm{Dom}(M)$$

is a function, each edge

$$e \in E_{\mathrm{tol}}$$

is assigned a unique uncertainty degree

$$\eta_M(e) \in \mathrm{Dom}(M).$$

Hence

$$(E_{\mathrm{tol}}, \eta_M)$$

is a well-defined Uncertain Set of type $M$ on $E_{\mathrm{tol}}$.

Consequently, all entries of the quadruple

$$\mathcal{G}_M^{\mathrm{tol}} = (V, E_{\mathrm{tol}}, \sigma_M, \eta_M)$$

are uniquely specified:

- $V$ is the vertex set induced by the tolerance representation;
- $E_{\mathrm{tol}}$ is the edge set of the support tolerance graph;
- $\sigma_M$ assigns a unique uncertainty degree to each vertex;
- $\eta_M$ assigns a unique uncertainty degree to each edge.

Therefore

$$\mathcal{G}_M^{\mathrm{tol}} = (V, E_{\mathrm{tol}}, \sigma_M, \eta_M)$$

defines a unique mathematical object. Hence the notion of an Uncertain Tolerance Graph is well-defined. □

**Remark 4.10.8.** If one takes

$$\mathrm{Dom}(M) = [0, 1],$$

then an uncertain tolerance graph becomes a scalar-valued uncertainty decoration of the support graph of a fuzzy tolerance representation. Thus the above construction separates the tolerance-incidence structure from the specific uncertainty formalism used to decorate it.

Representative tolerance-graph concepts under uncertainty-aware graph frameworks are listed in Table 4.8.

Table 4.8: Representative tolerance-graph concepts under uncertainty-aware graph frameworks, classified by the dimension $k$ of the information attached to vertices and/or edges.

| $k$ | Tolerance-graph concept | Typical coordinate form | Canonical information attached to vertices/edges |
|---|---|---|---|
| 1 | Fuzzy Tolerance Graph | $\mu$ | A tolerance graph studied in a fuzzy framework, where each vertex and edge is associated with a single membership degree in $[0, 1]$. |
| 2 | Intuitionistic Fuzzy Tolerance Graph [365] | $(\mu, \nu)$ | A tolerance graph defined in an intuitionistic fuzzy framework, where each vertex and edge carries a membership degree and a non-membership degree, usually satisfying $\mu+\nu \leq 1$. |
| 3 | Picture Fuzzy Tolerance Graph [366, 367] | $(\mu, \eta, \nu)$ | A tolerance graph defined in a picture fuzzy framework, where each vertex and edge is described by positive, neutral, and negative membership degrees, usually satisfying $\mu+\eta+\nu \leq 1$. |
| 3 | Neutrosophic Tolerance Graph | $(T, I, F)$ | A tolerance graph defined in a neutrosophic framework, where each vertex and edge is described by truth, indeterminacy, and falsity degrees. |

## 4.11 Uncertain Incidence graphs

Fuzzy incidence graph extends a fuzzy graph by assigning memberships to vertices, edges, and vertex-edge incidences, modeling how strongly each vertex influences each incident edge [368–370].

**Definition 4.11.1** (Fuzzy Incidence Graph). [368–370] Let

$$G^* = (V, E)$$

be a finite simple graph, where

$$E \subseteq \big\{\{u, v\} \subseteq V : u \neq v\big\}.$$

Define the incidence set of $G^*$ by

$$I(G^*) := \{(v, e) \in V \times E : v \in e\}.$$

A *fuzzy incidence graph* on $G^*$ is a triple

$$\widetilde{G} = (\sigma, \mu, \psi),$$

where

$$\sigma : V \to [0, 1], \qquad \mu : E \to [0, 1], \qquad \psi : I(G^*) \to [0, 1]$$

satisfy the following conditions:

1. for every edge

   $$e = \{u, v\} \in E,$$

   the edge-membership is bounded by the memberships of its end vertices:

   $$\mu(e) \leq \min\{\sigma(u), \sigma(v)\};$$

2. for every incidence pair

   $$(v, e) \in I(G^*),$$

   the incidence-membership is bounded by the memberships of the incident vertex and edge:

   $$\psi(v, e) \leq \min\{\sigma(v), \mu(e)\}.$$

The value $\sigma(v)$ is called the *membership degree* of the vertex $v$, the value $\mu(e)$ is called the *membership degree* of the edge $e$, and the value $\psi(v, e)$ is called the *incidence membership degree* of the incidence pair $(v, e)$.

Thus, a fuzzy incidence graph enriches an ordinary fuzzy graph by additionally assigning a degree to each vertex-edge incidence.

**Example 4.11.2** (Fuzzy incidence graph)**.** Let

$$V = \{v_1, v_2, v_3\}, \qquad E = \{e_1, e_2\},$$

where

$$e_1 = \{v_1, v_2\}, \qquad e_2 = \{v_2, v_3\}.$$

Then

$$G^* = (V, E)$$

is a finite simple graph.

The incidence set of $G^*$ is

$$I(G^*) = \{(v_1, e_1), (v_2, e_1), (v_2, e_2), (v_3, e_2)\}.$$

Define

$$\sigma : V \to [0, 1]$$

by

$$\sigma(v_1) = 0.8, \qquad \sigma(v_2) = 0.9, \qquad \sigma(v_3) = 0.7,$$

define

$$\mu : E \to [0, 1]$$

by

$$\mu(e_1) = 0.6, \qquad \mu(e_2) = 0.7,$$

and define

$$\psi : I(G^*) \to [0, 1]$$

by

$$\psi(v_1, e_1) = 0.5, \qquad \psi(v_2, e_1) = 0.6,$$
$$\psi(v_2, e_2) = 0.5, \qquad \psi(v_3, e_2) = 0.4.$$

We verify the defining conditions.

For the edge

$$e_1 = \{v_1, v_2\},$$

we have

$$\mu(e_1) = 0.6 \leq \min\{\sigma(v_1), \sigma(v_2)\} = \min\{0.8, 0.9\} = 0.8.$$

For the edge

$$e_2 = \{v_2, v_3\},$$

we have

$$\mu(e_2) = 0.7 \leq \min\{\sigma(v_2), \sigma(v_3)\} = \min\{0.9, 0.7\} = 0.7.$$

Next, for each incidence pair,

$$\psi(v_1, e_1) = 0.5 \leq \min\{\sigma(v_1), \mu(e_1)\} = \min\{0.8, 0.6\} = 0.6,$$
$$\psi(v_2, e_1) = 0.6 \leq \min\{\sigma(v_2), \mu(e_1)\} = \min\{0.9, 0.6\} = 0.6,$$
$$\psi(v_2, e_2) = 0.5 \leq \min\{\sigma(v_2), \mu(e_2)\} = \min\{0.9, 0.7\} = 0.7,$$

and

$$\psi(v_3, e_2) = 0.4 \leq \min\{\sigma(v_3), \mu(e_2)\} = \min\{0.7, 0.7\} = 0.7.$$

Therefore,

$$\widetilde{G} = (\sigma, \mu, \psi)$$

is a fuzzy incidence graph on $G^*$.

Here, the vertex $v_2$ has the largest vertex-membership degree, the edge $e_2$ has larger edge-membership than $e_1$, and the incidence pair

$$(v_2, e_1)$$

attains the maximum possible value allowed by

$$\min\{\sigma(v_2), \mu(e_1)\} = 0.6.$$

A schematic illustration is shown in Figure 4.5.

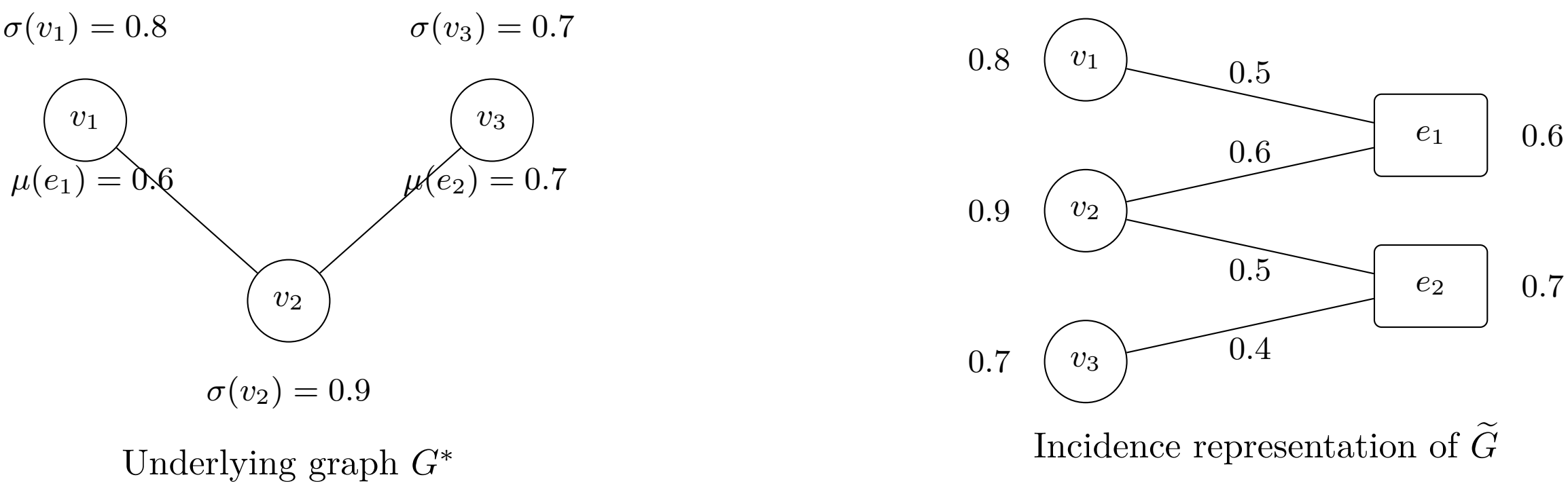

Figure 4.5: A fuzzy incidence graph on the simple graph $G^*$

The uncertain extension replaces these scalar degrees by general uncertainty degrees taken from a fixed uncertain model.

**Definition 4.11.3** (Uncertain Incidence Graph)**.** Let

$$G^* = (V, E)$$

be a finite simple graph, where

$$E \subseteq \big\{\{u, v\} \subseteq V : u \neq v\big\}.$$

Define the incidence set of $G^*$ by

$$I(G^*) := \{(v, e) \in V \times E : v \in e\}.$$

Let $M$ be a fixed uncertain model with degree-domain

$$\mathrm{Dom}(M) \subseteq [0, 1]^k.$$

An *Uncertain Incidence Graph of type $M$* on $G^*$ is a quintuple

$$\widetilde{G}_M = (V, E, \sigma_M, \mu_M, \psi_M),$$

where

$$\sigma_M : V \to \mathrm{Dom}(M), \qquad \mu_M : E \to \mathrm{Dom}(M), \qquad \psi_M : I(G^*) \to \mathrm{Dom}(M)$$

are uncertainty-degree functions.

Equivalently,

$$(V, \sigma_M), \qquad (E, \mu_M), \qquad \big(I(G^*), \psi_M\big)$$

are Uncertain Sets of type $M$ on the vertex set, edge set, and incidence set, respectively.

For each

$$v \in V,$$

the value

$$\sigma_M(v) \in \mathrm{Dom}(M)$$

is called the *uncertainty degree* of the vertex $v$. For each

$$e \in E,$$

the value

$$\mu_M(e) \in \mathrm{Dom}(M)$$

is called the *uncertainty degree* of the edge $e$. For each

$$(v, e) \in I(G^*),$$

the value

$$\psi_M(v, e) \in \mathrm{Dom}(M)$$

is called the *incidence uncertainty degree* of the incidence pair $(v, e)$.

Thus, an uncertain incidence graph enriches an ordinary graph by attaching uncertainty degrees to vertices, edges, and vertex-edge incidences.

**Theorem 4.11.4** (Well-definedness of the incidence set)**.** *Let*

$$G^* = (V, E)$$

*be a finite simple graph, where*

$$E \subseteq \big\{\{u, v\} \subseteq V : u \neq v\big\}.$$

*Then the set*

$$I(G^*) := \{(v, e) \in V \times E : v \in e\}$$

*is well-defined.*

*Proof.* Since $G^* = (V, E)$ is a finite simple graph, each edge

$$e \in E$$

is a two-element subset of $V$. Hence, for every pair

$$(v, e) \in V \times E,$$

the statement

$$v \in e$$

has a definite truth value.

Therefore the collection

$$I(G^*) = \{(v, e) \in V \times E : v \in e\}$$

is a well-defined subset of $V \times E$.

Moreover, because each edge $e = \{u, w\}$ has exactly two endpoints, the set

$$\{(v, e) \in V \times E : v \in e\}$$

contains exactly the two incidence pairs

$$(u, e) \quad \text{and} \quad (w, e).$$

Thus the incidence set is uniquely determined by the graph $G^*$. Hence $I(G^*)$ is well-defined. □

**Theorem 4.11.5** (Well-definedness of Uncertain Incidence Graph)**.** *Let*

$$G^* = (V, E)$$

*be a finite simple graph, let*

$$I(G^*) = \{(v, e) \in V \times E : v \in e\}$$

*be its incidence set, and let $M$ be an uncertain model with degree-domain*

$$\mathrm{Dom}(M) \subseteq [0, 1]^k.$$

*Suppose that*

$$\sigma_M : V \to \mathrm{Dom}(M), \qquad \mu_M : E \to \mathrm{Dom}(M), \qquad \psi_M : I(G^*) \to \mathrm{Dom}(M)$$

*are functions.*

*Then*

$$\widetilde{G}_M = (V, E, \sigma_M, \mu_M, \psi_M)$$

*is a well-defined Uncertain Incidence Graph of type $M$.*

*Moreover,*

$$(V, \sigma_M), \qquad (E, \mu_M), \qquad \big(I(G^*), \psi_M\big)$$

*are well-defined Uncertain Sets of type $M$.*

*Proof.* By the previous theorem, the incidence set

$$I(G^*)$$

is well-defined.

Since $M$ is a fixed uncertain model, its degree-domain

$$\mathrm{Dom}(M)$$

is fixed.

Because

$$\sigma_M : V \to \mathrm{Dom}(M)$$

is a function, each vertex

$$v \in V$$

is assigned a unique uncertainty degree

$$\sigma_M(v) \in \mathrm{Dom}(M).$$

Hence

$$(V, \sigma_M)$$

is a well-defined Uncertain Set of type $M$ on $V$.

Similarly, because

$$\mu_M : E \to \mathrm{Dom}(M)$$

is a function, each edge

$$e \in E$$

is assigned a unique uncertainty degree

$$\mu_M(e) \in \mathrm{Dom}(M).$$

Hence
$$(E, \mu_M)$$
is a well-defined Uncertain Set of type $M$ on $E$.

Likewise, because
$$\psi_M : I(G^*) \to \mathrm{Dom}(M)$$
is a function, each incidence pair
$$(v, e) \in I(G^*)$$
is assigned a unique uncertainty degree
$$\psi_M(v, e) \in \mathrm{Dom}(M).$$
Hence
$$\big(I(G^*), \psi_M\big)$$
is a well-defined Uncertain Set of type $M$ on the incidence set.

Consequently, all components of
$$\widetilde{G}_M = (V, E, \sigma_M, \mu_M, \psi_M)$$
are uniquely specified:

- $V$ is the vertex set of the underlying graph;
- $E$ is the edge set of the underlying graph;
- $\sigma_M$ assigns a unique uncertainty degree to each vertex;
- $\mu_M$ assigns a unique uncertainty degree to each edge;
- $\psi_M$ assigns a unique uncertainty degree to each incidence pair.

Therefore
$$\widetilde{G}_M = (V, E, \sigma_M, \mu_M, \psi_M)$$
defines a unique mathematical object. Hence the notion of an Uncertain Incidence Graph is well-defined. □

**Remark 4.11.6.** If one takes
$$\mathrm{Dom}(M) = [0, 1],$$
then an uncertain incidence graph becomes a scalar-valued incidence structure. With additional model-specific constraints such as
$$\mu_M(e) \le \min\{\sigma_M(u), \sigma_M(v)\} \qquad \text{for } e = \{u, v\} \in E,$$
and
$$\psi_M(v, e) \le \min\{\sigma_M(v), \mu_M(e)\} \qquad \text{for } (v, e) \in I(G^*),$$
the above construction reduces to the usual fuzzy incidence graph.

Representative incidence-graph concepts under uncertainty-aware graph frameworks are listed in Table 4.9.

Related concepts include the Levi graph [380–382], oriented incidence graph [383], signed incidence graph [384, 385], and weighted incidence graph [386, 387].

Table 4.9: Representative incidence-graph concepts under uncertainty-aware graph frameworks, classified by the dimension $k$ of the information attached to vertices, edges, and/or incidence relations.

| $k$ | **Incidence-graph concept** | **Typical coordinate form** | **Canonical information attached to vertices, edges, and/or incidence relations** |
|---|---|---|---|
| 1 | Fuzzy Incidence Graph | $\mu$ | An incidence graph studied in a fuzzy framework, where each vertex, edge, and/or incidence relation is associated with a single membership degree in $[0, 1]$. |
| 2 | Intuitionistic Fuzzy Incidence Graph [371, 372] | $(\mu, \nu)$ | An incidence graph defined in an intuitionistic fuzzy framework, where each vertex, edge, and/or incidence relation carries a membership degree and a non-membership degree, usually satisfying $\mu + \nu \leq 1$. |
| 2 | Vague Incidence Graph [373–375] | $(t, f)$ | An incidence graph defined in a vague framework, where each vertex, edge, and/or incidence relation is characterized by a truth-membership degree and a falsity-membership degree, typically with $t + f \leq 1$. |
| 3 | Picture Fuzzy Incidence Graph [376] | $(\mu, \eta, \nu)$ | An incidence graph defined in a picture fuzzy framework, where each vertex, edge, and/or incidence relation is described by positive, neutral, and negative membership degrees, usually satisfying $\mu + \eta + \nu \leq 1$. |
| 3 | Neutrosophic Incidence Graph [74, 377–379] | $(T, I, F)$ | An incidence graph defined in a neutrosophic framework, where each vertex, edge, and/or incidence relation is described by truth, indeterminacy, and falsity degrees. |

## 4.12 Uncertain Threshold Graphs

Threshold graphs are graphs constructible by repeatedly adding either an isolated vertex or a universal vertex, equivalently characterized by a single weight-threshold rule for adjacency [388, 389]. Fuzzy threshold graphs are fuzzy graphs where a vertex set is independent exactly when the sum of its membership weights does not exceed a threshold [390–393].

**Definition 4.12.1** (Fuzzy Threshold Graph)**.** [393] Let

$$\xi = (V, \sigma, \mu)$$

be a fuzzy graph, where

$$V \neq \varnothing, \qquad \sigma : V \to [0, 1], \qquad \mu : V \times V \to [0, 1],$$

such that for all $u, v \in V$,

$$\mu(u, v) = \mu(v, u), \qquad \mu(u, v) \leq \min\{\sigma(u), \sigma(v)\}.$$

A subset

$$U \subseteq V$$

is called a *stable set* (or *independent set*) of $\xi$ if

$$\mu(u, v) = 0 \qquad \text{for all distinct } u, v \in U.$$

Then $\xi$ is called a *fuzzy threshold graph* if there exists a nonnegative real number

$$T \geq 0$$

such that, for every subset $U \subseteq V$,

$$\sum_{u \in U} \sigma(u) \le T \quad \Longleftrightarrow \quad U \text{ is a stable set in } \xi.$$

In this case, $T$ is called a *threshold* of $\xi$, and the pair

$$(\sigma, T)$$

is called a *threshold representation* of $\xi$.

**Remark 4.12.2.** Equivalently, if one defines the support graph of $\xi$ by

$$G_\xi = (V, E_\xi), \qquad E_\xi := \big\{\{u,v\} \subseteq V : u \neq v,\ \mu(u,v) > 0\big\},$$

then $\xi$ is a fuzzy threshold graph if and only if there exists $T \ge 0$ such that

$$\sum_{u \in U} \sigma(u) \le T$$

holds exactly for those subsets $U \subseteq V$ that are independent in the crisp support graph $G_\xi$.

**Remark 4.12.3.** If

$$\sigma(u) = 1 \qquad (\forall\, u \in V),$$

then the above definition reduces to the classical threshold-graph condition: there exists a threshold $T$ such that a subset $U \subseteq V$ is stable if and only if

$$|U| \le T.$$

More generally, replacing the constant vertex weight 1 by $\sigma(u)$ yields the fuzzy extension.

The uncertain extension below replaces scalar vertex- and edge-memberships by general uncertainty degrees from a fixed uncertain model, and evaluates vertex-degrees through a model-dependent map.

**Definition 4.12.4** (Threshold-Evaluable Uncertain Model)**.** Let $M$ be an uncertain model with degree-domain

$$\mathrm{Dom}(M) \subseteq [0,1]^k.$$

We say that $M$ is *threshold-evaluable* if it is equipped with

1. a distinguished element
$$0_M \in \mathrm{Dom}(M),$$
called the *zero degree*, and
2. a map
$$\Delta_M : \mathrm{Dom}(M) \longrightarrow [0,\infty),$$
called the *threshold-evaluation map*,

such that

$$\Delta_M(0_M) = 0.$$

**Definition 4.12.5** (Stable Set in an Uncertain Graph)**.** Let

$$G^* = (V, E)$$

be a finite undirected loopless graph, and let $M$ be a threshold-evaluable uncertain model with degree-domain $\mathrm{Dom}(M)$, zero degree $0_M$, and threshold-evaluation map

$$\Delta_M : \mathrm{Dom}(M) \to [0,\infty).$$

Let

$$\mathcal{G}_M = (V, E, \sigma_M, \eta_M)$$

be an Uncertain Graph of type $M$, where

$$\sigma_M : V \to \mathrm{Dom}(M), \qquad \eta_M : E \to \mathrm{Dom}(M).$$

Define the support edge set of $\mathcal{G}_M$ by

$$E^*_{\mathrm{supp}}(\mathcal{G}_M) := \{e \in E : \eta_M(e) \neq 0_M\},$$

and define the support graph by

$$G^*_{\mathrm{supp}}(\mathcal{G}_M) := \big(V, E^*_{\mathrm{supp}}(\mathcal{G}_M)\big).$$

A subset

$$U \subseteq V$$

is called a *stable set* (or *independent set*) in $\mathcal{G}_M$ if $U$ is independent in the support graph

$$G^*_{\mathrm{supp}}(\mathcal{G}_M),$$

that is, if

$$\{u, v\} \notin E^*_{\mathrm{supp}}(\mathcal{G}_M) \qquad \text{for all distinct } u, v \in U.$$

Equivalently,

$$\eta_M(\{u, v\}) = 0_M \qquad \text{for all distinct } u, v \in U \text{ with } \{u, v\} \in E.$$

**Definition 4.12.6** (Uncertain Threshold Graph)**.** Let

$$G^* = (V, E)$$

be a finite undirected loopless graph, let $M$ be a threshold-evaluable uncertain model, and let

$$\mathcal{G}_M = (V, E, \sigma_M, \eta_M)$$

be an Uncertain Graph of type $M$.

For each vertex $v \in V$, define its evaluated vertex weight by

$$w_M(v) := \Delta_M\big(\sigma_M(v)\big) \in [0, \infty).$$

Then $\mathcal{G}_M$ is called an *Uncertain Threshold Graph* if there exists a nonnegative real number

$$T \geq 0$$

such that, for every subset

$$U \subseteq V,$$

one has

$$\sum_{u \in U} w_M(u) \leq T \iff U \text{ is a stable set in } \mathcal{G}_M.$$

Equivalently,

$$\sum_{u \in U} \Delta_M\big(\sigma_M(u)\big) \leq T \iff U \text{ is independent in } G^*_{\mathrm{supp}}(\mathcal{G}_M).$$

In this case, $T$ is called a *threshold* of $\mathcal{G}_M$, and the pair

$$(w_M, T)$$

is called a *threshold representation* of $\mathcal{G}_M$.

**Theorem 4.12.7** (Well-definedness of stable sets in an uncertain graph)**.** *Let*

$$G^* = (V, E)$$

*be a finite undirected loopless graph, let $M$ be a threshold-evaluable uncertain model with zero degree $0_M$, and let*

$$\mathcal{G}_M = (V, E, \sigma_M, \eta_M)$$

*be an Uncertain Graph of type $M$.*

*Then the support edge set*

$$E^*_{\text{supp}}(\mathcal{G}_M) = \{\, e \in E : \eta_M(e) \neq 0_M \,\}$$

*is well-defined. Consequently, the support graph*

$$G^*_{\text{supp}}(\mathcal{G}_M) = \bigl(V, E^*_{\text{supp}}(\mathcal{G}_M)\bigr)$$

*is well-defined, and hence the statement*

*"$U \subseteq V$ is a stable set in $\mathcal{G}_M$"*

*is well-defined for every subset $U \subseteq V$.*

*Proof.* Since $M$ is a threshold-evaluable uncertain model, its degree-domain

$$\mathrm{Dom}(M)$$

and zero degree

$$0_M \in \mathrm{Dom}(M)$$

are fixed.

Because

$$\eta_M : E \to \mathrm{Dom}(M)$$

is a function, for each edge

$$e \in E$$

the value

$$\eta_M(e) \in \mathrm{Dom}(M)$$

is uniquely determined. Therefore the predicate

$$\eta_M(e) \neq 0_M$$

has a definite truth value for every $e \in E$. Hence the set

$$E^*_{\text{supp}}(\mathcal{G}_M) = \{\, e \in E : \eta_M(e) \neq 0_M \,\}$$

is a well-defined subset of $E$.

Since $V$ is already fixed, it follows that

$$G^*_{\text{supp}}(\mathcal{G}_M) = \bigl(V, E^*_{\text{supp}}(\mathcal{G}_M)\bigr)$$

is a well-defined graph.

Now let

$$U \subseteq V.$$

The statement that $U$ is independent in

$$G^*_{\text{supp}}(\mathcal{G}_M)$$

means precisely that no two distinct vertices of $U$ are joined by an edge of

$$E^*_{\text{supp}}(\mathcal{G}_M).$$

Because the support graph is well-defined, this condition has a definite truth value. Hence the notion of a stable set in $\mathcal{G}_M$ is well-defined. ∎

**Theorem 4.12.8** (Well-definedness of Uncertain Threshold Graph)**.** *Let*

$$G^* = (V, E)$$

*be a finite undirected loopless graph, let $M$ be a threshold-evaluable uncertain model with degree-domain* $\mathrm{Dom}(M)$*, zero degree $0_M$, and threshold-evaluation map*

$$\Delta_M : \mathrm{Dom}(M) \to [0, \infty),$$

*and let*

$$\mathcal{G}_M = (V, E, \sigma_M, \eta_M)$$

*be an Uncertain Graph of type $M$.*

*Then, for every subset*

$$U \subseteq V,$$

*the sum*

$$\sum_{u \in U} \Delta_M\big(\sigma_M(u)\big)$$

*is a well-defined element of* $[0, \infty)$*.*

*Consequently, for every real number*

$$T \geq 0,$$

*the statement*

$$\sum_{u \in U} \Delta_M\big(\sigma_M(u)\big) \leq T \quad \Longleftrightarrow \quad U \textit{ is a stable set in } \mathcal{G}_M$$

*is well-defined for every subset $U \subseteq V$.*

*Hence the notion of an uncertain threshold graph is well-defined.*

*Proof.* Since

$$\sigma_M : V \to \mathrm{Dom}(M)$$

is a function and

$$\Delta_M : \mathrm{Dom}(M) \to [0, \infty)$$

is also a function, for each vertex

$$u \in V$$

the quantity

$$\Delta_M\big(\sigma_M(u)\big)$$

is uniquely determined and belongs to $[0, \infty)$.

Now let

$$U \subseteq V.$$

Because $V$ is finite, every subset $U$ is finite. Therefore

$$\sum_{u \in U} \Delta_M\big(\sigma_M(u)\big)$$

is a finite sum of nonnegative real numbers, and hence it is a well-defined element of

$$[0, \infty).$$

By the previous theorem, the statement

$$\text{“}U \text{ is a stable set in } \mathcal{G}_M\text{”}$$

is well-defined for every subset $U \subseteq V$. Thus, for each fixed

$$T \geq 0,$$

both sides of the biconditional

$$\sum_{u \in U} \Delta_M\big(\sigma_M(u)\big) \leq T \quad \Longleftrightarrow \quad U \text{ is a stable set in } \mathcal{G}_M$$

have definite truth values.

Hence the assertion

$$\exists\, T \geq 0 \text{ such that } \sum_{u \in U} \Delta_M\big(\sigma_M(u)\big) \leq T \quad \Longleftrightarrow \quad U \text{ is a stable set in } \mathcal{G}_M \quad (\forall U \subseteq V)$$

is well-defined.

Therefore the notion of an uncertain threshold graph is well-defined. □

**Remark 4.12.9.** If

$$\mathrm{Dom}(M) = [0,1], \qquad 0_M = 0, \qquad \Delta_M(a) = a,$$

then the above definition reduces to the usual fuzzy threshold graph:

$$\sum_{u \in U} \sigma(u) \leq T \quad \Longleftrightarrow \quad U \text{ is a stable set.}$$

Thus the uncertain threshold graph is a genuine extension of the fuzzy threshold graph.

Representative threshold-graph concepts under uncertainty-aware graph frameworks are listed in Table 4.10.

Table 4.10: Representative threshold-graph concepts under uncertainty-aware graph frameworks, classified by the dimension $k$ of the information attached to vertices and/or edges.

| $k$ | **Threshold-graph concept** | **Typical coordinate form** | **Canonical information attached to vertices/edges** |
|---|---|---|---|
| 1 | Fuzzy Threshold Graph | $\mu$ | A threshold graph studied in a fuzzy framework, where each vertex and edge is associated with a single membership degree in $[0,1]$. |
| 2 | Intuitionistic Fuzzy Threshold Graph [394–396] | $(\mu,\nu)$ | A threshold graph defined in an intuitionistic fuzzy framework, where each vertex and edge carries a membership degree and a non-membership degree, usually satisfying $\mu+\nu \leq 1$. |
| 3 | Neutrosophic Threshold Graph [397] | $(T,I,F)$ | A threshold graph defined in a neutrosophic framework, where each vertex and edge is described by truth, indeterminacy, and falsity degrees. |

In addition to uncertain threshold graphs, related concepts such as multithreshold graphs [398, 399] and threshold hypergraphs [400–402] are also known.

## 4.13 Random Uncertain Graph

Random Fuzzy Graph is a probability-indexed family of fuzzy graphs where vertex memberships and edge memberships vary randomly, while each realization satisfies the fuzzy-graph condition [403–405].

**Definition 4.13.1** (Random Fuzzy Graph)**.** Let $(\Omega, \mathcal{F}, \mathbb{P})$ be a probability space, and let $V$ be a nonempty finite set of vertices. Write

$$\binom{V}{2} := \{\{u, v\} \subseteq V : u \neq v\}.$$

A *random fuzzy graph* on $V$ is a pair

$$\mathbb{G} = (\Sigma, M),$$

where

$$\Sigma : \Omega \times V \to [0, 1], \qquad M : \Omega \times \binom{V}{2} \to [0, 1]$$

are measurable mappings such that, for every $\omega \in \Omega$ and every $\{u, v\} \in \binom{V}{2}$,

$$M(\omega, \{u, v\}) \leq \min\{\Sigma(\omega, u), \Sigma(\omega, v)\}.$$

For each $\omega \in \Omega$, define

$$\sigma_\omega(u) := \Sigma(\omega, u) \qquad (u \in V),$$

and

$$\mu_\omega(\{u, v\}) := M(\omega, \{u, v\}) \qquad (\{u, v\} \in \binom{V}{2}).$$

Then

$$G_\omega := (V, \sigma_\omega, \mu_\omega)$$

is a fuzzy graph in the usual sense.

The family

$$\mathbb{G} = \{G_\omega\}_{\omega \in \Omega}$$

is called a *random fuzzy graph*. Equivalently, a random fuzzy graph is a measurable map

$$\omega \longmapsto G_\omega$$

from $(\Omega, \mathcal{F}, \mathbb{P})$ into the class of fuzzy graphs on $V$.

The *support graph* of the realization $G_\omega$ is the crisp graph

$$\mathrm{supp}(G_\omega) = \big(V, E_\omega\big), \qquad E_\omega := \Big\{\{u, v\} \in \binom{V}{2} : \mu_\omega(\{u, v\}) > 0\Big\}.$$

Hence both the existence of edges and their membership grades may vary randomly.

The uncertain extension below replaces scalar-valued memberships by general uncertainty degrees from a fixed uncertain model, while preserving the idea that each realization is an uncertainty-aware graph on a fixed underlying finite graph.

**Definition 4.13.2** (Support-Evaluable Uncertain Model)**.** Let $M$ be an uncertain model with degree-domain

$$\mathrm{Dom}(M) \subseteq [0, 1]^k.$$

We say that $M$ is *support-evaluable* if it is equipped with a distinguished element

$$0_M \in \mathrm{Dom}(M),$$

called the *zero degree*.

**Definition 4.13.3** (Random Uncertain Graph)**.** Let

$$(\Omega, \mathcal{F}, \mathbb{P})$$

be a probability space, and let

$$G^* = (V, E)$$

be a finite undirected loopless graph.

Let $M$ be a support-evaluable uncertain model with degree-domain

$$\mathrm{Dom}(M) \subseteq [0, 1]^k$$

and zero degree

$$0_M \in \mathrm{Dom}(M).$$

A *random uncertain graph of type* $M$ on $G^*$ is a pair

$$\mathbb{G}_M = (\Sigma_M, H_M),$$

where

$$\Sigma_M : \Omega \times V \to \mathrm{Dom}(M), \qquad H_M : \Omega \times E \to \mathrm{Dom}(M)$$

satisfy the following measurability condition:

for every fixed

$$v \in V \qquad \text{and} \qquad e \in E,$$

the maps

$$\omega \longmapsto \Sigma_M(\omega, v) \qquad \text{and} \qquad \omega \longmapsto H_M(\omega, e)$$

are measurable.

For each

$$\omega \in \Omega,$$

define

$$\sigma_\omega(v) := \Sigma_M(\omega, v) \qquad (v \in V),$$

and

$$\eta_\omega(e) := H_M(\omega, e) \qquad (e \in E).$$

Then

$$\mathcal{G}_{M,\omega} := (V, E, \sigma_\omega, \eta_\omega)$$

is called the *realization of* $\mathbb{G}_M$ *at* $\omega$.

Equivalently, for each $\omega \in \Omega$,

$$(V, \sigma_\omega)$$

and

$$(E, \eta_\omega)$$

are Uncertain Sets of type $M$ on $V$ and $E$, respectively, and

$$\mathcal{G}_{M,\omega}$$

is an Uncertain Graph of type $M$ on the fixed crisp graph $G^*$.

The family

$$\mathbb{G}_M = \{\mathcal{G}_{M,\omega}\}_{\omega \in \Omega}$$

is called a *random uncertain graph*.

For each realization $\mathcal{G}_{M,\omega}$, define the support edge set by

$$E^*_\omega := \{\, e \in E : \eta_\omega(e) \neq 0_M \,\},$$

and the support graph by

$$\mathrm{supp}(\mathcal{G}_{M,\omega}) := (V, E^*_\omega).$$

Thus both the uncertainty degrees of edges and the support structure may vary randomly.

**Remark 4.13.4.** Because the underlying crisp graph

$$G^* = (V, E)$$

is fixed, randomness appears only in the uncertainty assignments

$$\sigma_\omega : V \to \mathrm{Dom}(M) \qquad \text{and} \qquad \eta_\omega : E \to \mathrm{Dom}(M),$$

not in the ambient vertex set or ambient edge set themselves. The support graph of each realization is obtained by deleting precisely those edges whose uncertainty degree equals the zero degree $0_M$.

**Theorem 4.13.5** (Well-definedness of each realization)**.** *Let*

$$(\Omega, \mathcal{F}, \mathbb{P})$$

*be a probability space, let*

$$G^* = (V, E)$$

*be a finite undirected loopless graph, let $M$ be a support-evaluable uncertain model with degree-domain*

$$\mathrm{Dom}(M) \subseteq [0, 1]^k$$

*and zero degree*

$$0_M \in \mathrm{Dom}(M),$$

*and let*

$$\mathbb{G}_M = (\Sigma_M, H_M)$$

*be a random uncertain graph of type $M$.*

*Then, for every*

$$\omega \in \Omega,$$

*the maps*

$$\sigma_\omega : V \to \mathrm{Dom}(M), \qquad \eta_\omega : E \to \mathrm{Dom}(M)$$

*are well-defined. Consequently,*

$$\mathcal{G}_{M,\omega} = (V, E, \sigma_\omega, \eta_\omega)$$

*is a well-defined Uncertain Graph of type $M$.*

*Moreover,*

$$(V, \sigma_\omega)$$

*and*

$$(E, \eta_\omega)$$

*are well-defined Uncertain Sets of type $M$.*

*Proof.* Fix

$$\omega \in \Omega.$$

Since

$$\Sigma_M : \Omega \times V \to \mathrm{Dom}(M)$$

is a function, for each

$$v \in V$$

the value

$$\Sigma_M(\omega, v) \in \mathrm{Dom}(M)$$

is uniquely determined. Hence the map

$$\sigma_\omega : V \to \mathrm{Dom}(M), \qquad \sigma_\omega(v) := \Sigma_M(\omega, v),$$

is well-defined.

Similarly, since

$$H_M : \Omega \times E \to \mathrm{Dom}(M)$$

is a function, for each

$$e \in E$$

the value

$$H_M(\omega, e) \in \mathrm{Dom}(M)$$

is uniquely determined. Hence the map

$$\eta_\omega : E \to \mathrm{Dom}(M), \qquad \eta_\omega(e) := H_M(\omega, e),$$

is well-defined.

Therefore

$$(V, \sigma_\omega)$$

is an Uncertain Set of type $M$ on $V$, and

$$(E, \eta_\omega)$$

is an Uncertain Set of type $M$ on $E$.

Since the underlying finite graph

$$G^* = (V, E)$$

is fixed, the quadruple

$$\mathcal{G}_{M,\omega} = (V, E, \sigma_\omega, \eta_\omega)$$

is uniquely determined. Hence it is a well-defined Uncertain Graph of type $M$. □

**Theorem 4.13.6** (Well-definedness of the support realization)**.** *Let*

$$\mathbb{G}_M = (\Sigma_M, H_M)$$

*be a random uncertain graph of type $M$, where $M$ is support-evaluable with zero degree*

$$0_M \in \mathrm{Dom}(M).$$

*For each*

$$\omega \in \Omega,$$

*define*

$$E^*_\omega := \{\, e \in E : \eta_\omega(e) \neq 0_M \,\}.$$

*Then*

$$E^*_\omega$$

*is well-defined, and hence*

$$\mathrm{supp}(\mathcal{G}_{M,\omega}) = (V, E^*_\omega)$$

*is a well-defined graph for every $\omega \in \Omega$.*

*Proof.* Fix
$$\omega \in \Omega.$$
By the previous theorem, the map
$$\eta_\omega : E \to \mathrm{Dom}(M)$$
is well-defined.

Since $M$ is support-evaluable, the zero degree
$$0_M \in \mathrm{Dom}(M)$$
is fixed. Therefore, for each
$$e \in E,$$
the statement
$$\eta_\omega(e) \neq 0_M$$
has a definite truth value. Hence
$$E^*_\omega = \{ e \in E : \eta_\omega(e) \neq 0_M \}$$
is a well-defined subset of $E$.

Because $V$ is already fixed, it follows that
$$\mathrm{supp}(\mathcal{G}_{M,\omega}) = (V, E^*_\omega)$$
is a well-defined graph. □

**Theorem 4.13.7** (Well-definedness of the random uncertain graph as a measurable family)**.** *Let*
$$(\Omega, \mathcal{F}, \mathbb{P})$$
*be a probability space, let*
$$G^* = (V, E)$$
*be a finite undirected loopless graph, and let $M$ be a support-evaluable uncertain model with degree-domain*
$$\mathrm{Dom}(M) \subseteq [0, 1]^k.$$

*Let*
$$\mathbb{G}_M = (\Sigma_M, H_M)$$
*be a random uncertain graph of type $M$. Define*
$$\Phi_M : \Omega \to \mathrm{Dom}(M)^V \times \mathrm{Dom}(M)^E$$
*by*
$$\Phi_M(\omega) := (\sigma_\omega, \eta_\omega).$$

*Then*
$$\Phi_M$$
*is a well-defined map. Moreover, if $\mathrm{Dom}(M)^V \times \mathrm{Dom}(M)^E$ is equipped with the product $\sigma$-algebra, then $\Phi_M$ is measurable.*

*Hence a random uncertain graph is equivalently a measurable map from*
$$(\Omega, \mathcal{F}, \mathbb{P})$$
*into the space of all uncertainty assignments on the fixed graph $G^*$.*

*Proof.* By the first theorem, for every

$$\omega \in \Omega,$$

the pair

$$(\sigma_\omega, \eta_\omega)$$

belongs to

$$\mathrm{Dom}(M)^V \times \mathrm{Dom}(M)^E.$$

Hence

$$\Phi_M(\omega) := (\sigma_\omega, \eta_\omega)$$

is well-defined for every $\omega \in \Omega$, so $\Phi_M$ is a well-defined map.

It remains to prove measurability. Since $V$ and $E$ are finite, the product space

$$\mathrm{Dom}(M)^V \times \mathrm{Dom}(M)^E$$

is a finite product of copies of $\mathrm{Dom}(M)$. Therefore a map into this product space is measurable if and only if each coordinate map is measurable.

The coordinate maps of $\Phi_M$ are precisely the maps

$$\omega \longmapsto \sigma_\omega(v) = \Sigma_M(\omega, v) \qquad (v \in V),$$

and

$$\omega \longmapsto \eta_\omega(e) = H_M(\omega, e) \qquad (e \in E).$$

These are measurable by the definition of a random uncertain graph.

Hence all coordinate maps of $\Phi_M$ are measurable, and therefore $\Phi_M$ is measurable. This proves the claim. □

**Remark 4.13.8.** If

$$\mathrm{Dom}(M) = [0,1] \qquad \text{and} \qquad 0_M = 0,$$

then the above definition reduces to a random scalar-valued graph model on a fixed crisp graph. In particular, when the ambient edge set is taken to be

$$E = \binom{V}{2},$$

the construction becomes the natural uncertain analogue of a random fuzzy graph on $V$.

## 4.14 Uncertain Oriented graph

A Fuzzy Oriented Graph is a fuzzy graph with directed edges, where each arc has a membership degree, expressing uncertain one-way relationships between vertices precisely [406–408].

**Definition 4.14.1** (Fuzzy Oriented Graph)**.** Let $V$ be a nonempty finite set. A *fuzzy oriented graph* is a triple

$$G = (V, \sigma, \mu),$$

where

$$\sigma : V \to [0,1]$$

is the vertex-membership function and

$$\mu : V \times V \to [0,1]$$

is the arc-membership function, satisfying the following conditions for all $u, v \in V$:

$$\mu(u,v) \le \min\{\sigma(u), \sigma(v)\}, \tag{4.1}$$

$$\mu(u,u) = 0, \tag{4.2}$$

$$\min\{\mu(u,v), \mu(v,u)\} = 0 \qquad (u \ne v). \tag{4.3}$$

Condition (4.2) excludes loops, and condition (4.3) means that for any two distinct vertices, at most one of the two opposite oriented arcs can have positive membership. Equivalently,

$$u \neq v,\ \mu(u,v) > 0 \implies \mu(v,u) = 0.$$

For each ordered pair $(u,v)$ with $u \neq v$, the value $\mu(u,v)$ is called the *membership degree of the oriented arc* from $u$ to $v$.

The *support digraph* of $G$ is defined by

$$\mathrm{Supp}(G) = (V,E), \qquad E := \{(u,v) \in V \times V : \mu(u,v) > 0\}.$$

Then $\mathrm{Supp}(G)$ is an oriented graph in the ordinary crisp sense.

An uncertain oriented graph extends a fuzzy oriented graph by replacing scalar-valued vertex and arc memberships with general uncertainty degrees taken from a fixed uncertain model.

**Definition 4.14.2** (Support-Evaluable Uncertain Model)**.** Let $M$ be an uncertain model with degree-domain

$$\mathrm{Dom}(M) \subseteq [0,1]^k.$$

We say that $M$ is *support-evaluable* if it is equipped with a distinguished element

$$0_M \in \mathrm{Dom}(M),$$

called the *zero degree.*

**Definition 4.14.3** (Uncertain Oriented Graph)**.** Let $V$ be a nonempty finite set, and let $M$ be a support-evaluable uncertain model with degree-domain

$$\mathrm{Dom}(M) \subseteq [0,1]^k$$

and zero degree

$$0_M \in \mathrm{Dom}(M).$$

An *Uncertain Oriented Graph of type $M$* on $V$ is a triple

$$\mathcal{G}_M = (V, \sigma_M, \mu_M),$$

where

$$\sigma_M : V \to \mathrm{Dom}(M)$$

is the vertex uncertainty-degree function and

$$\mu_M : V \times V \to \mathrm{Dom}(M)$$

is the arc uncertainty-degree function, satisfying the following conditions for all $u, v \in V$:

$$\mu_M(u,u) = 0_M, \tag{4.4}$$
$$u \neq v,\ \mu_M(u,v) \neq 0_M \implies \mu_M(v,u) = 0_M. \tag{4.5}$$

Equivalently, for every two distinct vertices $u, v \in V$, at most one of the two opposite arc-degrees

$$\mu_M(u,v), \qquad \mu_M(v,u)$$

can be different from $0_M$.

The *support digraph* of $\mathcal{G}_M$ is defined by

$$\mathrm{Supp}(\mathcal{G}_M) = (V, A_M),$$

where

$$A_M := \{(u,v) \in V \times V : u \neq v,\ \mu_M(u,v) \neq 0_M\}.$$

If desired, one may additionally impose model-specific compatibility conditions between

$$\mu_M(u,v) \quad \text{and} \quad \sigma_M(u), \sigma_M(v),$$

but such conditions depend on the chosen uncertain model $M$ and are not fixed at the level of this general definition.

**Remark 4.14.4.** Equivalently, one may regard

$$(V, \sigma_M)$$

as an Uncertain Set of type $M$ on the vertex set $V$, and

$$(V \times V, \mu_M)$$

as an Uncertain Set of type $M$ on the set of ordered vertex pairs. The oriented structure is then obtained by restricting to the support arcs

$$(u, v) \in V \times V \quad \text{with} \quad u \neq v,\ \mu_M(u, v) \neq 0_M.$$

**Theorem 4.14.5** (Well-definedness of the support digraph)**.** *Let*

$$\mathcal{G}_M = (V, \sigma_M, \mu_M)$$

*be an Uncertain Oriented Graph of type $M$. Then the support digraph*

$$\mathrm{Supp}(\mathcal{G}_M) = (V, A_M), \qquad A_M := \{(u, v) \in V \times V : u \neq v,\ \mu_M(u, v) \neq 0_M\},$$

*is a well-defined oriented graph in the ordinary crisp sense.*

*Proof.* Since

$$\mu_M : V \times V \to \mathrm{Dom}(M)$$

is a function and $0_M \in \mathrm{Dom}(M)$ is fixed, for every ordered pair

$$(u, v) \in V \times V$$

the statement

$$\mu_M(u, v) \neq 0_M$$

has a definite truth value. Therefore

$$A_M = \{(u, v) \in V \times V : u \neq v,\ \mu_M(u, v) \neq 0_M\}$$

is a well-defined subset of $V \times V$. Hence

$$\mathrm{Supp}(\mathcal{G}_M) = (V, A_M)$$

is a well-defined digraph.

Next, we show that $\mathrm{Supp}(\mathcal{G}_M)$ has no loops. Let $u \in V$. By (4.4),

$$\mu_M(u, u) = 0_M.$$

Therefore

$$(u, u) \notin A_M.$$

Hence $\mathrm{Supp}(\mathcal{G}_M)$ is loopless.

Finally, we prove that no two opposite arcs can occur simultaneously. Assume, for contradiction, that there exist distinct vertices $u, v \in V$ such that

$$(u, v) \in A_M \qquad \text{and} \qquad (v, u) \in A_M.$$

Then

$$\mu_M(u, v) \neq 0_M \qquad \text{and} \qquad \mu_M(v, u) \neq 0_M.$$

Since $u \neq v$, condition (4.5) applied to $(u, v)$ yields

$$\mu_M(v, u) = 0_M,$$

which is a contradiction.

Thus, for every two distinct vertices $u, v \in V$, at most one of

$$(u, v), \qquad (v, u)$$

belongs to $A_M$. Therefore $\mathrm{Supp}(\mathcal{G}_M)$ is an oriented graph. □

**Theorem 4.14.6** (Well-definedness of Uncertain Oriented Graph)**.** *Let $V$ be a nonempty finite set, let $M$ be a support-evaluable uncertain model with degree-domain* $\mathrm{Dom}(M)$ *and zero degree* $0_M$*, and let*

$$\sigma_M : V \to \mathrm{Dom}(M), \qquad \mu_M : V \times V \to \mathrm{Dom}(M)$$

*be functions satisfying*

$$\mu_M(u,u) = 0_M \qquad (\forall\, u \in V),$$

*and*

$$u \neq v,\ \mu_M(u,v) \neq 0_M \implies \mu_M(v,u) = 0_M \qquad (\forall\, u,v \in V).$$

*Then*

$$\mathcal{G}_M = (V, \sigma_M, \mu_M)$$

*is a well-defined Uncertain Oriented Graph of type $M$.*

*Moreover,*

$$(V, \sigma_M)$$

*and*

$$(V \times V, \mu_M)$$

*are well-defined Uncertain Sets of type $M$.*

*Proof.* Since $V$ is a fixed nonempty finite set and $M$ is a fixed uncertain model, the degree-domain

$$\mathrm{Dom}(M)$$

and the zero degree

$$0_M \in \mathrm{Dom}(M)$$

are fixed.

Because

$$\sigma_M : V \to \mathrm{Dom}(M)$$

is a function, each vertex

$$u \in V$$

is assigned a unique uncertainty degree

$$\sigma_M(u) \in \mathrm{Dom}(M).$$

Hence

$$(V, \sigma_M)$$

is a well-defined Uncertain Set of type $M$.

Similarly, because

$$\mu_M : V \times V \to \mathrm{Dom}(M)$$

is a function, each ordered pair

$$(u,v) \in V \times V$$

is assigned a unique uncertainty degree

$$\mu_M(u,v) \in \mathrm{Dom}(M).$$

Hence

$$(V \times V, \mu_M)$$

is a well-defined Uncertain Set of type $M$.

The conditions

$$\mu_M(u,u) = 0_M \qquad (\forall\, u \in V)$$

and

$$u \neq v,\ \mu_M(u,v) \neq 0_M \implies \mu_M(v,u) = 0_M \qquad (\forall\, u,v \in V)$$

are meaningful because all values involved belong to the fixed set $\mathrm{Dom}(M)$. Therefore the class of triples

$$(V, \sigma_M, \mu_M)$$

satisfying these conditions is well-defined.

By the previous theorem, the support digraph

$$\mathrm{Supp}(\mathcal{G}_M)$$

is a well-defined oriented graph. Consequently,

$$\mathcal{G}_M = (V, \sigma_M, \mu_M)$$

is a well-defined Uncertain Oriented Graph of type $M$. □

**Remark 4.14.7.** If

$$\mathrm{Dom}(M) = [0,1] \qquad \text{and} \qquad 0_M = 0,$$

and if one additionally imposes

$$\mu_M(u,v) \leq \min\{\sigma_M(u), \sigma_M(v)\} \qquad (\forall\, u,v \in V),$$

then the above definition reduces to the usual fuzzy oriented graph. Thus the uncertain oriented graph is a genuine extension of the fuzzy oriented graph.

## 4.15 Signed Uncertain Graph

A Signed Fuzzy Graph is a fuzzy graph whose vertices or edges carry positive or negative signs, modeling uncertain relationships with both magnitude and polarity [194, 409–412].

**Definition 4.15.1** (Signed Fuzzy Graph)**.** [409, 410] Let $V$ be a nonempty finite set, and let

$$E \subseteq \binom{V}{2}$$

be a set of undirected edges.

A *signed fuzzy graph* on $(V, E)$ is a quintuple

$$G^{\pm} = (V, E, \sigma, \mu, s),$$

where

$$\sigma : V \to [0,1]$$

is the vertex-membership function,

$$\mu : E \to [0,1]$$

is the edge-membership function, and

$$s : V \cup E \to \{+1, -1\}$$

is a sign function assigning to each vertex and each edge a positive or negative sign, such that

$$\mu(\{u,v\}) \leq \min\{\sigma(u), \sigma(v)\} \qquad (\forall\, \{u,v\} \in E).$$

The pair

$$(V, E, \sigma, \mu)$$

is called the *underlying fuzzy graph* of $G^{\pm}$.

For each vertex $v \in V$, the quantity

$$\widetilde{\sigma}(v) := s(v)\sigma(v) \in [-1, 1]$$

is called the *signed vertex membership* of $v$. For each edge $e = \{u, v\} \in E$, the quantity

$$\widetilde{\mu}(e) := s(e)\mu(e) \in [-1, 1]$$

is called the *signed edge membership* of $e$.

Thus, a signed fuzzy graph is a fuzzy graph together with a polarity structure on its vertices and edges, where the magnitude expresses the degree of presence and the sign expresses the nature of the relation (positive or negative).

The uncertain extension below replaces scalar-valued memberships by general uncertainty degrees from a fixed uncertain model, while the sign part is retained as an independent polarity structure.

**Definition 4.15.2** (Signed-Evaluable Uncertain Model)**.** Let $M$ be an uncertain model with degree-domain

$$\mathrm{Dom}(M) \subseteq [0, 1]^k.$$

We say that $M$ is *signed-evaluable* if it is equipped with

1. a distinguished element

   $$0_M \in \mathrm{Dom}(M),$$

   called the *zero degree*, and

2. a map

   $$\Delta_M : \mathrm{Dom}(M) \longrightarrow [0, \infty),$$

   called the *magnitude-evaluation map*,

such that

$$\Delta_M(0_M) = 0.$$

**Definition 4.15.3** (Signed Uncertain Graph)**.** Let $V$ be a nonempty finite set, and let

$$E \subseteq \binom{V}{2}$$

be a set of undirected edges.

Let $M$ be a signed-evaluable uncertain model with degree-domain

$$\mathrm{Dom}(M) \subseteq [0, 1]^k,$$

zero degree

$$0_M \in \mathrm{Dom}(M),$$

and magnitude-evaluation map

$$\Delta_M : \mathrm{Dom}(M) \to [0, \infty).$$

A *Signed Uncertain Graph of type $M$* on $(V, E)$ is a quintuple

$$\mathcal{G}^{\pm}_M = (V, E, \sigma_M, \eta_M, s),$$

where

$$\sigma_M : V \to \mathrm{Dom}(M)$$

is the vertex uncertainty-degree function,

$$\eta_M : E \to \mathrm{Dom}(M)$$

is the edge uncertainty-degree function, and

$$s : V \cup E \to \{+1, -1\}$$

is a sign function assigning to each vertex and each edge a positive or negative sign.

Equivalently,

$$(V, \sigma_M)$$

and

$$(E, \eta_M)$$

are Uncertain Sets of type $M$ on the vertex set and edge set, respectively.

The quadruple

$$(V, E, \sigma_M, \eta_M)$$

is called the *underlying uncertain graph* of $\mathcal{G}_M^{\pm}$.

For each vertex $v \in V$, define its *signed evaluated vertex degree* by

$$\widetilde{\sigma}_M(v) := s(v)\, \Delta_M\big(\sigma_M(v)\big) \in \mathbb{R}.$$

For each edge $e \in E$, define its *signed evaluated edge degree* by

$$\widetilde{\eta}_M(e) := s(e)\, \Delta_M\big(\eta_M(e)\big) \in \mathbb{R}.$$

The *support graph* of $\mathcal{G}_M^{\pm}$ is defined by

$$\mathrm{Supp}(\mathcal{G}_M^{\pm}) = (V, E_M^*), \qquad E_M^* := \{\, e \in E : \eta_M(e) \neq 0_M \,\}.$$

Thus, a signed uncertain graph is an uncertain graph together with a polarity structure on its vertices and edges, where the uncertainty degree represents magnitude at the model level, and the sign represents positive or negative polarity.

**Remark 4.15.4.** If one does not wish to evaluate uncertainty degrees into real magnitudes, then the sign function

$$s : V \cup E \to \{+1, -1\}$$

alone already provides a polarity assignment on the underlying uncertain graph

$$(V, E, \sigma_M, \eta_M).$$

The evaluated quantities

$$\widetilde{\sigma}_M(v) \quad \text{and} \quad \widetilde{\eta}_M(e)$$

are needed only when one wants signed real-valued magnitudes analogous to those in signed fuzzy graphs.

**Theorem 4.15.5** (Well-definedness of the support graph)**.** *Let*

$$\mathcal{G}_M^{\pm} = (V, E, \sigma_M, \eta_M, s)$$

*be a Signed Uncertain Graph of type $M$, where $M$ is signed-evaluable with zero degree*

$$0_M \in \mathrm{Dom}(M).$$

*Then the set*

$$E_M^* = \{\, e \in E : \eta_M(e) \neq 0_M \,\}$$

*is well-defined. Consequently,*

$$\mathrm{Supp}(\mathcal{G}_M^{\pm}) = (V, E_M^*)$$

*is a well-defined simple graph.*

*Proof.* Since

$$\eta_M : E \to \mathrm{Dom}(M)$$

is a function and

$$0_M \in \mathrm{Dom}(M)$$

is fixed, for every edge

$$e \in E$$

the statement

$$\eta_M(e) \neq 0_M$$

has a definite truth value. Hence

$$E^*_M = \{\, e \in E : \eta_M(e) \neq 0_M \,\}$$

is a well-defined subset of $E$.

Because $E \subseteq \binom{V}{2}$, every element of $E$ is an unordered pair of distinct vertices. Therefore every element of

$$E^*_M$$

is also an unordered pair of distinct vertices. Hence

$$\mathrm{Supp}(\mathcal{G}^{\pm}_M) = (V, E^*_M)$$

is a well-defined simple graph. □

**Theorem 4.15.6** (Well-definedness of Signed Uncertain Graph)**.** *Let $V$ be a nonempty finite set, let*

$$E \subseteq \binom{V}{2},$$

*let $M$ be a signed-evaluable uncertain model with degree-domain*

$$\mathrm{Dom}(M) \subseteq [0,1]^k,$$

*zero degree*

$$0_M \in \mathrm{Dom}(M),$$

*and magnitude-evaluation map*

$$\Delta_M : \mathrm{Dom}(M) \to [0,\infty).$$

*Suppose that*

$$\sigma_M : V \to \mathrm{Dom}(M), \qquad \eta_M : E \to \mathrm{Dom}(M), \qquad s : V \cup E \to \{+1,-1\}$$

*are functions.*

*Then*

$$\mathcal{G}^{\pm}_M = (V, E, \sigma_M, \eta_M, s)$$

*is a well-defined Signed Uncertain Graph of type $M$.*

*Moreover,*

$$(V, \sigma_M) \quad \text{and} \quad (E, \eta_M)$$

*are well-defined Uncertain Sets of type $M$, and for every*

$$v \in V, \qquad e \in E,$$

*the signed evaluated quantities*

$$\widetilde{\sigma}_M(v) = s(v)\, \Delta_M\big(\sigma_M(v)\big)$$

*and*

$$\widetilde{\eta}_M(e) = s(e)\, \Delta_M\big(\eta_M(e)\big)$$

*are well-defined real numbers.*

*Proof.* Since $V$ and $E$ are fixed sets, and

$$\sigma_M : V \to \mathrm{Dom}(M)$$

is a function, each vertex

$$v \in V$$

is assigned a unique uncertainty degree

$$\sigma_M(v) \in \mathrm{Dom}(M).$$

Hence

$$(V, \sigma_M)$$

is a well-defined Uncertain Set of type $M$.

Similarly, because

$$\eta_M : E \to \mathrm{Dom}(M)$$

is a function, each edge

$$e \in E$$

is assigned a unique uncertainty degree

$$\eta_M(e) \in \mathrm{Dom}(M).$$

Hence

$$(E, \eta_M)$$

is a well-defined Uncertain Set of type $M$.

Also, because

$$s : V \cup E \to \{+1, -1\}$$

is a function, each vertex and each edge is assigned a unique sign. Therefore the sign structure is well-defined.

Now let

$$v \in V.$$

Since

$$\sigma_M(v) \in \mathrm{Dom}(M)$$

and

$$\Delta_M : \mathrm{Dom}(M) \to [0, \infty)$$

is a function, the quantity

$$\Delta_M\big(\sigma_M(v)\big)$$

is a well-defined nonnegative real number. Since

$$s(v) \in \{+1, -1\},$$

the product

$$\widetilde{\sigma}_M(v) = s(v)\,\Delta_M\big(\sigma_M(v)\big)$$

is a well-defined real number.

Likewise, for each

$$e \in E,$$

the value

$$\Delta_M\big(\eta_M(e)\big) \in [0, \infty)$$

is well-defined, and since

$$s(e) \in \{+1, -1\},$$

the product

$$\widetilde{\eta}_M(e) = s(e)\,\Delta_M\big(\eta_M(e)\big)$$

is a well-defined real number.

Hence all components of

$$\mathcal{G}_M^{\pm} = (V, E, \sigma_M, \eta_M, s)$$

are uniquely specified, and its associated signed evaluated vertex and edge quantities are also uniquely determined.

Therefore

$$\mathcal{G}_M^{\pm}$$

is a well-defined Signed Uncertain Graph of type $M$. □

**Remark 4.15.7.** If

$$\mathrm{Dom}(M) = [0,1], \qquad 0_M = 0, \qquad \Delta_M(a) = a \quad (\forall\, a \in [0,1]),$$

then the above definition reduces to the ordinary signed fuzzy graph:

$$\widetilde{\sigma}_M(v) = s(v)\sigma_M(v), \qquad \widetilde{\eta}_M(e) = s(e)\eta_M(e).$$

Thus the signed uncertain graph is a genuine extension of the signed fuzzy graph.

Related signed graph concepts under fuzzy and uncertainty-aware frameworks are listed in Table 4.11.

Table 4.11: Related signed graph concepts under fuzzy and uncertainty-aware frameworks

| Concept | Reference(s) |
|---|---|
| Signed Fuzzy Graph | — |
| Signed Intuitionistic Fuzzy Graph | [413, 414] |
| Signed Neutrosophic Graph | [78, 415] |

## 4.16 Weighted Uncertain Graph

Weighted Fuzzy Graph is a fuzzy graph whose edges carry additional nonnegative weights, representing quantities like cost or length, while memberships still express uncertain connectivity [416].

**Definition 4.16.1** (Weighted Fuzzy Graph)**.** Let $V$ be a nonempty finite set. A *weighted fuzzy graph* on $V$ is a quadruple

$$G_w = (V, \sigma, \mu, w),$$

where

$$\sigma : V \to [0,1]$$

is the vertex-membership function,

$$\mu : V \times V \to [0,1]$$

is the edge-membership function, and

$$w : E_\mu \to \mathbb{R}_{\geq 0}$$

is the edge-weight function, such that the following conditions hold for all $u, v \in V$:

$$\mu(u,v) = \mu(v,u), \tag{4.6}$$

$$\mu(u,u) = 0, \tag{4.7}$$

$$\mu(u,v) \leq \min\{\sigma(u), \sigma(v)\}. \tag{4.8}$$

Here

$$E_\mu := \left\{ \{u,v\} \in \binom{V}{2} : \mu(u,v) > 0 \right\}$$

is the support edge set of the fuzzy graph.

For each $\{u,v\} \in E_\mu$, the number

$$w(\{u,v\})$$

is called the *weight* of the fuzzy edge $\{u,v\}$; it may represent, for example, length, cost, time, capacity, or resistance.

The triple
$$(V, \sigma, \mu)$$
is called the *underlying fuzzy graph* of $G_w$.

If, in addition,
$$\sigma(v) = 1 \quad (\forall v \in V) \qquad \text{and} \qquad \mu(u,v) \in \{0,1\} \quad (\forall u,v \in V),$$
then $G_w$ reduces to an ordinary weighted graph.

The uncertain extension below replaces scalar-valued memberships by general uncertainty degrees taken from a fixed uncertain model, while retaining ordinary nonnegative weights on the support edges.

**Definition 4.16.2** (Support-Evaluable Uncertain Model)**.** Let $M$ be an uncertain model with degree-domain
$$\mathrm{Dom}(M) \subseteq [0,1]^k.$$
We say that $M$ is *support-evaluable* if it is equipped with a distinguished element
$$0_M \in \mathrm{Dom}(M),$$
called the *zero degree*.

**Definition 4.16.3** (Uncertain Graph on a Finite Vertex Set)**.** Let $V$ be a nonempty finite set, and let $M$ be a support-evaluable uncertain model with degree-domain
$$\mathrm{Dom}(M) \subseteq [0,1]^k$$
and zero degree
$$0_M \in \mathrm{Dom}(M).$$

An *uncertain graph of type $M$* on $V$ is a triple
$$\mathcal{G}_M = (V, \sigma_M, \eta_M),$$
where
$$\sigma_M : V \to \mathrm{Dom}(M)$$
is the vertex uncertainty-degree function and
$$\eta_M : \binom{V}{2} \to \mathrm{Dom}(M)$$
is the edge uncertainty-degree function.

Equivalently,
$$(V, \sigma_M)$$
is an Uncertain Set of type $M$ on the vertex set $V$, and
$$\left(\binom{V}{2}, \eta_M\right)$$
is an Uncertain Set of type $M$ on the set of unordered pairs of distinct vertices.

**Definition 4.16.4** (Weighted Uncertain Graph)**.** Let $V$ be a nonempty finite set, let $M$ be a support-evaluable uncertain model, and let
$$\mathcal{G}_M = (V, \sigma_M, \eta_M)$$
be an uncertain graph of type $M$ on $V$.

Define the support edge set of $\mathcal{G}_M$ by

$$E_{\eta_M} := \left\{ e \in \binom{V}{2} : \eta_M(e) \neq 0_M \right\}.$$

A *weighted uncertain graph of type $M$ on $V$* is a quadruple

$$\mathcal{G}_{M,w} = (V, \sigma_M, \eta_M, w),$$

where

$$w : E_{\eta_M} \to \mathbb{R}_{\geq 0}$$

is a weight function on the support edge set.

For each

$$e \in E_{\eta_M},$$

the number

$$w(e)$$

is called the *weight* of the support edge $e$. It may represent, for example, length, cost, time, capacity, resistance, or any other nonnegative quantity.

The triple

$$(V, \sigma_M, \eta_M)$$

is called the *underlying uncertain graph* of $\mathcal{G}_{M,w}$.

The *support graph* of $\mathcal{G}_{M,w}$ is defined by

$$\mathrm{Supp}(\mathcal{G}_{M,w}) = (V, E_{\eta_M}),$$

and the *weighted support graph* is

$$\mathrm{Supp}_w(\mathcal{G}_{M,w}) = (V, E_{\eta_M}, w).$$

If desired, one may additionally impose model-specific compatibility conditions between

$$\eta_M(e) \quad \text{and} \quad w(e),$$

but such conditions depend on the chosen model and are not required in the general definition.

**Theorem 4.16.5** (Well-definedness of the support edge set)**.** *Let $V$ be a nonempty finite set, let $M$ be a support-evaluable uncertain model with zero degree*

$$0_M \in \mathrm{Dom}(M),$$

*and let*

$$\mathcal{G}_M = (V, \sigma_M, \eta_M)$$

*be an uncertain graph of type $M$ on $V$.*

*Then the set*

$$E_{\eta_M} := \left\{ e \in \binom{V}{2} : \eta_M(e) \neq 0_M \right\}$$

*is well-defined.*

*Consequently,*

$$\mathrm{Supp}(\mathcal{G}_M) = (V, E_{\eta_M})$$

*is a well-defined simple graph.*

*Proof.* Since

$$\eta_M : \binom{V}{2} \to \mathrm{Dom}(M)$$

is a function and

$$0_M \in \mathrm{Dom}(M)$$

is fixed, for every

$$e \in \binom{V}{2}$$

the statement

$$\eta_M(e) \neq 0_M$$

has a definite truth value. Therefore the subset

$$E_{\eta_M} = \left\{ e \in \binom{V}{2} : \eta_M(e) \neq 0_M \right\}$$

is well-defined.

Since every element of

$$\binom{V}{2}$$

is an unordered pair of distinct vertices, every element of

$$E_{\eta_M}$$

is also an unordered pair of distinct vertices. Hence

$$\mathrm{Supp}(\mathcal{G}_M) = (V, E_{\eta_M})$$

is a well-defined simple graph. □

**Theorem 4.16.6** (Well-definedness of Weighted Uncertain Graph)**.** *Let $V$ be a nonempty finite set, let $M$ be a support-evaluable uncertain model with degree-domain*

$$\mathrm{Dom}(M) \subseteq [0,1]^k$$

*and zero degree*

$$0_M \in \mathrm{Dom}(M),$$

*and let*

$$\sigma_M : V \to \mathrm{Dom}(M), \qquad \eta_M : \binom{V}{2} \to \mathrm{Dom}(M)$$

*be functions.*

*Define*

$$E_{\eta_M} := \left\{ e \in \binom{V}{2} : \eta_M(e) \neq 0_M \right\}.$$

*If*

$$w : E_{\eta_M} \to \mathbb{R}_{\geq 0}$$

*is a function, then*

$$\mathcal{G}_{M,w} = (V, \sigma_M, \eta_M, w)$$

*is a well-defined weighted uncertain graph of type $M$.*

*Moreover,*

$$(V, \sigma_M)$$

*and*

$$\left( \binom{V}{2}, \eta_M \right)$$

*are well-defined Uncertain Sets of type $M$, and*

$$\mathrm{Supp}_w(\mathcal{G}_{M,w}) = (V, E_{\eta_M}, w)$$

*is a well-defined weighted graph.*

*Proof.* Because

$$\sigma_M : V \to \mathrm{Dom}(M)$$

is a function, each vertex

$$v \in V$$

is assigned a unique uncertainty degree

$$\sigma_M(v) \in \mathrm{Dom}(M).$$

Hence

$$(V, \sigma_M)$$

is a well-defined Uncertain Set of type $M$.

Similarly, because

$$\eta_M : \binom{V}{2} \to \mathrm{Dom}(M)$$

is a function, each unordered pair

$$e \in \binom{V}{2}$$

is assigned a unique uncertainty degree

$$\eta_M(e) \in \mathrm{Dom}(M).$$

Hence

$$\left(\binom{V}{2}, \eta_M\right)$$

is a well-defined Uncertain Set of type $M$.

By the previous theorem, the support edge set

$$E_{\eta_M} = \left\{e \in \binom{V}{2} : \eta_M(e) \neq 0_M\right\}$$

is well-defined. Since

$$w : E_{\eta_M} \to \mathbb{R}_{\geq 0}$$

is a function, each support edge

$$e \in E_{\eta_M}$$

is assigned a unique nonnegative real number

$$w(e) \in \mathbb{R}_{\geq 0}.$$

Therefore the weighted support graph

$$\mathrm{Supp}_w(\mathcal{G}_{M,w}) = (V, E_{\eta_M}, w)$$

is well-defined.

Consequently, all components of

$$\mathcal{G}_{M,w} = (V, \sigma_M, \eta_M, w)$$

are uniquely specified:

- $V$ is the fixed vertex set;
- $\sigma_M$ gives a unique uncertainty degree to each vertex;
- $\eta_M$ gives a unique uncertainty degree to each unordered pair of distinct vertices;
- $E_{\eta_M}$ is the well-defined support edge set;
- $w$ gives a unique nonnegative weight to each support edge.

Hence

$$\mathcal{G}_{M,w}$$

is a well-defined weighted uncertain graph of type $M$. □

**Remark 4.16.7.** If

$$\mathrm{Dom}(M) = [0,1] \qquad \text{and} \qquad 0_M = 0,$$

then the above definition reduces to the usual weighted fuzzy graph:

$$E_{\eta_M} = \left\{ \{u,v\} \in \binom{V}{2} : \eta_M(\{u,v\}) > 0 \right\},$$

and

$$w : E_{\eta_M} \to \mathbb{R}_{\geq 0}$$

is the ordinary edge-weight function on the support edges.

If, in addition,

$$\sigma_M(v) = 1 \quad (\forall\, v \in V) \qquad \text{and} \qquad \eta_M(e) \in \{0,1\} \quad \left(\forall\, e \in \binom{V}{2}\right),$$

then one recovers an ordinary weighted graph.

## 4.17 Uncertain Connected graph

A Fuzzy Connected Graph is a fuzzy graph where every pair of vertices is joined by a path of positive strength, ensuring nonzero connectedness throughout.

**Definition 4.17.1** (Fuzzy Connected Graph)**.** Let

$$G = (V, \sigma, \mu)$$

be a fuzzy graph, where

$$\sigma : V \to [0,1], \qquad \mu : V \times V \to [0,1], \qquad \mu(u,v) \leq \min\{\sigma(u), \sigma(v)\} \quad (\forall\, u,v \in V).$$

Assume moreover that $G$ is undirected, that is,

$$\mu(u,v) = \mu(v,u) \qquad (\forall\, u,v \in V),$$

and loopless, that is,

$$\mu(v,v) = 0 \qquad (\forall\, v \in V).$$

Define the support vertex set by

$$V^* := \{v \in V : \sigma(v) > 0\}.$$

A *fuzzy path* from $u$ to $v$ is a finite sequence of distinct vertices

$$P : u = v_0, v_1, \ldots, v_n = v$$

such that

$$\mu(v_{i-1}, v_i) > 0 \qquad (i = 1, 2, \ldots, n).$$

The *strength* of the path $P$ is defined by

$$s(P) := \min_{1 \leq i \leq n} \mu(v_{i-1}, v_i).$$

For any $u, v \in V^*$, the *strength of connectedness* between $u$ and $v$ is

$$\mu^\infty(u,v) := \max\{\, s(P) : P \text{ is a fuzzy path from } u \text{ to } v \,\}.$$

Since $V$ is finite, the above maximum is well defined whenever at least one fuzzy path from $u$ to $v$ exists.

Then $G$ is called a *fuzzy connected graph* if

$$\mu^{\infty}(u,v) > 0 \qquad (\forall\, u,v \in V^*).$$

Equivalently, $G$ is fuzzy connected if for every two vertices in its support there exists at least one fuzzy path joining them.

The uncertain extension below replaces scalar-valued vertex and edge memberships by general uncertainty degrees from a fixed uncertain model, and defines connectedness through the support graph induced by the zero degree.

**Definition 4.17.2** (Support-Evaluable Uncertain Model)**.** Let $M$ be an uncertain model with degree-domain

$$\mathrm{Dom}(M) \subseteq [0,1]^k.$$

We say that $M$ is *support-evaluable* if it is equipped with a distinguished element

$$0_M \in \mathrm{Dom}(M),$$

called the *zero degree.*

**Definition 4.17.3** (Uncertain Connected Graph)**.** Let

$$V$$

be a nonempty finite set, and let $M$ be a support-evaluable uncertain model with degree-domain

$$\mathrm{Dom}(M) \subseteq [0,1]^k$$

and zero degree

$$0_M \in \mathrm{Dom}(M).$$

Let

$$\binom{V}{2} := \big\{\{u,v\} \subseteq V : u \neq v\big\}.$$

An *uncertain graph of type $M$ on $V$* is a triple

$$\mathcal{G}_M = (V, \sigma_M, \eta_M),$$

where

$$\sigma_M : V \to \mathrm{Dom}(M), \qquad \eta_M : \binom{V}{2} \to \mathrm{Dom}(M)$$

are uncertainty-degree functions on the vertex set and the unordered edge set, respectively.

Equivalently,

$$(V, \sigma_M)$$

and

$$\left(\binom{V}{2}, \eta_M\right)$$

are Uncertain Sets of type $M$.

Define the support vertex set by

$$V_M^* := \{\, v \in V : \sigma_M(v) \neq 0_M \,\},$$

and the support edge set by

$$E_M^* := \{\{u,v\} \in \binom{V_M^*}{2} : \eta_M(\{u,v\}) \neq 0_M\}.$$

The *support graph* of $\mathcal{G}_M$ is defined by

$$G_{\mathrm{supp}}(\mathcal{G}_M) := (V_M^*, E_M^*).$$

An *uncertain path* from $u$ to $v$, where

$$u, v \in V_M^*,$$

is a finite sequence of distinct vertices

$$P : u = v_0, v_1, \ldots, v_n = v$$

such that

$$\{v_{i-1}, v_i\} \in E_M^* \qquad (i = 1, 2, \ldots, n).$$

Then $\mathcal{G}_M$ is called an *uncertain connected graph* if for every two vertices

$$u, v \in V_M^*,$$

there exists an uncertain path from $u$ to $v$.

Equivalently, $\mathcal{G}_M$ is uncertain connected if and only if its support graph

$$G_{\mathrm{supp}}(\mathcal{G}_M)$$

is connected in the ordinary graph-theoretic sense.

**Theorem 4.17.4** (Well-definedness of the support graph)**.** *Let*

$$\mathcal{G}_M = (V, \sigma_M, \eta_M)$$

*be an uncertain graph of type $M$, where $M$ is support-evaluable with zero degree*

$$0_M \in \mathrm{Dom}(M).$$

*Then the support vertex set*

$$V_M^* = \{\, v \in V : \sigma_M(v) \neq 0_M \,\}$$

*and the support edge set*

$$E_M^* = \{\{u,v\} \in \binom{V_M^*}{2} : \eta_M(\{u,v\}) \neq 0_M\}$$

*are well-defined.*

*Consequently, the support graph*

$$G_{\mathrm{supp}}(\mathcal{G}_M) = (V_M^*, E_M^*)$$

*is a well-defined finite simple graph.*

*Proof.* Since

$$\sigma_M : V \to \mathrm{Dom}(M)$$

is a function and

$$0_M \in \mathrm{Dom}(M)$$

is fixed, for every

$$v \in V$$

the statement

$$\sigma_M(v) \neq 0_M$$

has a definite truth value. Therefore

$$V_M^* = \{ v \in V : \sigma_M(v) \neq 0_M \}$$

is a well-defined subset of $V$.

Next, since

$$\eta_M : \binom{V}{2} \to \mathrm{Dom}(M)$$

is a function, for every unordered pair

$$\{u, v\} \in \binom{V}{2}$$

the value

$$\eta_M(\{u, v\}) \in \mathrm{Dom}(M)$$

is uniquely determined. Hence the statement

$$\eta_M(\{u, v\}) \neq 0_M$$

also has a definite truth value.

Because

$$\binom{V_M^*}{2}$$

is a well-defined set of unordered pairs of distinct vertices, it follows that

$$E_M^* = \left\{ \{u, v\} \in \binom{V_M^*}{2} : \eta_M(\{u, v\}) \neq 0_M \right\}$$

is a well-defined subset of

$$\binom{V_M^*}{2}.$$

Therefore

$$G_{\mathrm{supp}}(\mathcal{G}_M) = (V_M^*, E_M^*)$$

is a well-defined graph. Since its edges are unordered pairs of distinct vertices, it is simple. Since $V$ is finite, the subset $V_M^*$ is finite, and hence the support graph is finite. □

**Theorem 4.17.5** (Well-definedness of uncertain paths)**.** *Let*

$$\mathcal{G}_M = (V, \sigma_M, \eta_M)$$

*be an uncertain graph of type $M$, and let*

$$G_{\mathrm{supp}}(\mathcal{G}_M) = (V_M^*, E_M^*)$$

*be its support graph.*

*Then, for any*

$$u, v \in V_M^*,$$

*the statement*

$$\text{“}P : u = v_0, v_1, \ldots, v_n = v \text{ is an uncertain path from } u \text{ to } v\text{”}$$

*is well-defined.*

*Hence the notion of an uncertain path is well-defined.*

*Proof.* By the previous theorem,

$$G_{\mathrm{supp}}(\mathcal{G}_M) = (V_M^*, E_M^*)$$

is a well-defined finite simple graph.

Now let

$$P : u = v_0, v_1, \ldots, v_n = v$$

be a finite sequence of vertices. The conditions

$$v_0 = u, \qquad v_n = v,$$

that the vertices

$$v_0, v_1, \ldots, v_n$$

are distinct, and that

$$\{v_{i-1}, v_i\} \in E_M^* \qquad (i = 1, 2, \ldots, n)$$

are all meaningful, because the vertex set

$$V_M^*$$

and the edge set

$$E_M^*$$

are well-defined.

Therefore the predicate

"$P$ is an uncertain path from $u$ to $v$"

has a definite truth value for every such sequence $P$. Hence the notion of an uncertain path is well-defined. □

**Theorem 4.17.6** (Well-definedness of Uncertain Connected Graph)**.** *Let*

$$V$$

*be a nonempty finite set, let $M$ be a support-evaluable uncertain model, and let*

$$\mathcal{G}_M = (V, \sigma_M, \eta_M)$$

*be an uncertain graph of type $M$.*

*Then the statement*

*"$\mathcal{G}_M$ is an uncertain connected graph"*

*is well-defined.*

*Equivalently, the statement*

*"$G_{\mathrm{supp}}(\mathcal{G}_M)$ is connected"*

*is well-defined.*

*Proof.* By Theorem 1, the support graph

$$G_{\mathrm{supp}}(\mathcal{G}_M) = (V_M^*, E_M^*)$$

is a well-defined finite simple graph.

By Theorem 2, for every pair of vertices

$$u, v \in V_M^*,$$

the statement

"there exists an uncertain path from $u$ to $v$"

is well-defined.

Therefore the universal statement

$$\forall\, u, v \in V_M^*,\ \exists \text{ an uncertain path from } u \text{ to } v$$

has a definite truth value.

Hence the assertion that $\mathcal{G}_M$ is an uncertain connected graph is well-defined.

The equivalence with connectedness of the support graph follows directly from the definition of connectedness in ordinary graph theory, applied to the well-defined graph

$$G_{\mathrm{supp}}(\mathcal{G}_M).$$

□

**Remark 4.17.7.** If

$$\mathrm{Dom}(M) = [0,1] \qquad \text{and} \qquad 0_M = 0,$$

then the support vertex set and support edge set become

$$V_M^* = \{\, v \in V : \sigma_M(v) > 0 \,\},$$

and

$$E_M^* = \Big\{ \{u, v\} \in \binom{V_M^*}{2} : \eta_M(\{u, v\}) > 0 \Big\}.$$

Hence the above definition reduces to the usual support-based notion of connectedness for a fuzzy graph.

## 4.18 Cayley Uncertain graph

A Fuzzy Cayley Graph is a Cayley graph endowed with fuzzy vertex or edge memberships, representing algebraic connections in groups under uncertainty or graded relations [417–419].

**Definition 4.18.1** (Fuzzy Cayley Graph)**.** (cf. [420–422]) Let $G$ be a group with identity element $e$, and let

$$\widetilde{S} : G \to [0,1]$$

be a fuzzy subset of $G$ such that

$$\widetilde{S}(e) = 0 \qquad \text{and} \qquad \widetilde{S}(g) = \widetilde{S}(g^{-1}) \quad (\forall\, g \in G).$$

Assume moreover that the support

$$\mathrm{supp}(\widetilde{S}) := \{g \in G : \widetilde{S}(g) > 0\}$$

generates $G$.

Then the *fuzzy Cayley graph* of $G$ with respect to $\widetilde{S}$, denoted by

$$\mathrm{Cay}_f(G, \widetilde{S}),$$

is the fuzzy graph

$$\mathrm{Cay}_f(G, \widetilde{S}) = (V, \sigma, \mu),$$

where

$$V := G, \qquad \sigma(x) := 1 \quad (\forall\, x \in G),$$

and

$$\mu(x, y) := \widetilde{S}(x^{-1}y) \quad (\forall\, x, y \in G).$$

Equivalently, two vertices $x, y \in G$ are joined with membership $\widetilde{S}(x^{-1}y)$, that is, the membership of the group element carrying $x$ to $y$.

**Remark 4.18.2.** The above definition is well posed as a fuzzy graph:

$$\mu(x,x) = \widetilde{S}(e) = 0 \quad (\forall\, x \in G),$$

so there are no loops of positive membership, and

$$\mu(x,y) = \widetilde{S}(x^{-1}y) = \widetilde{S}\big((x^{-1}y)^{-1}\big) = \widetilde{S}(y^{-1}x) = \mu(y,x),$$

so $\mu$ is symmetric. Also,

$$\mu(x,y) \le 1 = \min\{\sigma(x), \sigma(y)\} \quad (\forall\, x,y \in G).$$

Hence $\mathrm{Cay}_f(G,\widetilde{S})$ is indeed a fuzzy graph.

**Remark 4.18.3.** If $\widetilde{S} = \chi_S$ is the characteristic function of an ordinary inverse-closed generating set

$$S \subseteq G, \qquad e \notin S, \qquad S = S^{-1},$$

then $\mathrm{Cay}_f(G,\widetilde{S})$ reduces to the usual Cayley graph

$$\mathrm{Cay}(G,S).$$

**Remark 4.18.4.** If the condition

$$\widetilde{S}(g) = \widetilde{S}(g^{-1})$$

is omitted, then the same formula

$$\mu(x,y) = \widetilde{S}(x^{-1}y)$$

defines a *fuzzy Cayley digraph* in general.

The uncertain extension below replaces scalar-valued memberships by general uncertainty degrees taken from a fixed uncertain model.

**Definition 4.18.5** (Cayley-admissible uncertain model)**.** Let $M$ be an uncertain model with degree-domain

$$\mathrm{Dom}(M) \subseteq [0,1]^k.$$

We say that $M$ is *Cayley-admissible* if it is equipped with two distinguished elements

$$0_M,\ 1_M \in \mathrm{Dom}(M),$$

called the *zero degree* and the *unit degree*, respectively, such that

$$0_M \neq 1_M.$$

**Definition 4.18.6** (Uncertain generating subset)**.** Let $G$ be a group with identity element $e$, and let $M$ be a Cayley-admissible uncertain model.

An *uncertain subset of $G$ of type $M$* is a function

$$\widetilde{S}_M : G \to \mathrm{Dom}(M).$$

Its support is defined by

$$\mathrm{Supp}_M(\widetilde{S}_M) := \{\, g \in G : \widetilde{S}_M(g) \neq 0_M \,\}.$$

The uncertain subset $\widetilde{S}_M$ is called an *uncertain generating subset* of $G$ if

$$\widetilde{S}_M(e) = 0_M,$$

$$\widetilde{S}_M(g) = \widetilde{S}_M(g^{-1}) \qquad (\forall\, g \in G),$$

and

$$\langle \mathrm{Supp}_M(\widetilde{S}_M) \rangle = G.$$

**Definition 4.18.7** (Cayley uncertain graph)**.** Let $G$ be a group with identity element $e$, let $M$ be a Cayley-admissible uncertain model, and let

$$\widetilde{S}_M : G \to \mathrm{Dom}(M)$$

be an uncertain generating subset of $G$.

The *Cayley uncertain graph of* $G$ *with respect to* $\widetilde{S}_M$, denoted by

$$\mathrm{Cay}_M(G, \widetilde{S}_M),$$

is the uncertain graph

$$\mathrm{Cay}_M(G, \widetilde{S}_M) = (V, \sigma_M, \eta_M),$$

where

$$V := G,$$

$$\sigma_M(x) := 1_M \qquad (\forall\, x \in G),$$

and

$$\eta_M : \binom{G}{2} \to \mathrm{Dom}(M)$$

is defined by

$$\eta_M(\{x, y\}) := \widetilde{S}_M(x^{-1}y) \qquad (\forall\, \{x, y\} \in \binom{G}{2}).$$

Equivalently, two distinct vertices $x, y \in G$ are joined by the uncertainty degree of the group element $x^{-1}y$ carrying $x$ to $y$.

**Remark 4.18.8.** The above definition is intended to be the undirected uncertain analogue of the ordinary Cayley graph. The condition

$$\widetilde{S}_M(g) = \widetilde{S}_M(g^{-1})$$

ensures that the edge-degree assigned to $\{x, y\}$ is independent of whether one uses $x^{-1}y$ or $y^{-1}x$.

**Theorem 4.18.9** (Well-definedness of the edge-degree function)**.** *Let $G$ be a group with identity element $e$, let $M$ be a Cayley-admissible uncertain model, and let*

$$\widetilde{S}_M : G \to \mathrm{Dom}(M)$$

*be an uncertain subset satisfying*

$$\widetilde{S}_M(g) = \widetilde{S}_M(g^{-1}) \qquad (\forall\, g \in G).$$

*Then the formula*

$$\eta_M(\{x, y\}) := \widetilde{S}_M(x^{-1}y) \qquad (\forall\, \{x, y\} \in \binom{G}{2})$$

*defines a well-defined function*

$$\eta_M : \binom{G}{2} \to \mathrm{Dom}(M).$$

*Proof.* Let

$$\{x, y\} \in \binom{G}{2}.$$

Since $\{x, y\}$ is an unordered pair, the same edge may be represented either by the ordered pair $(x, y)$ or by $(y, x)$. Therefore it must be shown that

$$\widetilde{S}_M(x^{-1}y) = \widetilde{S}_M(y^{-1}x).$$

Now

$$y^{-1}x = (x^{-1}y)^{-1}.$$

Hence, by the inversion-symmetry of $\widetilde{S}_M$,

$$\widetilde{S}_M(y^{-1}x) = \widetilde{S}_M\big((x^{-1}y)^{-1}\big) = \widetilde{S}_M(x^{-1}y).$$

Therefore the value assigned to $\{x, y\}$ does not depend on the chosen ordering of its endpoints.

Since $\widetilde{S}_M$ takes values in $\mathrm{Dom}(M)$, it follows that

$$\eta_M(\{x, y\}) \in \mathrm{Dom}(M) \qquad (\forall\, \{x, y\} \in \binom{G}{2}),$$

and thus $\eta_M$ is a well-defined function

$$\eta_M : \binom{G}{2} \to \mathrm{Dom}(M).$$

□

**Theorem 4.18.10** (Well-definedness of Cayley uncertain graph)**.** *Let $G$ be a group with identity element $e$, let $M$ be a Cayley-admissible uncertain model, and let*

$$\widetilde{S}_M : G \to \mathrm{Dom}(M)$$

*be an uncertain generating subset of $G$.*

*Then*

$$\mathrm{Cay}_M(G, \widetilde{S}_M) = (V, \sigma_M, \eta_M)$$

*is a well-defined uncertain graph.*

*Moreover:*

1. $$(V, \sigma_M)$$

   *is a well-defined Uncertain Set of type $M$;*

2. $$\left(\binom{G}{2}, \eta_M\right)$$

   *is a well-defined Uncertain Set of type $M$;*

3. *the support graph of $\mathrm{Cay}_M(G, \widetilde{S}_M)$ is the ordinary Cayley graph*

   $$\mathrm{Cay}\big(G, \mathrm{Supp}_M(\widetilde{S}_M)\big).$$

*Proof.* Define

$$V := G, \qquad \sigma_M(x) := 1_M \quad (\forall\, x \in G).$$

Since $1_M \in \mathrm{Dom}(M)$ is fixed, $\sigma_M$ is a well-defined constant function

$$\sigma_M : V \to \mathrm{Dom}(M).$$

Hence

$$(V, \sigma_M)$$

is a well-defined Uncertain Set of type $M$.

By the previous theorem, the map

$$\eta_M : \binom{G}{2} \to \mathrm{Dom}(M), \qquad \eta_M(\{x, y\}) = \widetilde{S}_M(x^{-1}y),$$

is well-defined. Therefore

$$\left(\binom{G}{2}, \eta_M\right)$$

is also a well-defined Uncertain Set of type $M$.

Consequently,

$$\mathrm{Cay}_M(G, \widetilde{S}_M) = (V, \sigma_M, \eta_M)$$

is a well-defined uncertain graph.

It remains to identify the support graph. By definition, its support edge set is

$$E_M^* := \left\{\{x, y\} \in \binom{G}{2} : \eta_M(\{x, y\}) \neq 0_M\right\}.$$

Using the definition of $\eta_M$, one obtains

$$\eta_M(\{x, y\}) \neq 0_M \iff \widetilde{S}_M(x^{-1}y) \neq 0_M \iff x^{-1}y \in \mathrm{Supp}_M(\widetilde{S}_M).$$

Hence

$$E_M^* = \left\{\{x, y\} \in \binom{G}{2} : x^{-1}y \in \mathrm{Supp}_M(\widetilde{S}_M)\right\}.$$

This is exactly the edge set of the ordinary undirected Cayley graph

$$\mathrm{Cay}\big(G, \mathrm{Supp}_M(\widetilde{S}_M)\big).$$

Therefore the support graph of $\mathrm{Cay}_M(G, \widetilde{S}_M)$ is

$$\mathrm{Cay}\big(G, \mathrm{Supp}_M(\widetilde{S}_M)\big).$$

□

**Corollary 4.18.11** (Connectedness of the support graph)**.** *Let $G$ be a group, let $M$ be a Cayley-admissible uncertain model, and let*

$$\widetilde{S}_M : G \to \mathrm{Dom}(M)$$

*be an uncertain generating subset of $G$. Then the support graph of*

$$\mathrm{Cay}_M(G, \widetilde{S}_M)$$

*is connected.*

*Proof.* By the previous theorem, the support graph of

$$\mathrm{Cay}_M(G, \widetilde{S}_M)$$

is

$$\mathrm{Cay}\big(G, \mathrm{Supp}_M(\widetilde{S}_M)\big).$$

Since

$$\langle \mathrm{Supp}_M(\widetilde{S}_M)\rangle = G,$$

the set $\mathrm{Supp}_M(\widetilde{S}_M)$ generates $G$. It is a standard fact from group theory and graph theory that the Cayley graph of a group with respect to a generating set is connected. Hence

$$\mathrm{Cay}\big(G, \mathrm{Supp}_M(\widetilde{S}_M)\big)$$

is connected. Therefore the support graph of

$$\mathrm{Cay}_M(G, \widetilde{S}_M)$$

is connected. □

**Remark 4.18.12.** If

$$\mathrm{Dom}(M) = [0,1], \qquad 0_M = 0, \qquad 1_M = 1,$$

then the above definition reduces to the usual fuzzy Cayley graph:

$$\sigma_M(x) = 1 \qquad (\forall\, x \in G),$$

and

$$\eta_M(\{x,y\}) = \widetilde{S}_M(x^{-1}y).$$

Thus Cayley uncertain graphs genuinely extend fuzzy Cayley graphs.

**Remark 4.18.13.** If

$$\widetilde{S}_M(g) = 0_M \quad \Longleftrightarrow \quad g \notin S$$

for some inverse-closed generating set

$$S \subseteq G, \qquad e \notin S, \qquad S = S^{-1},$$

and if $\widetilde{S}_M$ takes the same nonzero degree on every element of $S$, then the support graph of

$$\mathrm{Cay}_M(G, \widetilde{S}_M)$$

coincides with the ordinary Cayley graph

$$\mathrm{Cay}(G, S).$$

Representative Cayley-graph concepts under uncertainty-aware graph frameworks are listed in Table 4.12.

Table 4.12: Representative Cayley-graph concepts under uncertainty-aware graph frameworks, classified by the dimension $k$ of the information attached to vertices and/or edges.

| $k$ | **Cayley-graph concept** | **Typical coordinate form** | **Canonical information attached to vertices/edges** |
|---|---|---|---|
| 1 | Fuzzy Cayley Graph | $\mu$ | A Cayley graph studied in a fuzzy framework, where each vertex and edge is associated with a single membership degree in $[0,1]$. |
| 2 | Intuitionistic Fuzzy Cayley Graph | $(\mu,\nu)$ | A Cayley graph defined in an intuitionistic fuzzy framework, where each vertex and edge carries a membership degree and a non-membership degree, usually satisfying $\mu+\nu \le 1$. |
| 3 | Picture Fuzzy Cayley Graph [423] | $(\mu,\eta,\nu)$ | A Cayley graph defined in a picture fuzzy framework, where each vertex and edge is described by positive, neutral, and negative membership degrees, usually satisfying $\mu+\eta+\nu \le 1$. |
| 3 | Neutrosophic Cayley Graph [424] | $(T,I,F)$ | A Cayley graph defined in a neutrosophic framework, where each vertex and edge is described by truth, indeterminacy, and falsity degrees. |

Related concepts such as the directed Cayley graph [425], weighted Cayley graph [426], signed Cayley graph [427], bi-Cayley graph [428, 429], and Cayley hypergraph [430] are also known. For reference, it may also be possible to define a MultiCayley graph as follows, as an extension of the bi-Cayley graph.

**Definition 4.18.14** (MultiCayley Graph)**.** Let $G$ be a group with identity element $e$, and let

$$m \ge 2$$

be an integer. For each pair

$$i, j \in \{1, 2, \ldots, m\},$$

let

$$S_{ij} \subseteq G$$

be a specified subset.

The *MultiCayley graph* associated with

$$G \quad \text{and} \quad \{S_{ij}\}_{1\le i,j\le m}$$

is the graph

$$\mathrm{MCay}\big(G;\{S_{ij}\}_{1\le i,j\le m}\big)$$

defined as follows:

- its vertex set is

  $$V := G \times \{1,2,\ldots,m\},$$

  so that each vertex is of the form

  $$(g,i),$$

  where $g \in G$ and $i$ indicates the layer;
- two vertices

  $$(g,i),\ (h,j) \in V$$

  are adjacent if and only if

  $$g^{-1}h \in S_{ij}.$$

If one wishes $\mathrm{MCay}(G;\{S_{ij}\})$ to be an undirected simple graph, it is natural to assume that

$$S_{ji} = S_{ij}^{-1} \qquad (\forall\, i,j),$$

and

$$e \notin S_{ii} \qquad (\forall\, i).$$

In the special case

$$m = 2,$$

this construction reduces to a natural generalization of the Bi-Cayley graph.

## 4.19 Fuzzy median graphs

A Median Graph is a connected graph where any three vertices have a unique median vertex lying on shortest paths between each pair of them [431–433]. A Fuzzy Median Graph is a connected fuzzy graph where every three distinct vertices have one unique median vertex belonging to all geodesic intervals simultaneously [434–436].

**Definition 4.19.1** (Fuzzy Median Graph)**.** Let

$$F = (V,\sigma,\mu)$$

be a connected fuzzy graph, where

$$\sigma : V \to [0,1], \qquad \mu : V \times V \to [0,1], \qquad \mu(x,y) = \mu(y,x), \qquad \mu(x,y) \le \min\{\sigma(x),\sigma(y)\}$$

for all $x, y \in V$.

Define the support vertex set by

$$V^* := \{x \in V : \sigma(x) > 0\}.$$

For two vertices $x, y \in V^*$, let

$$\mathrm{CONN}_F(x,y)$$

denote the strength of connectedness between $x$ and $y$, that is, the supremum of the strengths of all fuzzy paths joining $x$ and $y$, where the strength of a path is the minimum membership value among its edges.

An edge $xy$ with $\mu(x, y) > 0$ is called *strong* if

$$\mu(x, y) \geq \mathrm{CONN}_{F-xy}(x, y),$$

where $F - xy$ is the partial fuzzy subgraph obtained by deleting the edge $xy$.

A *strong path* is a path all of whose edges are strong. A *strong geodesic* between $x$ and $y$ is a shortest strong path joining $x$ and $y$. Its length is called the *geodesic distance* between $x$ and $y$, and is denoted by

$$d_g(x, y).$$

For $x, y \in V^*$, the *geodesic interval* between $x$ and $y$ is defined by

$$I_g(x, y) := \{u \in V^* : d_g(x, y) = d_g(x, u) + d_g(u, y)\}.$$

For $x, y, z \in V^*$, the *median set* of $x, y, z$ is defined by

$$\gamma(x, y, z) := I_g(x, y) \cap I_g(x, z) \cap I_g(y, z).$$

Then $F$ is called a *fuzzy median graph* if, for every triple of distinct vertices

$$x, y, z \in V^*,$$

the median set $\gamma(x, y, z)$ consists of exactly one vertex; equivalently,

$$|\gamma(x, y, z)| = 1 \qquad (\forall\, x, y, z \in V^* \text{ distinct}).$$

The unique vertex in $\gamma(x, y, z)$ is called the *median* of the triple $(x, y, z)$.

**Remark 4.19.2.** If

$$\gamma(x, y, z) = \{m\},$$

then the vertex $m$ lies simultaneously on a geodesic between $x$ and $y$, on a geodesic between $x$ and $z$, and on a geodesic between $y$ and $z$. Hence $m$ is the unique common geodesic mediator of the triple.

**Remark 4.19.3.** Every fuzzy tree is a basic example of a fuzzy median graph, since in a tree the geodesic structure forces the median set of any triple to be a singleton.

**Example 4.19.4** (A fuzzy median graph)**.** Let

$$V = \{v_1, v_2, v_3\},$$

and define a fuzzy graph

$$F = (V, \sigma, \mu)$$

by

$$\sigma(v_1) = 0.9, \qquad \sigma(v_2) = 0.8, \qquad \sigma(v_3) = 0.7,$$

and

$$\mu(v_1, v_2) = 0.6, \qquad \mu(v_2, v_3) = 0.5, \qquad \mu(v_1, v_3) = 0,$$

with

$$\mu(v_i, v_j) = \mu(v_j, v_i) \qquad (i, j \in \{1, 2, 3\}).$$

Then $F$ is a connected fuzzy graph, since its support graph is the path

$$v_1 - v_2 - v_3.$$

Moreover,

$$\mu(v_1, v_2) = 0.6 \le \min\{0.9, 0.8\} = 0.8, \qquad \mu(v_2, v_3) = 0.5 \le \min\{0.8, 0.7\} = 0.7,$$

so the defining condition of a fuzzy graph is satisfied.

Since all vertex-memberships are positive, we have

$$V^* = \{v_1, v_2, v_3\}.$$

We now verify that the two nonzero edges are strong.

For the edge $v_1v_2$, if we delete it, then $v_1$ and $v_2$ are disconnected in $F - v_1v_2$. Hence

$$\mathrm{CONN}_{F-v_1v_2}(v_1, v_2) = 0,$$

and therefore

$$\mu(v_1, v_2) = 0.6 \ge 0 = \mathrm{CONN}_{F-v_1v_2}(v_1, v_2).$$

Thus $v_1v_2$ is a strong edge.

Similarly, after deleting $v_2v_3$, the vertices $v_2$ and $v_3$ become disconnected, so

$$\mathrm{CONN}_{F-v_2v_3}(v_2, v_3) = 0,$$

and hence

$$\mu(v_2, v_3) = 0.5 \ge 0 = \mathrm{CONN}_{F-v_2v_3}(v_2, v_3).$$

Thus $v_2v_3$ is also a strong edge.

Therefore the path

$$v_1 - v_2 - v_3$$

is a strong path. Since it is the only path joining $v_1$ and $v_3$, it is also the strong geodesic between them. Hence the geodesic distances are

$$d_g(v_1, v_2) = 1, \qquad d_g(v_2, v_3) = 1, \qquad d_g(v_1, v_3) = 2.$$

Now compute the geodesic intervals.

First,

$$I_g(v_1, v_2) = \{v_1, v_2\},$$

because

$$d_g(v_1, v_2) = 1 = d_g(v_1, v_1) + d_g(v_1, v_2)$$

and

$$d_g(v_1, v_2) = 1 = d_g(v_1, v_2) + d_g(v_2, v_2),$$

while

$$d_g(v_1, v_2) \ne d_g(v_1, v_3) + d_g(v_3, v_2) = 2 + 1 = 3.$$

Similarly,

$$I_g(v_2, v_3) = \{v_2, v_3\}.$$

For the pair $v_1, v_3$, we have

$$d_g(v_1, v_3) = 2,$$

and

$$2 = d_g(v_1, v_1) + d_g(v_1, v_3) = 0 + 2,$$
$$2 = d_g(v_1, v_2) + d_g(v_2, v_3) = 1 + 1,$$
$$2 = d_g(v_1, v_3) + d_g(v_3, v_3) = 2 + 0.$$

Hence

$$I_g(v_1, v_3) = \{v_1, v_2, v_3\}.$$

Therefore, for the only triple of distinct vertices $(v_1, v_2, v_3)$,

$$\gamma(v_1, v_2, v_3) = I_g(v_1, v_2) \cap I_g(v_1, v_3) \cap I_g(v_2, v_3) = \{v_1, v_2\} \cap \{v_1, v_2, v_3\} \cap \{v_2, v_3\} = \{v_2\}.$$

Thus the median set consists of exactly one vertex, namely $v_2$. Since $(v_1, v_2, v_3)$ is the only triple of distinct vertices in $V^*$, it follows that $F$ is a fuzzy median graph.

For uncertainty-aware graph models, the most natural extension is obtained by defining medianity on the support graph induced by nonzero uncertainty degrees.

**Definition 4.19.5** (Support-evaluable uncertain model)**.** Let $M$ be an uncertain model with degree-domain

$$\mathrm{Dom}(M) \subseteq [0, 1]^k.$$

We say that $M$ is *support-evaluable* if it is equipped with a distinguished element

$$0_M \in \mathrm{Dom}(M),$$

called the *zero degree.*

**Definition 4.19.6** (Support graph)**.** Let

$$\mathcal{G}_M = (V, \sigma_M, \eta_M)$$

be an uncertain graph of type $M$.

Define the support vertex set by

$$V_M^* := \{\, v \in V : \sigma_M(v) \neq 0_M \,\},$$

and define the support edge set by

$$E_M^* := \left\{ \{u, v\} \in \binom{V_M^*}{2} : \eta_M(\{u, v\}) \neq 0_M \right\}.$$

The graph

$$G_{\mathrm{supp}}(\mathcal{G}_M) := (V_M^*, E_M^*)$$

is called the *support graph* of $\mathcal{G}_M$.

**Definition 4.19.7** (Geodesic interval and median set)**.** Let

$$\mathcal{G}_M = (V, \sigma_M, \eta_M)$$

be an uncertain graph of type $M$, and assume that its support graph

$$G_{\mathrm{supp}}(\mathcal{G}_M) = (V_M^*, E_M^*)$$

is connected.

For any two vertices

$$x, y \in V_M^*,$$

let

$$d_M(x, y)$$

denote the ordinary graph distance between $x$ and $y$ in the support graph

$$G_{\text{supp}}(\mathcal{G}_M).$$

The *geodesic interval* between $x$ and $y$ is defined by

$$I_M(x, y) := \{u \in V_M^* : d_M(x, y) = d_M(x, u) + d_M(u, y)\}.$$

For any three vertices

$$x, y, z \in V_M^*,$$

the *median set* of $x, y, z$ is defined by

$$\gamma_M(x, y, z) := I_M(x, y) \cap I_M(x, z) \cap I_M(y, z).$$

**Definition 4.19.8** (Uncertain median graph)**.** Let

$$\mathcal{G}_M = (V, \sigma_M, \eta_M)$$

be an uncertain graph of type $M$. Then $\mathcal{G}_M$ is called an *uncertain median graph* if its support graph

$$G_{\text{supp}}(\mathcal{G}_M)$$

is connected and, for every triple of distinct vertices

$$x, y, z \in V_M^*,$$

the median set $\gamma_M(x, y, z)$ consists of exactly one vertex; equivalently,

$$|\gamma_M(x, y, z)| = 1 \qquad (\forall\, x, y, z \in V_M^* \text{ distinct}).$$

The unique vertex in $\gamma_M(x, y, z)$ is called the *median* of the triple

$$(x, y, z).$$

**Theorem 4.19.9** (Well-definedness of the support graph)**.** *Let*

$$\mathcal{G}_M = (V, \sigma_M, \eta_M)$$

*be an uncertain graph of type $M$, where $M$ is support-evaluable with zero degree*

$$0_M \in \text{Dom}(M).$$

*Then the support vertex set*

$$V_M^* = \{\, v \in V : \sigma_M(v) \neq 0_M \,\}$$

*and the support edge set*

$$E_M^* = \left\{ \{u, v\} \in \binom{V_M^*}{2} : \eta_M(\{u, v\}) \neq 0_M \right\}$$

*are well-defined. Consequently,*

$$G_{\text{supp}}(\mathcal{G}_M) = (V_M^*, E_M^*)$$

*is a well-defined finite simple graph.*

*Proof.* Since

$$\sigma_M : V \to \mathrm{Dom}(M)$$

is a function and

$$0_M \in \mathrm{Dom}(M)$$

is fixed, for every

$$v \in V$$

the statement

$$\sigma_M(v) \neq 0_M$$

has a definite truth value. Hence

$$V_M^* = \{\, v \in V : \sigma_M(v) \neq 0_M \,\}$$

is a well-defined subset of $V$.

Similarly, since

$$\eta_M : \binom{V}{2} \to \mathrm{Dom}(M)$$

is a function, for every unordered pair

$$\{u, v\} \in \binom{V}{2}$$

the statement

$$\eta_M(\{u, v\}) \neq 0_M$$

has a definite truth value. Therefore the set

$$E_M^* = \left\{ \{u, v\} \in \binom{V_M^*}{2} : \eta_M(\{u, v\}) \neq 0_M \right\}$$

is well-defined.

By construction,

$$E_M^* \subseteq \binom{V_M^*}{2},$$

so every edge of $E_M^*$ is an unordered pair of distinct vertices of $V_M^*$. Hence

$$G_{\mathrm{supp}}(\mathcal{G}_M) = (V_M^*, E_M^*)$$

is a simple graph. Since $V$ is finite, the subset $V_M^*$ is finite, and therefore

$$G_{\mathrm{supp}}(\mathcal{G}_M)$$

is a finite simple graph. □

**Theorem 4.19.10** (Well-definedness of geodesic distance, interval, and median set)**.** *Let*

$$\mathcal{G}_M = (V, \sigma_M, \eta_M)$$

*be an uncertain graph of type $M$, and assume that its support graph*

$$G_{\mathrm{supp}}(\mathcal{G}_M) = (V_M^*, E_M^*)$$

*is connected.*

*Then, for every*

$$x, y \in V_M^*,$$

*the graph distance*

$$d_M(x, y)$$

*is well-defined. Consequently, for every*

$$x, y, z \in V_M^*,$$

*the geodesic interval*

$$I_M(x, y)$$

*and the median set*

$$\gamma_M(x, y, z)$$

*are well-defined.*

*Proof.* Since

$$G_{\text{supp}}(\mathcal{G}_M)$$

is a connected finite graph, for every two vertices

$$x, y \in V_M^*$$

there exists at least one path in

$$G_{\text{supp}}(\mathcal{G}_M)$$

joining $x$ and $y$.

If $x = y$, then the distance is

$$d_M(x, x) = 0.$$

Assume now that

$$x \neq y.$$

Because the graph is finite and simple, every shortest $x$-$y$ walk can be taken to be a simple path. Hence the set of lengths of all $x$-$y$ paths is a nonempty subset of

$$\{1, 2, \ldots, |V_M^*| - 1\}.$$

Therefore this set has a minimum element, and so

$$d_M(x, y)$$

is well-defined.

Now let

$$x, y \in V_M^*.$$

For each

$$u \in V_M^*,$$

the quantities

$$d_M(x, y), \qquad d_M(x, u), \qquad d_M(u, y)$$

are well-defined integers. Hence the predicate

$$d_M(x, y) = d_M(x, u) + d_M(u, y)$$

has a definite truth value. Therefore

$$I_M(x, y) = \{u \in V_M^* : d_M(x, y) = d_M(x, u) + d_M(u, y)\}$$

is a well-defined subset of $V_M^*$.

Finally, for

$$x, y, z \in V_M^*,$$

the three sets

$$I_M(x, y), \qquad I_M(x, z), \qquad I_M(y, z)$$

are well-defined subsets of $V_M^*$. Hence their intersection

$$\gamma_M(x, y, z) = I_M(x, y) \cap I_M(x, z) \cap I_M(y, z)$$

is also a well-defined subset of $V_M^*$. □

**Theorem 4.19.11** (Well-definedness of the notion of uncertain median graph)**.** *Let*

$$\mathcal{G}_M = (V, \sigma_M, \eta_M)$$

*be an uncertain graph of type $M$, where $M$ is support-evaluable. Then the statement*

*"$\mathcal{G}_M$ is an uncertain median graph"*

*is well-defined.*

*Proof.* By Theorem 1, the support graph

$$G_{\mathrm{supp}}(\mathcal{G}_M)$$

is well-defined. Hence the statement

$$\text{“}G_{\mathrm{supp}}(\mathcal{G}_M) \text{ is connected”}$$

has a definite truth value.

If the support graph is not connected, then by definition

$$\mathcal{G}_M$$

is not an uncertain median graph, so the notion is already determined.

Assume now that

$$G_{\mathrm{supp}}(\mathcal{G}_M)$$

is connected. Then, by Theorem 2, for every triple

$$x, y, z \in V_M^*$$

the median set

$$\gamma_M(x, y, z)$$

is well-defined.

Therefore, for every triple of distinct vertices

$$x, y, z \in V_M^*,$$

the statement

$$|\gamma_M(x, y, z)| = 1$$

has a definite truth value. Consequently, the universal statement

$$|\gamma_M(x, y, z)| = 1 \qquad (\forall\, x, y, z \in V_M^* \text{ distinct})$$

is well-defined.

Hence the assertion that $\mathcal{G}_M$ is an uncertain median graph is well-defined. □

**Remark 4.19.12.** The above definition is support-graph based. In particular, medianity is determined by the ordinary graph-metric structure of

$$G_{\mathrm{supp}}(\mathcal{G}_M).$$

This is the natural uncertainty-aware extension of classical median graphs obtained from Uncertain Sets.

**Remark 4.19.13.** If

$$\mathrm{Dom}(M) = [0, 1] \qquad \text{and} \qquad 0_M = 0,$$

then

$$V_M^* = \{\, v \in V : \sigma_M(v) > 0 \,\},$$

and

$$E_M^* = \Big\{ \{u, v\} \in \binom{V_M^*}{2} : \eta_M(\{u, v\}) > 0 \Big\}.$$

Hence the above construction reduces to the median-graph condition on the support graph of a scalar-valued uncertainty graph.

Related concepts include the modular graph [437, 438, 438], quasi-median graph [439–441], and pseudo-median graph [442–444]

## 4.20 Fuzzy chordal graphs

Fuzzy chordal graph is fuzzy graph where every cycle of length at least four contains a chord with membership not below the cycle's weakest edge [445–447].

**Definition 4.20.1** (Fuzzy Chordal Graph)**.** Let

$$G = (V, \sigma, \mu)$$

be a fuzzy graph, where

$$\sigma : V \to [0, 1], \qquad \mu : V \times V \to [0, 1],$$

such that

$$\mu(x, y) = \mu(y, x) \qquad \text{and} \qquad \mu(x, y) \leq \min\{\sigma(x), \sigma(y)\}$$

for all $x, y \in V$.

A *cycle* in $G$ is a sequence

$$C : x_0, x_1, \ldots, x_n$$

with $n \geq 3$, $x_0 = x_n$, $x_0, x_1, \ldots, x_{n-1}$ distinct, and

$$\mu(x_{i-1}, x_i) > 0 \qquad (i = 1, 2, \ldots, n).$$

Let

$$w(C) := \min_{1 \leq i \leq n} \mu(x_{i-1}, x_i)$$

denote the minimum edge-membership on the cycle $C$.

A *chord* of $C$ is an edge joining two nonconsecutive vertices of the cycle; that is, an edge

$$x_j x_k$$

such that

$$0 \leq j < k - 1 < n - 1$$

(or equivalently, $x_j$ and $x_k$ are not adjacent on the cycle).

Then $G$ is called a *fuzzy chordal graph* if for every cycle

$$C : x_0, x_1, \ldots, x_n$$

of length $n \geq 4$, there exist indices $j, k$ with

$$0 \leq j < k - 1 < n, \qquad (j, k) \neq (0, n - 1),$$

such that

$$\mu(x_j, x_k) \geq w(C) = \min_{1 \leq i \leq n} \mu(x_{i-1}, x_i).$$

In other words, every cycle of length at least 4 has a chord whose membership value is at least the weakest edge-membership on that cycle.

**Example 4.20.2** (A fuzzy chordal graph)**.** Let

$$V = \{v_1, v_2, v_3, v_4\},$$

and define a fuzzy graph

$$G = (V, \sigma, \mu)$$

by

$$\sigma(v_1) = 0.9, \qquad \sigma(v_2) = 0.8, \qquad \sigma(v_3) = 0.7, \qquad \sigma(v_4) = 0.8,$$

and

$$\mu(v_1, v_2) = 0.6, \qquad \mu(v_2, v_3) = 0.5, \qquad \mu(v_3, v_4) = 0.4, \qquad \mu(v_4, v_1) = 0.5,$$

$$\mu(v_1, v_3) = 0.4, \qquad \mu(v_2, v_4) = 0,$$

with

$$\mu(v_i, v_j) = \mu(v_j, v_i) \qquad (i, j \in \{1, 2, 3, 4\}).$$

First, $G$ is a fuzzy graph, because each edge-membership is bounded by the minimum of the memberships of its endpoints. Indeed,

$$\mu(v_1, v_2) = 0.6 \le \min\{0.9, 0.8\} = 0.8,$$

$$\mu(v_2, v_3) = 0.5 \le \min\{0.8, 0.7\} = 0.7,$$

$$\mu(v_3, v_4) = 0.4 \le \min\{0.7, 0.8\} = 0.7,$$

$$\mu(v_4, v_1) = 0.5 \le \min\{0.8, 0.9\} = 0.8,$$

and

$$\mu(v_1, v_3) = 0.4 \le \min\{0.9, 0.7\} = 0.7.$$

Now consider the cycle

$$C : v_1, v_2, v_3, v_4, v_1.$$

Its edge-memberships are

$$\mu(v_1, v_2) = 0.6, \qquad \mu(v_2, v_3) = 0.5, \qquad \mu(v_3, v_4) = 0.4, \qquad \mu(v_4, v_1) = 0.5,$$

so

$$w(C) = \min\{0.6, 0.5, 0.4, 0.5\} = 0.4.$$

The vertices $v_1$ and $v_3$ are nonconsecutive on the cycle, and

$$\mu(v_1, v_3) = 0.4 \ge w(C).$$

Hence $v_1v_3$ is a chord of $C$ whose membership value is at least the weakest edge-membership on the cycle.

Since the support graph of $G$ has only four vertices, this is the only cycle of length at least 4. Therefore every cycle of length at least 4 in $G$ has a suitable chord, and thus $G$ is a fuzzy chordal graph.

The notion of chordality for uncertain graphs is naturally defined through the support graph induced by nonzero uncertainty degrees.

**Definition 4.20.3** (Uncertain Chordal Graph)**.** Let

$$\mathcal{G}_M = (V, \sigma_M, \eta_M)$$

be an uncertain graph of type $M$, and let

$$G_{\text{supp}}(\mathcal{G}_M) = (V_M^*, E_M^*)$$

be its support graph.

A *chord* of a cycle

$$C : x_0, x_1, \ldots, x_n$$

in

$$G_{\text{supp}}(\mathcal{G}_M)$$

is an edge

$$\{x_j, x_k\} \in E_M^*$$

joining two nonconsecutive vertices of the cycle; that is,

$$0 \leq j < k \leq n - 1,$$

$$k \neq j + 1,$$

and

$$(j, k) \neq (0, n - 1).$$

Then

$$\mathcal{G}_M$$

is called an *uncertain chordal graph* if every cycle of length at least 4 in the support graph

$$G_{\text{supp}}(\mathcal{G}_M)$$

has a chord.

Equivalently,

$$\mathcal{G}_M$$

is an uncertain chordal graph if and only if

$$G_{\text{supp}}(\mathcal{G}_M)$$

is a chordal graph in the ordinary crisp sense.

**Theorem 4.20.4** (Well-definedness of chords in the support graph)**.** *Let*

$$\mathcal{G}_M = (V, \sigma_M, \eta_M)$$

*be an uncertain graph of type $M$, and let*

$$G_{\text{supp}}(\mathcal{G}_M) = (V_M^*, E_M^*)$$

*be its support graph.*

*Then, for every cycle*

$$C : x_0, x_1, \ldots, x_n$$

*in*

$$G_{\text{supp}}(\mathcal{G}_M),$$

*the statement*

$$\text{“}\{x_j, x_k\} \textit{ is a chord of } C\text{”}$$

*is well-defined.*

*Proof.* Since
$$G_{\text{supp}}(\mathcal{G}_M) = (V_M^*, E_M^*)$$
is already defined, the vertex set
$$V_M^*$$
and edge set
$$E_M^*$$
are well-defined.

Let
$$C : x_0, x_1, \dots, x_n$$
be a cycle in
$$G_{\text{supp}}(\mathcal{G}_M).$$
Then the vertices
$$x_0, x_1, \dots, x_{n-1}$$
are distinct, with
$$x_0 = x_n,$$
and each cycle edge
$$\{x_{i-1}, x_i\} \in E_M^* \qquad (i = 1, 2, \dots, n).$$

Now fix indices
$$0 \le j < k \le n - 1.$$
Because the cycle is already specified, the condition that $x_j$ and $x_k$ are nonconsecutive on the cycle is determined entirely by the indices $j$ and $k$, namely,
$$k \ne j + 1 \qquad \text{and} \qquad (j, k) \ne (0, n - 1).$$
Also, since
$$E_M^*$$
is well-defined, the statement
$$\{x_j, x_k\} \in E_M^*$$
has a definite truth value.

Therefore the conjunction of the two conditions
$$\{x_j, x_k\} \in E_M^*$$
and
$$x_j,\ x_k \text{ are nonconsecutive on } C$$
has a definite truth value. Hence the statement
$$\text{“}\{x_j, x_k\} \text{ is a chord of } C\text{”}$$
is well-defined. □

**Theorem 4.20.5** (Well-definedness of Uncertain Chordal Graph)**.** *Let*
$$\mathcal{G}_M = (V, \sigma_M, \eta_M)$$
*be an uncertain graph of type $M$, and let*
$$G_{\text{supp}}(\mathcal{G}_M)$$
*be its support graph.*

*Then the statement*
$$\text{“}\mathcal{G}_M \textit{ is an uncertain chordal graph”}$$
*is well-defined.*

*Proof.* Because

$$G_{\mathrm{supp}}(\mathcal{G}_M)$$

is a well-defined finite simple graph, the collection of all cycles in

$$G_{\mathrm{supp}}(\mathcal{G}_M)$$

is well-defined.

For every cycle

$$C$$

of length at least 4, by the previous theorem the statement

$$\text{“}C \text{ has a chord”}$$

is well-defined.

Therefore the universal statement

$$\text{“every cycle of length at least 4 in } G_{\mathrm{supp}}(\mathcal{G}_M) \text{ has a chord”}$$

has a definite truth value.

Hence the assertion

$$\text{“}\mathcal{G}_M \text{ is an uncertain chordal graph”}$$

is well-defined. □

**Remark 4.20.6.** The above definition is support-based. Thus chordality is determined by the ordinary graph structure of

$$G_{\mathrm{supp}}(\mathcal{G}_M),$$

rather than by direct comparison among uncertainty degrees on the edges of a cycle.

This is the natural model-independent extension obtained from Uncertain Sets. If one wishes to compare edge degrees quantitatively, then additional order or evaluation structures on the uncertainty model would be required.

**Remark 4.20.7.** If the uncertain model is the ordinary fuzzy model with zero degree 0, then

$$G_{\mathrm{supp}}(\mathcal{G}_M)$$

is exactly the support graph of the fuzzy graph. Hence the above notion reduces to the ordinary chordal-graph condition on the support graph of a fuzzy graph.

## 4.21 Uncertain Line Graph

Fuzzy line graph transforms each edge of a fuzzy graph into a vertex, with adjacency and memberships induced by incidence and original edge strengths naturally [155, 448, 449].

**Definition 4.21.1** (Fuzzy Line Graph)**.** Let

$$G = (V, \sigma, \mu)$$

be a fuzzy graph, where

$$\sigma : V \to [0, 1], \qquad \mu : V \times V \to [0, 1],$$

such that, for all $u, v \in V$,

$$\mu(u, v) = \mu(v, u), \qquad \mu(u, v) \le \min\{\sigma(u), \sigma(v)\}.$$

Define the support edge set of $G$ by

$$E^* := \{\{u,v\} \subseteq V : \ u \neq v, \ \mu(u,v) > 0\}.$$

The *fuzzy line graph* of $G$, denoted by

$$L_f(G) = (E^*, \tau, \eta),$$

is the fuzzy graph whose vertex set is $E^*$, whose vertex-membership function

$$\tau : E^* \to [0,1]$$

is defined by

$$\tau(\{u,v\}) := \mu(u,v) \qquad (\forall \{u,v\} \in E^*),$$

and whose edge-membership function

$$\eta : E^* \times E^* \to [0,1]$$

is defined by

$$\eta(e,f) := \begin{cases} \min\{\tau(e), \tau(f)\}, & e \neq f \text{ and } e \cap f \neq \varnothing, \\ 0, & \text{otherwise.} \end{cases}$$

Equivalently, if

$$e = \{u,v\}, \qquad f = \{v,w\}$$

are two distinct support edges of $G$ sharing a common endpoint, then

$$\eta(e,f) = \min\{\mu(u,v), \mu(v,w)\}.$$

Thus each support edge of $G$ becomes a vertex of $L_f(G)$, and two such vertices are adjacent in $L_f(G)$ precisely when the corresponding edges of $G$ are incident in the underlying support graph.

**Remark 4.21.2.** The above construction is a direct fuzzy extension of the classical line graph. Moreover, $L_f(G)$ is again a fuzzy graph, because for all $e, f \in E^*$,

$$\eta(e,f) \leq \min\{\tau(e), \tau(f)\},$$

and $\eta$ is symmetric.

**Remark 4.21.3.** If $G$ is crisp in the sense that

$$\sigma(v) = 1 \quad (\forall v \in V), \qquad \mu(u,v) \in \{0,1\} \quad (\forall u,v \in V),$$

then $L_f(G)$ reduces to the ordinary line graph of the support graph of $G$.

**Definition 4.21.4** (Line-admissible uncertain model)**.** Let $M$ be a support-evaluable uncertain model with degree-domain

$$\mathrm{Dom}(M)$$

and zero degree

$$0_M \in \mathrm{Dom}(M).$$

We say that $M$ is *line-admissible* if it is equipped with a map

$$\Lambda_M : \mathrm{Dom}(M) \times \mathrm{Dom}(M) \to \mathrm{Dom}(M),$$

called the *line-adjacency operator*, such that:

1. $$\Lambda_M(a,b) = \Lambda_M(b,a) \qquad (\forall a,b \in \mathrm{Dom}(M));$$

2. $$\Lambda_M(a,b) = 0_M \iff a = 0_M \text{ or } b = 0_M \qquad (\forall a,b \in \mathrm{Dom}(M)).$$

**Definition 4.21.5** (Uncertain Line Graph)**.** Let

$$\mathcal{G}_M = (V, \sigma_M, \eta_M)$$

be an uncertain graph of type $M$, and let

$$E_M^*$$

be its support edge set.

Assume that $M$ is line-admissible with line-adjacency operator

$$\Lambda_M : \mathrm{Dom}(M) \times \mathrm{Dom}(M) \to \mathrm{Dom}(M).$$

Define

$$E_L^* := \big\{\{e, f\} \in \binom{E_M^*}{2} : e \cap f \neq \varnothing\big\}.$$

The *uncertain line graph* of $\mathcal{G}_M$, denoted by

$$L_M(\mathcal{G}_M),$$

is the uncertain graph

$$L_M(\mathcal{G}_M) = (E_M^*, \tau_M, \lambda_M),$$

where the vertex uncertainty-degree function

$$\tau_M : E_M^* \to \mathrm{Dom}(M)$$

is defined by

$$\tau_M(e) := \eta_M(e) \qquad (\forall\, e \in E_M^*),$$

and the edge uncertainty-degree function

$$\lambda_M : \binom{E_M^*}{2} \to \mathrm{Dom}(M)$$

is defined by

$$\lambda_M(\{e, f\}) := \begin{cases} \Lambda_M\big(\tau_M(e), \tau_M(f)\big), & \text{if } e \cap f \neq \varnothing, \\ 0_M, & \text{if } e \cap f = \varnothing. \end{cases}$$

Equivalently, each support edge of $\mathcal{G}_M$ becomes a vertex of $L_M(\mathcal{G}_M)$, and two such vertices are adjacent precisely when the corresponding support edges of $\mathcal{G}_M$ are incident in the support graph.

**Theorem 4.21.6** (Well-definedness of the line-edge set)**.** *Let*

$$\mathcal{G}_M = (V, \sigma_M, \eta_M)$$

*be an uncertain graph of type $M$, and let*

$$E_M^*$$

*be its support edge set. Then the set*

$$E_L^* = \big\{\{e, f\} \in \binom{E_M^*}{2} : e \cap f \neq \varnothing\big\}$$

*is well-defined.*

*Proof.* Since

$$E_M^*$$

is the support edge set of $\mathcal{G}_M$, it is a well-defined set of unordered pairs of distinct vertices of $V$. Hence

$$\binom{E_M^*}{2}$$

is a well-defined set of unordered pairs of distinct support edges.

Now let

$$\{e, f\} \in \binom{E_M^*}{2}.$$

Because $e$ and $f$ are sets, the statement

$$e \cap f \neq \varnothing$$

has a definite truth value. Therefore the subset

$$E_L^* = \left\{\{e, f\} \in \binom{E_M^*}{2} : e \cap f \neq \varnothing\right\}$$

is well-defined. □

**Theorem 4.21.7** (Well-definedness of Uncertain Line Graph)**.** *Let*

$$\mathcal{G}_M = (V, \sigma_M, \eta_M)$$

*be an uncertain graph of type $M$, and assume that $M$ is line-admissible with line-adjacency operator*

$$\Lambda_M : \mathrm{Dom}(M) \times \mathrm{Dom}(M) \to \mathrm{Dom}(M).$$

*Then*

$$L_M(\mathcal{G}_M) = (E_M^*, \tau_M, \lambda_M)$$

*is a well-defined uncertain graph.*

*Proof.* Since

$$E_M^*$$

is a well-defined set, the function

$$\tau_M : E_M^* \to \mathrm{Dom}(M), \qquad \tau_M(e) := \eta_M(e),$$

is well-defined because

$$\eta_M$$

is already a well-defined function on the support edge set of $\mathcal{G}_M$.

Next, consider

$$\lambda_M : \binom{E_M^*}{2} \to \mathrm{Dom}(M).$$

Let

$$\{e, f\} \in \binom{E_M^*}{2}.$$

If

$$e \cap f = \varnothing,$$

then

$$\lambda_M(\{e, f\}) = 0_M$$

is uniquely determined.

If

$$e \cap f \neq \varnothing,$$

then

$$\tau_M(e), \tau_M(f) \in \mathrm{Dom}(M),$$

so

$$\Lambda_M\big(\tau_M(e), \tau_M(f)\big) \in \mathrm{Dom}(M)$$

is well-defined. Because the argument $\{e, f\}$ is an unordered pair, one must verify independence of ordering. But

$$\Lambda_M\big(\tau_M(e), \tau_M(f)\big) = \Lambda_M\big(\tau_M(f), \tau_M(e)\big)$$

by symmetry of $\Lambda_M$. Hence

$$\lambda_M(\{e, f\})$$

does not depend on the order in which $e$ and $f$ are written.

Therefore

$$\lambda_M$$

is a well-defined function on

$$\binom{E_M^*}{2}.$$

Consequently,

$$L_M(\mathcal{G}_M) = (E_M^*, \tau_M, \lambda_M)$$

is a well-defined uncertain graph. □

**Theorem 4.21.8** (Support graph of the uncertain line graph)**.** *Let*

$$\mathcal{G}_M = (V, \sigma_M, \eta_M)$$

*be an uncertain graph of type $M$, let*

$$G_{\mathrm{supp}}(\mathcal{G}_M) = (V_M^*, E_M^*)$$

*be its support graph, and let $M$ be line-admissible.*

*Then the support graph of*

$$L_M(\mathcal{G}_M)$$

*is exactly the ordinary line graph of*

$$G_{\mathrm{supp}}(\mathcal{G}_M).$$

*That is,*

$$G_{\mathrm{supp}}\big(L_M(\mathcal{G}_M)\big) = L\big(G_{\mathrm{supp}}(\mathcal{G}_M)\big).$$

*Proof.* The vertex set of

$$L_M(\mathcal{G}_M)$$

is

$$E_M^*.$$

For each

$$e \in E_M^*,$$

one has

$$\tau_M(e) = \eta_M(e) \neq 0_M$$

by the definition of support edge set. Hence every vertex of

$$L_M(\mathcal{G}_M)$$

belongs to its support vertex set.

Now let

$$\{e,f\} \in \binom{E_M^*}{2}.$$

By definition,

$$\lambda_M(\{e,f\}) = \begin{cases} \Lambda_M\big(\tau_M(e),\tau_M(f)\big), & e\cap f \neq \varnothing, \\ 0_M, & e\cap f = \varnothing. \end{cases}$$

Since

$$\tau_M(e) \neq 0_M \qquad \text{and} \qquad \tau_M(f) \neq 0_M,$$

the support-preserving property of $\Lambda_M$ implies that

$$\Lambda_M\big(\tau_M(e),\tau_M(f)\big) \neq 0_M.$$

Therefore

$$\lambda_M(\{e,f\}) \neq 0_M \iff e\cap f \neq \varnothing.$$

Hence the support edge set of

$$L_M(\mathcal{G}_M)$$

is precisely

$$\Big\{\{e,f\} \in \binom{E_M^*}{2} : e\cap f \neq \varnothing\Big\},$$

which is exactly the edge set of the ordinary line graph of

$$G_{\mathrm{supp}}(\mathcal{G}_M) = (V_M^*, E_M^*).$$

Thus

$$G_{\mathrm{supp}}\big(L_M(\mathcal{G}_M)\big) = L\big(G_{\mathrm{supp}}(\mathcal{G}_M)\big).$$

□

**Remark 4.21.9.** If

$$\mathrm{Dom}(M) = [0,1], \qquad 0_M = 0,$$

and

$$\Lambda_M(a,b) = \min\{a,b\},$$

then the above construction reduces to the usual fuzzy line graph.

Representative line-graph concepts under uncertainty-aware graph frameworks are listed in Table 4.13.

Besides uncertain line graphs, several related concepts are also known, including iterated line graphs [461–464], total graphs [465, 466], iterated total graphs [467, 468], line hypergraphs [469, 470], and line superhypergraphs [471].

## 4.22 Uncertain HyperGraph

A fuzzy hypergraph generalizes a hypergraph by assigning membership degrees to vertices or hyperedges, thereby modeling uncertain higher-order relationships among multiple entities simultaneously in networks [36].

**Definition 4.22.1** (Uncertain HyperGraph)**.** Let $H = (V,E)$ be a hypergraph and let $M$ be an uncertain model with degree–domain $\mathrm{Dom}(M)$. An *Uncertain HyperGraph of type* $M$ is a triple

$$\mathcal{H}_M = (V,E,\mu_M),$$

where

$$\mu_M : V \cup E \longrightarrow \mathrm{Dom}(M)$$

assigns an uncertainty degree to each vertex $v \in V$ and each hyperedge $e \in E$.

As in the graph case, possible relations between vertex and hyperedge degrees (for instance, bounds of $\mu_M(e)$ in terms of $\mu_M(v)$ for $v \in e$) are governed by the chosen model $M$ and its constraints.

**Remark 4.22.2.** For suitable choices of $M$, this framework yields fuzzy hypergraphs, intuitionistic fuzzy hypergraphs, neutrosophic hypergraphs, plithogenic hypergraphs, and many further extensions. We present the catalogue of uncertainty-hypergraph families (Uncertain HyperGraphs) by the dimension $k$ of the degree-domain $\mathrm{Dom}(M) \subseteq [0,1]^k$ in Table 4.14.

Table 4.13: Representative line-graph concepts under uncertainty-aware graph frameworks, classified by the dimension $k$ of the information attached to vertices and/or edges.

| $k$ | **Line-graph concept** | **Typical coordinate form** | **Canonical information attached to vertices/edges** |
|---|---|---|---|
| 1 | Fuzzy Line Graph | $\mu$ | A line graph studied in a fuzzy framework, where each vertex and edge is associated with a single membership degree in $[0, 1]$. |
| 2 | Intuitionistic Fuzzy Line Graph [450–453] | $(\mu, \nu)$ | A line graph defined in an intuitionistic fuzzy framework, where each vertex and edge carries a membership degree and a non-membership degree, usually satisfying $\mu + \nu \leq 1$. |
| 2 | Bipolar Fuzzy Line Graph [454–456] | $(\mu^+, \mu^-)$ | A line graph defined in a bipolar fuzzy framework, where each vertex and edge is described by a positive membership degree and a negative membership degree. |
| 3 | Picture Fuzzy Line Graph [457, 458] | $(\mu, \eta, \nu)$ | A line graph defined in a picture fuzzy framework, where each vertex and edge is described by positive, neutral, and negative membership degrees, usually satisfying $\mu + \eta + \nu \leq 1$. |
| 3 | Neutrosophic Line Graph [459, 460] | $(T, I, F)$ | A line graph defined in a neutrosophic framework, where each vertex and edge is described by truth, indeterminacy, and falsity degrees. |

Table 4.14: A catalogue of uncertainty-hypergraph families (Uncertain HyperGraphs) by the dimension $k$ of the degree-domain $\mathrm{Dom}(M) \subseteq [0, 1]^k$.

| $k$ | Representative uncertainty-hypergraph family (type $M$ with $\mathrm{Dom}(M) \subseteq [0, 1]^k$) |
|---|---|
| 1 | *Fuzzy HyperGraph [472–474]:* $\mu_M : V \cup E \to [0, 1]$. |
| 2 | *Intuitionistic-fuzzy HyperGraph [475–477]:* $\mu_M : V \cup E \to [0, 1]^2$ (e.g., (membership, non-membership)). |
| 3 | *Neutrosophic HyperGraph [68, 478–480]:* $\mu_M : V \cup E \to [0, 1]^3$ (e.g., $(T, I, F)$). |
| 4 | *Quadripartitioned Neutrosophic / four-component uncertainty HyperGraph:* $\mu_M : V \cup E \to [0, 1]^4$. |
| 5 | *Pentapartitioned Neutrosophic / five-component uncertainty HyperGraph:* $\mu_M : V \cup E \to [0, 1]^5$. |
| $k$ | *k-component uncertainty HyperGraph:* $\mu_M : V \cup E \to \mathrm{Dom}(M) \subseteq [0, 1]^k$ (model-specific semantics). |

## 4.23 Uncertain SuperHyperGraph

A SuperHyperGraph generalizes graphs and hypergraphs by allowing vertices and hyperedges to themselves be sets, enabling hierarchical, multilevel, higher-order, and recursively nested relations [90, 481–485]. A fuzzy superhypergraph generalizes a hypergraph by assigning membership degrees to supervertices and superhyperedges, thereby modeling uncertain hierarchical higher-order relationships in complex networked systems effectively [486, 487].

**Definition 4.23.1** (Uncertain $n$-SuperHyperGraph)**.** [486] Let $V_0$ be a finite base set and let $n \in \mathbb{N}_0$. Assume that an $n$-SuperHyperGraph on $V_0$ is given by

$$\mathrm{SHG}^{(n)} = (V_n, E),$$

where

$$\emptyset \neq V_n \subseteq \mathcal{P}^n(V_0) \quad \text{and} \quad \emptyset \neq E \subseteq \mathcal{P}(V_n) \setminus \{\emptyset\},$$

so that each $n$-superedge $e \in E$ is a nonempty subset of the $n$-supervertex set $V_n$.

Let $M$ be a fixed uncertain model with degree–domain $\mathrm{Dom}(M) \subseteq [0, 1]^k$. An *Uncertain n-SuperHyperGraph of type M* is a triple

$$\mathcal{S}_M^{(n)} = (V_n, E, \mu_M),$$

where

$$\mu_M : V_n \cup E \longrightarrow \mathrm{Dom}(M)$$

assigns to each $n$-supervertex $v \in V_n$ and each $n$-superedge $e \in E$ an uncertainty degree $\mu_M(v)$ or $\mu_M(e)$ in $\mathrm{Dom}(M)$.

Any additional relations between the degrees of $n$-superedges and the degrees of the $n$-supervertices they contain (for example, model- specific bounds or aggregations) are imposed by the chosen uncertain model $M$ and are not fixed at the level of this general definition.

For $n = 0$ and $V_0 = V_n$, the above notion reduces to an Uncertain HyperGraph of type $M$.

**Remark 4.23.2.** Particular choices of the model $M$ recover well–known uncertain SuperHyperGraph types:

- Fuzzy $n$-SuperHyperGraphs (when $M$ is fuzzy);
- Intuitionistic fuzzy, neutrosophic, and plithogenic $n$-SuperHyperGraphs for the corresponding models $M$;
- More exotic variants (e.g. $q$-rung orthopair, picture fuzzy, refined neutrosophic) are obtained by choosing the appropriate degree–domain $\mathrm{Dom}(M)$.

Regarding the catalogue of uncertainty-superhypergraph families (Uncertain $n$-SuperHyperGraphs) by the dimension $k$ of the degree-domain $\mathrm{Dom}(M) \subseteq [0,1]^k$, we list them in Table 4.15.

Table 4.15: A catalogue of uncertainty-superhypergraph families (Uncertain $n$-SuperHyperGraphs) by the dimension $k$ of the degree-domain $\mathrm{Dom}(M) \subseteq [0,1]^k$.

| $k$ | Representative uncertainty-superhypergraph family (type $M$ with $\mathrm{Dom}(M) \subseteq [0,1]^k$) |
|---|---|
| 1 | *Fuzzy n-SuperHyperGraph [488]:* $\mu_M : V_n \cup E \to [0,1]$. |
| 2 | *Intuitionistic-fuzzy n-SuperHyperGraph [488, 489]:* $\mu_M : V_n \cup E \to [0,1]^2$ (e.g., $(\text{membership}, \text{non-membership})$). |
| 3 | *Neutrosophic n-SuperHyperGraph [490–492]:* $\mu_M : V_n \cup E \to [0,1]^3$ (e.g., $(T, I, F)$). |
| 4 | *Quadripartitioned / four-component uncertainty n-SuperHyperGraph:* $\mu_M : V_n \cup E \to [0,1]^4$. |
| $k$ | *k-component uncertainty n-SuperHyperGraph:* $\mu_M : V_n \cup E \to \mathrm{Dom}(M) \subseteq [0,1]^k$ (model-specific semantics). |

## 4.24 Meta-Uncertain Graph

A meta-fuzzy graph is a fuzzy graph whose vertices are themselves fuzzy graphs, with meta-edge memberships representing uncertain higher-level relations between them.

**Definition 4.24.1** (Meta-Fuzzy Graph)**.** Let $\mathcal{FG}$ be a nonempty universe of fuzzy graphs, and let

$$\mathcal{R}$$

be a nonempty family of fuzzy relations on $\mathcal{FG}$, that is, each

$$R \in \mathcal{R}$$

is a map

$$R : \mathcal{FG} \times \mathcal{FG} \to [0,1].$$

A *Meta-Fuzzy Graph* over $(\mathcal{FG}, \mathcal{R})$ is a triple

$$M = (\sigma_M, \mu_M, L_M),$$

where

- $\sigma_M : \mathcal{FG} \to [0,1]$ is the *meta-vertex membership function*,
- $\mu_M : \mathcal{FG} \times \mathcal{FG} \to [0,1]$ is the *meta-edge membership function*,
- $L_M : \mathcal{FG} \times \mathcal{FG} \to \mathcal{P}_{\mathrm{fin}}(\mathcal{R})$ is a *label selector*,

such that, for all $F, G \in \mathcal{FG}$,

$$\mu_M(F,G) \le \min\{\sigma_M(F), \sigma_M(G)\},$$

and

$$\mu_M(F,G) \le \sup_{R \in L_M(F,G)} R(F,G),$$

with the convention that

$$\sup \varnothing := 0.$$

The *support* of $M$ is

$$V(M) := \{F \in \mathcal{FG} : \sigma_M(F) > 0\},$$

and the associated crisp underlying meta-graph has vertex set $V(M)$ and arc set

$$A(M) := \{(F,G) \in V(M) \times V(M) : \mu_M(F,G) > 0\}.$$

A meta-uncertain graph is an uncertain graph whose vertices are themselves uncertain graphs, and whose meta-edge degrees describe higher-level uncertain relations among them.

**Definition 4.24.2** (Meta-Evaluable Uncertain Model)**.** Let $N$ be an uncertain model with degree-domain

$$\mathrm{Dom}(N) \subseteq [0,1]^{\ell}.$$

We say that $N$ is *meta-evaluable* if it is equipped with the following additional data:

1. a distinguished element
$$0_N \in \mathrm{Dom}(N),$$
called the *zero degree*;
2. a partial order
$$\preceq_N \subseteq \mathrm{Dom}(N) \times \mathrm{Dom}(N);$$
3. a symmetric binary operator
$$\Gamma_N : \mathrm{Dom}(N) \times \mathrm{Dom}(N) \longrightarrow \mathrm{Dom}(N),$$
called the *meta-edge compatibility operator*, satisfying
$$\Gamma_N(a,b) = \Gamma_N(b,a) \qquad (\forall\, a,b \in \mathrm{Dom}(N));$$
4. a finite aggregation operator
$$\Omega_N : \mathcal{P}_{\mathrm{fin}}(\mathrm{Dom}(N)) \longrightarrow \mathrm{Dom}(N),$$
satisfying
$$\Omega_N(\varnothing) = 0_N.$$

**Definition 4.24.3** (Meta-Uncertain Graph)**.** Let $M$ be an uncertain model, and let

$$\mathfrak{UG}_M$$

be a nonempty universe of uncertain graphs of type $M$.

Let $N$ be a meta-evaluable uncertain model, and let

$$\mathcal{R}$$

be a nonempty family of symmetric $N$-valued binary relations on $\mathfrak{UG}_M$, that is, each

$$R \in \mathcal{R}$$

is a map

$$R : \mathfrak{UG}_M \times \mathfrak{UG}_M \to \mathrm{Dom}(N)$$

such that

$$R(F,G) = R(G,F) \qquad (\forall\, F,G \in \mathfrak{UG}_M).$$

A *Meta-Uncertain Graph* over $(\mathfrak{UG}_M, \mathcal{R})$ of meta-type $N$ is a triple

$$\mathfrak{M}_N = (\Sigma_N, \mathrm{H}_N, L_N),$$

where

- $$\Sigma_N : \mathfrak{UG}_M \to \mathrm{Dom}(N)$$
  is the *meta-vertex uncertainty-degree function*;
- $$\mathrm{H}_N : \mathfrak{UG}_M \times \mathfrak{UG}_M \to \mathrm{Dom}(N)$$
  is the *meta-edge uncertainty-degree function*;
- $$L_N : \mathfrak{UG}_M \times \mathfrak{UG}_M \to \mathcal{P}_{\mathrm{fin}}(\mathcal{R})$$
  is a *label selector.*

These data are required to satisfy, for all $F, G \in \mathfrak{UG}_M$,

$$\mathrm{H}_N(F,G) = \mathrm{H}_N(G,F),$$

$$L_N(F,G) = L_N(G,F),$$

$$\mathrm{H}_N(F,G) \preceq_N \Gamma_N\big(\Sigma_N(F), \Sigma_N(G)\big),$$

and

$$\mathrm{H}_N(F,G) \preceq_N \Omega_N\big(\{\, R(F,G) \mid R \in L_N(F,G)\,\}\big).$$

The *support vertex set* of $\mathfrak{M}_N$ is

$$V(\mathfrak{M}_N) := \{\, F \in \mathfrak{UG}_M \mid \Sigma_N(F) \neq 0_N \,\},$$

and the associated crisp underlying meta-graph has edge set

$$E(\mathfrak{M}_N) := \big\{\{F,G\} \subseteq V(\mathfrak{M}_N) \mid F \neq G,\ \mathrm{H}_N(F,G) \neq 0_N\big\}.$$

## 4.25 Uncertain MultiGraph

A fuzzy multigraph assigns membership degrees to vertices and multiple parallel edges, modeling uncertain relationships where several distinct fuzzy connections may exist between two vertices [493, 494].

**Definition 4.25.1** (Fuzzy Multigraph)**.** [493, 494] Let $V$ be a finite nonempty set of vertices, and let $L$ be a finite nonempty set of edge labels. A *fuzzy multigraph* is a triple

$$\Omega = (\sigma, \mu, \iota),$$

where

$$\sigma : V \to [0,1]$$

is a fuzzy subset of vertices,

$$\mu : L \to [0,1]$$

is a fuzzy subset of edges, and

$$\iota : L \to \big\{\{u,v\} \mid u, v \in V\big\}$$

is an incidence map assigning to each edge label $e \in L$ its unordered pair of end vertices, such that for every edge $e \in L$ with

$$\iota(e) = \{u, v\},$$

we have

$$\mu(e) \leq \min\{\sigma(u), \sigma(v)\}.$$

Two distinct fuzzy edges $e_1, e_2 \in L$ are said to be *parallel* if

$$\iota(e_1) = \iota(e_2).$$

The fuzzy multigraph $\Omega$ is called a *fuzzy multigraph* whenever parallel edges are allowed; that is, there may exist distinct edges $e_1 \neq e_2$ such that

$$\iota(e_1) = \iota(e_2).$$

If no such pair exists, then $\Omega$ is a *fuzzy simple graph.*

An uncertain multigraph assigns uncertainty degrees to vertices and to edge-identifiers, allowing several distinct parallel uncertain edges between the same pair of vertices.

**Definition 4.25.2** (Uncertain MultiGraph)**.** Let

$$V$$

be a finite nonempty set of vertices, and let

$$L$$

be a finite nonempty set of edge identifiers.

Let

$$\iota : L \longrightarrow \big\{\{u, v\} \mid u, v \in V\big\}$$

be an incidence map assigning to each edge identifier $e \in L$ an unordered pair of end vertices.

Let $M$ be a fixed uncertain model with degree-domain

$$\mathrm{Dom}(M) \subseteq [0, 1]^k.$$

An *Uncertain MultiGraph of type M* is a quadruple

$$\Omega_M = (V, L, \sigma_M, \eta_M),$$

or, when the incidence map is displayed explicitly,

$$\Omega_M = (V, L, \iota, \sigma_M, \eta_M),$$

where

$$\sigma_M : V \longrightarrow \mathrm{Dom}(M)$$

and

$$\eta_M : L \longrightarrow \mathrm{Dom}(M)$$

are uncertainty-degree functions on the vertex set and the edge-identifier set, respectively.

Equivalently,

$$(V, \sigma_M)$$

is an Uncertain Set of type $M$ on $V$, and

$$(L, \eta_M)$$

is an Uncertain Set of type $M$ on $L$.

For each vertex $v \in V$, the value

$$\sigma_M(v) \in \mathrm{Dom}(M)$$

represents the uncertainty degree of $v$, and for each edge identifier $e \in L$, the value

$$\eta_M(e) \in \mathrm{Dom}(M)$$

represents the uncertainty degree of the edge whose endpoints are given by $\iota(e)$.

Two distinct edges $e_1, e_2 \in L$ are said to be *parallel* if

$$\iota(e_1) = \iota(e_2).$$

Thus parallel uncertain edges are allowed whenever there exist distinct $e_1, e_2 \in L$ such that

$$\iota(e_1) = \iota(e_2).$$

If desired, one may additionally impose model-specific compatibility conditions between

$$\eta_M(e) \quad \text{and} \quad \sigma_M(u),\ \sigma_M(v) \quad \text{for} \quad \iota(e) = \{u, v\},$$

but such conditions depend on the chosen uncertain model $M$ and are not fixed at the level of this general definition.

Table 4.16: Representative multigraph concepts under uncertainty-aware graph frameworks, classified by the dimension $k$ of the information attached to vertices and/or edges.

| $k$ | Multigraph concept | Typical coordinate form | Canonical information attached to vertices/edges |
|---|---|---|---|
| 1 | Fuzzy Multigraph | $\mu$ | A multigraph studied in a fuzzy framework, where each vertex and edge is associated with a single membership degree in $[0, 1]$. |
| 2 | Vague Multigraph [495, 496] | $(t, f)$ | A multigraph defined in a vague framework, where each vertex and edge is characterized by a truth-membership degree and a falsity-membership degree, typically with $t + f \leq 1$. |
| 2 | Intuitionistic Fuzzy Multigraph [497, 498] | $(\mu, \nu)$ | A multigraph defined in an intuitionistic fuzzy framework, where each vertex and edge carries a membership degree and a non-membership degree, usually satisfying $\mu + \nu \leq 1$. |
| 2 | Bipolar Fuzzy Multigraph [499–501] | $(\mu^+, \mu^-)$ | A multigraph defined in a bipolar fuzzy framework, where each vertex and edge is described by a positive membership degree and a negative membership degree. |
| 3 | Picture Fuzzy Multigraph [457, 502] | $(\mu, \eta, \nu)$ | A multigraph defined in a picture fuzzy framework, where each vertex and edge is described by positive, neutral, and negative membership degrees, usually satisfying $\mu + \eta + \nu \leq 1$. |
| 3 | Neutrosophic Multigraph [503, 504] | $(T, I, F)$ | A multigraph defined in a neutrosophic framework, where each vertex and edge is described by truth, indeterminacy, and falsity degrees. |

Representative multigraph concepts under uncertainty-aware graph frameworks are listed in Table 4.16.

Besides uncertain multigraphs, several related concepts are also known, including directed multigraphs [505], weighted multigraphs [506], complete multigraphs [507–509], bipartite multigraphs [510, 511], regular multigraphs [512, 513], soft multigraphs [514], and multihypergraphs [8, 515–517].

## 4.26 Uncertain Bipartite Graph

A fuzzy bipartite graph partitions vertices into two disjoint fuzzy sets, allowing positive edge memberships only between the parts, thereby faithfully modeling uncertain bipartite relationships [518, 519].

**Definition 4.26.1** (Fuzzy Bipartite Graph)**.** [518, 519] Let

$$G = (V, \sigma, \mu)$$

be a fuzzy graph, where

$$\sigma : V \to [0, 1], \qquad \mu : V \times V \to [0, 1], \qquad \mu(u, v) \leq \min\{\sigma(u), \sigma(v)\} \quad (\forall\, u, v \in V),$$

and assume that $\mu$ is symmetric.

Then $G$ is called a *fuzzy bipartite graph* if there exist two nonempty disjoint sets

$$V_1, V_2 \subseteq V$$

such that

$$V = V_1 \cup V_2, \qquad V_1 \cap V_2 = \varnothing,$$

and

$$\mu(u,v) = 0$$

whenever either

$$u, v \in V_1 \qquad \text{or} \qquad u, v \in V_2.$$

Equivalently, every edge with positive membership joins a vertex of $V_1$ to a vertex of $V_2$.

An uncertain bipartite graph is an interval-valued generalization of a fuzzy bipartite graph, in which the uncertainty of vertices and edges is represented by closed subintervals of $[0,1]$, and nonzero edge-memberships are allowed only between two disjoint parts.

First, let

$$\mathbb{I}([0,1]) := \{[a^-, a^+] \subseteq [0,1] : \ 0 \le a^- \le a^+ \le 1\}.$$

For

$$A = [a^-, a^+], \qquad B = [b^-, b^+] \in \mathbb{I}([0,1]),$$

define the partial order

$$A \preceq B \iff a^- \le b^- \ \text{ and } \ a^+ \le b^+,$$

and define the interval meet by

$$A \wedge B := \left[\min\{a^-, b^-\}, \min\{a^+, b^+\}\right].$$

**Definition 4.26.2** (Uncertain Bipartite Graph)**.** Let $V$ be a nonempty set. An *uncertain bipartite graph* is a triple

$$G = (V, \Sigma, M),$$

where

$$\Sigma : V \to \mathbb{I}([0,1]), \qquad M : V \times V \to \mathbb{I}([0,1]),$$

such that the following conditions hold:

1. $M$ is symmetric, that is,
$$M(u,v) = M(v,u) \qquad (\forall\, u, v \in V).$$
2. For all $u, v \in V$,
$$M(u,v) \preceq \Sigma(u) \wedge \Sigma(v).$$
3. There exist two nonempty disjoint subsets
$$V_1, V_2 \subseteq V$$
such that
$$V = V_1 \cup V_2, \qquad V_1 \cap V_2 = \varnothing,$$
and
$$M(u,v) = [0,0]$$
whenever either
$$u, v \in V_1 \qquad \text{or} \qquad u, v \in V_2.$$

In this case, $(V_1, V_2)$ is called a *bipartition* of $G$. Equivalently, every edge with nonzero uncertain membership joins a vertex of $V_1$ to a vertex of $V_2$.

**Theorem 4.26.3** (Well-definedness of uncertain bipartite graphs)**.** *Let $V_1$ and $V_2$ be two nonempty disjoint sets, and put*

$$V := V_1 \cup V_2.$$

*Let*

$$\Sigma : V \to \mathbb{I}([0,1])$$

*be any interval-valued vertex-membership function, and let*

$$M_{12} : V_1 \times V_2 \to \mathbb{I}([0,1])$$

*satisfy*

$$M_{12}(u,v) \preceq \Sigma(u) \wedge \Sigma(v) \qquad (\forall\, u \in V_1,\ \forall\, v \in V_2).$$

*Define*

$$M : V \times V \to \mathbb{I}([0,1])$$

*by*

$$M(x,y) := \begin{cases} M_{12}(x,y), & x \in V_1,\ y \in V_2, \\ M_{12}(y,x), & x \in V_2,\ y \in V_1, \\ [0,0], & x,y \in V_1 \text{ or } x,y \in V_2. \end{cases}$$

*Then $M$ is well-defined, and*

$$G = (V, \Sigma, M)$$

*is an uncertain bipartite graph with bipartition $(V_1, V_2)$.*

*Proof.* We first show that $M$ is well-defined. Since $V_1 \cap V_2 = \varnothing$ and $V = V_1 \cup V_2$, every ordered pair $(x,y) \in V \times V$ falls into exactly one of the following mutually exclusive cases:

$$(x,y) \in V_1 \times V_2, \qquad (x,y) \in V_2 \times V_1, \qquad (x,y) \in V_1 \times V_1, \qquad (x,y) \in V_2 \times V_2.$$

Hence the above piecewise definition assigns a unique value $M(x,y)$ to every $(x,y) \in V \times V$.

Next, we verify that $M(x,y) \in \mathbb{I}([0,1])$ for all $x,y \in V$. If $(x,y) \in V_1 \times V_2$, then

$$M(x,y) = M_{12}(x,y) \in \mathbb{I}([0,1]).$$

If $(x,y) \in V_2 \times V_1$, then

$$M(x,y) = M_{12}(y,x) \in \mathbb{I}([0,1]).$$

If $x,y$ belong to the same part, then

$$M(x,y) = [0,0] \in \mathbb{I}([0,1]).$$

Therefore $M : V \times V \to \mathbb{I}([0,1])$ is a well-defined interval-valued function.

We now prove symmetry. If $x \in V_1$ and $y \in V_2$, then

$$M(x,y) = M_{12}(x,y)$$

and

$$M(y,x) = M_{12}(x,y),$$

so $M(x,y) = M(y,x)$. The same conclusion holds when $x \in V_2$ and $y \in V_1$. If $x,y$ lie in the same part, then both

$$M(x,y) = [0,0] \qquad \text{and} \qquad M(y,x) = [0,0].$$

Thus $M$ is symmetric on $V \times V$.

It remains to verify the membership constraint

$$M(x,y) \preceq \Sigma(x) \wedge \Sigma(y) \qquad (\forall\, x,y \in V).$$

If $x \in V_1$ and $y \in V_2$, then by hypothesis,

$$M(x, y) = M_{12}(x, y) \preceq \Sigma(x) \wedge \Sigma(y).$$

If $x \in V_2$ and $y \in V_1$, then

$$M(x, y) = M_{12}(y, x) \preceq \Sigma(y) \wedge \Sigma(x) = \Sigma(x) \wedge \Sigma(y).$$

If $x, y$ lie in the same part, then

$$M(x, y) = [0, 0].$$

Since $\Sigma(x), \Sigma(y) \in \mathbb{I}([0, 1])$, their meet

$$\Sigma(x) \wedge \Sigma(y) = \big[\min\{\Sigma^-(x), \Sigma^-(y)\}, \min\{\Sigma^+(x), \Sigma^+(y)\}\big]$$

has both endpoints in $[0, 1]$, and therefore

$$[0, 0] \preceq \Sigma(x) \wedge \Sigma(y).$$

Hence the constraint holds in all cases.

Finally, by construction,

$$M(x, y) = [0, 0]$$

whenever $x, y \in V_1$ or $x, y \in V_2$. Therefore all nonzero uncertain edge-memberships occur only between the two parts. Thus

$$G = (V, \Sigma, M)$$

satisfies all conditions, and so it is an uncertain bipartite graph with bipartition $(V_1, V_2)$. □

Representative bipartite-graph concepts under uncertainty-aware graph frameworks are listed in Table 4.17.

Table 4.17: Representative bipartite-graph concepts under uncertainty-aware graph frameworks, classified by the dimension $k$ of the information attached to vertices and/or edges.

| $k$ | **Bipartite-graph concept** | **Typical coordinate form** | **Canonical information attached to vertices/edges** |
|---|---|---|---|
| 1 | Fuzzy Bipartite Graph | $\mu$ | A bipartite graph studied in a fuzzy framework, where each vertex and edge is associated with a single membership degree in $[0, 1]$. |
| 2 | Intuitionistic Fuzzy Bipartite Graph | $(\mu, \nu)$ | A bipartite graph defined in an intuitionistic fuzzy framework, where each vertex and edge carries a membership degree and a non-membership degree, usually satisfying $\mu + \nu \leq 1$. |
| 3 | Neutrosophic Bipartite Graph [520–522] | $(T, I, F)$ | A bipartite graph defined in a neutrosophic framework, where each vertex and edge is described by truth, indeterminacy, and falsity degrees. |

In addition to the uncertain bipartite graph, related concepts such as the tripartite graph [335, 523], multipartite graph [524, 525], complete bipartite graph [526], soft bipartite graph [527, 528], and weighted bipartite graph [529–531] are also known.

## 4.27 Dombi fuzzy graphs

Dombi fuzzy graphs assign fuzzy memberships to vertices and symmetric edges, requiring each edge membership to be bounded by a Dombi t-norm of endpoint memberships [82, 532–534].

**Definition 4.27.1** (Dombi fuzzy graph). [535] Let

$$V$$

be a finite nonempty set, and let

$$T_D : [0,1]^2 \to [0,1]$$

be the Dombi-type binary operator defined by

$$T_D(a,b) := \frac{ab}{a+b-ab} \qquad (a,b \in [0,1]).$$

A *Dombi fuzzy graph* on $V$ is a pair

$$G_D = (\sigma, \mu),$$

where

$$\sigma : V \to [0,1]$$

is a fuzzy vertex-membership function and

$$\mu : V \times V \to [0,1]$$

is a symmetric fuzzy edge-membership function satisfying

$$\mu(u,v) = \mu(v,u) \qquad (\forall\, u,v \in V),$$

and

$$\mu(u,v) \le T_D\big(\sigma(u),\sigma(v)\big) = \frac{\sigma(u)\sigma(v)}{\sigma(u)+\sigma(v)-\sigma(u)\sigma(v)} \qquad (\forall\, u,v \in V).$$

If one additionally assumes the loopless condition

$$\mu(v,v) = 0 \qquad (\forall\, v \in V),$$

then $G_D$ is called a *loopless Dombi fuzzy graph*.

The function $\sigma$ is called the *Dombi fuzzy vertex set* of $G_D$, and $\mu$ is called the *Dombi fuzzy edge set* of $G_D$.

The support edge set of $G_D$ is defined by

$$E^*(G_D) := \big\{\{u,v\} \subseteq V :\ u \neq v,\ \mu(u,v) > 0\big\}.$$

**Remark 4.27.2.** Since

$$0 \le T_D(a,b) \le \min\{a,b\} \qquad (\forall\, a,b \in [0,1]),$$

every Dombi fuzzy graph is, in particular, a fuzzy graph in the usual Rosenfeld sense. However, the Dombi bound is generally stricter than the classical minimum bound.

**Remark 4.27.3.** More generally, one may use the full Dombi $t$-norm family

$$T_{D,\lambda}(a,b) := \frac{1}{1+\Big(\big(\frac{1-a}{a}\big)^{\lambda}+\big(\frac{1-b}{b}\big)^{\lambda}\Big)^{1/\lambda}} \qquad (\lambda > 0,\ a,b \in (0,1]),$$

with the standard boundary extension at $a=0$ or $b=0$. Then a $\lambda$-*Dombi fuzzy graph* may be defined by

$$\mu(u,v) \le T_{D,\lambda}\big(\sigma(u),\sigma(v)\big).$$

The above definition corresponds to the special case

$$\lambda = 1,$$

for which

$$T_{D,1}(a,b) = \frac{ab}{a+b-ab}.$$

**Definition 4.27.4** (Dombi-admissible uncertain model)**.** Let $M$ be a support-evaluable uncertain model with degree-domain

$$\mathrm{Dom}(M)$$

and zero degree

$$0_M \in \mathrm{Dom}(M).$$

We say that $M$ is *Dombi-admissible* if there exists a bijection

$$\Phi_M : \mathrm{Dom}(M) \longrightarrow [0,1]$$

such that

$$\Phi_M(0_M) = 0.$$

For a fixed parameter

$$\lambda > 0,$$

define the Dombi $t$-norm

$$T_{D,\lambda} : [0,1]^2 \to [0,1]$$

by

$$T_{D,\lambda}(x,y) := \begin{cases} \dfrac{1}{1 + \left( \left( \dfrac{1-x}{x} \right)^{\lambda} + \left( \dfrac{1-y}{y} \right)^{\lambda} \right)^{1/\lambda}}, & x, y \in (0,1], \\ 0, & x = 0 \text{ or } y = 0. \end{cases}$$

Using $\Phi_M$, define an induced order $\preceq_M$ on $\mathrm{Dom}(M)$ by

$$a \preceq_M b \iff \Phi_M(a) \le \Phi_M(b),$$

and define the *Dombi conjunction* on $\mathrm{Dom}(M)$ by

$$a \odot^M_{D,\lambda} b := \Phi_M^{-1}(T_{D,\lambda}(\Phi_M(a), \Phi_M(b))) \qquad (\forall\, a, b \in \mathrm{Dom}(M)).$$

**Definition 4.27.5** (Dombi uncertain graph)**.** Let

$$\mathcal{G}_M = (V, \sigma_M, \eta_M)$$

be an uncertain graph of type $M$ on a finite nonempty vertex set $V$, where $M$ is Dombi-admissible.

Then

$$\mathcal{G}_M$$

is called a *Dombi uncertain graph* (with parameter $\lambda > 0$) if

$$\eta_M(\{u,v\}) \preceq_M \sigma_M(u) \odot^M_{D,\lambda} \sigma_M(v) \qquad \left( \forall\, \{u,v\} \in \binom{V}{2} \right).$$

Equivalently, in the coordinate system induced by $\Phi_M$,

$$\Phi_M(\eta_M(\{u,v\})) \le T_{D,\lambda}(\Phi_M(\sigma_M(u)), \Phi_M(\sigma_M(v))) \qquad \left( \forall\, \{u,v\} \in \binom{V}{2} \right).$$

**Definition 4.27.6** (Strong Dombi uncertain graph)**.** A Dombi uncertain graph

$$\mathcal{G}_M = (V, \sigma_M, \eta_M)$$

is called *strong* if

$$\eta_M(\{u,v\}) = \sigma_M(u) \odot^M_{D,\lambda} \sigma_M(v) \qquad (\forall\, \{u,v\} \in E^*_M),$$

where

$$E^*_M = \left\{ \{u,v\} \in \binom{V^*_M}{2} : \eta_M(\{u,v\}) \ne 0_M \right\}$$

is the support edge set of $\mathcal{G}_M$.

**Theorem 4.27.7** (Well-definedness of the induced Dombi structure)**.** *Let $M$ be a Dombi-admissible uncertain model. Then:*

1. *the relation*
$$\preceq_M$$
*is a well-defined total order on* $\mathrm{Dom}(M)$*;*
2. *the operation*
$$\odot^M_{D,\lambda} : \mathrm{Dom}(M) \times \mathrm{Dom}(M) \to \mathrm{Dom}(M)$$
*is well-defined;*
3. *for all*
$$a, b \in \mathrm{Dom}(M),$$
*one has*
$$a \odot^M_{D,\lambda} b = b \odot^M_{D,\lambda} a;$$
4. *for all*
$$a \in \mathrm{Dom}(M),$$
*one has*
$$a \odot^M_{D,\lambda} 0_M = 0_M.$$

*Proof.* Since
$$\Phi_M : \mathrm{Dom}(M) \to [0, 1]$$
is a bijection, the rule
$$a \preceq_M b \iff \Phi_M(a) \le \Phi_M(b)$$
transports the usual total order on $[0, 1]$ to $\mathrm{Dom}(M)$. Hence $\preceq_M$ is a well-defined total order on $\mathrm{Dom}(M)$.

Next, let
$$a, b \in \mathrm{Dom}(M).$$
Then
$$\Phi_M(a), \Phi_M(b) \in [0, 1].$$
Since
$$T_{D,\lambda} : [0, 1]^2 \to [0, 1]$$
is well-defined, the value
$$T_{D,\lambda}(\Phi_M(a), \Phi_M(b))$$
belongs to $[0, 1]$. Because $\Phi_M$ is bijective, its inverse
$$\Phi_M^{-1} : [0, 1] \to \mathrm{Dom}(M)$$
is well-defined, and thus
$$a \odot^M_{D,\lambda} b = \Phi_M^{-1}(T_{D,\lambda}(\Phi_M(a), \Phi_M(b)))$$
is a well-defined element of $\mathrm{Dom}(M)$. This proves (2).

Since the Dombi $t$-norm $T_{D,\lambda}$ is symmetric, we have
$$T_{D,\lambda}(\Phi_M(a), \Phi_M(b)) = T_{D,\lambda}(\Phi_M(b), \Phi_M(a)),$$
and hence
$$a \odot^M_{D,\lambda} b = b \odot^M_{D,\lambda} a.$$
Thus (3) holds.

Finally, because
$$\Phi_M(0_M) = 0,$$
one obtains
$$a \odot^M_{D,\lambda} 0_M = \Phi_M^{-1}(T_{D,\lambda}(\Phi_M(a), 0)) = \Phi_M^{-1}(0) = 0_M.$$
Hence (4) also holds. □

**Theorem 4.27.8** (Well-definedness of Dombi uncertain graphs)**.** *Let*

$$\mathcal{G}_M = (V, \sigma_M, \eta_M)$$

*be an uncertain graph of type $M$, and assume that $M$ is Dombi-admissible.*

*Then the statement*

$$\text{“}\mathcal{G}_M \text{ is a Dombi uncertain graph”}$$

*is well-defined.*

*Moreover, the statement*

$$\text{“}\mathcal{G}_M \text{ is a strong Dombi uncertain graph”}$$

*is also well-defined.*

*Proof.* For every vertex

$$u \in V,$$

the value

$$\sigma_M(u) \in \mathrm{Dom}(M)$$

is well-defined, since

$$\sigma_M : V \to \mathrm{Dom}(M)$$

is a function. Likewise, for every unordered pair

$$\{u, v\} \in \binom{V}{2},$$

the value

$$\eta_M(\{u, v\}) \in \mathrm{Dom}(M)$$

is well-defined, since

$$\eta_M : \binom{V}{2} \to \mathrm{Dom}(M)$$

is a function.

By the previous theorem, for every

$$u, v \in V,$$

the element

$$\sigma_M(u) \odot^M_{D,\lambda} \sigma_M(v) \in \mathrm{Dom}(M)$$

is well-defined, and the comparison relation

$$\eta_M(\{u, v\}) \preceq_M \sigma_M(u) \odot^M_{D,\lambda} \sigma_M(v)$$

has a definite truth value, because $\preceq_M$ is a well-defined total order on $\mathrm{Dom}(M)$.

Therefore, for each

$$\{u, v\} \in \binom{V}{2},$$

the Dombi edge condition has a definite truth value. Since

$$\binom{V}{2}$$

is finite, the universal statement

$$\eta_M(\{u, v\}) \preceq_M \sigma_M(u) \odot^M_{D,\lambda} \sigma_M(v) \qquad \left(\forall \{u, v\} \in \binom{V}{2}\right)$$

is well-defined. Hence the notion of Dombi uncertain graph is well-defined.

For the strong case, the support edge set

$$E_M^*$$

is already well-defined from the previously introduced support construction for uncertain graphs. For every

$$\{u, v\} \in E_M^*,$$

both sides of

$$\eta_M(\{u, v\}) = \sigma_M(u) \odot_{D,\lambda}^M \sigma_M(v)$$

are well-defined elements of $\mathrm{Dom}(M)$, so the equality has a definite truth value. Since

$$E_M^*$$

is finite, the universal statement over all support edges is well-defined. Hence the notion of strong Dombi uncertain graph is also well-defined. □

**Remark 4.27.9.** If

$$\mathrm{Dom}(M) = [0, 1], \qquad 0_M = 0, \qquad \Phi_M = \mathrm{id}_{[0,1]},$$

then

$$a \odot_{D,\lambda}^M b = T_{D,\lambda}(a, b),$$

and the definition reduces to the ordinary scalar Dombi graph condition

$$\eta_M(\{u, v\}) \leq T_{D,\lambda}\big(\sigma_M(u), \sigma_M(v)\big).$$

In particular, when

$$\lambda = 1,$$

one obtains

$$T_{D,1}(a, b) = \frac{ab}{a + b - ab},$$

so the above definition reduces to the usual Dombi fuzzy graph.

Related Dombi graph concepts under fuzzy and uncertainty-aware frameworks are listed in Table 4.18.

Table 4.18: Related Dombi graph concepts under fuzzy and uncertainty-aware frameworks

| **Concept** | **Reference(s)** |
|---|---|
| Dombi Fuzzy Graph | [534–536] |
| Intuitionistic Dombi Fuzzy Graph | — |
| Pythagorean Dombi Fuzzy Graph | [533, 537] |
| Picture Dombi Fuzzy Graph | [82] |
| Dombi Neutrosophic Graph | [538–540] |

## 4.28 Balanced Uncertain Graph

Balanced fuzzy graph is a fuzzy graph whose every nonempty fuzzy subgraph has density not exceeding that of the whole graph, preserving relative structural balance [541–544].

**Definition 4.28.1** (Balanced Fuzzy Graph)**.** [544] Let

$$G = (V, \sigma, \mu)$$

be a finite fuzzy graph, where

$$\sigma : V \to [0, 1], \qquad \mu : V \times V \to [0, 1], \qquad \mu(u, v) = \mu(v, u), \qquad \mu(u, v) \leq \min\{\sigma(u), \sigma(v)\}$$

for all $u, v \in V$.

Define the support vertex set and support edge set of $G$ by

$$V^* := \{u \in V : \sigma(u) > 0\}, \qquad E^* := \big\{\{u,v\} \subseteq V : u \neq v,\ \mu(u,v) > 0\big\}.$$

Assume that $V^* \neq \varnothing$. The *density* of $G$ is defined by

$$D(G) := \frac{2 \displaystyle\sum_{\{u,v\} \in E^*} \mu(u,v)}{\displaystyle\sum_{u,v \in V^*} \min\{\sigma(u), \sigma(v)\}}.$$

Now let

$$H = (X, \sigma_H, \mu_H)$$

be a fuzzy subgraph of $G$, where $X \subseteq V$,

$$\sigma_H : X \to [0,1], \qquad \mu_H : X \times X \to [0,1],$$

and

$$\sigma_H(x) \leq \sigma(x), \qquad \mu_H(x,y) \leq \mu(x,y)$$

for all $x, y \in X$, with

$$\mu_H(x,y) \leq \min\{\sigma_H(x), \sigma_H(y)\}.$$

If $H$ is nonempty, define its density $D(H)$ analogously by

$$D(H) := \frac{2 \displaystyle\sum_{\{x,y\} \in E^*(H)} \mu_H(x,y)}{\displaystyle\sum_{x,y \in V^*(H)} \min\{\sigma_H(x), \sigma_H(y)\}},$$

where

$$V^*(H) := \{x \in X : \sigma_H(x) > 0\}, \qquad E^*(H) := \big\{\{x,y\} \subseteq X : x \neq y,\ \mu_H(x,y) > 0\big\}.$$

Then $G$ is called a *balanced fuzzy graph* if

$$D(H) \leq D(G)$$

for every nonempty fuzzy subgraph $H$ of $G$.

**Definition 4.28.2** (Balance-Evaluable Uncertain Model)**.** Let $M$ be an uncertain model with degree-domain

$$\mathrm{Dom}(M) \subseteq [0,1]^k.$$

We say that $M$ is *balance-evaluable* if it is equipped with:

$$0_M \in \mathrm{Dom}(M), \qquad \Delta_M : \mathrm{Dom}(M) \to [0,\infty), \qquad \Lambda_M : \mathrm{Dom}(M) \times \mathrm{Dom}(M) \to [0,\infty),$$

such that:

1. $\Delta_M(0_M) = 0$;
2. $\Lambda_M(a,b) = \Lambda_M(b,a)$ for all $a, b \in \mathrm{Dom}(M)$;
3. $\Lambda_M(a,a) > 0$ for every $a \in \mathrm{Dom}(M) \setminus \{0_M\}$.

Here $\Delta_M$ is called the *edge-evaluation map*, and $\Lambda_M$ is called the *pair-capacity map*.

Extensions based on Uncertain Graph are presented below.

**Definition 4.28.3** (Balanced Uncertain Graph)**.** Let $V$ be a finite nonempty set, let $M$ be a balance-evaluable uncertain model, and let

$$G_M = (V, \sigma_M, \eta_M)$$

be an uncertain graph of type $M$ on $V$, where

$$\sigma_M : V \to \mathrm{Dom}(M), \qquad \eta_M : \binom{V}{2} \to \mathrm{Dom}(M).$$

Equivalently,

$$(V, \sigma_M)$$

is an Uncertain Set of type $M$ on $V$, and

$$\left(\binom{V}{2}, \eta_M\right)$$

is an Uncertain Set of type $M$ on the set of unordered pairs of distinct vertices.

Define the support vertex set and support edge set of $G_M$ by

$$V_M^* := \{u \in V : \sigma_M(u) \neq 0_M\}, \qquad E_M^* := \left\{\{u,v\} \in \binom{V}{2} : \eta_M(\{u,v\}) \neq 0_M\right\}.$$

Assume that $V_M^* \neq \varnothing$. The *density* of $G_M$ is defined by

$$D_M(G_M) := \frac{2 \sum_{\{u,v\} \in E_M^*} \Delta_M\big(\eta_M(\{u,v\})\big)}{\sum_{u,v \in V_M^*} \Lambda_M\big(\sigma_M(u), \sigma_M(v)\big)}.$$

Now let

$$H_M = (X, F, \sigma_H, \eta_H)$$

be an uncertain subgraph of $G_M$, where $X \subseteq V$,

$$F \subseteq E_M^* \cap \binom{X}{2}, \qquad \sigma_H = \sigma_M|_X, \qquad \eta_H = \eta_M|_F.$$

Equivalently,

$$(X, \sigma_H)$$

and

$$(F, \eta_H)$$

are Uncertain Sets of type $M$ obtained by restricting the uncertainty-degree functions of $G_M$ to $X$ and $F$, respectively.

Define

$$V_M^*(H) := \{x \in X : \sigma_H(x) \neq 0_M\}.$$

If $V_M^*(H) \neq \varnothing$, define the density of $H_M$ by

$$D_M(H_M) := \frac{2 \sum_{e \in F} \Delta_M\big(\eta_H(e)\big)}{\sum_{x,y \in V_M^*(H)} \Lambda_M\big(\sigma_H(x), \sigma_H(y)\big)}.$$

Then $G_M$ is called a *balanced uncertain graph* if

$$D_M(H_M) \leq D_M(G_M)$$

for every uncertain subgraph $H_M$ of $G_M$ with $V_M^*(H) \neq \varnothing$.

**Theorem 4.28.4** (Well-definedness of Balanced Uncertain Graph)**.** *Let $V$ be a finite nonempty set, let $M$ be a balance-evaluable uncertain model, and let*

$$G_M = (V, \sigma_M, \eta_M)$$

*be an uncertain graph of type $M$ on $V$. Then:*

1. *the support vertex set $V_M^*$ and the support edge set $E_M^*$ are well-defined;*
2. *the density $D_M(G_M)$ is a well-defined nonnegative real number;*
3. *for every uncertain subgraph*
$$H_M = (X, F, \sigma_H, \eta_H)$$
*with $V_M^*(H) \neq \varnothing$, the density $D_M(H_M)$ is a well-defined nonnegative real number;*
4. *consequently, the statement*
$$\text{“}G_M \text{ is a balanced uncertain graph”}$$
*is well-defined.*

*Proof.* Since $M$ is an uncertain model, its degree-domain $\mathrm{Dom}(M)$ is fixed. Because

$$\sigma_M : V \to \mathrm{Dom}(M), \qquad \eta_M : \binom{V}{2} \to \mathrm{Dom}(M)$$

are functions, the pairs

$$(V, \sigma_M) \qquad \text{and} \qquad \left(\binom{V}{2}, \eta_M\right)$$

are well-defined Uncertain Sets of type $M$.

Therefore the predicates

$$\sigma_M(u) \neq 0_M \qquad (u \in V)$$

and

$$\eta_M(e) \neq 0_M \qquad \left(e \in \binom{V}{2}\right)$$

have definite truth values, because both $\sigma_M(u)$ and $\eta_M(e)$ belong to $\mathrm{Dom}(M)$, and $0_M \in \mathrm{Dom}(M)$ is fixed. Hence the sets

$$V_M^* = \{u \in V : \sigma_M(u) \neq 0_M\}$$

and

$$E_M^* = \left\{e \in \binom{V}{2} : \eta_M(e) \neq 0_M\right\}$$

are well-defined. This proves (1).

Next, since $V$ is finite, the set $\binom{V}{2}$ is finite, and hence $E_M^*$ is finite. Because $\Delta_M : \mathrm{Dom}(M) \to [0, \infty)$, for each edge

$$e \in E_M^*$$

the quantity

$$\Delta_M\big(\eta_M(e)\big)$$

is a well-defined nonnegative real number. Therefore

$$\sum_{e \in E_M^*} \Delta_M\big(\eta_M(e)\big)$$

is a well-defined finite sum in $[0, \infty)$.

Also, since $V_M^* \subseteq V$, the set $V_M^*$ is finite. Because

$$\Lambda_M : \mathrm{Dom}(M) \times \mathrm{Dom}(M) \to [0, \infty),$$

for every $u, v \in V_M^*$ the quantity

$$\Lambda_M\big(\sigma_M(u), \sigma_M(v)\big)$$

is a well-defined nonnegative real number. Hence

$$\sum_{u,v \in V_M^*} \Lambda_M\big(\sigma_M(u), \sigma_M(v)\big)$$

is also a well-defined finite sum in $[0, \infty)$.

It remains to show that the denominator is strictly positive. Since $V_M^* \neq \varnothing$, choose $u_0 \in V_M^*$. Then

$$\sigma_M(u_0) \neq 0_M.$$

By condition (3) in the definition of a balance-evaluable uncertain model,

$$\Lambda_M\big(\sigma_M(u_0), \sigma_M(u_0)\big) > 0.$$

Since this term appears in the denominator, we obtain

$$\sum_{u,v \in V_M^*} \Lambda_M\big(\sigma_M(u), \sigma_M(v)\big) > 0.$$

Therefore the quotient

$$D_M(G_M) = \frac{2 \displaystyle\sum_{e \in E_M^*} \Delta_M\big(\eta_M(e)\big)}{\displaystyle\sum_{u,v \in V_M^*} \Lambda_M\big(\sigma_M(u), \sigma_M(v)\big)}$$

is a well-defined nonnegative real number. This proves (2).

Now let

$$H_M = (X, F, \sigma_H, \eta_H)$$

be an uncertain subgraph of $G_M$ with $V_M^*(H) \neq \varnothing$. By definition,

$$X \subseteq V, \qquad F \subseteq E_M^* \cap \binom{X}{2}, \qquad \sigma_H = \sigma_M|_X, \qquad \eta_H = \eta_M|_F.$$

Hence $\sigma_H$ and $\eta_H$ are restrictions of well-defined functions, so they are well-defined functions themselves. Therefore

$$(X, \sigma_H) \qquad \text{and} \qquad (F, \eta_H)$$

are well-defined Uncertain Sets of type $M$.

Since $X$ and $F$ are finite, the sums

$$\sum_{e \in F} \Delta_M\big(\eta_H(e)\big) \qquad \text{and} \qquad \sum_{x,y \in V_M^*(H)} \Lambda_M\big(\sigma_H(x), \sigma_H(y)\big)$$

are finite sums of well-defined nonnegative real numbers. Because $V_M^*(H) \neq \varnothing$, choose $x_0 \in V_M^*(H)$. Then

$$\sigma_H(x_0) \neq 0_M,$$

and again by condition (3),

$$\Lambda_M\big(\sigma_H(x_0), \sigma_H(x_0)\big) > 0.$$

Hence the denominator in $D_M(H_M)$ is strictly positive, so

$$D_M(H_M) = \frac{2\sum_{e\in F} \Delta_M\big(\eta_H(e)\big)}{\sum_{x,y\in V_M^*(H)} \Lambda_M\big(\sigma_H(x), \sigma_H(y)\big)}$$

is a well-defined nonnegative real number. This proves (3).

Finally, since both $D_M(H_M)$ and $D_M(G_M)$ are well-defined real numbers, the comparison

$$D_M(H_M) \le D_M(G_M)$$

has a definite truth value for every uncertain subgraph $H_M$ with $V_M^*(H) \neq \varnothing$. Therefore the statement

"$G_M$ is a balanced uncertain graph"

is well-defined. This proves (4). □

**Remark 4.28.5.** If

$$\mathrm{Dom}(M) = [0,1], \qquad 0_M = 0, \qquad \Delta_M(t) = t, \qquad \Lambda_M(a,b) = \min\{a,b\},$$

then the above density becomes

$$D_M(G_M) = \frac{2\sum_{\{u,v\}\in E^*} \mu(u,v)}{\sum_{u,v\in V^*} \min\{\sigma(u), \sigma(v)\}},$$

and the definition reduces exactly to the usual notion of a balanced fuzzy graph.

Representative balanced-graph concepts under uncertainty-aware graph frameworks are listed in Table 4.19.

## 4.29 Product Uncertain Graph

Product fuzzy graph assigns vertex memberships and edge memberships bounded by the product of endpoint memberships, modeling uncertainty through multiplicative interaction rather than minimum-based constraints [554–557].

**Definition 4.29.1** (Product Fuzzy Graph)**.** [558–560] Let $V$ be a finite nonempty set. A *product fuzzy graph* on $V$ is a pair

$$G = (\sigma, \mu),$$

where

$$\sigma : V \to [0,1]$$

is a vertex-membership function and

$$\mu : V \times V \to [0,1]$$

is an edge-membership function such that, for all $x, y \in V$,

$$\mu(x,y) = \mu(y,x)$$

and

$$\mu(x,y) \le \sigma(x)\sigma(y).$$

The support vertex set and support edge set of $G$ are defined by

$$V^* := \{x \in V : \sigma(x) > 0\},$$

and

$$E^* := \big\{\{x,y\} \subseteq V^* : x \neq y,\ \mu(x,y) > 0\big\},$$

respectively. The crisp graph

$$G^* := (V^*, E^*)$$

is called the *support graph* of $G$.

Representative product-graph concepts under uncertainty-aware graph frameworks are listed in Table 4.20.

Table 4.19: Representative balanced-graph concepts under uncertainty-aware graph frameworks, classified by the dimension $k$ of the information attached to vertices and/or edges.

| $k$ | **Balanced-graph concept** | **Typical coordinate form** | **Canonical information attached to vertices/edges** |
|---|---|---|---|
| 1 | Balanced Fuzzy Graph | $\mu$ | A balanced graph studied in a fuzzy framework, where each vertex and edge is associated with a single membership degree in $[0, 1]$. |
| 2 | Balanced Vague Graph [545, 546] | $(t, f)$ | A balanced graph defined in a vague framework, where each vertex and edge is characterized by a truth-membership degree and a falsity-membership degree, typically with $t + f \leq 1$. |
| 2 | Balanced Intuitionistic Fuzzy Graph [547] | $(\mu, \nu)$ | A balanced graph defined in an intuitionistic fuzzy framework, where each vertex and edge carries a membership degree and a non-membership degree, usually satisfying $\mu+\nu \leq 1$. |
| 2 | Balanced Bipolar Fuzzy Graph [548] | $(\mu^+, \mu^-)$ | A balanced graph defined in a bipolar fuzzy framework, where each vertex and edge is described by a positive membership degree and a negative membership degree. |
| 3 | Balanced Picture Fuzzy Graph [157] | $(\mu, \eta, \nu)$ | A balanced graph defined in a picture fuzzy framework, where each vertex and edge is described by positive, neutral, and negative membership degrees, usually satisfying $\mu+\eta+\nu \leq 1$. |
| 3 | Balanced Spherical Fuzzy Graph [549] | $(\mu, \eta, \nu)$ | A balanced graph defined in a spherical fuzzy framework, where each vertex and edge is described by positive, neutral, and negative membership degrees, usually satisfying $\mu^2+\eta^2+\nu^2 \leq 1$. |
| 3 | Balanced Neutrosophic Graph [550–553] | $(T, I, F)$ | A balanced graph defined in a neutrosophic framework, where each vertex and edge is described by truth, indeterminacy, and falsity degrees. |

## 4.30 Dynamic Uncertain Graph

Dynamic fuzzy graph models time-varying uncertainty by assigning time-dependent membership values to vertices and edges, so each time instant yields a fuzzy graph distinct snapshot (cf. [566–568]).

**Definition 4.30.1** (Dynamic Fuzzy Graph). [566, 567] Let $T$ be a nonempty time set (discrete or continuous), and let $V$ be a finite nonempty set of potential vertices.

A *dynamic fuzzy graph* on $(T, V)$ is a pair

$$\mathcal{G} = (\sigma, \mu),$$

where

$$\sigma : T \times V \to [0, 1]$$

is the time-dependent vertex membership function, and

$$\mu : T \times V \times V \to [0, 1]$$

is the time-dependent edge membership function, such that for every fixed time $t \in T$ and all $u, v \in V$,

$$\mu(t, u, v) \leq \min\{\sigma(t, u), \sigma(t, v)\}.$$

For each $t \in T$, the snapshot

$$\mathcal{G}(t) = \big(V, \sigma_t, \mu_t\big),$$

Table 4.20: Representative product-graph concepts under uncertainty-aware graph frameworks, classified by the dimension $k$ of the information attached to vertices and/or edges.

| $k$ | **Product-graph concept** | **Typical coordinate form** | **Canonical information attached to vertices/edges** |
|---|---|---|---|
| 1 | Product Fuzzy Graph | $\mu$ | A product graph defined in a fuzzy framework, where each vertex and edge is associated with a single membership degree in $[0, 1]$, and edge-membership values are typically constrained by the product of the memberships of their incident vertices. |
| 2 | Product Bipolar Fuzzy Graph [561, 562] | $(\mu^+, \mu^-)$ | A product graph defined in a bipolar fuzzy framework, where each vertex and edge is described by a positive membership degree and a negative membership degree. |
| 2 | Product Intuitionistic Fuzzy Graph [563, 564] | $(\mu, \nu)$ | A product graph defined in an intuitionistic fuzzy framework, where each vertex and edge carries a membership degree and a non-membership degree, usually satisfying $\mu+\nu \leq 1$. |
| 3 | Product Picture Fuzzy Graph | $(\mu, \eta, \nu)$ | A product graph defined in a picture fuzzy framework, where each vertex and edge is described by positive, neutral, and negative membership degrees, usually satisfying $\mu+\eta+\nu \leq 1$. |
| 3 | Product Neutrosophic Graph [565] | $(T, I, F)$ | A product graph defined in a neutrosophic framework, where each vertex and edge is described by truth, indeterminacy, and falsity degrees. |

defined by

$$\sigma_t(u) := \sigma(t, u), \qquad \mu_t(u, v) := \mu(t, u, v),$$

is a fuzzy graph. The family

$$\{\mathcal{G}(t) : t \in T\}$$

is called the *temporal evolution* of the dynamic fuzzy graph.

The support vertex set and support edge set at time $t$ are defined by

$$V_t^* := \{u \in V : \sigma(t, u) > 0\},$$

and

$$E_t^* := \big\{\{u, v\} \subseteq V_t^* : u \neq v,\ \mu(t, u, v) > 0\big\}.$$

Thus the underlying crisp graph at time $t$ is

$$G_t^* = (V_t^*, E_t^*).$$

A dynamic uncertain graph models time-dependent uncertainty by assigning interval-valued memberships to vertices and edges at each time instant, so that every snapshot is an uncertain graph.

First, let

$$\mathbb{I}([0, 1]) := \big\{[a^-, a^+] \subseteq [0, 1] :\ 0 \leq a^- \leq a^+ \leq 1\big\}$$

denote the family of all closed subintervals of $[0, 1]$. For

$$A = [a^-, a^+], \qquad B = [b^-, b^+] \in \mathbb{I}([0, 1]),$$

define the componentwise partial order

$$A \preceq B \iff a^- \leq b^- \ \text{ and } \ a^+ \leq b^+,$$

and define

$$A \wedge B := \bigl[\min\{a^-, b^-\}, \min\{a^+, b^+\}\bigr].$$

Recall that an *uncertain graph* is a triple

$$G = (V, \Sigma, M),$$

where

$$\Sigma : V \to \mathbb{I}([0,1]), \qquad M : V \times V \to \mathbb{I}([0,1]),$$

such that $M$ is symmetric and

$$M(u,v) \preceq \Sigma(u) \wedge \Sigma(v) \qquad (\forall\, u, v \in V).$$

**Definition 4.30.2** (Dynamic Uncertain Graph)**.** Let $T$ be a nonempty time set, and let $V$ be a finite nonempty set of potential vertices.

A *dynamic uncertain graph* on $(T, V)$ is a pair

$$\mathcal{G} = (\Sigma, M),$$

where

$$\Sigma : T \times V \to \mathbb{I}([0,1])$$

is the time-dependent uncertain vertex-membership function, and

$$M : T \times V \times V \to \mathbb{I}([0,1])$$

is the time-dependent uncertain edge-membership function, satisfying the following conditions for every fixed time $t \in T$ and all $u, v \in V$:

1. $$M(t,u,v) = M(t,v,u),$$

   that is, $M$ is symmetric with respect to the vertex variables.

2. $$M(t,u,v) \preceq \Sigma(t,u) \wedge \Sigma(t,v).$$

For each $t \in T$, define

$$\Sigma_t : V \to \mathbb{I}([0,1]), \qquad \Sigma_t(u) := \Sigma(t,u),$$

and

$$M_t : V \times V \to \mathbb{I}([0,1]), \qquad M_t(u,v) := M(t,u,v).$$

Then the uncertain graph

$$\mathcal{G}(t) := (V, \Sigma_t, M_t)$$

is called the *snapshot* of $\mathcal{G}$ at time $t$. The family

$$\{\mathcal{G}(t) : t \in T\}$$

is called the *temporal evolution* of the dynamic uncertain graph.

The support vertex set at time $t$ is defined by

$$V_t^* := \{u \in V : \Sigma(t,u) \neq [0,0]\},$$

and the support edge set at time $t$ is defined by

$$E_t^* := \bigl\{\{u,v\} \subseteq V : \ u \neq v, \ M(t,u,v) \neq [0,0]\bigr\}.$$

Hence the support crisp graph at time $t$ is

$$G_t^* = (V_t^*, E_t^*).$$

**Theorem 4.30.3** (Well-definedness of dynamic uncertain graph snapshots)**.** *Let*

$$\mathcal{G} = (\Sigma, M)$$

*be a dynamic uncertain graph on* $(T, V)$*. Then, for every* $t \in T$*:*

1. *the mappings*
$$\Sigma_t : V \to \mathbb{I}([0,1]), \qquad M_t : V \times V \to \mathbb{I}([0,1]),$$
*are well-defined;*
2. *the snapshot*
$$\mathcal{G}(t) = (V, \Sigma_t, M_t)$$
*is an uncertain graph;*
3. *the support crisp graph*
$$G_t^* = (V_t^*, E_t^*)$$
*is well-defined, and every edge in* $E_t^*$ *has both endpoints in* $V_t^*$*. Equivalently,*
$$E_t^* \subseteq \big\{\{u, v\} \subseteq V_t^* : u \neq v\big\}.$$

*Proof.* Fix an arbitrary time $t \in T$.

First, since
$$\Sigma : T \times V \to \mathbb{I}([0,1]),$$
for every $u \in V$ the value $\Sigma(t, u)$ belongs to $\mathbb{I}([0,1])$. Hence the assignment
$$\Sigma_t(u) := \Sigma(t, u)$$
defines a unique function
$$\Sigma_t : V \to \mathbb{I}([0,1]).$$
Thus $\Sigma_t$ is well-defined.

Similarly, since
$$M : T \times V \times V \to \mathbb{I}([0,1]),$$
for every $(u, v) \in V \times V$ the value $M(t, u, v)$ belongs to $\mathbb{I}([0,1])$. Hence the assignment
$$M_t(u, v) := M(t, u, v)$$
defines a unique function
$$M_t : V \times V \to \mathbb{I}([0,1]).$$
Thus $M_t$ is well-defined. This proves part (1).

Next, we verify that
$$\mathcal{G}(t) = (V, \Sigma_t, M_t)$$
is an uncertain graph. By the defining symmetry of the dynamic uncertain graph,
$$M(t, u, v) = M(t, v, u) \qquad (\forall\, u, v \in V),$$
and therefore
$$M_t(u, v) = M_t(v, u) \qquad (\forall\, u, v \in V).$$
So $M_t$ is symmetric.

Also, for all $u, v \in V$,

$$M_t(u,v) = M(t,u,v) \preceq \Sigma(t,u) \wedge \Sigma(t,v) = \Sigma_t(u) \wedge \Sigma_t(v).$$

Therefore the edge-membership condition of an uncertain graph is satisfied. Hence

$$\mathcal{G}(t) = (V, \Sigma_t, M_t)$$

is an uncertain graph. This proves part (2).

Finally, we prove part (3). By definition,

$$V_t^* = \{u \in V : \Sigma(t,u) \neq [0,0]\}$$

is a subset of $V$, and

$$E_t^* = \big\{\{u,v\} \subseteq V : \ u \neq v, \ M(t,u,v) \neq [0,0]\big\}$$

is a family of two-element subsets of $V$. Thus $G_t^* = (V_t^*, E_t^*)$ is set-theoretically well-defined.

It remains to show that every edge in $E_t^*$ has both endpoints in $V_t^*$. Take any

$$\{u,v\} \in E_t^*.$$

Then $u \neq v$ and

$$M(t,u,v) \neq [0,0].$$

Suppose, for contradiction, that $u \notin V_t^*$. Then

$$\Sigma(t,u) = [0,0].$$

Hence

$$\Sigma(t,u) \wedge \Sigma(t,v) = [0,0].$$

Since

$$M(t,u,v) \preceq \Sigma(t,u) \wedge \Sigma(t,v),$$

we obtain

$$M(t,u,v) \preceq [0,0].$$

But the only interval in $\mathbb{I}([0,1])$ that is $\preceq [0,0]$ is $[0,0]$ itself. Thus

$$M(t,u,v) = [0,0],$$

which contradicts the choice of $\{u,v\} \in E_t^*$. Therefore $u \in V_t^*$. By the same argument, $v \in V_t^*$.

Consequently,

$$E_t^* \subseteq \big\{\{u,v\} \subseteq V_t^* : u \neq v\big\},$$

so $G_t^*$ is indeed a well-defined crisp support graph associated with the snapshot at time $t$. This completes the proof. □

## 4.31 Uncertain Soft Graph

A fuzzy soft graph is a parameterized family of fuzzy subgraphs, assigning vertex and edge memberships under parameters to represent uncertainty in graph structures flexibly [569–571].

**Definition 4.31.1** (Fuzzy Soft Graph)**.** [569, 570] Let

$$G^* = (V, E)$$

be a simple graph, and let

$$A$$

be a nonempty set of parameters.

A *fuzzy soft graph* over $G^*$ is a quadruple

$$\widetilde{G} = (G^*, \widetilde{F}, \widetilde{K}, A),$$

where

- $\widetilde{F} : A \to \mathcal{F}(V)$ is a fuzzy soft set over the vertex set $V$,
- $\widetilde{K} : A \to \mathcal{F}(E)$ is a fuzzy soft set over the edge set $E$,

such that, for every parameter $a \in A$, the pair

$$H(a) = \big(\widetilde{F}(a), \widetilde{K}(a)\big)$$

is a fuzzy subgraph of $G^*$. Equivalently, for every $a \in A$ and every edge

$$xy \in E,$$

we have

$$\widetilde{K}(a)(xy) \le \min\{\widetilde{F}(a)(x), \widetilde{F}(a)(y)\}.$$

Thus, a fuzzy soft graph may be viewed as a parameterized family of fuzzy graphs associated with the underlying crisp graph $G^*$.

An uncertain soft graph is a parameterized family of uncertain graphs, where the uncertainty of vertices and edges is represented by closed subintervals of $[0, 1]$.

First, let

$$\mathbb{I}([0,1]) := \big\{[a^-, a^+] \subseteq [0,1] : \ 0 \le a^- \le a^+ \le 1\big\}$$

denote the family of all closed subintervals of $[0, 1]$. For

$$A = [a^-, a^+], \qquad B = [b^-, b^+] \in \mathbb{I}([0,1]),$$

define the componentwise partial order

$$A \preceq B \iff a^- \le b^- \ \text{ and } \ a^+ \le b^+,$$

and define

$$A \wedge B := \big[\min\{a^-, b^-\}, \min\{a^+, b^+\}\big].$$

Let

$$G^* = (V, E)$$

be a simple graph, where

$$E \subseteq \big\{\{u, v\} \subseteq V : u \ne v\big\},$$

and let

$$P(V) := \{f : V \to \mathbb{I}([0,1])\}, \qquad P(E) := \{g : E \to \mathbb{I}([0,1])\}.$$

**Definition 4.31.2** (Uncertain Soft Graph)**.** Let

$$G^* = (V, E)$$

be a simple graph, and let

$$A$$

be a nonempty set of parameters.

An *uncertain soft graph* over $G^*$ is a quadruple

$$\widetilde{G} = (G^*, \widetilde{\Sigma}, \widetilde{M}, A),$$

where

$$\widetilde{\Sigma} : A \to P(V)$$

is an uncertain soft set on the vertex set $V$, and

$$\widetilde{M} : A \to P(E)$$

is an uncertain soft set on the edge set $E$, such that for every parameter

$$a \in A$$

and every edge

$$e = \{u, v\} \in E,$$

the following condition holds:

$$\widetilde{M}(a)(e) \preceq \widetilde{\Sigma}(a)(u) \wedge \widetilde{\Sigma}(a)(v).$$

For each parameter $a \in A$, define

$$\Sigma_a : V \to \mathbb{I}([0,1]), \qquad \Sigma_a(u) := \widetilde{\Sigma}(a)(u),$$

and define

$$M_a : V \times V \to \mathbb{I}([0,1])$$

by

$$M_a(u,v) := \begin{cases} \widetilde{M}(a)(\{u,v\}), & u \neq v \text{ and } \{u,v\} \in E, \\ [0,0], & u = v \text{ or } \{u,v\} \notin E. \end{cases}$$

Then

$$H(a) := (V, \Sigma_a, M_a)$$

is called the *uncertain graph associated with the parameter $a$.*

Thus, an uncertain soft graph may be regarded as a parameterized family

$$\{H(a) : a \in A\}$$

of uncertain graphs associated with the underlying crisp graph $G^*$.

**Theorem 4.31.3** (Well-definedness of uncertain soft graphs)**.** *Let*

$$\widetilde{G} = (G^*, \widetilde{\Sigma}, \widetilde{M}, A)$$

*be an uncertain soft graph over the simple graph*

$$G^* = (V, E).$$

*Then, for every parameter $a \in A$, the following statements hold:*

1. *the mappings*

   $$\Sigma_a : V \to \mathbb{I}([0,1]) \qquad \text{and} \qquad M_a : V \times V \to \mathbb{I}([0,1])$$

   *are well-defined;*

2. *$M_a$ is symmetric, that is,*

   $$M_a(u,v) = M_a(v,u) \qquad (\forall\, u, v \in V);$$

3. *for all $u, v \in V$,*

   $$M_a(u,v) \preceq \Sigma_a(u) \wedge \Sigma_a(v);$$

4. *therefore,*

   $$H(a) = (V, \Sigma_a, M_a)$$

   *is an uncertain graph.*

   *Hence every uncertain soft graph determines a well-defined family of uncertain graphs indexed by the parameter set $A$.*

*Proof.* Fix an arbitrary parameter

$$a \in A.$$

We first prove that $\Sigma_a$ is well-defined. Since

$$\widetilde{\Sigma} : A \to P(V),$$

the value

$$\widetilde{\Sigma}(a)$$

is a function from $V$ into $\mathbb{I}([0,1])$. Therefore, for each

$$u \in V,$$

the value

$$\Sigma_a(u) := \widetilde{\Sigma}(a)(u)$$

is uniquely determined and belongs to $\mathbb{I}([0,1])$. Hence

$$\Sigma_a : V \to \mathbb{I}([0,1])$$

is well-defined.

Next, we prove that $M_a$ is well-defined. Let

$$(u,v) \in V \times V.$$

Since $G^* = (V,E)$ is a simple graph, exactly one of the following mutually exclusive cases occurs:

$$u = v, \qquad u \neq v \text{ and } \{u,v\} \in E, \qquad u \neq v \text{ and } \{u,v\} \notin E.$$

Thus the piecewise formula

$$M_a(u,v) := \begin{cases} \widetilde{M}(a)(\{u,v\}), & u \neq v \text{ and } \{u,v\} \in E, \\ [0,0], & u = v \text{ or } \{u,v\} \notin E \end{cases}$$

assigns exactly one value to each ordered pair $(u,v) \in V \times V$.

Moreover, if

$$u \neq v \text{ and } \{u,v\} \in E,$$

then

$$\widetilde{M}(a)(\{u,v\}) \in \mathbb{I}([0,1])$$

because

$$\widetilde{M}(a) \in P(E).$$

In the remaining cases,

$$M_a(u,v) = [0,0] \in \mathbb{I}([0,1]).$$

Therefore

$$M_a : V \times V \to \mathbb{I}([0,1])$$

is well-defined. This proves part (1).

We now prove symmetry. Take arbitrary

$$u, v \in V.$$

If

$$u = v,$$

then clearly

$$M_a(u,v) = M_a(v,u) = [0,0].$$

Assume next that

$$u \neq v.$$

If

$$\{u,v\} \notin E,$$

then

$$\{v,u\} \notin E$$

as well, because $\{u,v\} = \{v,u\}$ as unordered pairs. Hence

$$M_a(u,v) = M_a(v,u) = [0,0].$$

If

$$\{u,v\} \in E,$$

then again

$$\{u,v\} = \{v,u\},$$

so

$$M_a(u,v) = \widetilde{M}(a)(\{u,v\}) = \widetilde{M}(a)(\{v,u\}) = M_a(v,u).$$

Thus

$$M_a(u,v) = M_a(v,u) \qquad (\forall\, u,v \in V),$$

and part (2) follows.

Next, we verify the uncertain graph condition. Let

$$u,v \in V.$$

If

$$u \neq v \text{ and } \{u,v\} \in E,$$

then by the defining condition of uncertain soft graphs,

$$\widetilde{M}(a)(\{u,v\}) \preceq \widetilde{\Sigma}(a)(u) \wedge \widetilde{\Sigma}(a)(v).$$

Hence

$$M_a(u,v) = \widetilde{M}(a)(\{u,v\}) \preceq \Sigma_a(u) \wedge \Sigma_a(v).$$

If

$$u = v \quad \text{or} \quad \{u,v\} \notin E,$$

then

$$M_a(u,v) = [0,0].$$

Since

$$\Sigma_a(u), \Sigma_a(v) \in \mathbb{I}([0,1]),$$

their meet

$$\Sigma_a(u) \wedge \Sigma_a(v)$$

also belongs to $\mathbb{I}([0,1])$, and therefore

$$[0,0] \preceq \Sigma_a(u) \wedge \Sigma_a(v).$$

Thus, in all cases,

$$M_a(u,v) \preceq \Sigma_a(u) \wedge \Sigma_a(v).$$

This proves part (3).

By parts (1), (2), and (3), the triple

$$H(a) = (V, \Sigma_a, M_a)$$

satisfies the defining axioms of an uncertain graph. Therefore $H(a)$ is an uncertain graph. This proves part (4).

Since $a \in A$ was arbitrary, the conclusion holds for every parameter in $A$. Hence an uncertain soft graph determines a well-defined parameterized family of uncertain graphs. □

Representative soft-graph concepts under uncertainty-aware graph frameworks are listed in Table 4.21.

Related extension concepts such as Fuzzy HyperSoft Graphs [584, 585], Neutrosophic HyperSoft Graphs [586, 587], SuperHyperSoft Graphs [98, 588], MultiSoft Graphs [589], and TreeSoft Graphs [590] are also known.

Table 4.21: Representative soft-graph concepts under uncertainty-aware graph frameworks, classified by the dimension $k$ of the information attached to vertices and/or edges for each parameter.

| $k$ | **Soft-graph concept** | **Typical coordinate form** | **Canonical information attached to vertices/edges** |
|---|---|---|---|
| 2 | Intuitionistic Fuzzy Soft Graph [572, 573] | $e \mapsto (\mu, \nu)$ | A soft graph in which, for each parameter $e$, the associated graph is described by intuitionistic fuzzy information; each vertex and edge carries a membership degree and a non-membership degree, usually satisfying $\mu + \nu \leq 1$. |
| 2 | Vague Soft Graph [60, 574] | $e \mapsto (t, f)$ | A soft graph in which, for each parameter $e$, the associated graph is described by a truth-membership degree and a falsity-membership degree, typically with $t + f \leq 1$. |
| 2 | Bipolar Fuzzy Soft Graph [575] | $e \mapsto (\mu^+, \mu^-)$ | A soft graph in which, for each parameter $e$, the associated graph is described by a positive membership degree and a negative membership degree on vertices and edges. |
| $k \in \mathbb{N}$ | Hesitant Fuzzy Soft Graph | $e \mapsto \{\mu_1, \ldots, \mu_k\} \subseteq [0, 1]$ | A soft graph in which, for each parameter $e$, each vertex and edge is associated with a finite set of possible membership degrees rather than a single value; hence the information dimension is variable. |
| 3 | Spherical Fuzzy Soft Graph [576] | $e \mapsto (\mu, \eta, \nu)$ | A soft graph in which, for each parameter $e$, each vertex and edge is described by positive, neutral, and negative membership degrees, usually satisfying $\mu^2 + \eta^2 + \nu^2 \leq 1$. |
| 3 | Picture Fuzzy Soft Graph [577–579] | $e \mapsto (\mu, \eta, \nu)$ | A soft graph in which, for each parameter $e$, each vertex and edge is described by positive, neutral, and negative membership degrees, usually satisfying $\mu + \eta + \nu \leq 1$. |
| 3 | Neutrosophic Soft Graph [122, 580–583] | $e \mapsto (T, I, F)$ | A soft graph in which, for each parameter $e$, each vertex and edge is described by truth, indeterminacy, and falsity degrees. |

## 4.32 Uncertain Rough Graph

A fuzzy rough graph combines fuzzy and rough approximations, representing vertices and edges through lower and upper fuzzy graphs to model uncertainty, vagueness, and incompleteness [591–593].

**Definition 4.32.1** (Fuzzy Rough Graph)**.** [591, 592] Let $U$ be a nonempty finite set, and let

$$T : U \times U \to [0, 1]$$

be a fuzzy tolerance relation on $U$. Let

$$A : U \to [0, 1]$$

be a fuzzy set on $U$.

Define the lower and upper approximations of $A$ with respect to $T$ by

$$\underline{T}(A)(x) = \inf_{y \in U} \big((1 - T(x, y)) \vee A(y)\big), \qquad \overline{T}(A)(x) = \sup_{y \in U} \big(T(x, y) \wedge A(y)\big)$$

for all $x \in U$. Then

$$T(A) := \big(\underline{T}(A), \overline{T}(A)\big)$$

is called a fuzzy rough set on $U$.

Let
$$E^* \subseteq \big\{\{x,y\} \mid x,y \in U,\ x \neq y\big\}$$
be a set of potential edges, and let
$$P : E^* \to [0,1]$$
be a fuzzy set on $E^*$ such that
$$P(\{x,y\}) \leq \min\big\{\overline{T}(A)(x), \overline{T}(A)(y)\big\} \qquad (\forall\, \{x,y\} \in E^*).$$

Let
$$H : E^* \times E^* \to [0,1]$$
be a fuzzy tolerance relation on $E^*$. Define the lower and upper approximations of $P$ with respect to $H$ by
$$\underline{H}(P)(e) = \inf_{f \in E^*} \big((1 - H(e,f)) \vee P(f)\big), \qquad \overline{H}(P)(e) = \sup_{f \in E^*} \big(H(e,f) \wedge P(f)\big)$$
for all $e \in E^*$. Then
$$H(P) := \big(\underline{H}(P), \overline{H}(P)\big)$$
is called a fuzzy rough relation on $E^*$.

A *fuzzy rough graph* on $U$ is the pair
$$G = \big(\underline{G}, \overline{G}\big) = \big((\underline{T}(A), \underline{H}(P)),\ (\overline{T}(A), \overline{H}(P))\big),$$
where
$$\underline{G} = (\underline{T}(A), \underline{H}(P)) \qquad \text{and} \qquad \overline{G} = (\overline{T}(A), \overline{H}(P))$$
are fuzzy graphs, that is,
$$\underline{H}(P)(\{x,y\}) \leq \min\big\{\underline{T}(A)(x), \underline{T}(A)(y)\big\},$$
and
$$\overline{H}(P)(\{x,y\}) \leq \min\big\{\overline{T}(A)(x), \overline{T}(A)(y)\big\}$$
for all $\{x,y\} \in E^*$.

Here, $\underline{G}$ is called the *lower approximation graph* and $\overline{G}$ is called the *upper approximation graph* of $G$.

An uncertain rough graph represents a graph by a lower and an upper uncertain graph, obtained from rough approximations of interval-valued vertex and edge uncertainty.

First, let
$$\mathbb{I}([0,1]) := \big\{[a^-, a^+] \subseteq [0,1] :\ 0 \leq a^- \leq a^+ \leq 1\big\}$$
denote the family of all closed subintervals of $[0,1]$. For
$$A = [a^-, a^+], \qquad B = [b^-, b^+] \in \mathbb{I}([0,1]),$$
define the componentwise partial order
$$A \preceq B \iff a^- \leq b^- \ \text{ and } \ a^+ \leq b^+,$$
and define
$$A \wedge B := \big[\min\{a^-, b^-\}, \min\{a^+, b^+\}\big].$$

If
$$\mathcal{A} = \{[a_i^-, a_i^+] : i \in I\} \subseteq \mathbb{I}([0,1])$$
is a finite nonempty family of intervals, define
$$\inf \mathcal{A} := \left[\min_{i \in I} a_i^-, \min_{i \in I} a_i^+\right], \qquad \sup \mathcal{A} := \left[\max_{i \in I} a_i^-, \max_{i \in I} a_i^+\right].$$

**Definition 4.32.2** (Uncertain Rough Graph)**.** Let $V$ be a finite nonempty set, and let

$$E^* \subseteq \big\{\{x, y\} \subseteq V : x \neq y\big\}$$

be a set of potential edges. Let

$$R_V$$

be an equivalence relation on $V$, and let

$$R_E$$

be an equivalence relation on $E^*$.

Let

$$\Sigma : V \to \mathbb{I}([0, 1])$$

be an uncertain vertex-membership function, and let

$$P : E^* \to \mathbb{I}([0, 1])$$

be an uncertain edge-membership function.

For each $x \in V$, define the lower and upper vertex approximations by

$$\underline{\Sigma}(x) := \inf\{\Sigma(u) : u \in [x]_{R_V}\}, \qquad \overline{\Sigma}(x) := \sup\{\Sigma(u) : u \in [x]_{R_V}\},$$

where $[x]_{R_V}$ denotes the $R_V$-equivalence class of $x$.

For each $e \in E^*$, define the lower and upper edge approximations by

$$\underline{P}(e) := \inf\{P(f) : f \in [e]_{R_E}\}, \qquad \overline{P}(e) := \sup\{P(f) : f \in [e]_{R_E}\},$$

where $[e]_{R_E}$ denotes the $R_E$-equivalence class of $e$.

Now define

$$\underline{M}, \overline{M} : V \times V \to \mathbb{I}([0, 1])$$

by

$$\underline{M}(x, y) := \begin{cases} \underline{P}(\{x, y\}), & x \neq y \text{ and } \{x, y\} \in E^*, \\ [0, 0], & x = y \text{ or } \{x, y\} \notin E^*, \end{cases}$$

and

$$\overline{M}(x, y) := \begin{cases} \overline{P}(\{x, y\}), & x \neq y \text{ and } \{x, y\} \in E^*, \\ [0, 0], & x = y \text{ or } \{x, y\} \notin E^*. \end{cases}$$

Assume further that, for every edge

$$\{x, y\} \in E^*,$$

the following compatibility conditions hold:

$$\underline{P}(\{x, y\}) \preceq \underline{\Sigma}(x) \wedge \underline{\Sigma}(y),$$

and

$$\overline{P}(\{x, y\}) \preceq \overline{\Sigma}(x) \wedge \overline{\Sigma}(y).$$

Then the pair

$$\mathcal{G}^R = \big(\underline{G}, \overline{G}\big) = \big((V, \underline{\Sigma}, \underline{M}),\ (V, \overline{\Sigma}, \overline{M})\big)$$

is called an *uncertain rough graph.*

Here,

$$\underline{G} = (V, \underline{\Sigma}, \underline{M})$$

is called the *lower approximation uncertain graph*, and

$$\overline{G} = (V, \overline{\Sigma}, \overline{M})$$

is called the *upper approximation uncertain graph.*

**Theorem 4.32.3** (Well-definedness of uncertain rough graphs)**.** *Let the notation and assumptions be as in Definition 1. Then:*

1. *the mappings*
$$\underline{\Sigma}, \overline{\Sigma} : V \to \mathbb{I}([0,1])$$
*and*
$$\underline{P}, \overline{P} : E^* \to \mathbb{I}([0,1])$$
*are well-defined;*
2. *for all $x \in V$ and $e \in E^*$,*
$$\underline{\Sigma}(x) \preceq \overline{\Sigma}(x), \qquad \underline{P}(e) \preceq \overline{P}(e);$$
3. *the mappings*
$$\underline{M}, \overline{M} : V \times V \to \mathbb{I}([0,1])$$
*are well-defined and symmetric;*
4. *both*
$$\underline{G} = (V, \underline{\Sigma}, \underline{M}) \qquad \text{and} \qquad \overline{G} = (V, \overline{\Sigma}, \overline{M})$$
*are uncertain graphs;*
5. *therefore,*
$$\mathcal{G}^R = \big(\underline{G}, \overline{G}\big)$$
*is a well-defined uncertain rough graph.*

*Proof.* Since $V$ is finite and nonempty, every equivalence class
$$[x]_{R_V}$$
is finite and nonempty. Hence, for each fixed $x \in V$, the set
$$\{\Sigma(u) : u \in [x]_{R_V}\} \subseteq \mathbb{I}([0,1])$$
is a finite nonempty family of intervals. Therefore the quantities
$$\underline{\Sigma}(x) = \inf\{\Sigma(u) : u \in [x]_{R_V}\}$$
and
$$\overline{\Sigma}(x) = \sup\{\Sigma(u) : u \in [x]_{R_V}\}$$
exist.

Write
$$\Sigma(u) = [\Sigma^-(u), \Sigma^+(u)] \qquad (u \in V).$$
Then
$$\underline{\Sigma}(x) = \left[\min_{u \in [x]_{R_V}} \Sigma^-(u), \min_{u \in [x]_{R_V}} \Sigma^+(u)\right],$$
and
$$\overline{\Sigma}(x) = \left[\max_{u \in [x]_{R_V}} \Sigma^-(u), \max_{u \in [x]_{R_V}} \Sigma^+(u)\right].$$
Because
$$\Sigma^-(u) \le \Sigma^+(u) \qquad (\forall\, u \in [x]_{R_V}),$$
we obtain
$$\min_{u \in [x]_{R_V}} \Sigma^-(u) \le \min_{u \in [x]_{R_V}} \Sigma^+(u),$$
and
$$\max_{u \in [x]_{R_V}} \Sigma^-(u) \le \max_{u \in [x]_{R_V}} \Sigma^+(u).$$

Thus both

$$\underline{\Sigma}(x), \overline{\Sigma}(x) \in \mathbb{I}([0,1]).$$

Hence

$$\underline{\Sigma}, \overline{\Sigma} : V \to \mathbb{I}([0,1])$$

are well-defined.

Exactly the same argument applies to the edge approximations. Indeed, since $E^*$ is finite, every equivalence class

$$[e]_{R_E}$$

is finite and nonempty. Therefore, for each $e \in E^*$, the intervals

$$\underline{P}(e) = \inf\{P(f) : f \in [e]_{R_E}\}$$

and

$$\overline{P}(e) = \sup\{P(f) : f \in [e]_{R_E}\}$$

exist and belong to $\mathbb{I}([0,1])$. Thus

$$\underline{P}, \overline{P} : E^* \to \mathbb{I}([0,1])$$

are well-defined. This proves part (1).

Next, let $x \in V$. For every $u \in [x]_{R_V}$,

$$\min_{v \in [x]_{R_V}} \Sigma^-(v) \le \Sigma^-(u) \le \max_{v \in [x]_{R_V}} \Sigma^-(v),$$

and similarly,

$$\min_{v \in [x]_{R_V}} \Sigma^+(v) \le \Sigma^+(u) \le \max_{v \in [x]_{R_V}} \Sigma^+(v).$$

Hence

$$\underline{\Sigma}(x) \preceq \overline{\Sigma}(x).$$

Likewise, for every $e \in E^*$,

$$\underline{P}(e) \preceq \overline{P}(e).$$

Thus part (2) follows.

We now show that

$$\underline{M}, \overline{M} : V \times V \to \mathbb{I}([0,1])$$

are well-defined. Fix $(x,y) \in V \times V$. Exactly one of the following cases occurs:

$$x = y, \qquad x \ne y \text{ and } \{x,y\} \in E^*, \qquad x \ne y \text{ and } \{x,y\} \notin E^*.$$

Hence the piecewise formulas defining $\underline{M}$ and $\overline{M}$ assign a unique value to each ordered pair $(x,y)$.

If

$$x \ne y \text{ and } \{x,y\} \in E^*,$$

then

$$\underline{M}(x,y) = \underline{P}(\{x,y\}) \in \mathbb{I}([0,1]), \qquad \overline{M}(x,y) = \overline{P}(\{x,y\}) \in \mathbb{I}([0,1]).$$

In the remaining cases,

$$\underline{M}(x,y) = \overline{M}(x,y) = [0,0] \in \mathbb{I}([0,1]).$$

Therefore both mappings are well-defined.

To prove symmetry, let $x, y \in V$. If

$$x = y,$$

then trivially

$$\underline{M}(x,y) = \underline{M}(y,x) = [0,0], \qquad \overline{M}(x,y) = \overline{M}(y,x) = [0,0].$$

If
$$x \neq y,$$
then
$$\{x, y\} = \{y, x\}.$$
Therefore, if $\{x, y\} \in E^*$,
$$\underline{M}(x, y) = \underline{P}(\{x, y\}) = \underline{P}(\{y, x\}) = \underline{M}(y, x),$$
and similarly,
$$\overline{M}(x, y) = \overline{P}(\{x, y\}) = \overline{P}(\{y, x\}) = \overline{M}(y, x).$$
If $\{x, y\} \notin E^*$, then both values are $[0, 0]$ in either order. Hence $\underline{M}$ and $\overline{M}$ are symmetric. This proves part (3).

Next, we show that $\underline{G}$ is an uncertain graph. We already know that
$$\underline{\Sigma} : V \to \mathbb{I}([0, 1])$$
and
$$\underline{M} : V \times V \to \mathbb{I}([0, 1])$$
are well-defined, and that $\underline{M}$ is symmetric. It remains to verify that
$$\underline{M}(x, y) \preceq \underline{\Sigma}(x) \wedge \underline{\Sigma}(y) \qquad (\forall\, x, y \in V).$$

If
$$x \neq y \text{ and } \{x, y\} \in E^*,$$
then, by the compatibility assumption in Definition 1,
$$\underline{M}(x, y) = \underline{P}(\{x, y\}) \preceq \underline{\Sigma}(x) \wedge \underline{\Sigma}(y).$$

If
$$x = y \quad \text{or} \quad \{x, y\} \notin E^*,$$
then
$$\underline{M}(x, y) = [0, 0].$$
Since
$$\underline{\Sigma}(x), \underline{\Sigma}(y) \in \mathbb{I}([0, 1]),$$
their meet also belongs to $\mathbb{I}([0, 1])$, and hence
$$[0, 0] \preceq \underline{\Sigma}(x) \wedge \underline{\Sigma}(y).$$
Therefore
$$\underline{M}(x, y) \preceq \underline{\Sigma}(x) \wedge \underline{\Sigma}(y) \qquad (\forall\, x, y \in V),$$
so
$$\underline{G} = (V, \underline{\Sigma}, \underline{M})$$
is an uncertain graph.

The proof for
$$\overline{G} = (V, \overline{\Sigma}, \overline{M})$$
is identical. Indeed, by the upper compatibility condition,
$$\overline{P}(\{x, y\}) \preceq \overline{\Sigma}(x) \wedge \overline{\Sigma}(y) \qquad (\forall\, \{x, y\} \in E^*),$$
and outside $E^*$ the value is $[0, 0]$. Thus
$$\overline{M}(x, y) \preceq \overline{\Sigma}(x) \wedge \overline{\Sigma}(y) \qquad (\forall\, x, y \in V),$$
so $\overline{G}$ is also an uncertain graph. This proves part (4).

Finally, since both $\underline{G}$ and $\overline{G}$ are uncertain graphs, the ordered pair
$$\mathcal{G}^R = (\underline{G}, \overline{G})$$
is well-defined. Hence $\mathcal{G}^R$ is a well-defined uncertain rough graph. This proves part (5). □

Representative rough-graph concepts under uncertainty-aware graph frameworks are listed in Table 4.22.

Table 4.22: Representative rough-graph concepts under uncertainty-aware graph frameworks, classified by the dimension $k$ of the information attached to vertices and/or edges.

| $k$ | **Rough-graph concept** | **Typical coordinate form** | **Canonical information attached to vertices/edges** |
|---|---|---|---|
| 1 | Fuzzy Rough Graph | $\mu$ | A rough graph studied in a fuzzy framework, where each vertex and edge is associated with a single membership degree in $[0,1]$. |
| 2 | Intuitionistic Fuzzy Rough Graph [593–596] | $(\mu,\nu)$ | A rough graph defined in an intuitionistic fuzzy framework, where each vertex and edge carries a membership degree and a non-membership degree, usually satisfying $\mu+\nu\le 1$. |
| 3 | Neutrosophic Rough Graph [597] | $(T,I,F)$ | A rough graph defined in a neutrosophic framework, where each vertex and edge is described by truth, indeterminacy, and falsity degrees. |

## 4.33 Uncertain Soft Expert Graph

Fuzzy soft expert sets combine fuzzy sets, soft sets, and expert opinions to represent uncertain, parameterized information, enabling decision-making through graded membership and expert-based evaluations [598–601]. A fuzzy soft expert graph is a parameterized family of fuzzy graphs indexed by experts, attributes, and opinions, modeling uncertain relationships under expert-based evaluations [588, 602].

**Definition 4.33.1** (Fuzzy Soft Expert Graph)**.** [588, 602] Let

$$G^* = (V,E)$$

be a simple graph, let

$$Y$$

be a set of parameters,

$$X$$

a set of experts, and

$$O = \{1,0\}$$

the set of opinions, where 1 denotes agreement and 0 denotes disagreement.

Set

$$Z := Y \times X \times O, \qquad A \subseteq Z.$$

A *Fuzzy Soft Expert Graph* (briefly, *FSEG*) over $G^*$ is a quadruple

$$\mathcal{G} = (G^*, A, f, g),$$

where

$$f : A \to \mathcal{F}(V), \qquad g : A \to \mathcal{F}(V \times V),$$

are fuzzy soft sets over $V$ and $V \times V$, respectively, such that for every

$$\alpha \in A,$$

the pair

$$\mathcal{H}(\alpha) = \big(f(\alpha), g(\alpha)\big)$$

is a fuzzy subgraph of $G^*$. Equivalently, if

$$f(\alpha) = f_\alpha, \qquad g(\alpha) = g_\alpha,$$

then for all $x, y \in V$,

$$g_\alpha(x,y) \le \min\{f_\alpha(x), f_\alpha(y)\}.$$

Here,

$$f_\alpha : V \to [0,1]$$

gives the membership degree of each vertex under the expert-parameter-opinion triple $\alpha$, and

$$g_\alpha : V \times V \to [0,1]$$

gives the membership degree of each edge under $\alpha$.

Thus, a fuzzy soft expert graph can be viewed as a parameterized family of fuzzy graphs indexed by expert opinions and parameters.

An uncertain soft expert graph is a family of uncertain graphs indexed by parameters, experts, and opinions, where the uncertainty of vertices and edges is represented by closed subintervals of $[0,1]$.

First, let

$$\mathbb{I}([0,1]) := \big\{[a^-, a^+] \subseteq [0,1] : \ 0 \le a^- \le a^+ \le 1\big\}$$

denote the family of all closed subintervals of $[0,1]$. For

$$I = [a^-, a^+], \qquad J = [b^-, b^+] \in \mathbb{I}([0,1]),$$

define the componentwise partial order

$$I \preceq J \iff a^- \le b^- \ \text{ and } \ a^+ \le b^+,$$

and define

$$I \wedge J := \big[\min\{a^-, b^-\}, \min\{a^+, b^+\}\big].$$

Recall that an *uncertain graph* is a triple

$$G = (V, \Sigma, M),$$

where

$$\Sigma : V \to \mathbb{I}([0,1]), \qquad M : V \times V \to \mathbb{I}([0,1]),$$

such that $M$ is symmetric and

$$M(u,v) \preceq \Sigma(u) \wedge \Sigma(v) \qquad (\forall\, u, v \in V).$$

Let

$$G^* = (V, E)$$

be a simple graph, where

$$E \subseteq \big\{\{u,v\} \subseteq V : u \ne v\big\}.$$

Let

$$Y$$

be a nonempty set of parameters,

$$X$$

a nonempty set of experts, and

$$O = \{1, 0\}$$

the set of opinions, where 1 denotes agreement and 0 denotes disagreement. Set

$$Z := Y \times X \times O, \qquad A \subseteq Z, \qquad A \ne \varnothing.$$

Also define

$$P(V) := \{f : V \to \mathbb{I}([0,1])\}, \qquad P(E) := \{g : E \to \mathbb{I}([0,1])\}.$$

**Definition 4.33.2** (Uncertain Soft Expert Graph)**.** An *uncertain soft expert graph* over $G^*$ is a quadruple

$$\mathcal{G} = (G^*, A, \widetilde{\Sigma}, \widetilde{M}),$$

where

$$\widetilde{\Sigma} : A \to P(V)$$

is an uncertain soft expert set on the vertex set $V$, and

$$\widetilde{M} : A \to P(E)$$

is an uncertain soft expert set on the edge set $E$, such that for every

$$\alpha = (y, x, o) \in A$$

and every edge

$$e = \{u, v\} \in E,$$

the following condition holds:

$$\widetilde{M}(\alpha)(e) \preceq \widetilde{\Sigma}(\alpha)(u) \wedge \widetilde{\Sigma}(\alpha)(v).$$

For each

$$\alpha \in A,$$

define

$$\Sigma_\alpha : V \to \mathbb{I}([0,1]), \qquad \Sigma_\alpha(u) := \widetilde{\Sigma}(\alpha)(u),$$

and define

$$M_\alpha : V \times V \to \mathbb{I}([0,1])$$

by

$$M_\alpha(u, v) := \begin{cases} \widetilde{M}(\alpha)(\{u, v\}), & u \neq v \text{ and } \{u, v\} \in E, \\ [0, 0], & u = v \text{ or } \{u, v\} \notin E. \end{cases}$$

Then

$$H(\alpha) := (V, \Sigma_\alpha, M_\alpha)$$

is called the *uncertain graph associated with the expert-parameter-opinion triple* $\alpha$.

Thus, an uncertain soft expert graph may be regarded as a family

$$\{H(\alpha) : \alpha \in A\}$$

of uncertain graphs indexed by parameters, experts, and opinions.

**Theorem 4.33.3** (Well-definedness of uncertain soft expert graphs)**.** *Let*

$$\mathcal{G} = (G^*, A, \widetilde{\Sigma}, \widetilde{M})$$

*be an uncertain soft expert graph over the simple graph*

$$G^* = (V, E).$$

*Then, for every*

$$\alpha \in A,$$

*the following statements hold:*

1. *the mappings*

$$\Sigma_\alpha : V \to \mathbb{I}([0,1]) \qquad \text{and} \qquad M_\alpha : V \times V \to \mathbb{I}([0,1])$$

   *are well-defined;*

2. $M_\alpha$ *is symmetric, that is,*

$$M_\alpha(u, v) = M_\alpha(v, u) \qquad (\forall\, u, v \in V);$$

3. *for all*
$$u, v \in V,$$
*we have*
$$M_\alpha(u, v) \preceq \Sigma_\alpha(u) \wedge \Sigma_\alpha(v);$$

4. *therefore,*
$$H(\alpha) = (V, \Sigma_\alpha, M_\alpha)$$
*is an uncertain graph.*

*Hence every uncertain soft expert graph determines a well-defined family of uncertain graphs indexed by* $A \subseteq Y \times X \times O$*.*

*Proof.* Fix an arbitrary
$$\alpha \in A.$$

We first prove that $\Sigma_\alpha$ is well-defined. Since
$$\widetilde{\Sigma} : A \to P(V),$$
the value
$$\widetilde{\Sigma}(\alpha)$$
is a function from $V$ into $\mathbb{I}([0, 1])$. Hence, for every
$$u \in V,$$
the value
$$\Sigma_\alpha(u) := \widetilde{\Sigma}(\alpha)(u)$$
is uniquely determined and belongs to $\mathbb{I}([0, 1])$. Therefore
$$\Sigma_\alpha : V \to \mathbb{I}([0, 1])$$
is well-defined.

Next, we prove that $M_\alpha$ is well-defined. Let
$$(u, v) \in V \times V.$$
Since $G^* = (V, E)$ is a simple graph, exactly one of the following mutually exclusive cases occurs:
$$u = v, \qquad u \neq v \text{ and } \{u, v\} \in E, \qquad u \neq v \text{ and } \{u, v\} \notin E.$$
Hence the piecewise formula
$$M_\alpha(u, v) := \begin{cases} \widetilde{M}(\alpha)(\{u, v\}), & u \neq v \text{ and } \{u, v\} \in E, \\ [0, 0], & u = v \text{ or } \{u, v\} \notin E \end{cases}$$
assigns exactly one value to each ordered pair
$$(u, v) \in V \times V.$$

Moreover, if
$$u \neq v \text{ and } \{u, v\} \in E,$$
then
$$\widetilde{M}(\alpha)(\{u, v\}) \in \mathbb{I}([0, 1]),$$
because
$$\widetilde{M}(\alpha) \in P(E).$$
In the remaining cases,
$$M_\alpha(u, v) = [0, 0] \in \mathbb{I}([0, 1]).$$

Therefore

$$M_\alpha : V \times V \to \mathbb{I}([0,1])$$

is well-defined. This proves part (1).

We now prove symmetry. Take arbitrary

$$u, v \in V.$$

If

$$u = v,$$

then clearly

$$M_\alpha(u,v) = M_\alpha(v,u) = [0,0].$$

Assume next that

$$u \neq v.$$

If

$$\{u,v\} \notin E,$$

then

$$\{v,u\} \notin E$$

as well, since

$$\{u,v\} = \{v,u\}$$

as unordered pairs. Hence

$$M_\alpha(u,v) = M_\alpha(v,u) = [0,0].$$

If

$$\{u,v\} \in E,$$

then again

$$\{u,v\} = \{v,u\},$$

so

$$M_\alpha(u,v) = \widetilde{M}(\alpha)(\{u,v\}) = \widetilde{M}(\alpha)(\{v,u\}) = M_\alpha(v,u).$$

Thus

$$M_\alpha(u,v) = M_\alpha(v,u) \qquad (\forall\, u,v \in V),$$

and part (2) follows.

Next, we verify the uncertain graph condition. Let

$$u, v \in V.$$

If

$$u \neq v \text{ and } \{u,v\} \in E,$$

then by the defining condition of uncertain soft expert graphs,

$$\widetilde{M}(\alpha)(\{u,v\}) \preceq \widetilde{\Sigma}(\alpha)(u) \wedge \widetilde{\Sigma}(\alpha)(v).$$

Hence

$$M_\alpha(u,v) = \widetilde{M}(\alpha)(\{u,v\}) \preceq \Sigma_\alpha(u) \wedge \Sigma_\alpha(v).$$

If

$$u = v \quad \text{or} \quad \{u,v\} \notin E,$$

then

$$M_\alpha(u,v) = [0,0].$$

Since

$$\Sigma_\alpha(u), \Sigma_\alpha(v) \in \mathbb{I}([0,1]),$$

their meet

$$\Sigma_\alpha(u) \wedge \Sigma_\alpha(v)$$

also belongs to $\mathbb{I}([0,1])$, and therefore

$$[0,0] \preceq \Sigma_\alpha(u) \wedge \Sigma_\alpha(v).$$

Thus, in all cases,

$$M_\alpha(u,v) \preceq \Sigma_\alpha(u) \wedge \Sigma_\alpha(v).$$

This proves part (3).

By parts (1), (2), and (3), the triple

$$H(\alpha) = (V, \Sigma_\alpha, M_\alpha)$$

satisfies the defining axioms of an uncertain graph. Therefore

$$H(\alpha)$$

is an uncertain graph. This proves part (4).

Since

$$\alpha \in A$$

was arbitrary, the conclusion holds for every expert-parameter-opinion triple in $A$. Hence the family

$$\{H(\alpha) : \alpha \in A\}$$

is well-defined. Therefore every uncertain soft expert graph determines a well-defined family of uncertain graphs indexed by

$$A \subseteq Y \times X \times O.$$

□

Representative soft-expert-graph concepts under uncertainty-aware graph frameworks are listed in Table 4.23.

Table 4.23: Representative soft-expert-graph concepts under uncertainty-aware graph frameworks, classified by the dimension $k$ of the information attached to vertices and/or edges for each parameter–expert–opinion instance.

| $k$ | **Soft-expert-graph concept** | **Typical coordinate form** | **Canonical information attached to vertices/edges** |
|---|---|---|---|
| 1 | Fuzzy Soft Expert Graph | $(e,x,o) \mapsto \mu$ | A soft expert graph in which, for each parameter–expert–opinion instance $(e,x,o)$, the associated graph is described by a single membership degree on vertices and edges. |
| 2 | Intuitionistic Fuzzy Soft Expert Graph [603] | $(e,x,o) \mapsto (\mu,\nu)$ | A soft expert graph in which, for each parameter–expert–opinion instance $(e,x,o)$, each vertex and edge carries a membership degree and a non-membership degree, usually satisfying $\mu + \nu \leq 1$. |
| 3 | Neutrosophic Soft Expert Graph [604, 605] | $(e,x,o) \mapsto (T,I,F)$ | A soft expert graph in which, for each parameter–expert–opinion instance $(e,x,o)$, each vertex and edge is described by truth, indeterminacy, and falsity degrees. |
| $s+t$ | Plithogenic Soft Expert Graph [50] | $(e,x,o) \mapsto (\mathbf{a},\mathbf{c}) \in [0,1]^s \times [0,1]^t$ | A soft expert graph in which, for each parameter–expert–opinion instance $(e,x,o)$, each vertex and edge is described by attribute-based information together with an $s$-dimensional appurtenance vector and a $t$-dimensional contradiction vector. |

## 4.34 Uncertain Eulerian Graph

A fuzzy Eulerian graph is a connected fuzzy graph whose support admits a closed trail traversing every positive-membership edge exactly once and returning to start [606].

**Definition 4.34.1** (Fuzzy Eulerian Circuit)**.** [606] Let

$$G = (V, \sigma, \mu)$$

be a fuzzy graph, where

$$\sigma : V \to [0,1], \qquad \mu : V \times V \to [0,1], \qquad \mu(x,y) = \mu(y,x), \qquad \mu(x,y) \le \min\{\sigma(x), \sigma(y)\}$$

for all $x, y \in V$.

Define the support vertex set and support edge set by

$$V^* := \{x \in V : \sigma(x) > 0\},$$

and

$$E^* := \big\{\{x,y\} \subseteq V^* : x \neq y,\ \mu(x,y) > 0\big\}.$$

The crisp graph

$$G^* := (V^*, E^*)$$

is called the *support graph* of $G$.

A *fuzzy Eulerian circuit* in $G$ is a closed trail

$$W : x_0, e_1, x_1, e_2, \dots, e_m, x_m$$

in the support graph $G^*$ such that

$$x_0 = x_m,$$

all edges $e_1, e_2, \dots, e_m$ are pairwise distinct, and

$$\{e_1, e_2, \dots, e_m\} = E^*.$$

That is, $W$ traverses every support edge of $G$ exactly once and returns to its initial vertex.

**Definition 4.34.2** (Fuzzy Eulerian Graph)**.** A fuzzy graph

$$G = (V, \sigma, \mu)$$

is called a *fuzzy Eulerian graph* if its support graph $G^*$ is connected and $G$ admits a fuzzy Eulerian circuit.

Equivalently, $G$ is fuzzy Eulerian if and only if the crisp support graph

$$G^* = (V^*, E^*)$$

is an Eulerian graph in the ordinary sense.

## 4.35 Uncertain Hamiltonian Graph

A fuzzy Hamiltonian graph is a fuzzy graph whose support contains a cycle visiting every positive-membership vertex exactly once, except for repeating the initial vertex [607–609].

**Definition 4.35.1** (Fuzzy Hamiltonian Cycle)**.** Let

$$G = (V, \sigma, \mu)$$

be a fuzzy graph with support graph

$$G^* = (V^*, E^*).$$

A *fuzzy Hamiltonian cycle* in $G$ is a cycle

$$C : x_0, x_1, \ldots, x_{n-1}, x_n$$

in $G^*$ such that

$$x_0 = x_n,$$

the vertices

$$x_0, x_1, \ldots, x_{n-1}$$

are pairwise distinct, and

$$\{x_0, x_1, \ldots, x_{n-1}\} = V^*.$$

In other words, $C$ visits every support vertex of $G$ exactly once, except that the initial vertex is repeated at the end.

**Definition 4.35.2** (Fuzzy Hamiltonian Graph)**.** A fuzzy graph

$$G = (V, \sigma, \mu)$$

is called a *fuzzy Hamiltonian graph* if its support graph $G^*$ contains a fuzzy Hamiltonian cycle.

Equivalently, $G$ is fuzzy Hamiltonian if and only if the support graph

$$G^* = (V^*, E^*)$$

is Hamiltonian in the ordinary graph-theoretic sense.

Representative Hamiltonian-cycle-related concepts in uncertainty-aware graph frameworks are listed in Table 4.24.

Table 4.24: Representative Hamiltonian-cycle-related concepts under uncertainty-aware graph frameworks, classified by the dimension $k$ of the information attached to vertices and/or edges.

| $k$ | **Hamiltonian-cycle-related concept** | **Typical coordinate form** | **Canonical information attached to vertices/edges** |
|---|---|---|---|
| 1 | Fuzzy Hamiltonian Cycle | $\mu$ | A Hamiltonian cycle studied in a fuzzy framework, where each vertex and edge is associated with a single membership degree in $[0, 1]$. |
| 2 | Intuitionistic Fuzzy Hamiltonian Cycle [610] | $(\mu, \nu)$ | A Hamiltonian cycle defined in an intuitionistic fuzzy framework, where each vertex and edge carries a membership degree and a non-membership degree, usually satisfying $\mu+\nu \leq 1$. |
| 3 | Neutrosophic Hamiltonian Cycle [611, 612] | $(T, I, F)$ | A Hamiltonian cycle defined in a neutrosophic framework, where each vertex and edge is described by truth, indeterminacy, and falsity degrees. |

## 4.36 Uncertain Spanning Tree

Fuzzy spanning tree is a spanning subgraph of a fuzzy graph whose support contains all vertices, is connected, and acyclic [613–615].

**Definition 4.36.1** (Fuzzy Spanning Subgraph)**.** Let

$$G = (V, \sigma, \mu)$$

be a fuzzy graph, where

$$\sigma : V \to [0,1], \qquad \mu : V \times V \to [0,1], \qquad \mu(x,y) = \mu(y,x), \qquad \mu(x,y) \le \min\{\sigma(x), \sigma(y)\}$$

for all $x, y \in V$.

A fuzzy graph

$$H = (V, \sigma, \nu)$$

is called a *fuzzy spanning subgraph* of $G$ if

$$\nu(x,y) \le \mu(x,y) \qquad (\forall\, x, y \in V).$$

Thus, $H$ has the same vertex set and the same vertex-membership function as $G$, while its edge-memberships are obtained by deleting or weakening some edges of $G$.

**Definition 4.36.2** (Support Graph)**.** Let

$$G = (V, \sigma, \mu)$$

be a fuzzy graph. Its *support graph* is the crisp graph

$$G^* = (V^*, E^*),$$

where

$$V^* := \{x \in V : \sigma(x) > 0\},$$

and

$$E^* := \big\{\{x,y\} \subseteq V^* : x \ne y,\ \mu(x,y) > 0\big\}.$$

**Definition 4.36.3** (Fuzzy Spanning Tree)**.** Let

$$G = (V, \sigma, \mu)$$

be a connected fuzzy graph, and let

$$T = (V, \sigma, \tau)$$

be a fuzzy spanning subgraph of $G$.

Then $T$ is called a *fuzzy spanning tree* of $G$ if the support graph

$$T^* = \big(V^*, E_T^*\big)$$

is a spanning tree of the support graph

$$G^* = \big(V^*, E_G^*\big).$$

Equivalently, $T$ is a fuzzy spanning tree of $G$ if:

1. $$\tau(x,y) \le \mu(x,y) \qquad (\forall\, x, y \in V),$$
   so that $T$ is a fuzzy spanning subgraph of $G$;
2. $$V(T^*) = V(G^*) = V^*;$$
3. $T^*$ is connected;
4. $T^*$ contains no cycle.

An uncertain spanning tree is an uncertain spanning subgraph whose support graph is a spanning tree of the support graph of the original uncertain graph.

**Definition 4.36.4** (Spanning-Tree-Evaluable Uncertain Model)**.** Let $M$ be an uncertain model with degree-domain

$$\mathrm{Dom}(M) \subseteq [0,1]^k.$$

We say that $M$ is *spanning-tree-evaluable* if it is equipped with a distinguished element

$$0_M \in \mathrm{Dom}(M),$$

called the *zero degree.*

**Definition 4.36.5** (Uncertain Spanning Subgraph)**.** Let $V$ be a finite nonempty set, let $M$ be a spanning-tree-evaluable uncertain model, and let

$$G_M = (V, \sigma_M, \eta_M)$$

be an uncertain graph of type $M$, where

$$\sigma_M : V \to \mathrm{Dom}(M), \qquad \eta_M : \binom{V}{2} \to \mathrm{Dom}(M).$$

Equivalently,

$$(V, \sigma_M)$$

is an Uncertain Set of type $M$ on the vertex set $V$, and

$$\left(\binom{V}{2}, \eta_M\right)$$

is an Uncertain Set of type $M$ on the set of all unordered pairs of distinct vertices.

An uncertain graph

$$H_M = (V, \sigma_M, \tau_M)$$

is called an *uncertain spanning subgraph* of $G_M$ if

$$\tau_M : \binom{V}{2} \to \mathrm{Dom}(M)$$

satisfies

$$\tau_M(e) \in \{0_M, \eta_M(e)\} \qquad \left(\forall\, e \in \binom{V}{2}\right).$$

Thus, $H_M$ has the same vertex set and the same vertex-degree function as $G_M$, while each edge-degree is either kept unchanged or deleted by replacing it with the zero degree.

**Definition 4.36.6** (Support Graph)**.** Let

$$G_M = (V, \sigma_M, \eta_M)$$

be an uncertain graph of type $M$, where

$$\sigma_M : V \to \mathrm{Dom}(M), \qquad \eta_M : \binom{V}{2} \to \mathrm{Dom}(M).$$

Its *support vertex set* is defined by

$$V_M^* := \{x \in V : \sigma_M(x) \neq 0_M\},$$

and its *support edge set* is defined by

$$E_M^* := \left\{ \{x,y\} \in \binom{V_M^*}{2} : \eta_M(\{x,y\}) \neq 0_M \right\}.$$

The *support graph* of $G_M$ is the crisp graph

$$G_M^* := (V_M^*, E_M^*).$$

**Definition 4.36.7** (Connected Uncertain Graph)**.** Let

$$G_M = (V, \sigma_M, \eta_M)$$

be an uncertain graph of type $M$. We say that $G_M$ is *connected* if its support graph

$$G_M^*$$

is connected in the ordinary graph-theoretic sense.

**Definition 4.36.8** (Uncertain Spanning Tree)**.** Let

$$G_M = (V, \sigma_M, \eta_M)$$

be a connected uncertain graph of type $M$, and let

$$T_M = (V, \sigma_M, \tau_M)$$

be an uncertain spanning subgraph of $G_M$.

Then $T_M$ is called an *uncertain spanning tree* of $G_M$ if the support graph

$$T_M^* = (V_M^*, E_T^*)$$

is a spanning tree of the support graph

$$G_M^* = (V_M^*, E_G^*).$$

Equivalently, $T_M$ is an uncertain spanning tree of $G_M$ if:

1. $$\tau_M(e) \in \{0_M, \eta_M(e)\} \qquad \left(\forall\, e \in \binom{V}{2}\right),$$
   so that $T_M$ is an uncertain spanning subgraph of $G_M$;
2. $$V(T_M^*) = V(G_M^*) = V_M^*;$$
3. $T_M^*$ is connected;
4. $T_M^*$ contains no cycle.

**Theorem 4.36.9** (Well-definedness of Uncertain Spanning Tree)**.** *Let $V$ be a finite nonempty set, let $M$ be a spanning-tree-evaluable uncertain model with degree-domain*

$$\mathrm{Dom}(M) \subseteq [0,1]^k$$

*and zero degree*

$$0_M \in \mathrm{Dom}(M),$$

*and let*

$$G_M = (V, \sigma_M, \eta_M)$$

*be an uncertain graph of type $M$. Then:*

1. *the support graph $G_M^*$ is well-defined;*
2. *every uncertain spanning subgraph*
   $$H_M = (V, \sigma_M, \tau_M)$$
   *of $G_M$ is well-defined;*
3. *the support graph $H_M^*$ of every uncertain spanning subgraph $H_M$ is well-defined;*

4. *consequently, the statement*

$$\text{“}T_M \textit{ is an uncertain spanning tree of } G_M\text{”}$$

*is well-defined.*

*Proof.* Since $M$ is an uncertain model, the degree-domain

$$\mathrm{Dom}(M)$$

is fixed. Since $M$ is spanning-tree-evaluable, the element

$$0_M \in \mathrm{Dom}(M)$$

is also fixed.

Because

$$\sigma_M : V \to \mathrm{Dom}(M) \qquad \text{and} \qquad \eta_M : \binom{V}{2} \to \mathrm{Dom}(M)$$

are functions, the pairs

$$(V, \sigma_M) \qquad \text{and} \qquad \left(\binom{V}{2}, \eta_M\right)$$

are well-defined Uncertain Sets of type $M$.

Hence, for each $x \in V$, the statement

$$\sigma_M(x) \neq 0_M$$

has a definite truth value, and therefore

$$V_M^* := \{x \in V : \sigma_M(x) \neq 0_M\}$$

is a well-defined subset of $V$.

Likewise, for each unordered pair

$$\{x, y\} \in \binom{V_M^*}{2},$$

the statement

$$\eta_M(\{x, y\}) \neq 0_M$$

has a definite truth value, and therefore

$$E_M^* := \left\{ \{x, y\} \in \binom{V_M^*}{2} : \eta_M(\{x, y\}) \neq 0_M \right\}$$

is a well-defined subset of $\binom{V_M^*}{2}$. Consequently,

$$G_M^* = (V_M^*, E_M^*)$$

is a well-defined crisp graph. This proves (1).

Now let

$$H_M = (V, \sigma_M, \tau_M)$$

be an uncertain spanning subgraph of $G_M$. By definition,

$$\tau_M : \binom{V}{2} \to \mathrm{Dom}(M)$$

is a function such that

$$\tau_M(e) \in \{0_M, \eta_M(e)\} \qquad \left(\forall\, e \in \binom{V}{2}\right).$$

Since both $0_M$ and $\eta_M(e)$ belong to $\mathrm{Dom}(M)$, every value $\tau_M(e)$ belongs to $\mathrm{Dom}(M)$. Hence

$$\left(\binom{V}{2}, \tau_M\right)$$

is a well-defined Uncertain Set of type $M$, and therefore

$$H_M = (V, \sigma_M, \tau_M)$$

is a well-defined uncertain graph of type $M$. This proves (2).

Next, the support vertex set of $H_M$ is

$$V_H^* := \{x \in V : \sigma_M(x) \neq 0_M\} = V_M^*,$$

which is already known to be well-defined. Its support edge set is

$$E_H^* := \left\{ \{x, y\} \in \binom{V_H^*}{2} : \tau_M(\{x, y\}) \neq 0_M \right\}.$$

Since $\tau_M$ is a function into $\mathrm{Dom}(M)$ and $0_M$ is fixed, the predicate

$$\tau_M(\{x, y\}) \neq 0_M$$

has a definite truth value for every $\{x, y\} \in \binom{V_H^*}{2}$. Therefore $E_H^*$ is well-defined, and so

$$H_M^* = (V_H^*, E_H^*)$$

is a well-defined crisp graph. This proves (3).

Finally, the statement

"$T_M$ is an uncertain spanning tree of $G_M$"

means exactly that:

1. $T_M$ is an uncertain spanning subgraph of $G_M$;
2. the support graph $T_M^*$ has the same vertex set as $G_M^*$;
3. $T_M^*$ is connected;
4. $T_M^*$ is acyclic.

By parts (1)–(3), both support graphs are well-defined crisp graphs. Connectedness, acyclicity, and the property of being a spanning tree are standard graph-theoretic properties with definite truth values for crisp graphs. Hence the above statement is well-defined.

Therefore, the notion of an uncertain spanning tree is well-defined. □

For reference, representative spanning-tree extensions classified by the dimension k are listed in Table 4.25.

Besides uncertain spanning trees, various related concepts are also known. For example, related extensions and generalizations of spanning trees include minimum spanning trees, maximum spanning trees [622], rooted spanning trees [623], directed spanning trees (arborescences) [624], spanning forests [625], and Steiner trees [626].

Table 4.25: Representative spanning-tree extensions classified by the dimension $k$ of the uncertainty information attached to vertices and/or edges.

| $k$ | **Spanning Tree Type** | **Typical coordinate form** | **Canonical information attached to vertices/edges** |
|---|---|---|---|
| 1 | Fuzzy Spanning Tree | $\mu$ | A spanning tree in a fuzzy graph, where each vertex and edge is associated with a single membership degree in $[0, 1]$. |
| 2 | Intuitionistic Fuzzy Spanning Tree [616, 617] | $(\mu, \nu)$ | A spanning tree in an intuitionistic fuzzy graph, where each vertex and edge carries a membership degree and a non-membership degree, usually satisfying $\mu + \nu \leq 1$. |
| 3 | Neutrosophic Spanning Tree [618–621] | $(T, I, F)$ | A spanning tree in a neutrosophic graph, where each vertex and edge is described by three mutually distinguished components: truth, indeterminacy, and falsity. |
| 4 | Double-valued Neutrosophic Spanning Tree [123] | $(T, I_1, I_2, F)$ | A spanning tree in a double-valued neutrosophic framework, where each vertex and edge is represented by four primary coordinates, typically one truth degree, two distinct indeterminacy degrees, and one falsity degree. |

# Chapter 5

# Uncertain Graph Parameters

In this chapter, we discuss graph parameters. Graph parameters are numerical or structural invariants, such as degree, chromatic number, or diameter, that quantify, classify, and compare essential abstract mathematical properties of graphs. Uncertain graph parameters generalize classical invariants to graphs with uncertain vertex or edge information, measuring structure, connectivity, domination, and complexity under uncertainty.

## 5.1 Domination Number in Uncertain Graph

Domination number in a fuzzy graph is the minimum cardinality of a vertex set whose neighborhood memberships collectively sufficiently dominate every vertex to required degrees [627–629].

**Definition 5.1.1** (Domination Number in a Fuzzy Graph)**.** [627–629] Let $G = (V, \sigma, \mu)$ be a finite fuzzy graph, where

$$\sigma : V \to [0, 1], \qquad \mu : V \times V \to [0, 1], \qquad \mu(u, v) \le \min\{\sigma(u), \sigma(v)\} \quad (\forall\, u, v \in V).$$

For $u, v \in V$, the *strength of connectedness* between $u$ and $v$ is defined by

$$\mu_G^{\infty}(u, v) = \sup\Big\{ \min_{0 \le i < m} \mu(v_i, v_{i+1}) : v_0 = u,\ v_m = v,\ (v_0, v_1, \dots, v_m) \text{ is a } u\text{–}v \text{ path in } G \Big\}.$$

An edge $(u, v)$ is called a *strong arc* if

$$\mu(u, v) = \mu_G^{\infty}(u, v).$$

A subset $D \subseteq V$ is called a *dominating set* of $G$ if for every $v \in V \setminus D$, there exists $u \in D$ such that

$$(u, v) \text{ is a strong arc} \qquad \text{and} \qquad \sigma(u) \ge \sigma(v).$$

The *fuzzy cardinality* of $D$ is

$$|D|_f := \sum_{u \in D} \sigma(u).$$

The *domination number* of $G$ is defined by

$$\gamma_f(G) = \min\big\{ |D|_f : D \subseteq V \text{ is a dominating set of } G \big\}.$$

Any dominating set $D$ satisfying $|D|_f = \gamma_f(G)$ is called a *minimum dominating set.*

Table 5.1: Representative domination-related concepts under uncertainty-aware graph frameworks, classified by the dimension $k$ of the information attached to vertices and/or edges.

| $k$ | **Domination-related concept** | **Typical coordinate form** | **Canonical information attached to vertices/edges** |
|---|---|---|---|
| 2 | Domination Number in Intuitionistic Fuzzy Graph [630, 631] | $(\mu, \nu)$ | Domination is defined on an intuitionistic fuzzy graph, where each vertex and edge carries a membership degree and a non-membership degree, usually satisfying $\mu + \nu \leq 1$. |
| 2 | Domination in Bipolar Fuzzy Graphs [632–634] | $(\mu^+, \mu^-)$ | Domination is studied in a bipolar fuzzy graph, where each vertex and edge is described by a positive membership degree and a negative membership degree, representing supportive and counteractive information simultaneously. |
| 2 | Domination Number in Vague Graph [635–637] | $(t, f)$ | Domination is defined on a vague graph, where each vertex and edge is characterized by a truth-membership degree and a falsity-membership degree, typically with $t + f \leq 1$. |
| 3 | Domination Number in Picture Fuzzy Graph [638, 639] | $(\mu, \eta, \nu)$ | Domination is defined on a picture fuzzy graph, where each vertex and edge has positive, neutral, and negative membership degrees, usually satisfying $\mu + \eta + \nu \leq 1$. |
| $k \in \mathbb{N}$ | Domination Number in Hesitant Fuzzy Graph [640] | $\{\mu_1, \ldots, \mu_k\} \subseteq [0, 1]$ | Domination is defined on a hesitant fuzzy graph, where each vertex and edge is associated with a finite set of possible membership degrees rather than a single value; thus the information dimension is variable. |
| 3 | Domination Number in Neutrosophic Graph [641–643] | $(T, I, F)$ | Domination is defined on a neutrosophic graph, where each vertex and edge is described by truth, indeterminacy, and falsity degrees. |

Representative domination-related concepts under uncertainty-aware graph frameworks are listed in Table 5.1.

Besides uncertain domination number, various related concepts are also known. Related extensions of the domination number in graphs include total domination number [644], connected domination number [645], independent domination number [646], paired domination number [647], Roman domination number [648], double domination number [649], distance domination number [650], restrained domination number [651], broadcast domination number [652], rainbow domination number [653], and fractional domination number [654].

## 5.2 Secure Domination Number in Uncertain Graph

Secure domination number in a fuzzy graph is the minimum size of a dominating set where each outside vertex can replace defender without losing domination [655].

**Definition 5.2.1** (Secure Domination Number in a Fuzzy Graph)**.** [656] Let $G = (V, \sigma, \mu)$ be a finite fuzzy graph, where

$$\sigma : V \to [0, 1], \qquad \mu : V \times V \to [0, 1]$$

is symmetric and satisfies

$$\mu(u, v) \leq \min\{\sigma(u), \sigma(v)\} \qquad (\forall\, u, v \in V).$$

For any subset $S \subseteq V$, define its *fuzzy cardinality* by

$$|S|_f := \sum_{v \in S} \sigma(v).$$

A vertex $u \in V$ is said to *dominate* a vertex $v \in V$ if

$$\mu(u, v) = \min\{\sigma(u), \sigma(v)\}.$$

A subset $S \subseteq V$ is called a *dominating set* of $G$ if for every

$$x \in V \setminus S,$$

there exists some

$$y \in S$$

such that $y$ dominates $x$.

A dominating set $S \subseteq V$ is called a *secure dominating set* if for every

$$x \in V \setminus S,$$

there exists a vertex

$$y \in S \quad \text{with} \quad \mu(x, y) > 0$$

such that the exchanged set

$$(S \setminus \{y\}) \cup \{x\}$$

is again a dominating set of $G$.

The *secure domination number* of $G$ is defined by

$$\gamma_s(G) := \min\{|S|_f : S \subseteq V \text{ is a secure dominating set of } G\}.$$

Any secure dominating set $S$ satisfying

$$|S|_f = \gamma_s(G)$$

is called a *minimum secure dominating set.*

For convenience, Table 5.2 summarizes representative secure-domination-related concepts according to the dimension $k$ of the information associated with vertices and/or edges.

Table 5.2: Representative secure-domination-related concepts under uncertainty-aware graph frameworks, classified by the dimension $k$ of the information attached to vertices and/or edges.

| $k$ | **Secure domination-related concept** | **Typical coordinate form** | **Canonical information attached to vertices/edges** |
|---|---|---|---|
| 1 | Secure Domination Number in Fuzzy Graph | $\mu$ | Secure domination is studied in a fuzzy graph, where each vertex and edge is associated with a single membership degree in $[0, 1]$. |
| 2 | Secure Domination Number in Intuitionistic Fuzzy Graph [655, 657] | $(\mu, \nu)$ | Secure domination is defined on an intuitionistic fuzzy graph, where each vertex and edge carries a membership degree and a non-membership degree, usually satisfying $\mu + \nu \leq 1$. |
| 3 | Secure Domination Number in Neutrosophic Graph [658, 659] | $(T, I, F)$ | Secure domination is defined on a neutrosophic graph, where each vertex and edge is described by truth, indeterminacy, and falsity degrees. |

Related concepts such as Secure Total Domination Number [660], Secure Connected Domination Number [661], and Perfect Secure Domination Number [662] are also known.

## 5.3 Regularity in Uncertain Graph

Regularity in an uncertain graph means every vertex has the same uncertain degree, so the network exhibits uniform connectivity structure despite interval-valued or uncertain memberships [663].

**Definition 5.3.1** (Regularity in a Fuzzy Graph)**.** Let

$$G = (V, \sigma, \mu)$$

be a finite fuzzy graph, where

$$\sigma : V \to [0,1], \qquad \mu : V \times V \to [0,1]$$

satisfies

$$\mu(u,v) \leq \min\{\sigma(u), \sigma(v)\} \qquad (\forall\, u, v \in V),$$

and $\mu$ is symmetric.

The *degree* of a vertex $v \in V$ is defined by

$$d_G(v) := \sum_{\substack{u \in V \\ u \neq v}} \mu(v,u).$$

The fuzzy graph $G$ is called *$r$-regular* if there exists a constant $r \geq 0$ such that

$$d_G(v) = r \qquad \text{for all } v \in V.$$

Equivalently, $G$ is called *regular* if all vertices of $G$ have the same degree.

**Definition 5.3.2** (Degree-Evaluable Uncertain Model)**.** Let $M$ be an uncertain model with degree-domain

$$\mathrm{Dom}(M) \subseteq [0,1]^k.$$

We say that $M$ is *degree-evaluable* if it is equipped with a map

$$\Delta_M : \mathrm{Dom}(M) \to [0, \infty),$$

called the *degree-evaluation map.*

**Definition 5.3.3** (Uncertain Graph)**.** Let $V$ be a finite nonempty set, and let $M$ be a degree-evaluable uncertain model. An *uncertain graph* of type $M$ on $V$ is a triple

$$G_M = (V, \sigma_M, \eta_M),$$

where

$$\sigma_M : V \to \mathrm{Dom}(M), \qquad \eta_M : \binom{V}{2} \to \mathrm{Dom}(M)$$

are functions.

Equivalently,

$$(V, \sigma_M)$$

is an Uncertain Set of type $M$ on the vertex set $V$, and

$$\left(\binom{V}{2}, \eta_M\right)$$

is an Uncertain Set of type $M$ on the set of unordered pairs of distinct vertices.

**Definition 5.3.4** (Regularity in an Uncertain Graph)**.** Let

$$G_M = (V, \sigma_M, \eta_M)$$

be a finite uncertain graph of type $M$, where $M$ is degree-evaluable with degree-evaluation map

$$\Delta_M : \mathrm{Dom}(M) \to [0, \infty).$$

For each vertex $v \in V$, define the *degree* of $v$ in $G_M$ by

$$d_{G_M}(v) := \sum_{\substack{u \in V \\ u \neq v}} \Delta_M\big(\eta_M(\{u, v\})\big).$$

The uncertain graph $G_M$ is called *$r$-regular* if there exists a constant

$$r \in [0, \infty)$$

such that

$$d_{G_M}(v) = r \qquad \text{for all } v \in V.$$

Equivalently, $G_M$ is called *regular* if all vertices have the same degree.

**Theorem 5.3.5** (Well-definedness of Regularity in an Uncertain Graph)**.** *Let $V$ be a finite nonempty set, let $M$ be a degree-evaluable uncertain model, and let*

$$G_M = (V, \sigma_M, \eta_M)$$

*be an uncertain graph of type $M$ on $V$. Then:*

1. *for each $v \in V$, the degree*
$$d_{G_M}(v) = \sum_{\substack{u \in V \\ u \neq v}} \Delta_M\big(\eta_M(\{u, v\})\big)$$
*is a well-defined element of $[0, \infty)$;*
2. *for every $r \in [0, \infty)$, the statement*
$$d_{G_M}(v) = r \qquad (\forall\, v \in V)$$
*is meaningful;*
3. *consequently, the statement*
$$\text{“}G_M \text{ is regular”}$$
*is well-defined;*
4. *if $G_M$ is regular, then the regularity constant $r$ is unique.*

*Proof.* Since $M$ is an uncertain model, its degree-domain

$$\mathrm{Dom}(M) \subseteq [0, 1]^k$$

is fixed. Since $M$ is degree-evaluable, the map

$$\Delta_M : \mathrm{Dom}(M) \to [0, \infty)$$

is also fixed.

Because

$$\eta_M : \binom{V}{2} \to \mathrm{Dom}(M)$$

is a function, for each unordered pair

$$\{u, v\} \in \binom{V}{2},$$

the value

$$\eta_M(\{u, v\}) \in \mathrm{Dom}(M)$$

is uniquely determined. Hence, for every such pair,

$$\Delta_M\big(\eta_M(\{u, v\})\big) \in [0, \infty)$$

is a well-defined nonnegative real number.

Now fix $v \in V$. Since $V$ is finite, the set

$$\{u \in V : u \neq v\}$$

is finite. Therefore

$$\sum_{\substack{u \in V \\ u \neq v}} \Delta_M\big(\eta_M(\{u, v\})\big)$$

is a finite sum of well-defined nonnegative real numbers. Thus

$$d_{G_M}(v) \in [0, \infty)$$

is well-defined. This proves (1).

Since $d_{G_M}(v)$ is a well-defined real number for every $v \in V$, the equality

$$d_{G_M}(v) = r$$

has a definite truth value for every $r \in [0, \infty)$. Hence the universal statement

$$d_{G_M}(v) = r \qquad (\forall\, v \in V)$$

is meaningful. This proves (2).

By definition, $G_M$ is regular if and only if

$$\exists\, r \in [0, \infty) \quad \text{such that} \quad d_{G_M}(v) = r \text{ for all } v \in V.$$

Since the predicate

$$d_{G_M}(v) = r \qquad (\forall\, v \in V)$$

is meaningful for each $r \in [0, \infty)$, the above existential statement is also meaningful. Therefore the notion of regularity in an uncertain graph is well-defined. This proves (3).

Finally, assume that $G_M$ is regular and that both $r, s \in [0, \infty)$ satisfy

$$d_{G_M}(v) = r \qquad \text{for all } v \in V,$$

and

$$d_{G_M}(v) = s \qquad \text{for all } v \in V.$$

Because $V \neq \varnothing$, choose any $v_0 \in V$. Then

$$r = d_{G_M}(v_0) = s.$$

Hence the regularity constant is unique. This proves (4). □

## 5.4 Planarity in Uncertain Graph

Planarity in a fuzzy graph measures whether its underlying structure admits an embedding with no crossings or quantifies crossing tolerance through membership-weighted intersections between edges (cf. [493]).

**Definition 5.4.1** (Planarity in a Fuzzy Graph)**.** Let

$$G = (V, \sigma, \mu)$$

be a finite fuzzy graph, where

$$\sigma : V \to [0,1], \qquad \mu : V \times V \to [0,1], \qquad \mu(u,v) \le \min\{\sigma(u), \sigma(v)\} \quad (\forall\, u, v \in V).$$

Let

$$E^* := \big\{\{u,v\} \subseteq V : u \neq v,\ \mu(u,v) > 0\big\},$$

and let

$$G^* = (V, E^*)$$

be the underlying crisp graph of $G$.

For an edge $e = \{u,v\} \in E^*$, define its *fuzzy edge strength* by

$$s_G(e) := \frac{\mu(u,v)}{\min\{\sigma(u), \sigma(v)\}}.$$

This is well-defined for every $e \in E^*$.

Now fix a plane drawing $D$ of the crisp graph $G^*$. If two drawn edges $e_1, e_2 \in E^*$ intersect at a crossing point $\theta$ in $D$, define the *crossing value* of $\theta$ by

$$\Lambda_D(\theta) := \frac{s_G(e_1) + s_G(e_2)}{2}.$$

Let

$$\Theta_D = \{\theta_1, \theta_2, \ldots, \theta_m\}$$

be the set of all crossing points in the drawing $D$. The *planarity value* of $G$ with respect to $D$ is defined by

$$\vartheta_D(G) := \frac{1}{1 + \sum_{i=1}^{m} \Lambda_D(\theta_i)}.$$

The *planarity* of the fuzzy graph $G$ is then defined by

$$\vartheta(G) := \sup_{D} \vartheta_D(G),$$

where the supremum is taken over all plane drawings $D$ of the underlying crisp graph $G^*$.

We say that $G$ is *planar* if

$$\vartheta(G) = 1.$$

Equivalently, $G$ is planar if and only if its underlying crisp graph $G^*$ admits a crossing-free plane embedding.

More generally, for $\varepsilon \in (0,1)$, one may call $G$ *$\varepsilon$-planar* if

$$\vartheta(G) > \varepsilon.$$

**Definition 5.4.2** (Planarity-Evaluable Uncertain Model)**.** Let $M$ be an uncertain model with degree-domain

$$\mathrm{Dom}(M) \subseteq [0,1]^k.$$

We say that $M$ is *planarity-evaluable* if it is equipped with:

1. a distinguished element
$$0_M \in \mathrm{Dom}(M),$$
called the *zero degree*;
2. an *edge-activity map*
$$\alpha_M : \mathrm{Dom}(M) \to [0,\infty),$$
such that
$$\alpha_M(d) = 0 \iff d = 0_M;$$
3. a *vertex-capacity map*
$$\beta_M : \mathrm{Dom}(M) \times \mathrm{Dom}(M) \to [0,\infty),$$
such that
$$\beta_M(a,b) = \beta_M(b,a) \qquad (\forall\, a,b \in \mathrm{Dom}(M)),$$
and
$$\beta_M(a,b) > 0 \qquad \text{whenever } a \neq 0_M \text{ and } b \neq 0_M.$$

**Definition 5.4.3** (Uncertain Graph)**.** Let $V$ be a finite nonempty set, and let $M$ be a planarity-evaluable uncertain model. An *uncertain graph* of type $M$ on $V$ is a triple

$$G_M = (V, \sigma_M, \eta_M),$$

where

$$\sigma_M : V \to \mathrm{Dom}(M), \qquad \eta_M : \binom{V}{2} \to \mathrm{Dom}(M)$$

are functions.

Equivalently,

$$(V, \sigma_M)$$

is an Uncertain Set of type $M$ on the vertex set $V$, and

$$\left(\binom{V}{2}, \eta_M\right)$$

is an Uncertain Set of type $M$ on the set of unordered pairs of distinct vertices.

**Definition 5.4.4** (Planarity in an Uncertain Graph)**.** Let

$$G_M = (V, \sigma_M, \eta_M)$$

be a finite uncertain graph of type $M$, where $M$ is planarity-evaluable.

Define the *support vertex set* of $G_M$ by

$$V_M^* := \{u \in V : \sigma_M(u) \neq 0_M\},$$

and define the *support edge set* of $G_M$ by

$$E_M^* := \left\{ \{u,v\} \in \binom{V_M^*}{2} : \eta_M(\{u,v\}) \neq 0_M \right\}.$$

The associated *support graph* is the crisp graph

$$G_M^* := (V_M^*, E_M^*).$$

For an edge

$$e = \{u, v\} \in E_M^*,$$

define its *uncertain edge strength* by

$$s_{G_M}(e) := \frac{\alpha_M\big(\eta_M(e)\big)}{\beta_M\big(\sigma_M(u), \sigma_M(v)\big)}.$$

Now let $D$ be a plane drawing of the crisp graph $G_M^*$ in general position; that is, vertices are drawn as distinct points, edges are drawn as Jordan arcs joining their endpoints, no edge passes through a nonincident vertex, no edge intersects itself, any two edges intersect in only finitely many points, and no three edges meet at one interior crossing point.

If two drawn edges

$$e_1, e_2 \in E_M^*$$

intersect at a crossing point $\theta$ in $D$, define the *crossing value* of $\theta$ by

$$\Lambda_D(\theta) := \frac{s_{G_M}(e_1) + s_{G_M}(e_2)}{2}.$$

Let

$$\Theta_D$$

denote the set of all crossing points in the drawing $D$. The *planarity value* of $G_M$ with respect to $D$ is defined by

$$\vartheta_D(G_M) := \frac{1}{1 + \sum_{\theta \in \Theta_D} \Lambda_D(\theta)}.$$

The *planarity* of the uncertain graph $G_M$ is then defined by

$$\vartheta(G_M) := \sup_D \vartheta_D(G_M),$$

where the supremum is taken over all plane drawings $D$ of $G_M^*$ in general position.

We say that $G_M$ is *planar* if

$$\vartheta(G_M) = 1.$$

Equivalently, $G_M$ is planar if and only if its support graph $G_M^*$ admits a crossing-free plane embedding.

More generally, for $\varepsilon \in (0, 1)$, one may call $G_M$ *$\varepsilon$-planar* if

$$\vartheta(G_M) > \varepsilon.$$

**Theorem 5.4.5** (Well-definedness of Planarity in an Uncertain Graph)**.** *Let $V$ be a finite nonempty set, let $M$ be a planarity-evaluable uncertain model, and let*

$$G_M = (V, \sigma_M, \eta_M)$$

*be an uncertain graph of type $M$ on $V$. Then:*

1. *the support graph*

$$G_M^* = (V_M^*, E_M^*)$$

*is well-defined;*

2. *for every edge*

$$e = \{u, v\} \in E_M^*,$$

*the uncertain edge strength*

$$s_{G_M}(e) = \frac{\alpha_M(\eta_M(e))}{\beta_M(\sigma_M(u), \sigma_M(v))}$$

*is a well-defined positive real number;*

3. *for every plane drawing $D$ of $G_M^*$ in general position, the set*

$$\Theta_D$$

*of crossing points is finite, each crossing value $\Lambda_D(\theta)$ is well-defined, and the quantity*

$$\vartheta_D(G_M)$$

*is a well-defined real number satisfying*

$$0 < \vartheta_D(G_M) \leq 1;$$

4. *the global planarity value*

$$\vartheta(G_M) = \sup_D \vartheta_D(G_M)$$

*is well-defined and satisfies*

$$0 < \vartheta(G_M) \leq 1;$$

5. *the statement*

$$\text{“}G_M \text{ is planar”}$$

*is well-defined, and*

$$\vartheta(G_M) = 1 \iff G_M^* \text{ is planar.}$$

6. *for every $\varepsilon \in (0, 1)$, the statement*

$$\text{“}G_M \text{ is } \varepsilon\text{-planar”}$$

*is well-defined.*

*Proof.* Since $M$ is an uncertain model, its degree-domain

$$\mathrm{Dom}(M) \subseteq [0, 1]^k$$

is fixed. Since $M$ is planarity-evaluable, the element

$$0_M \in \mathrm{Dom}(M)$$

and the maps

$$\alpha_M : \mathrm{Dom}(M) \to [0, \infty), \qquad \beta_M : \mathrm{Dom}(M) \times \mathrm{Dom}(M) \to [0, \infty)$$

are fixed as part of the data.

Because

$$\sigma_M : V \to \mathrm{Dom}(M) \qquad \text{and} \qquad \eta_M : \binom{V}{2} \to \mathrm{Dom}(M)$$

are functions, the pairs

$$(V, \sigma_M) \qquad \text{and} \qquad \left(\binom{V}{2}, \eta_M\right)$$

are well-defined Uncertain Sets of type $M$.

Therefore, for each $u \in V$, the statement

$$\sigma_M(u) \neq 0_M$$

has a definite truth value, and hence

$$V_M^* = \{u \in V : \sigma_M(u) \neq 0_M\}$$

is a well-defined subset of $V$.

Likewise, for each unordered pair

$$\{u,v\} \in \binom{V_M^*}{2},$$

the statement

$$\eta_M(\{u,v\}) \neq 0_M$$

has a definite truth value, and hence

$$E_M^* = \left\{ \{u,v\} \in \binom{V_M^*}{2} : \eta_M(\{u,v\}) \neq 0_M \right\}$$

is a well-defined subset of $\binom{V_M^*}{2}$. Consequently,

$$G_M^* = (V_M^*, E_M^*)$$

is a well-defined finite crisp graph. This proves (1).

Now let

$$e = \{u,v\} \in E_M^*.$$

Then $u, v \in V_M^*$, so

$$\sigma_M(u) \neq 0_M \qquad \text{and} \qquad \sigma_M(v) \neq 0_M.$$

By the defining property of $\beta_M$,

$$\beta_M(\sigma_M(u), \sigma_M(v)) > 0.$$

Also, since $e \in E_M^*$, we have

$$\eta_M(e) \neq 0_M,$$

and by the defining property of $\alpha_M$,

$$\alpha_M(\eta_M(e)) > 0.$$

Hence

$$s_{G_M}(e) = \frac{\alpha_M(\eta_M(e))}{\beta_M(\sigma_M(u), \sigma_M(v))}$$

is a well-defined positive real number. This proves (2).

Fix a plane drawing $D$ of $G_M^*$ in general position. Since $G_M^*$ is finite, it has only finitely many edges. Because each pair of drawn edges intersects in only finitely many points and there are only finitely many pairs of edges, the set

$$\Theta_D$$

of crossing points is finite.

If $\theta \in \Theta_D$, then by definition $\theta$ arises from two distinct crossed edges

$$e_1, e_2 \in E_M^*.$$

By part (2), both

$$s_{G_M}(e_1) \qquad \text{and} \qquad s_{G_M}(e_2)$$

are well-defined positive real numbers. Therefore

$$\Lambda_D(\theta) = \frac{s_{G_M}(e_1) + s_{G_M}(e_2)}{2}$$

is also a well-defined positive real number.

Since $\Theta_D$ is finite, the sum

$$\sum_{\theta \in \Theta_D} \Lambda_D(\theta)$$

is a well-defined finite nonnegative real number. Hence

$$\vartheta_D(G_M) = \frac{1}{1 + \sum_{\theta \in \Theta_D} \Lambda_D(\theta)}$$

is a well-defined real number. Moreover,

$$1 + \sum_{\theta \in \Theta_D} \Lambda_D(\theta) \geq 1,$$

so

$$0 < \vartheta_D(G_M) \leq 1.$$

This proves (3).

Let

$$S := \{\vartheta_D(G_M) : D \text{ is a plane drawing of } G_M^* \text{ in general position}\}.$$

The set $S$ is nonempty, because every finite graph admits a plane drawing in general position. By part (3),

$$S \subseteq (0, 1].$$

Therefore $S$ is nonempty and bounded above by 1, so by completeness of $\mathbb{R}$, its supremum exists. Hence

$$\vartheta(G_M) := \sup S$$

is well-defined and satisfies

$$0 < \vartheta(G_M) \leq 1.$$

This proves (4).

We now prove (5). Assume first that $G_M^*$ is planar. Then there exists a crossing-free plane embedding $D_0$ of $G_M^*$. Thus

$$\Theta_{D_0} = \varnothing,$$

so

$$\vartheta_{D_0}(G_M) = \frac{1}{1+0} = 1.$$

Hence

$$\vartheta(G_M) \geq 1.$$

Together with $\vartheta(G_M) \leq 1$, this gives

$$\vartheta(G_M) = 1.$$

Conversely, assume that $G_M^*$ is nonplanar. Then every plane drawing $D$ of $G_M^*$ has at least one crossing point. Consider the finite set

$$\mathcal{C} := \left\{ \frac{s_{G_M}(e) + s_{G_M}(f)}{2} : e, f \in E_M^*,\ e \neq f \right\}.$$

Every element of $\mathcal{C}$ is a positive real number by part (2), and $\mathcal{C}$ is finite. Hence, if $\mathcal{C} \neq \varnothing$, it has a minimum

$$m := \min \mathcal{C} > 0.$$

Since $G_M^*$ is nonplanar, every drawing $D$ has at least one crossing $\theta$, and thus

$$\sum_{\theta \in \Theta_D} \Lambda_D(\theta) \geq m.$$

Therefore

$$\vartheta_D(G_M) \leq \frac{1}{1+m} < 1$$

for every such drawing $D$. Hence

$$\vartheta(G_M) \leq \frac{1}{1+m} < 1.$$

Thus $\vartheta(G_M) \neq 1$.

Therefore

$$\vartheta(G_M) = 1 \iff G_M^* \text{ is planar.}$$

Since planarity of a finite crisp graph is a definite graph-theoretic property, the statement

"$G_M$ is planar"

is well-defined.

Finally, for any fixed $\varepsilon \in (0,1)$, the value $\vartheta(G_M)$ is well-defined by (4), so the inequality

$$\vartheta(G_M) > \varepsilon$$

has a definite truth value. Thus the statement

"$G_M$ is $\varepsilon$-planar"

is well-defined. This proves (6). □

## 5.5 Uncertain Tree-width

A tree-decomposition represents a graph by overlapping vertex bags arranged in a tree, preserving edge containment and connected vertex occurrence for structural analysis and algorithms [664–667]. A fuzzy tree-decomposition represents a fuzzy graph by tree-structured fuzzy bags, preserving vertex connectedness and edge coverage, thereby measuring how the graph resembles a tree.

**Definition 5.5.1** (Fuzzy Tree-Decomposition and Fuzzy Tree-Width)**.** Let

$$G = (V, \sigma, \mu)$$

be a finite fuzzy graph, where

$$\sigma : V \to [0,1], \qquad \mu : V \times V \to [0,1], \qquad \mu(u,v) \leq \min\{\sigma(u), \sigma(v)\} \quad (\forall\, u, v \in V),$$

and assume that $\mu$ is symmetric and $G$ has no loops.

A *fuzzy tree-decomposition* of $G$ is a pair

$$\big(T, \{\beta_t\}_{t \in I}\big),$$

where

$$T = (I, F)$$

is a finite tree and, for each $t \in I$,

$$\beta_t : V \to [0,1]$$

is a fuzzy subset of $V$ (called a *fuzzy bag*), satisfying the following conditions:

(FT1) **Bag domination by vertex memberships:** for every $t \in I$ and every $v \in V$,

$$\beta_t(v) \leq \sigma(v).$$

(FT2) **Vertex coverage:** for every $v \in V$,

$$\sigma(v) = \max_{t \in I} \beta_t(v).$$

(FT3) **Running-intersection property:** for every vertex $v \in V$, the index set

$$I_v := \{\, t \in I : \beta_t(v) > 0 \,\}$$

induces a connected subtree of $T$.

(FT4) **Edge coverage:** for every pair $u, v \in V$ with $\mu(u,v) > 0$, there exists some node $t \in I$ such that

$$\min\{\beta_t(u), \beta_t(v)\} \geq \mu(u,v).$$

The *fuzzy cardinality* of a fuzzy bag $\beta_t$ is defined by

$$|\beta_t|_f := \sum_{v \in V} \beta_t(v).$$

The *width* of the fuzzy tree-decomposition

$$\big(T, \{\beta_t\}_{t\in I}\big)$$

is

$$\operatorname{width}\big(T, \{\beta_t\}_{t\in I}\big) := \max_{t \in I}\big(|\beta_t|_f - 1\big).$$

The *fuzzy tree-width* of $G$ is defined by

$$\operatorname{ftw}(G) := \inf\Big\{\operatorname{width}\big(T, \{\beta_t\}_{t\in I}\big) : \big(T, \{\beta_t\}_{t\in I}\big) \text{ is a fuzzy tree-decomposition of } G\Big\}.$$

**Definition 5.5.2** (Tree-Decomposition-Evaluable Uncertain Model)**.** Let $M$ be an uncertain model with degree-domain

$$\operatorname{Dom}(M) \subseteq [0,1]^k.$$

We say that $M$ is *tree-decomposition-evaluable* if it is equipped with:

1. a distinguished element
$$0_M \in \operatorname{Dom}(M),$$
called the *zero degree*;
2. a partial order
$$\preceq_M \subseteq \operatorname{Dom}(M) \times \operatorname{Dom}(M);$$
3. for every finite nonempty subset $A \subseteq \operatorname{Dom}(M)$, an element
$$\bigvee_M A \in \operatorname{Dom}(M),$$
called the *finite join* of $A$, which is the least upper bound of $A$ with respect to $\preceq_M$;
4. a symmetric map
$$\Gamma_M : \operatorname{Dom}(M) \times \operatorname{Dom}(M) \to \operatorname{Dom}(M),$$
called the *pair-capacity map*;
5. a map
$$\omega_M : \operatorname{Dom}(M) \to [0, \infty),$$
called the *bag-size evaluation map*.

**Definition 5.5.3** (Uncertain Graph of Type $M$)**.** Let $V$ be a finite nonempty set, and let $M$ be a tree-decomposition-evaluable uncertain model. An *uncertain graph* of type $M$ on $V$ is a triple

$$G_M = (V, \sigma_M, \eta_M),$$

where

$$\sigma_M : V \to \operatorname{Dom}(M), \qquad \eta_M : \binom{V}{2} \to \operatorname{Dom}(M)$$

are functions satisfying

$$\eta_M(\{u,v\}) \preceq_M \Gamma_M\big(\sigma_M(u), \sigma_M(v)\big) \qquad \left(\forall\, \{u,v\} \in \binom{V}{2}\right).$$

Equivalently,

$$(V, \sigma_M)$$

is an Uncertain Set of type $M$ on the vertex set $V$, and

$$\left(\binom{V}{2}, \eta_M\right)$$

is an Uncertain Set of type $M$ on the set of unordered pairs of distinct vertices.

**Definition 5.5.4** (Support Edge Set)**.** Let

$$G_M = (V, \sigma_M, \eta_M)$$

be an uncertain graph of type $M$. Its *support edge set* is defined by

$$E^*_M(G_M) := \left\{ e \in \binom{V}{2} : \eta_M(e) \neq 0_M \right\}.$$

**Definition 5.5.5** (Uncertain Tree-Decomposition)**.** Let

$$G_M = (V, \sigma_M, \eta_M)$$

be a finite uncertain graph of type $M$, where $M$ is tree-decomposition-evaluable.

An *uncertain tree-decomposition* of $G_M$ is a pair

$$\big(T, \{\beta_t\}_{t \in I}\big),$$

where

$$T = (I, F)$$

is a finite tree and, for each $t \in I$,

$$\beta_t : V \to \mathrm{Dom}(M)$$

is a function, called an *uncertain bag*, such that the following conditions hold:

(UT1) **Bag domination by vertex degrees:** for every $t \in I$ and every $v \in V$,

$$\beta_t(v) \preceq_M \sigma_M(v).$$

(UT2) **Vertex coverage:** for every $v \in V$,

$$\sigma_M(v) = \bigvee_M \{\beta_t(v) : t \in I\}.$$

(UT3) **Running-intersection property:** for every $v \in V$, the index set

$$I_v := \{\, t \in I : \beta_t(v) \neq 0_M \,\}$$

is either empty or induces a connected subtree of $T$.

(UT4) **Edge coverage:** for every support edge

$$e = \{u, v\} \in E^*_M(G_M),$$

there exists some node $t \in I$ such that

$$\eta_M(\{u, v\}) \preceq_M \Gamma_M\big(\beta_t(u), \beta_t(v)\big).$$

For each $t \in I$, the pair

$$B_t := (V, \beta_t)$$

is called the *uncertain bag* at $t$.

**Definition 5.5.6** (Uncertain Bag Size and Width)**.** Let

$$\big(T, \{\beta_t\}_{t\in I}\big)$$

be an uncertain tree-decomposition of an uncertain graph $G_M$.

For each $t \in I$, define the *uncertain size* of the bag $B_t = (V, \beta_t)$ by

$$|B_t|_M := \sum_{v\in V} \omega_M\big(\beta_t(v)\big).$$

The *width* of the uncertain tree-decomposition

$$\big(T, \{\beta_t\}_{t\in I}\big)$$

is defined by

$$\mathrm{width}_M\big(T, \{\beta_t\}_{t\in I}\big) := \max_{t\in I}\big(|B_t|_M - 1\big).$$

**Definition 5.5.7** (Uncertain Tree-Width)**.** Let

$$G_M = (V, \sigma_M, \eta_M)$$

be a finite uncertain graph of type $M$.

The *uncertain tree-width* of $G_M$ is defined by

$$\mathrm{utw}_M(G_M) := \inf\Big\{\mathrm{width}_M\big(T, \{\beta_t\}_{t\in I}\big) : \big(T, \{\beta_t\}_{t\in I}\big) \text{ is an uncertain tree-decomposition of } G_M\Big\}.$$

**Theorem 5.5.8** (Well-definedness of Uncertain Tree-Decomposition and Uncertain Tree-Width)**.** *Let $V$ be a finite nonempty set, let $M$ be a tree-decomposition-evaluable uncertain model, and let*

$$G_M = (V, \sigma_M, \eta_M)$$

*be an uncertain graph of type $M$ on $V$. Then:*

1. *the support edge set*
$$E^*_M(G_M) = \{\, e \in \binom{V}{2} : \eta_M(e) \neq 0_M \,\}$$
*is well-defined;*
2. *for every finite tree $T = (I, F)$ and every family of maps*
$$\beta_t : V \to \mathrm{Dom}(M) \qquad (t \in I),$$
*each bag*
$$B_t = (V, \beta_t)$$
*is a well-defined Uncertain Set of type $M$;*
3. *the conditions* (UT1)–(UT4) *are meaningful and have definite truth values;*
4. *for every uncertain tree-decomposition*
$$\big(T, \{\beta_t\}_{t\in I}\big),$$
*the bag sizes*
$$|B_t|_M$$
*and the width*
$$\mathrm{width}_M\big(T, \{\beta_t\}_{t\in I}\big)$$
*are well-defined real numbers;*
5. *the class of all uncertain tree-decompositions of $G_M$ is nonempty;*

6. *the uncertain tree-width*

$$\mathrm{utw}_M(G_M)$$

*is a well-defined real number in the interval* $[-1, \infty)$.

*Proof.* Since $M$ is an uncertain model, its degree-domain

$$\mathrm{Dom}(M) \subseteq [0,1]^k$$

is fixed. Since $M$ is tree-decomposition-evaluable, the element

$$0_M \in \mathrm{Dom}(M),$$

the partial order $\preceq_M$, the finite join operator $\bigvee_M$, the pair-capacity map $\Gamma_M$, and the bag-size evaluation map $\omega_M$ are all fixed.

Because

$$\eta_M : \binom{V}{2} \to \mathrm{Dom}(M)$$

is a function, for each

$$e \in \binom{V}{2},$$

the statement

$$\eta_M(e) \neq 0_M$$

has a definite truth value. Therefore

$$E^*_M(G_M) = \{\, e \in \binom{V}{2} : \eta_M(e) \neq 0_M \,\}$$

is a well-defined subset of $\binom{V}{2}$. This proves (1).

Now fix a finite tree

$$T = (I, F)$$

and a family of maps

$$\beta_t : V \to \mathrm{Dom}(M) \qquad (t \in I).$$

Since each $\beta_t$ is an ordinary function with codomain $\mathrm{Dom}(M)$, the pair

$$B_t = (V, \beta_t)$$

is a well-defined Uncertain Set of type $M$ on $V$. This proves (2).

We next verify that conditions (UT1)–(UT4) are meaningful.

Condition (UT1) consists only of comparisons

$$\beta_t(v) \preceq_M \sigma_M(v),$$

which are meaningful because $\preceq_M$ is a fixed partial order on $\mathrm{Dom}(M)$.

For (UT2), since $I$ is the vertex set of a finite tree, $I$ is finite and nonempty. Hence for each fixed $v \in V$, the set

$$\{\beta_t(v) : t \in I\}$$

is a finite nonempty subset of $\mathrm{Dom}(M)$, so its finite join

$$\bigvee_M \{\beta_t(v) : t \in I\}$$

is well-defined.

For (UT3), the set

$$I_v := \{\, t \in I : \beta_t(v) \neq 0_M \,\}$$

is well-defined because each statement $\beta_t(v) \neq 0_M$ has a definite truth value. Since $I_v \subseteq I$, the induced subgraph $T[I_v]$ is a well-defined graph, and thus the statement that $I_v$ is empty or induces a connected subtree of $T$ is meaningful.

For (UT4), if

$$e = \{u, v\} \in E_M^*(G_M),$$

then $\eta_M(e) \in \mathrm{Dom}(M)$, $\beta_t(u) \in \mathrm{Dom}(M)$, and $\beta_t(v) \in \mathrm{Dom}(M)$ for every $t \in I$. Hence

$$\Gamma_M\big(\beta_t(u), \beta_t(v)\big) \in \mathrm{Dom}(M),$$

and therefore the comparison

$$\eta_M(e) \preceq_M \Gamma_M\big(\beta_t(u), \beta_t(v)\big)$$

is meaningful. Thus (UT4) is meaningful as well.

Consequently, each of (UT1)–(UT4) has a definite truth value. This proves (3).

Assume now that

$$\big(T, \{\beta_t\}_{t\in I}\big)$$

is an uncertain tree-decomposition of $G_M$. For each $t \in I$ and each $v \in V$, the value

$$\omega_M\big(\beta_t(v)\big)$$

is a well-defined element of $[0, \infty)$, because $\beta_t(v) \in \mathrm{Dom}(M)$ and

$$\omega_M : \mathrm{Dom}(M) \to [0, \infty).$$

Since $V$ is finite,

$$|B_t|_M = \sum_{v\in V} \omega_M\big(\beta_t(v)\big)$$

is a finite sum of well-defined nonnegative real numbers, so it is itself well-defined.

Because $I$ is finite and nonempty, the set

$$\{\, |B_t|_M - 1 : t \in I \,\}$$

is a finite nonempty set of real numbers. Hence

$$\mathrm{width}_M\big(T, \{\beta_t\}_{t\in I}\big) = \max_{t\in I}\big(|B_t|_M - 1\big)$$

is a well-defined real number. This proves (4).

We now prove (5). Consider the one-vertex tree

$$T_0 = (\{t_0\}, \varnothing).$$

Define

$$\beta_{t_0} : V \to \mathrm{Dom}(M)$$

by

$$\beta_{t_0}(v) := \sigma_M(v) \qquad (\forall\, v \in V).$$

We claim that

$$\big(T_0, \{\beta_{t_0}\}\big)$$

is an uncertain tree-decomposition of $G_M$.

Condition (UT1) holds trivially, since

$$\beta_{t_0}(v) = \sigma_M(v) \preceq_M \sigma_M(v).$$

Condition (UT2) holds because for each $v \in V$,

$$\{\beta_{t_0}(v)\} = \{\sigma_M(v)\},$$

and the least upper bound of a singleton is the element itself; hence

$$\bigvee_M \{\beta_{t_0}(v)\} = \sigma_M(v).$$

Condition (UT3) holds because, for each $v \in V$, the set

$$I_v = \begin{cases} \{t_0\}, & \text{if } \sigma_M(v) \neq 0_M, \\ \varnothing, & \text{if } \sigma_M(v) = 0_M, \end{cases}$$

is either empty or a connected subtree of the one-vertex tree $T_0$.

Finally, let

$$e = \{u, v\} \in E^*_M(G_M).$$

Since $G_M$ is an uncertain graph of type $M$, we have

$$\eta_M(\{u, v\}) \preceq_M \Gamma_M\big(\sigma_M(u), \sigma_M(v)\big).$$

Because

$$\beta_{t_0}(u) = \sigma_M(u), \qquad \beta_{t_0}(v) = \sigma_M(v),$$

it follows that

$$\eta_M(\{u, v\}) \preceq_M \Gamma_M\big(\beta_{t_0}(u), \beta_{t_0}(v)\big).$$

Thus (UT4) holds.

Therefore

$$\big(T_0, \{\beta_{t_0}\}\big)$$

is an uncertain tree-decomposition of $G_M$, so the class of all uncertain tree-decompositions of $G_M$ is nonempty. This proves (5).

Let

$$\mathcal{W}_M(G_M) := \big\{\mathrm{width}_M\big(T, \{\beta_t\}_{t\in I}\big) : \big(T, \{\beta_t\}_{t\in I}\big) \text{ is an uncertain tree-decomposition of } G_M\big\}.$$

By (5), this set is nonempty. By (4), every element of $\mathcal{W}_M(G_M)$ is a real number.

Moreover, for every uncertain tree-decomposition and every $t \in I$,

$$|B_t|_M \geq 0,$$

because it is a sum of nonnegative real numbers. Hence

$$|B_t|_M - 1 \geq -1,$$

and therefore every width satisfies

$$\mathrm{width}_M\big(T, \{\beta_t\}_{t\in I}\big) \geq -1.$$

Thus $\mathcal{W}_M(G_M)$ is bounded below by $-1$.

Since $\mathcal{W}_M(G_M) \subseteq \mathbb{R}$ is nonempty and bounded below, its infimum exists in $\mathbb{R}$. Hence

$$\mathrm{utw}_M(G_M) = \inf \mathcal{W}_M(G_M)$$

is well-defined, and

$$\mathrm{utw}_M(G_M) \in [-1, \infty).$$

This proves (6). □

As related concepts of tree-width, notions such as clique-width [668], hypertree-width [669, 670], superhypertree-width [486], and bandwidth [671] are also known.

## 5.6 Independence number in Uncertain graphs

The independence number of a fuzzy graph is the maximum fuzzy cardinality of a vertex set whose distinct vertices are pairwise nonadjacent through strong edges (cf. [672–675]).

**Definition 5.6.1** (Independence Number in a Fuzzy Graph)**.** Let

$$G = (V, \sigma, \mu)$$

be a finite fuzzy graph, where

$$\sigma : V \to [0,1], \qquad \mu : V \times V \to [0,1], \qquad \mu(u,v) \le \min\{\sigma(u), \sigma(v)\} \quad (\forall\, u,v \in V),$$

and assume that $\mu$ is symmetric.

For $u, v \in V$, let

$$\mu_G^{\infty}(u,v) = \sup\Big\{ \min_{0 \le i < m} \mu(v_i, v_{i+1}) : v_0 = u,\ v_m = v,\ (v_0, v_1, \ldots, v_m) \text{ is a } u\text{–}v \text{ path in } G \Big\}.$$

An edge $(u, v)$ is called a *strong arc* if

$$\mu(u,v) \ge \mu_G^{\infty}(u,v).$$

Two vertices $u, v \in V$ are said to be *fuzzy independent* if there is no strong arc between them.

A subset

$$S \subseteq V$$

is called a *fuzzy independent set* if every two distinct vertices in $S$ are fuzzy independent; equivalently, for all distinct $u, v \in S$,

$$(u,v) \text{ is not a strong arc.}$$

The *fuzzy cardinality* of $S$ is defined by

$$|S|_f := \sum_{v \in S} \sigma(v).$$

The *independence number* of the fuzzy graph $G$ is defined by

$$\beta(G) := \max\big\{ |S|_f : S \subseteq V \text{ is a fuzzy independent set of } G \big\}.$$

Any fuzzy independent set $S$ satisfying

$$|S|_f = \beta(G)$$

is called a *maximum fuzzy independent set*.

**Definition 5.6.2** (Independence-Evaluable Uncertain Model)**.** Let $M$ be an uncertain model with degree-domain

$$\mathrm{Dom}(M) \subseteq [0,1]^k.$$

We say that $M$ is *independence-evaluable* if it is equipped with:

1. a distinguished element

   $$0_M \in \mathrm{Dom}(M),$$

   called the *zero degree*;

2. a total order
$$\preceq_M \subseteq \operatorname{Dom}(M) \times \operatorname{Dom}(M),$$
called the *strength order*;

3. for each integer $n \geq 1$, a map
$$\Psi_M^{(n)} : \operatorname{Dom}(M)^n \to \operatorname{Dom}(M),$$
called the *path-strength operator of length* $n$;

4. a map
$$\omega_M : \operatorname{Dom}(M) \to [0, \infty),$$
called the *vertex-weight evaluation map*.

In addition, we assume that
$$\Psi_M^{(1)}(d) = d \qquad (\forall\, d \in \operatorname{Dom}(M)).$$

**Definition 5.6.3** (Uncertain Graph of Type $M$)**.** Let $V$ be a finite nonempty set, and let $M$ be an independence-evaluable uncertain model. An *uncertain graph* of type $M$ on $V$ is a triple
$$G_M = (V, \sigma_M, \eta_M),$$
where
$$\sigma_M : V \to \operatorname{Dom}(M), \qquad \eta_M : \binom{V}{2} \to \operatorname{Dom}(M)$$
are functions.

Equivalently,
$$(V, \sigma_M)$$
is an Uncertain Set of type $M$ on the vertex set $V$, and
$$\left(\binom{V}{2}, \eta_M\right)$$
is an Uncertain Set of type $M$ on the set of unordered pairs of distinct vertices.

**Definition 5.6.4** (Support Graph)**.** Let
$$G_M = (V, \sigma_M, \eta_M)$$
be an uncertain graph of type $M$. Its *support vertex set* is
$$V_M^* := \{v \in V : \sigma_M(v) \neq 0_M\},$$
its *support edge set* is
$$E_M^* := \left\{ \{u, v\} \in \binom{V_M^*}{2} : \eta_M(\{u, v\}) \neq 0_M \right\},$$
and its *support graph* is the crisp graph
$$G_M^* := (V_M^*, E_M^*).$$

**Definition 5.6.5** (Path Strength and Strong Support Edge)**.** Let
$$G_M = (V, \sigma_M, \eta_M)$$
be a finite uncertain graph of type $M$, and let
$$e = \{u, v\} \in E_M^*$$
be a support edge.

A *simple u-v path* in $G_M^*$ is a sequence
$$P = (v_0, v_1, \ldots, v_m)$$

such that

$$v_0 = u, \qquad v_m = v,$$

the vertices $v_0, \dots, v_m$ are pairwise distinct, and

$$\{v_{i-1}, v_i\} \in E_M^* \qquad (i = 1, \dots, m).$$

For such a path $P$, define its *uncertain path strength* by

$$\mathrm{Str}_M(P) := \Psi_M^{(m)}\Big(\eta_M(\{v_0, v_1\}), \eta_M(\{v_1, v_2\}), \dots, \eta_M(\{v_{m-1}, v_m\})\Big).$$

Let

$$\mathcal{P}_{u,v}(G_M)$$

denote the set of all simple $u$-$v$ paths in $G_M^*$. Since $e = \{u, v\} \in E_M^*$, this set is nonempty.

Define the *uncertain connectedness strength* between $u$ and $v$ by

$$\eta_{G_M}^{\infty}(u, v) := \max_{\preceq_M}\big\{\mathrm{Str}_M(P) : P \in \mathcal{P}_{u,v}(G_M)\big\}.$$

The support edge

$$e = \{u, v\} \in E_M^*$$

is called a *strong support edge* if

$$\eta_M(\{u, v\}) \succeq_M \eta_{G_M}^{\infty}(u, v).$$

**Definition 5.6.6** (Independence Number in an Uncertain Graph)**.** Let

$$G_M = (V, \sigma_M, \eta_M)$$

be a finite uncertain graph of type $M$.

Two distinct vertices

$$u, v \in V$$

are said to be *uncertain independent* if there is no strong support edge joining them.

A subset

$$S \subseteq V$$

is called an *uncertain independent set* if every two distinct vertices in $S$ are uncertain independent; equivalently, for all distinct $u, v \in S$,

$$\{u, v\} \notin E_M^* \quad \text{or} \quad \{u, v\} \text{ is not a strong support edge.}$$

The *uncertain cardinality* of $S$ is defined by

$$|S|_M := \sum_{v \in S} \omega_M\big(\sigma_M(v)\big).$$

The *independence number* of $G_M$ is defined by

$$\beta_M(G_M) := \max\big\{|S|_M : S \subseteq V \text{ is an uncertain independent set of } G_M\big\}.$$

Any uncertain independent set $S \subseteq V$ satisfying

$$|S|_M = \beta_M(G_M)$$

is called a *maximum uncertain independent set*.

**Theorem 5.6.7** (Well-definedness of Independence Number in an Uncertain Graph)**.** *Let $V$ be a finite nonempty set, let $M$ be an independence-evaluable uncertain model, and let*

$$G_M = (V, \sigma_M, \eta_M)$$

*be an uncertain graph of type $M$ on $V$. Then:*

1. *the support graph*
$$G^*_M = (V^*_M, E^*_M)$$
*is well-defined;*
2. *for every support edge*
$$e = \{u, v\} \in E^*_M,$$
*the set*
$$\mathcal{P}_{u,v}(G_M)$$
*of simple u-v paths is a well-defined finite nonempty set;*
3. *for every path*
$$P \in \mathcal{P}_{u,v}(G_M),$$
*the path strength $\mathrm{Str}_M(P)$ is well-defined, and hence the connectedness strength*
$$\eta^{\infty}_{G_M}(u, v)$$
*is well-defined;*
4. *the statement*
$$\text{“}\{u, v\} \text{ is a strong support edge”}$$
*is well-defined for every $\{u, v\} \in E^*_M$;*
5. *for every subset $S \subseteq V$, the statements*
$$\text{“}S \text{ is an uncertain independent set”} \qquad \text{and} \qquad |S|_M \in [0, \infty)$$
*are well-defined;*
6. *the independence number*
$$\beta_M(G_M)$$
*is a well-defined element of $[0, \infty)$, and there exists at least one maximum uncertain independent set.*

*Proof.* Since $M$ is an uncertain model, its degree-domain

$$\mathrm{Dom}(M) \subseteq [0, 1]^k$$

is fixed. Since $M$ is independence-evaluable, the element

$$0_M \in \mathrm{Dom}(M),$$

the total order $\preceq_M$, the maps

$$\Psi^{(n)}_M : \mathrm{Dom}(M)^n \to \mathrm{Dom}(M) \qquad (n \geq 1),$$

and the evaluation map

$$\omega_M : \mathrm{Dom}(M) \to [0, \infty)$$

are all fixed.

Because

$$\sigma_M : V \to \mathrm{Dom}(M) \qquad \text{and} \qquad \eta_M : \binom{V}{2} \to \mathrm{Dom}(M)$$

are functions, the pairs

$$(V, \sigma_M) \qquad \text{and} \qquad \left(\binom{V}{2}, \eta_M\right)$$

are well-defined Uncertain Sets of type $M$.

For each $v \in V$, the statement

$$\sigma_M(v) \neq 0_M$$

has a definite truth value, and hence

$$V_M^* := \{v \in V : \sigma_M(v) \neq 0_M\}$$

is well-defined. Likewise, for each

$$\{u, v\} \in \binom{V_M^*}{2},$$

the statement

$$\eta_M(\{u, v\}) \neq 0_M$$

has a definite truth value, and hence

$$E_M^* = \left\{\{u, v\} \in \binom{V_M^*}{2} : \eta_M(\{u, v\}) \neq 0_M\right\}$$

is well-defined. Therefore the crisp graph

$$G_M^* = (V_M^*, E_M^*)$$

is well-defined. This proves (1).

Now fix a support edge

$$e = \{u, v\} \in E_M^*.$$

Because $G_M^*$ is a finite graph, the family of simple $u$-$v$ paths in $G_M^*$ is finite. Moreover, since $\{u, v\} \in E_M^*$, the length-one path

$$(u, v)$$

is a simple $u$-$v$ path. Hence

$$\mathcal{P}_{u,v}(G_M)$$

is a well-defined finite nonempty set. This proves (2).

Let

$$P = (v_0, v_1, \ldots, v_m) \in \mathcal{P}_{u,v}(G_M).$$

For each $i = 1, \ldots, m$,

$$\{v_{i-1}, v_i\} \in E_M^*,$$

so

$$\eta_M(\{v_{i-1}, v_i\}) \in \mathrm{Dom}(M).$$

Therefore the $m$-tuple

$$\Big(\eta_M(\{v_0, v_1\}), \eta_M(\{v_1, v_2\}), \ldots, \eta_M(\{v_{m-1}, v_m\})\Big)$$

belongs to $\mathrm{Dom}(M)^m$, and hence

$$\mathrm{Str}_M(P) = \Psi_M^{(m)}\Big(\eta_M(\{v_0, v_1\}), \ldots, \eta_M(\{v_{m-1}, v_m\})\Big)$$

is a well-defined element of $\mathrm{Dom}(M)$.

Since $\mathcal{P}_{u,v}(G_M)$ is finite and nonempty, the set

$$\big\{\mathrm{Str}_M(P) : P \in \mathcal{P}_{u,v}(G_M)\big\}$$

is a well-defined finite nonempty subset of $\mathrm{Dom}(M)$. Because $\preceq_M$ is a total order on $\mathrm{Dom}(M)$, this set has a unique maximum. Therefore

$$\eta_{G_M}^{\infty}(u, v) = \max\nolimits_{\preceq_M}\big\{\mathrm{Str}_M(P) : P \in \mathcal{P}_{u,v}(G_M)\big\}$$

is well-defined. This proves (3).

Now let $\{u, v\} \in E_M^*$. Both

$$\eta_M(\{u, v\}) \in \mathrm{Dom}(M) \qquad \text{and} \qquad \eta_{G_M}^{\infty}(u, v) \in \mathrm{Dom}(M)$$

are well-defined. Since $\preceq_M$ is a total order, the comparison

$$\eta_M(\{u, v\}) \succeq_M \eta_{G_M}^{\infty}(u, v)$$

has a definite truth value. Hence the statement

$$\text{“}\{u, v\} \text{ is a strong support edge”}$$

is well-defined. This proves (4).

Let $S \subseteq V$. For any two distinct vertices $u, v \in S$, the statement

$$\{u, v\} \notin E_M^* \quad \text{or} \quad \{u, v\} \text{ is not a strong support edge}$$

has a definite truth value by (1) and (4). Since $S$ is finite, the universal condition over all distinct pairs $u, v \in S$ is meaningful. Therefore the statement

$$\text{“}S \text{ is an uncertain independent set”}$$

is well-defined.

Also, for each $v \in S$, since $\sigma_M(v) \in \mathrm{Dom}(M)$ and

$$\omega_M : \mathrm{Dom}(M) \to [0, \infty),$$

the number

$$\omega_M(\sigma_M(v))$$

is well-defined. Because $S$ is finite,

$$|S|_M = \sum_{v \in S} \omega_M(\sigma_M(v))$$

is a finite sum of well-defined nonnegative real numbers. Hence $|S|_M$ is well-defined. This proves (5).

Finally, since $V$ is finite, its power set

$$\mathcal{P}(V)$$

is finite. Let

$$\mathcal{I}(G_M) := \{\, S \subseteq V : S \text{ is an uncertain independent set of } G_M \,\}.$$

By (5), this is a well-defined subset of the finite set $\mathcal{P}(V)$, hence it is finite. Moreover, $\varnothing \in \mathcal{I}(G_M)$, so $\mathcal{I}(G_M) \neq \varnothing$.

Therefore the set

$$\big\{ |S|_M : S \in \mathcal{I}(G_M) \big\}$$

is a finite nonempty subset of $[0, \infty)$. Hence it has a maximum. Consequently,

$$\beta_M(G_M) = \max\big\{ |S|_M : S \subseteq V \text{ is an uncertain independent set of } G_M \big\}$$

is well-defined.

Since this maximum is attained by at least one set $S \in \mathcal{I}(G_M)$, there exists at least one maximum uncertain independent set. This proves (6). □

## 5.7 Connectivity in Uncertain graphs

Connectivity in a fuzzy graph measures the maximum path strength between vertices and indicates whether every vertex pair remains linked through edges with positive membership (cf. [676–678]).

**Definition 5.7.1** (Connectivity in a Fuzzy Graph)**.** Let

$$G = (V, \sigma, \mu)$$

be a finite fuzzy graph, where

$$\sigma : V \to [0,1], \qquad \mu : V \times V \to [0,1], \qquad \mu(u,v) \le \min\{\sigma(u), \sigma(v)\} \quad (\forall\, u, v \in V),$$

and assume that $\mu$ is symmetric.

Define the support vertex set by

$$V^* := \{v \in V : \sigma(v) > 0\}.$$

A *path* from $x$ to $y$ in $G$ is a sequence of distinct vertices

$$P : x = u_0, u_1, \ldots, u_n = y$$

such that

$$\mu(u_{i-1}, u_i) > 0 \qquad (i = 1, 2, \ldots, n).$$

The *strength* of the path $P$ is defined by

$$s_G(P) := \min_{1 \le i \le n} \mu(u_{i-1}, u_i).$$

For two distinct vertices $x, y \in V^*$, the *strength of connectedness* between $x$ and $y$ is defined by

$$\mathrm{CONN}_G(x,y) := \max\big\{\, s_G(P) : P \text{ is a path from } x \text{ to } y \,\big\}.$$

A path $P$ from $x$ to $y$ is called a *strongest $x$–$y$ path* if

$$s_G(P) = \mathrm{CONN}_G(x,y).$$

The fuzzy graph $G$ is said to be *connected* if

$$\mathrm{CONN}_G(x,y) > 0 \qquad \text{for every distinct } x, y \in V^*.$$

**Definition 5.7.2** (Connectivity-Evaluable Uncertain Model)**.** Let $M$ be an uncertain model with degree-domain

$$\mathrm{Dom}(M) \subseteq [0,1]^k.$$

We say that $M$ is *connectivity-evaluable* if it is equipped with:

1. a distinguished element

   $$0_M \in \mathrm{Dom}(M),$$

   called the *zero degree*;

2. a total order
$$\preceq_M \subseteq \operatorname{Dom}(M) \times \operatorname{Dom}(M),$$
called the *strength order*, such that $0_M$ is its least element;

3. for each integer $n \geq 1$, a map
$$\Psi_M^{(n)} : \operatorname{Dom}(M)^n \to \operatorname{Dom}(M),$$
called the *path-strength operator of length n*;

4. the following positivity condition: for every $n \geq 1$ and every $d_1, \dots, d_n \in \operatorname{Dom}(M) \setminus \{0_M\}$,
$$\Psi_M^{(n)}(d_1, \dots, d_n) \neq 0_M.$$

**Definition 5.7.3** (Uncertain Graph of Type $M$)**.** Let $V$ be a finite nonempty set, and let $M$ be a connectivity-evaluable uncertain model. An *uncertain graph* of type $M$ on $V$ is a triple
$$G_M = (V, \sigma_M, \eta_M),$$
where
$$\sigma_M : V \to \operatorname{Dom}(M), \qquad \eta_M : \binom{V}{2} \to \operatorname{Dom}(M)$$
are functions.

Equivalently,
$$(V, \sigma_M)$$
is an Uncertain Set of type $M$ on the vertex set $V$, and
$$\left(\binom{V}{2}, \eta_M\right)$$
is an Uncertain Set of type $M$ on the set of unordered pairs of distinct vertices.

**Definition 5.7.4** (Connectivity in an Uncertain Graph)**.** Let
$$G_M = (V, \sigma_M, \eta_M)$$
be a finite uncertain graph of type $M$, where $M$ is connectivity-evaluable.

Define the *support vertex set* by
$$V_M^* := \{v \in V : \sigma_M(v) \neq 0_M\}.$$

Define the *support edge set* by
$$E_M^* := \left\{ \{u, v\} \in \binom{V_M^*}{2} : \eta_M(\{u, v\}) \neq 0_M \right\}.$$

A *path* from $x$ to $y$ in $G_M$ is a sequence of distinct vertices
$$P : x = u_0, u_1, \dots, u_n = y$$
such that
$$\{u_{i-1}, u_i\} \in E_M^* \qquad (i = 1, 2, \dots, n).$$

The *strength* of the path $P$ is defined by
$$s_{G_M}(P) := \Psi_M^{(n)}\Big(\eta_M(\{u_0, u_1\}), \eta_M(\{u_1, u_2\}), \dots, \eta_M(\{u_{n-1}, u_n\})\Big).$$

For two distinct vertices $x, y \in V_M^*$, let

$$\mathcal{P}_{x,y}(G_M)$$

denote the set of all paths from $x$ to $y$ in $G_M$. Whenever $\mathcal{P}_{x,y}(G_M) \neq \varnothing$, define the *strength of connectedness* between $x$ and $y$ by

$$\mathrm{CONN}_{G_M}(x, y) := \max_{\preceq_M}\{s_{G_M}(P) : P \in \mathcal{P}_{x,y}(G_M)\}.$$

A path $P$ from $x$ to $y$ is called a *strongest $x$–$y$ path* if

$$s_{G_M}(P) = \mathrm{CONN}_{G_M}(x, y).$$

The uncertain graph $G_M$ is said to be *connected* if

$$\mathrm{CONN}_{G_M}(x, y) \neq 0_M \qquad \text{for every distinct } x, y \in V_M^*.$$

**Theorem 5.7.5** (Well-definedness of Connectivity in an Uncertain Graph)**.** *Let $V$ be a finite nonempty set, let $M$ be a connectivity-evaluable uncertain model, and let*

$$G_M = (V, \sigma_M, \eta_M)$$

*be an uncertain graph of type $M$ on $V$. Then:*

1. *the support vertex set*

$$V_M^* = \{v \in V : \sigma_M(v) \neq 0_M\}$$

*and the support edge set*

$$E_M^* = \left\{\{u, v\} \in \binom{V_M^*}{2} : \eta_M(\{u, v\}) \neq 0_M\right\}$$

*are well-defined;*

2. *for every distinct $x, y \in V_M^*$, the set*

$$\mathcal{P}_{x,y}(G_M)$$

*of paths from $x$ to $y$ is a well-defined finite set;*

3. *for every path*

$$P \in \mathcal{P}_{x,y}(G_M),$$

*the path strength*

$$s_{G_M}(P)$$

*is well-defined;*

4. *whenever*

$$\mathcal{P}_{x,y}(G_M) \neq \varnothing,$$

*the connectedness strength*

$$\mathrm{CONN}_{G_M}(x, y)$$

*is well-defined, and there exists at least one strongest $x$–$y$ path;*

5. *the statement*

*"$G_M$ is connected"*

*is well-defined;*

6. *moreover, $G_M$ is connected if and only if the crisp support graph*

$$G_M^* := (V_M^*, E_M^*)$$

*is connected in the ordinary graph-theoretic sense.*

*Proof.* Since $M$ is an uncertain model, its degree-domain

$$\mathrm{Dom}(M) \subseteq [0,1]^k$$

is fixed. Since $M$ is connectivity-evaluable, the element

$$0_M \in \mathrm{Dom}(M),$$

the total order $\preceq_M$, and the path-strength operators

$$\Psi_M^{(n)} : \mathrm{Dom}(M)^n \to \mathrm{Dom}(M) \qquad (n \geq 1)$$

are all fixed.

Because

$$\sigma_M : V \to \mathrm{Dom}(M) \qquad \text{and} \qquad \eta_M : \binom{V}{2} \to \mathrm{Dom}(M)$$

are functions, the pairs

$$(V, \sigma_M) \qquad \text{and} \qquad \left(\binom{V}{2}, \eta_M\right)$$

are well-defined Uncertain Sets of type $M$.

Hence, for each $v \in V$, the statement

$$\sigma_M(v) \neq 0_M$$

has a definite truth value, and therefore

$$V_M^* := \{v \in V : \sigma_M(v) \neq 0_M\}$$

is a well-defined subset of $V$.

Likewise, for each unordered pair

$$\{u, v\} \in \binom{V_M^*}{2},$$

the statement

$$\eta_M(\{u, v\}) \neq 0_M$$

has a definite truth value. Therefore

$$E_M^* = \left\{ \{u, v\} \in \binom{V_M^*}{2} : \eta_M(\{u, v\}) \neq 0_M \right\}$$

is a well-defined subset of $\binom{V_M^*}{2}$. This proves (1).

Now fix distinct vertices $x, y \in V_M^*$. A path from $x$ to $y$ is, by definition, a finite sequence of distinct vertices

$$x = u_0, u_1, \ldots, u_n = y$$

such that each consecutive unordered pair belongs to $E_M^*$. Because $V_M^*$ is finite, there are only finitely many sequences of distinct vertices in $V_M^*$. Hence the collection

$$\mathcal{P}_{x,y}(G_M)$$

of all such paths is a well-defined finite set. This proves (2).

Let

$$P : x = u_0, u_1, \ldots, u_n = y$$

be a path in $G_M$. For each $i = 1, \ldots, n$, we have

$$\{u_{i-1}, u_i\} \in E_M^*,$$

so

$$\eta_M(\{u_{i-1}, u_i\}) \in \mathrm{Dom}(M).$$

Therefore the $n$-tuple

$$\Big(\eta_M(\{u_0, u_1\}), \eta_M(\{u_1, u_2\}), \ldots, \eta_M(\{u_{n-1}, u_n\})\Big) \in \mathrm{Dom}(M)^n$$

is well-defined. Since

$$\Psi_M^{(n)} : \mathrm{Dom}(M)^n \to \mathrm{Dom}(M),$$

the quantity

$$s_{G_M}(P) = \Psi_M^{(n)}\Big(\eta_M(\{u_0, u_1\}), \eta_M(\{u_1, u_2\}), \ldots, \eta_M(\{u_{n-1}, u_n\})\Big)$$

is a well-defined element of $\mathrm{Dom}(M)$. This proves (3).

Assume now that

$$\mathcal{P}_{x,y}(G_M) \neq \varnothing.$$

Then

$$\{\, s_{G_M}(P) : P \in \mathcal{P}_{x,y}(G_M) \,\}$$

is a finite nonempty subset of $\mathrm{Dom}(M)$. Because $\preceq_M$ is a total order on $\mathrm{Dom}(M)$, every finite nonempty subset of $\mathrm{Dom}(M)$ has a unique maximum with respect to $\preceq_M$. Hence

$$\mathrm{CONN}_{G_M}(x, y) = \max\nolimits_{\preceq_M} \big\{ s_{G_M}(P) : P \in \mathcal{P}_{x,y}(G_M) \big\}$$

is well-defined.

Since the maximum is attained by some element of a finite nonempty set, there exists at least one path $P_0 \in \mathcal{P}_{x,y}(G_M)$ such that

$$s_{G_M}(P_0) = \mathrm{CONN}_{G_M}(x, y).$$

Thus a strongest $x$–$y$ path exists. This proves (4).

We now prove (5). By definition, $G_M$ is connected if

$$\mathrm{CONN}_{G_M}(x, y) \neq 0_M \qquad \text{for every distinct } x, y \in V_M^*.$$

For a given pair $x, y \in V_M^*$, if there is no path from $x$ to $y$, then

$$\mathcal{P}_{x,y}(G_M) = \varnothing,$$

so $\mathrm{CONN}_{G_M}(x, y)$ is not defined. Thus the defining statement for connectedness is true exactly when every distinct pair $x, y \in V_M^*$ has at least one path, and for each such pair the corresponding $\mathrm{CONN}_{G_M}(x, y)$ is a well-defined element of $\mathrm{Dom}(M) \setminus \{0_M\}$. Hence the statement

"$G_M$ is connected"

has a definite truth value. This proves (5).

Finally, we prove (6).

Assume first that $G_M$ is connected in the above sense. Take distinct vertices $x, y \in V_M^*$. Then $\mathrm{CONN}_{G_M}(x, y) \neq 0_M$, so in particular $\mathrm{CONN}_{G_M}(x, y)$ is defined. Therefore

$$\mathcal{P}_{x,y}(G_M) \neq \varnothing,$$

which means that there exists a path from $x$ to $y$ using edges from $E_M^*$. Hence the crisp graph

$$G_M^* = (V_M^*, E_M^*)$$

is connected.

Conversely, assume that the support graph $G_M^*$ is connected. Let $x, y \in V_M^*$ be distinct. Then there exists a path

$$P : x = u_0, u_1, \dots, u_n = y$$

in $G_M^*$. By definition of $E_M^*$,

$$\eta_M(\{u_{i-1}, u_i\}) \neq 0_M \qquad (i = 1, \dots, n).$$

By the positivity condition in the definition of a connectivity-evaluable uncertain model,

$$s_{G_M}(P) = \Psi_M^{(n)}\Big(\eta_M(\{u_0, u_1\}), \dots, \eta_M(\{u_{n-1}, u_n\})\Big) \neq 0_M.$$

Since

$$\mathrm{CONN}_{G_M}(x, y) = \max_{\preceq_M}\{s_{G_M}(Q) : Q \in \mathcal{P}_{x,y}(G_M)\},$$

we have

$$s_{G_M}(P) \preceq_M \mathrm{CONN}_{G_M}(x, y).$$

Because $0_M$ is the least element of $(\mathrm{Dom}(M), \preceq_M)$ and

$$s_{G_M}(P) \neq 0_M,$$

it follows that

$$\mathrm{CONN}_{G_M}(x, y) \neq 0_M.$$

Since $x, y \in V_M^*$ were arbitrary distinct vertices, $G_M$ is connected.

Therefore

$$G_M \text{ is connected} \quad \Longleftrightarrow \quad G_M^* \text{ is connected.}$$

This proves (6). □

## 5.8 Chromatic number in Uncertain graphs

Chromatic number of a fuzzy graph is the minimum number of colors needed so strongly adjacent vertices receive different colors under fuzzy adjacency constraints (cf. [679–682]).

**Definition 5.8.1** (Chromatic Number of a Fuzzy Graph)**.** [679, 680] Let

$$G = (V, \sigma, \mu)$$

be a finite fuzzy graph, where

$$\sigma : V \to [0, 1], \qquad \mu : V \times V \to [0, 1], \qquad \mu(u, v) \leq \min\{\sigma(u), \sigma(v)\} \quad (\forall\, u, v \in V),$$

and assume that $\mu$ is symmetric and $G$ has no loops.

Define the level sets

$$L_\sigma := \{\sigma(u) : u \in V,\ \sigma(u) > 0\}, \qquad L_\mu := \{\mu(u, v) : u, v \in V,\ \mu(u, v) > 0\},$$

and let

$$L := L_\sigma \cup L_\mu.$$

For each $\alpha \in L$, the *$\alpha$-cut graph* (or *threshold graph*) of $G$ is the crisp graph

$$G_\alpha = (V_\alpha, E_\alpha),$$

where

$$V_\alpha := \{u \in V : \sigma(u) \geq \alpha\}, \qquad E_\alpha := \big\{\{u, v\} \subseteq V_\alpha : u \neq v,\ \mu(u, v) \geq \alpha\big\}.$$

Let

$$\chi(G_\alpha)$$

denote the ordinary chromatic number of the crisp graph $G_\alpha$.

Then the *chromatic number* of the fuzzy graph $G$ is defined by

$$\chi(G) := \max_{\alpha \in L} \chi(G_\alpha).$$

**Definition 5.8.2** (Chromatic-Evaluable Uncertain Model)**.** Let $M$ be an uncertain model with degree-domain

$$\mathrm{Dom}(M) \subseteq [0,1]^k.$$

We say that $M$ is *chromatic-evaluable* if it is equipped with:

1. a distinguished element
$$0_M \in \mathrm{Dom}(M),$$
called the *zero degree*;
2. a total order
$$\preceq_M \subseteq \mathrm{Dom}(M) \times \mathrm{Dom}(M),$$
called the *threshold order*, such that $0_M$ is its least element.

**Definition 5.8.3** (Uncertain Graph of Type $M$)**.** Let $V$ be a finite set, and let $M$ be a chromatic-evaluable uncertain model. An *uncertain graph* of type $M$ on $V$ is a triple

$$G_M = (V, \sigma_M, \eta_M),$$

where

$$\sigma_M : V \to \mathrm{Dom}(M), \qquad \eta_M : \binom{V}{2} \to \mathrm{Dom}(M)$$

are functions.

Equivalently,

$$(V, \sigma_M)$$

is an Uncertain Set of type $M$ on the vertex set $V$, and

$$\left(\binom{V}{2}, \eta_M\right)$$

is an Uncertain Set of type $M$ on the set of unordered pairs of distinct vertices.

**Definition 5.8.4** (Chromatic Number in an Uncertain Graph)**.** Let

$$G_M = (V, \sigma_M, \eta_M)$$

be a finite uncertain graph of type $M$, where $M$ is chromatic-evaluable.

Define the realized positive vertex-level set by

$$L_\sigma^M(G_M) := \{\sigma_M(u) : u \in V,\ \sigma_M(u) \neq 0_M\},$$

and the realized positive edge-level set by

$$L_\eta^M(G_M) := \{\eta_M(e) : e \in \binom{V}{2},\ \eta_M(e) \neq 0_M\}.$$

Set

$$L_M(G_M) := L_\sigma^M(G_M) \cup L_\eta^M(G_M).$$

For each

$$\lambda \in L_M(G_M),$$

the $\lambda$*-cut graph* (or *threshold graph*) of $G_M$ is the crisp graph

$$G_M^\lambda = (V_\lambda, E_\lambda),$$

where

$$V_\lambda := \{u \in V : \lambda \preceq_M \sigma_M(u)\},$$

and

$$E_\lambda := \big\{\{u,v\} \in \binom{V_\lambda}{2} : \lambda \preceq_M \eta_M(\{u,v\})\big\}.$$

Let

$$\chi(G_M^\lambda)$$

denote the ordinary chromatic number of the crisp graph $G_M^\lambda$.

The *chromatic number* of the uncertain graph $G_M$ is defined by

$$\chi_M(G_M) := \begin{cases} 0, & \text{if } L_M(G_M) = \varnothing, \\ \max_{\lambda \in L_M(G_M)} \chi(G_M^\lambda), & \text{if } L_M(G_M) \neq \varnothing. \end{cases}$$

**Theorem 5.8.5** (Well-definedness of Chromatic Number in an Uncertain Graph)**.** *Let $V$ be a finite set, let $M$ be a chromatic-evaluable uncertain model, and let*

$$G_M = (V, \sigma_M, \eta_M)$$

*be an uncertain graph of type $M$ on $V$. Then:*

1. *the sets*
$$L_\sigma^M(G_M), \qquad L_\eta^M(G_M), \qquad L_M(G_M)$$
*are well-defined finite subsets of* $\mathrm{Dom}(M)$*;*
2. *for every*
$$\lambda \in L_M(G_M),$$
*the $\lambda$-cut graph*
$$G_M^\lambda = (V_\lambda, E_\lambda)$$
*is a well-defined finite crisp graph;*
3. *for every*
$$\lambda \in L_M(G_M),$$
*the ordinary chromatic number*
$$\chi(G_M^\lambda)$$
*is well-defined;*
4. *the number*
$$\chi_M(G_M)$$
*is a well-defined nonnegative integer.*

*Proof.* Since $M$ is an uncertain model, its degree-domain

$$\mathrm{Dom}(M) \subseteq [0,1]^k$$

is fixed. Since $M$ is chromatic-evaluable, the element

$$0_M \in \mathrm{Dom}(M)$$

and the total order

$$\preceq_M$$

on $\mathrm{Dom}(M)$ are fixed as part of the structure.

Because

$$\sigma_M : V \to \mathrm{Dom}(M)$$

is a function, for every $u \in V$ the value $\sigma_M(u)$ is a well-defined element of $\mathrm{Dom}(M)$, and hence the statement

$$\sigma_M(u) \neq 0_M$$

has a definite truth value. Therefore

$$L_\sigma^M(G_M) = \{\sigma_M(u) : u \in V,\ \sigma_M(u) \neq 0_M\}$$

is a well-defined subset of $\mathrm{Dom}(M)$.

Likewise, because

$$\eta_M : \binom{V}{2} \to \mathrm{Dom}(M)$$

is a function, for every edge candidate

$$e \in \binom{V}{2}$$

the value $\eta_M(e)$ is a well-defined element of $\mathrm{Dom}(M)$, and hence the statement

$$\eta_M(e) \neq 0_M$$

has a definite truth value. Therefore

$$L_\eta^M(G_M) = \{\eta_M(e) : e \in \binom{V}{2},\ \eta_M(e) \neq 0_M\}$$

is a well-defined subset of $\mathrm{Dom}(M)$.

Since $V$ is finite, both $V$ and $\binom{V}{2}$ are finite sets. Hence the images

$$\{\sigma_M(u) : u \in V\} \qquad \text{and} \qquad \{\eta_M(e) : e \in \binom{V}{2}\}$$

are finite. It follows that

$$L_\sigma^M(G_M), \qquad L_\eta^M(G_M)$$

are finite, and so

$$L_M(G_M) = L_\sigma^M(G_M) \cup L_\eta^M(G_M)$$

is also a well-defined finite subset of $\mathrm{Dom}(M)$. This proves (1).

Now fix

$$\lambda \in L_M(G_M).$$

Since $\preceq_M$ is a total order on $\mathrm{Dom}(M)$, for each $u \in V$ the comparison

$$\lambda \preceq_M \sigma_M(u)$$

has a definite truth value. Hence

$$V_\lambda = \{u \in V : \lambda \preceq_M \sigma_M(u)\}$$

is a well-defined subset of $V$.

Similarly, for each unordered pair

$$\{u, v\} \in \binom{V_\lambda}{2},$$

the comparison

$$\lambda \preceq_M \eta_M(\{u, v\})$$

has a definite truth value. Hence

$$E_\lambda = \{\{u, v\} \in \binom{V_\lambda}{2} : \lambda \preceq_M \eta_M(\{u, v\})\}$$

is a well-defined subset of $\binom{V_\lambda}{2}$.

Therefore
$$G_M^\lambda = (V_\lambda, E_\lambda)$$
is a well-defined crisp graph. Because $V$ is finite, the set $V_\lambda \subseteq V$ is finite, and thus $G_M^\lambda$ is a finite crisp graph. This proves (2).

For every
$$\lambda \in L_M(G_M),$$
the graph $G_M^\lambda$ is a finite crisp graph by (2). The ordinary chromatic number of a finite crisp graph is a well-defined positive integer if the vertex set is nonempty, and equals 0 when the vertex set is empty. Hence
$$\chi(G_M^\lambda)$$
is well-defined for every
$$\lambda \in L_M(G_M).$$
This proves (3).

Finally, consider the definition of
$$\chi_M(G_M).$$

If
$$L_M(G_M) = \varnothing,$$
then by definition
$$\chi_M(G_M) = 0,$$
so $\chi_M(G_M)$ is well-defined.

Assume now that
$$L_M(G_M) \neq \varnothing.$$
Since $L_M(G_M)$ is finite by (1), the set
$$\{\chi(G_M^\lambda) : \lambda \in L_M(G_M)\}$$
is a finite nonempty set of nonnegative integers. Every finite nonempty set of integers has a maximum. Therefore
$$\max_{\lambda \in L_M(G_M)} \chi(G_M^\lambda)$$
is well-defined, and it is a nonnegative integer.

Hence, in all cases,
$$\chi_M(G_M)$$
is a well-defined nonnegative integer. This proves (4). □

Representative chromatic-number-related concepts under uncertainty-aware graph frameworks are listed in Table 5.3.

Besides uncertain chromatic number, several related concepts are also known, including edge chromatic number [686], total chromatic number [687, 688], list chromatic number [689], equitable chromatic number [690, 691], circular chromatic number [692, 693], fractional chromatic number [694], and star chromatic number [695, 696].

Table 5.3: Representative chromatic-number-related concepts under uncertainty-aware graph frameworks, classified by the dimension $k$ of the information attached to vertices and/or edges.

| $k$ | **Chromatic-number-related concept** | **Typical coordinate form** | **Canonical information attached to vertices/edges** |
|---|---|---|---|
| 1 | Chromatic Number of a Fuzzy Graph | $\mu$ | The chromatic number is studied in a fuzzy graph, where each vertex and edge is associated with a single membership degree in $[0, 1]$. |
| 2 | Chromatic Number of an Intuitionistic Fuzzy Graph [683] | $(\mu, \nu)$ | The chromatic number is defined on an intuitionistic fuzzy graph, where each vertex and edge carries a membership degree and a non-membership degree, usually satisfying $\mu + \nu \leq 1$. |
| 3 | Chromatic Number of a Neutrosophic Graph [684, 685] | $(T, I, F)$ | The chromatic number is defined on a neutrosophic graph, where each vertex and edge is described by truth, indeterminacy, and falsity degrees. |

## 5.9 Matching number in Uncertain graphs

Matching number in a fuzzy graph is the maximum total membership of edges forming a fuzzy matching, subject to each vertex respecting its membership capacity [697, 698].

**Definition 5.9.1** (Matching Number in a Fuzzy Graph)**.** Let

$$G = (V, \sigma, \mu)$$

be a finite fuzzy graph, where

$$\sigma : V \to [0, 1], \qquad \mu : V \times V \to [0, 1], \qquad \mu(u, v) \leq \min\{\sigma(u), \sigma(v)\} \quad (\forall\, u, v \in V),$$

and $\mu$ is symmetric.

Define the support edge set by

$$E^*(G) := \big\{\{u, v\} \subseteq V : u \neq v,\ \mu(u, v) > 0\big\}.$$

A subset

$$M \subseteq E^*(G)$$

is called a *fuzzy matching* of $G$ if for every vertex $u \in V$,

$$\sum_{\substack{v \in V \\ \{u,v\} \in M}} \mu(u, v) \leq \sigma(u).$$

The *weight* of a fuzzy matching $M$ is defined by

$$w(M) := \sum_{\{u,v\} \in M} \mu(u, v).$$

The *matching number* of the fuzzy graph $G$ is defined by

$$\nu_f(G) := \max\big\{\, w(M) : M \subseteq E^*(G) \text{ is a fuzzy matching of } G \,\big\}.$$

Any fuzzy matching $M$ satisfying

$$w(M) = \nu_f(G)$$

is called a *maximum fuzzy matching*.

Matching number in an uncertain graph is the maximum total edge-weight of an uncertain matching, where the incident edge-weights at each vertex do not exceed the capacity induced by the corresponding uncertain vertex degree.

**Definition 5.9.2** (Matching-Evaluable Uncertain Model)**.** Let $M$ be an uncertain model with degree-domain

$$\mathrm{Dom}(M) \subseteq [0,1]^k.$$

We say that $M$ is *matching-evaluable* if it is equipped with:

1. a distinguished element
$$0_M \in \mathrm{Dom}(M),$$
called the *zero degree*;
2. an *edge-weight evaluation map*
$$\delta_M : \mathrm{Dom}(M) \to [0,\infty);$$
3. a *vertex-capacity evaluation map*
$$\omega_M : \mathrm{Dom}(M) \to [0,\infty);$$

such that

$$\delta_M(0_M) = 0, \qquad \omega_M(0_M) = 0.$$

**Definition 5.9.3** (Uncertain Graph of Type $M$)**.** Let $V$ be a finite set, and let $M$ be a matching-evaluable uncertain model. An *uncertain graph* of type $M$ on $V$ is a triple

$$G_M = (V, \sigma_M, \eta_M),$$

where

$$\sigma_M : V \to \mathrm{Dom}(M), \qquad \eta_M : \binom{V}{2} \to \mathrm{Dom}(M)$$

are functions.

Equivalently,

$$(V, \sigma_M)$$

is an Uncertain Set of type $M$ on the vertex set $V$, and

$$\left(\binom{V}{2}, \eta_M\right)$$

is an Uncertain Set of type $M$ on the set of unordered pairs of distinct vertices.

**Definition 5.9.4** (Matching Number in an Uncertain Graph)**.** Let

$$G_M = (V, \sigma_M, \eta_M)$$

be a finite uncertain graph of type $M$, where $M$ is matching-evaluable.

Define the support edge set of $G_M$ by

$$E^*_M(G_M) := \left\{ e \in \binom{V}{2} : \eta_M(e) \neq 0_M \right\}.$$

A subset

$$\mathcal{M} \subseteq E^*_M(G_M)$$

is called an *uncertain matching* of $G_M$ if for every vertex $u \in V$,

$$\sum_{\substack{v \in V \\ \{u,v\} \in \mathcal{M}}} \delta_M\big(\eta_M(\{u,v\})\big) \leq \omega_M\big(\sigma_M(u)\big).$$

The *weight* of an uncertain matching $\mathcal{M}$ is defined by

$$w_M(\mathcal{M}) := \sum_{e \in \mathcal{M}} \delta_M\big(\eta_M(e)\big).$$

The *matching number* of the uncertain graph $G_M$ is defined by

$$\nu_M(G_M) := \max\big\{w_M(\mathcal{M}) : \mathcal{M} \subseteq E^*_M(G_M) \text{ is an uncertain matching of } G_M\big\}.$$

Any uncertain matching $\mathcal{M}$ satisfying

$$w_M(\mathcal{M}) = \nu_M(G_M)$$

is called a *maximum uncertain matching.*

**Theorem 5.9.5** (Well-definedness of Matching Number in an Uncertain Graph)**.** *Let $V$ be a finite set, let $M$ be a matching-evaluable uncertain model, and let*

$$G_M = (V, \sigma_M, \eta_M)$$

*be an uncertain graph of type $M$ on $V$. Then:*

1. *the support edge set*

$$E^*_M(G_M) = \left\{ e \in \binom{V}{2} : \eta_M(e) \neq 0_M \right\}$$

*is a well-defined finite set;*

2. *for every subset*

$$\mathcal{M} \subseteq E^*_M(G_M),$$

*the statement*

*"$\mathcal{M}$ is an uncertain matching of $G_M$"*

*is well-defined;*

3. *for every uncertain matching $\mathcal{M}$, the weight*

$$w_M(\mathcal{M})$$

*is a well-defined element of $[0, \infty)$;*

4. *the class of all uncertain matchings of $G_M$ is nonempty;*

5. *the matching number*

$$\nu_M(G_M)$$

*is a well-defined element of $[0, \infty)$, and there exists at least one maximum uncertain matching.*

*Proof.* Since $M$ is an uncertain model, its degree-domain

$$\mathrm{Dom}(M) \subseteq [0, 1]^k$$

is fixed. Since $M$ is matching-evaluable, the element

$$0_M \in \mathrm{Dom}(M)$$

and the maps

$$\delta_M : \mathrm{Dom}(M) \to [0, \infty), \qquad \omega_M : \mathrm{Dom}(M) \to [0, \infty)$$

are fixed as part of the structure.

Because

$$\sigma_M : V \to \mathrm{Dom}(M) \qquad \text{and} \qquad \eta_M : \binom{V}{2} \to \mathrm{Dom}(M)$$

are functions, the pairs

$$(V, \sigma_M) \qquad \text{and} \qquad \left(\binom{V}{2}, \eta_M\right)$$

are well-defined Uncertain Sets of type $M$.

For each

$$e \in \binom{V}{2},$$

the value

$$\eta_M(e) \in \mathrm{Dom}(M)$$

is well-defined, and therefore the statement

$$\eta_M(e) \neq 0_M$$

has a definite truth value. Hence

$$E_M^*(G_M) = \left\{ e \in \binom{V}{2} : \eta_M(e) \neq 0_M \right\}$$

is a well-defined subset of $\binom{V}{2}$. Since $V$ is finite, the set $\binom{V}{2}$ is finite, and therefore $E_M^*(G_M)$ is finite. This proves (1).

Now let

$$\mathcal{M} \subseteq E_M^*(G_M).$$

For each vertex $u \in V$ and each $v \in V$ such that

$$\{u, v\} \in \mathcal{M},$$

we have

$$\eta_M(\{u, v\}) \in \mathrm{Dom}(M).$$

Hence

$$\delta_M\big(\eta_M(\{u, v\})\big) \in [0, \infty)$$

is well-defined. Also,

$$\sigma_M(u) \in \mathrm{Dom}(M),$$

so

$$\omega_M\big(\sigma_M(u)\big) \in [0, \infty)$$

is well-defined.

Because $\mathcal{M}$ is finite, for each fixed $u \in V$ the sum

$$\sum_{\substack{v \in V \\ \{u,v\} \in \mathcal{M}}} \delta_M\big(\eta_M(\{u, v\})\big)$$

is a finite sum of well-defined nonnegative real numbers. Therefore the inequality

$$\sum_{\substack{v \in V \\ \{u,v\} \in \mathcal{M}}} \delta_M\big(\eta_M(\{u, v\})\big) \leq \omega_M\big(\sigma_M(u)\big)$$

has a definite truth value for every $u \in V$. Hence the universal statement

"$\mathcal{M}$ is an uncertain matching of $G_M$"

is well-defined. This proves (2).

Assume that $\mathcal{M} \subseteq E_M^*(G_M)$ is an uncertain matching. For each edge $e \in \mathcal{M}$, the value

$$\delta_M(\eta_M(e))$$

is a well-defined element of $[0, \infty)$. Since $\mathcal{M}$ is finite, the sum

$$w_M(\mathcal{M}) = \sum_{e \in \mathcal{M}} \delta_M(\eta_M(e))$$

is a finite sum of well-defined nonnegative real numbers. Hence

$$w_M(\mathcal{M}) \in [0, \infty)$$

is well-defined. This proves (3).

We next prove (4). Consider the empty set

$$\varnothing \subseteq E_M^*(G_M).$$

For every vertex $u \in V$,

$$\sum_{\substack{v \in V \\ \{u,v\} \in \varnothing}} \delta_M(\eta_M(\{u, v\})) = 0.$$

Since

$$\omega_M(\sigma_M(u)) \in [0, \infty),$$

we have

$$0 \leq \omega_M(\sigma_M(u)).$$

Therefore $\varnothing$ satisfies the defining inequality at every vertex $u \in V$, so $\varnothing$ is an uncertain matching of $G_M$. Hence the class of all uncertain matchings is nonempty. This proves (4).

Finally, let

$$\mathfrak{M}(G_M) := \{\mathcal{M} \subseteq E_M^*(G_M) : \mathcal{M} \text{ is an uncertain matching of } G_M\}.$$

By (4), the set $\mathfrak{M}(G_M)$ is nonempty. Since $E_M^*(G_M)$ is finite, its power set is finite, and therefore $\mathfrak{M}(G_M)$ is a finite nonempty set.

By (3), for every

$$\mathcal{M} \in \mathfrak{M}(G_M),$$

the weight

$$w_M(\mathcal{M}) \in [0, \infty)$$

is well-defined. Hence the set

$$\{w_M(\mathcal{M}) : \mathcal{M} \in \mathfrak{M}(G_M)\}$$

is a finite nonempty subset of $[0, \infty)$. Every finite nonempty subset of $\mathbb{R}$ has a maximum. Therefore

$$\nu_M(G_M) = \max\{w_M(\mathcal{M}) : \mathcal{M} \subseteq E_M^*(G_M) \text{ is an uncertain matching of } G_M\}$$

is well-defined.

Since the maximum of a finite set is attained by some element of that set, there exists

$$\mathcal{M}_{\max} \in \mathfrak{M}(G_M)$$

such that

$$w_M(\mathcal{M}_{\max}) = \nu_M(G_M).$$

Thus a maximum uncertain matching exists. This proves (5). □

Representative matching-number-related concepts under uncertainty-aware graph frameworks are listed in Table 5.4.

Table 5.4: Representative matching-number-related concepts under uncertainty-aware graph frameworks, classified by the dimension $k$ of the information attached to vertices and/or edges.

| $k$ | **Matching-number-related concept** | **Typical coordinate form** | **Canonical information attached to vertices/edges** |
|---|---|---|---|
| 1 | Matching Number in a Fuzzy Graph | $\mu$ | The matching number is studied in a fuzzy graph, where each vertex and edge is associated with a single membership degree in $[0, 1]$. |
| 2 | Matching Number in an Intuitionistic Fuzzy Graph | $(\mu, \nu)$ | The matching number is defined on an intuitionistic fuzzy graph, where each vertex and edge carries a membership degree and a non-membership degree, usually satisfying $\mu + \nu \leq 1$. |
| 3 | Matching Number in a Neutrosophic Graph | $(T, I, F)$ | The matching number is defined on a neutrosophic graph, where each vertex and edge is described by truth, indeterminacy, and falsity degrees. |

## 5.10 Vertex cover number in Uncertain graphs

A vertex cover in a fuzzy graph is a vertex set covering every positive edge, and its vertex cover number is the minimum fuzzy cardinality [699].

**Definition 5.10.1** (Vertex Cover and Vertex Cover Number in a Fuzzy Graph)**.** Let

$$G = (V, \sigma, \mu)$$

be a finite fuzzy graph, where

$$\sigma : V \to [0, 1], \qquad \mu : V \times V \to [0, 1], \qquad \mu(u, v) \leq \min\{\sigma(u), \sigma(v)\} \quad (\forall\, u, v \in V),$$

and $\mu$ is symmetric.

Define the support vertex set and support edge set by

$$V^* := \{v \in V : \sigma(v) > 0\}, \qquad E^* := \big\{\{u, v\} \subseteq V^* : u \neq v,\ \mu(u, v) > 0\big\}.$$

A subset

$$C \subseteq V^*$$

is called a *vertex cover* of $G$ if for every edge

$$\{u, v\} \in E^*,$$

at least one of its end vertices belongs to $C$, that is,

$$\{u, v\} \cap C \neq \varnothing.$$

The *fuzzy cardinality* of $C$ is defined by

$$|C|_f := \sum_{v \in C} \sigma(v).$$

The *vertex cover number* of the fuzzy graph $G$ is defined by

$$\tau_f(G) := \min\big\{|C|_f : C \subseteq V^* \text{ is a vertex cover of } G\big\}.$$

Any vertex cover $C$ satisfying

$$|C|_f = \tau_f(G)$$

is called a *minimum vertex cover* of $G$.

A vertex cover in an uncertain graph is a subset of support vertices that meets every support edge, and its vertex cover number is the minimum uncertain cardinality of such a subset.

**Definition 5.10.2** (Vertex-Cover-Evaluable Uncertain Model)**.** Let $M$ be an uncertain model with degree-domain

$$\mathrm{Dom}(M) \subseteq [0,1]^k.$$

We say that $M$ is *vertex-cover-evaluable* if it is equipped with:

1. a distinguished element
$$0_M \in \mathrm{Dom}(M),$$
called the *zero degree*;
2. a *vertex-weight evaluation map*
$$\omega_M : \mathrm{Dom}(M) \to [0,\infty)$$
such that
$$\omega_M(0_M) = 0.$$

**Definition 5.10.3** (Uncertain Graph of Type $M$)**.** Let $V$ be a finite set, and let $M$ be a vertex-cover-evaluable uncertain model. An *uncertain graph* of type $M$ on $V$ is a triple

$$G_M = (V, \sigma_M, \eta_M),$$

where

$$\sigma_M : V \to \mathrm{Dom}(M), \qquad \eta_M : \binom{V}{2} \to \mathrm{Dom}(M)$$

are functions.

Equivalently,

$$(V, \sigma_M)$$

is an Uncertain Set of type $M$ on the vertex set $V$, and

$$\left(\binom{V}{2}, \eta_M\right)$$

is an Uncertain Set of type $M$ on the set of unordered pairs of distinct vertices.

**Definition 5.10.4** (Vertex Cover and Vertex Cover Number in an Uncertain Graph)**.** Let

$$G_M = (V, \sigma_M, \eta_M)$$

be a finite uncertain graph of type $M$, where $M$ is vertex-cover-evaluable.

Define the support vertex set and the support edge set by

$$V_M^* := \{v \in V : \sigma_M(v) \neq 0_M\},$$

and

$$E_M^* := \{\{u,v\} \subseteq V_M^* : u \neq v,\ \eta_M(\{u,v\}) \neq 0_M\}.$$

A subset

$$C \subseteq V_M^*$$

is called an *uncertain vertex cover* of $G_M$ if for every edge

$$\{u,v\} \in E_M^*,$$

at least one of its end vertices belongs to $C$, that is,

$$\{u,v\} \cap C \neq \varnothing.$$

The *uncertain cardinality* of $C$ is defined by

$$|C|_M := \sum_{v \in C} \omega_M\big(\sigma_M(v)\big).$$

The *vertex cover number* of the uncertain graph $G_M$ is defined by

$$\tau_M(G_M) := \min\big\{|C|_M : C \subseteq V_M^* \text{ is an uncertain vertex cover of } G_M\big\}.$$

Any uncertain vertex cover $C$ satisfying

$$|C|_M = \tau_M(G_M)$$

is called a *minimum uncertain vertex cover* of $G_M$.

**Theorem 5.10.5** (Well-definedness of Vertex Cover Number in an Uncertain Graph)**.** *Let $V$ be a finite set, let $M$ be a vertex-cover-evaluable uncertain model, and let*

$$G_M = (V, \sigma_M, \eta_M)$$

*be an uncertain graph of type $M$ on $V$. Then:*

1. *the support vertex set*
$$V_M^* = \{v \in V : \sigma_M(v) \neq 0_M\}$$
*and the support edge set*
$$E_M^* = \{\{u, v\} \subseteq V_M^* : u \neq v,\ \eta_M(\{u, v\}) \neq 0_M\}$$
*are well-defined finite sets;*
2. *for every subset*
$$C \subseteq V_M^*,$$
*the statement*
$$\text{“}C \text{ is an uncertain vertex cover of } G_M\text{”}$$
*is well-defined;*
3. *for every subset*
$$C \subseteq V_M^*,$$
*the uncertain cardinality*
$$|C|_M$$
*is a well-defined element of $[0, \infty)$;*
4. *the family of all uncertain vertex covers of $G_M$ is nonempty;*
5. *the vertex cover number*
$$\tau_M(G_M)$$
*is a well-defined element of $[0, \infty)$, and there exists at least one minimum uncertain vertex cover of $G_M$.*

*Proof.* Since $M$ is an uncertain model, its degree-domain

$$\mathrm{Dom}(M) \subseteq [0, 1]^k$$

is fixed. Since $M$ is vertex-cover-evaluable, the distinguished element

$$0_M \in \mathrm{Dom}(M)$$

and the map

$$\omega_M : \mathrm{Dom}(M) \to [0, \infty)$$

are fixed as part of the structure.

Because

$$\sigma_M : V \to \mathrm{Dom}(M) \qquad \text{and} \qquad \eta_M : \binom{V}{2} \to \mathrm{Dom}(M)$$

are functions, the pairs

$$(V, \sigma_M) \qquad \text{and} \qquad \left(\binom{V}{2}, \eta_M\right)$$

are well-defined Uncertain Sets of type $M$.

For each $v \in V$, the value $\sigma_M(v)$ is well-defined in $\mathrm{Dom}(M)$, so the statement

$$\sigma_M(v) \neq 0_M$$

has a definite truth value. Hence

$$V_M^* := \{v \in V : \sigma_M(v) \neq 0_M\}$$

is a well-defined subset of $V$.

Likewise, for each unordered pair

$$\{u, v\} \in \binom{V_M^*}{2},$$

the value $\eta_M(\{u, v\})$ is well-defined in $\mathrm{Dom}(M)$, so the statement

$$\eta_M(\{u, v\}) \neq 0_M$$

has a definite truth value. Hence

$$E_M^* := \{\{u, v\} \subseteq V_M^* : u \neq v,\ \eta_M(\{u, v\}) \neq 0_M\}$$

is a well-defined subset of $\binom{V_M^*}{2}$.

Since $V$ is finite, the set $V_M^* \subseteq V$ is finite, and therefore $\binom{V_M^*}{2}$ is finite. Thus $E_M^*$ is finite as well. This proves (1).

Now let

$$C \subseteq V_M^*.$$

For every edge

$$e = \{u, v\} \in E_M^*,$$

the statement

$$e \cap C \neq \varnothing$$

has a definite truth value, because $e$ and $C$ are well-defined sets. Therefore the universal statement

$$\forall \{u, v\} \in E_M^*, \quad \{u, v\} \cap C \neq \varnothing$$

has a definite truth value. Hence the statement

$$\text{“}C \text{ is an uncertain vertex cover of } G_M\text{”}$$

is well-defined. This proves (2).

Let

$$C \subseteq V_M^*.$$

For each $v \in C$, we have

$$\sigma_M(v) \in \mathrm{Dom}(M),$$

and therefore

$$\omega_M(\sigma_M(v)) \in [0, \infty)$$

is well-defined. Since $C$ is finite, the sum

$$|C|_M = \sum_{v \in C} \omega_M(\sigma_M(v))$$

is a finite sum of well-defined nonnegative real numbers. Hence

$$|C|_M \in [0, \infty)$$

is well-defined. This proves (3).

We next show that the family of uncertain vertex covers is nonempty. Consider the subset

$$V_M^* \subseteq V_M^*.$$

For every edge

$$\{u, v\} \in E_M^*,$$

both $u$ and $v$ belong to $V_M^*$, and therefore

$$\{u, v\} \cap V_M^* \neq \varnothing.$$

Hence $V_M^*$ is an uncertain vertex cover of $G_M$. Thus the family of uncertain vertex covers is nonempty. This proves (4).

Finally, let

$$\mathcal{C}(G_M) := \{\, C \subseteq V_M^* : C \text{ is an uncertain vertex cover of } G_M \,\}.$$

By (4), $\mathcal{C}(G_M) \neq \varnothing$. Since $V_M^*$ is finite, its power set $\mathcal{P}(V_M^*)$ is finite, and thus $\mathcal{C}(G_M)$ is a finite nonempty set.

By (3), for every

$$C \in \mathcal{C}(G_M),$$

the quantity $|C|_M$ is a well-defined element of $[0, \infty)$. Therefore

$$\{\, |C|_M : C \in \mathcal{C}(G_M) \,\}$$

is a finite nonempty subset of $[0, \infty)$. Every finite nonempty subset of $\mathbb{R}$ has a minimum. Hence

$$\tau_M(G_M) = \min\big\{ |C|_M : C \subseteq V_M^* \text{ is an uncertain vertex cover of } G_M \big\}$$

is well-defined.

Since this minimum is attained by some element of the finite set $\mathcal{C}(G_M)$, there exists at least one subset

$$C_{\min} \in \mathcal{C}(G_M)$$

such that

$$|C_{\min}|_M = \tau_M(G_M).$$

Therefore a minimum uncertain vertex cover exists. This proves (5). □

## 5.11 Wiener index in Uncertain graphs

Wiener index in a fuzzy graph measures the total fuzzy shortest-path distance between all vertex pairs, quantifying global structural closeness and network compactness under uncertainty [700–702].

**Definition 5.11.1** (Wiener Index in a Fuzzy Graph)**.** Let

$$G = (V, \sigma, \mu)$$

be a finite connected fuzzy graph, where

$$\sigma : V \to [0,1], \qquad \mu : V \times V \to [0,1], \qquad \mu(u,v) \le \min\{\sigma(u), \sigma(v)\} \quad (\forall\, u, v \in V),$$

and $\mu$ is symmetric.

An edge $uv$ is called *strong* if it is a strong arc of $G$. A path

$$P : u_0 u_1 \cdots u_k$$

is called a *strong path* if every edge $u_{i-1}u_i$ $(1 \le i \le k)$ is strong.

For a strong path

$$P : u_0 u_1 \cdots u_k,$$

its *length* is $k$, and its *weight* is defined by

$$w(P) := \sum_{i=1}^{k} \mu(u_{i-1}, u_i).$$

For two distinct vertices $u, v \in V$ with $\sigma(u) > 0$ and $\sigma(v) > 0$, a *geodesic* from $u$ to $v$ is a strong $u$–$v$ path having minimum length among all strong $u$–$v$ paths.

Let

$$d_s(u,v) := \min\{\, w(P) : P \text{ is a geodesic from } u \text{ to } v \,\},$$

and set

$$d_s(u,u) := 0 \qquad (\forall\, u \in V).$$

Then the *Wiener index* of $G$ is defined by

$$WI(G) := \sum_{\{u,v\} \subseteq V,\ u \ne v} \sigma(u)\sigma(v)\, d_s(u,v),$$

where the sum is taken over all unordered pairs of distinct vertices with positive vertex-membership values.

Equivalently, if

$$V^* := \{u \in V : \sigma(u) > 0\},$$

then

$$WI(G) = \sum_{\{u,v\} \subseteq V^*} \sigma(u)\sigma(v)\, d_s(u,v).$$

Wiener index in an uncertain graph measures the total uncertainty-weighted shortest-path distance between all pairs of support vertices, thereby quantifying global structural closeness and compactness under uncertainty.

**Definition 5.11.2** (Wiener-Evaluable Uncertain Model)**.** Let $M$ be an uncertain model with degree-domain

$$\mathrm{Dom}(M) \subseteq [0,1]^k.$$

We say that $M$ is *Wiener-evaluable* if it is equipped with:

1. a distinguished element
$$0_M \in \mathrm{Dom}(M),$$
called the *zero degree*;
2. an *edge-length evaluation map*
$$\Lambda_M : \mathrm{Dom}(M) \setminus \{0_M\} \to (0,\infty);$$
3. a *vertex-weight evaluation map*
$$\omega_M : \mathrm{Dom}(M) \to [0,\infty).$$

**Definition 5.11.3** (Uncertain Graph of Type $M$)**.** Let $V$ be a finite set, and let $M$ be a Wiener-evaluable uncertain model. An *uncertain graph* of type $M$ on $V$ is a triple

$$G_M = (V, \sigma_M, \eta_M),$$

where

$$\sigma_M : V \to \mathrm{Dom}(M), \qquad \eta_M : \binom{V}{2} \to \mathrm{Dom}(M)$$

are functions.

Equivalently,

$$(V, \sigma_M)$$

is an Uncertain Set of type $M$ on the vertex set $V$, and

$$\left(\binom{V}{2}, \eta_M\right)$$

is an Uncertain Set of type $M$ on the set of unordered pairs of distinct vertices.

**Definition 5.11.4** (Wiener Index in an Uncertain Graph)**.** Let

$$G_M = (V, \sigma_M, \eta_M)$$

be a finite uncertain graph of type $M$, where $M$ is Wiener-evaluable.

Define the support vertex set by

$$V_M^* := \{u \in V : \sigma_M(u) \neq 0_M\},$$

and the support edge set by

$$E_M^*(G_M) := \left\{ \{u,v\} \in \binom{V_M^*}{2} : \eta_M(\{u,v\}) \neq 0_M \right\}.$$

Let

$$G_M^* := (V_M^*, E_M^*(G_M))$$

be the support graph, and assume that $G_M^*$ is connected.

A path from $u$ to $v$ in $G_M$ is a sequence of distinct vertices

$$P : u_0, u_1, \ldots, u_n$$

in the support graph $G_M^*$ such that

$$u_0 = u, \qquad u_n = v,$$

and

$$\{u_{i-1}, u_i\} \in E_M^*(G_M) \qquad (i = 1, \ldots, n).$$

The *uncertain length* of such a path $P$ is defined by

$$\ell_M(P) := \sum_{i=1}^{n} \Lambda_M\big(\eta_M(\{u_{i-1}, u_i\})\big).$$

For two vertices $u, v \in V_M^*$, the *uncertain distance* between $u$ and $v$ is defined by

$$d_M(u, v) := \min\big\{\ell_M(P) : P \text{ is a path from } u \text{ to } v \text{ in } G_M\big\},$$

and

$$d_M(u, u) := 0 \qquad (\forall\, u \in V_M^*).$$

A path $P$ from $u$ to $v$ is called an *uncertain geodesic* if

$$\ell_M(P) = d_M(u, v).$$

Then the *Wiener index* of $G_M$ is defined by

$$WI_M(G_M) := \sum_{\{u,v\} \in \binom{V_M^*}{2}} \omega_M\big(\sigma_M(u)\big)\, \omega_M\big(\sigma_M(v)\big)\, d_M(u, v).$$

**Theorem 5.11.5** (Well-definedness of Wiener Index in an Uncertain Graph)**.** *Let $V$ be a finite set, let $M$ be a Wiener-evaluable uncertain model, and let*

$$G_M = (V, \sigma_M, \eta_M)$$

*be an uncertain graph of type $M$ on $V$. Assume that the support graph*

$$G_M^* = (V_M^*, E_M^*(G_M))$$

*is connected. Then:*

1. *the support vertex set $V_M^*$, the support edge set $E_M^*(G_M)$, and the support graph $G_M^*$ are well-defined;*
2. *for every pair of vertices $u, v \in V_M^*$, the uncertain distance*

   $$d_M(u, v)$$

   *is a well-defined nonnegative real number;*
3. *for every pair of distinct vertices $u, v \in V_M^*$, the product*

   $$\omega_M(\sigma_M(u))\, \omega_M(\sigma_M(v))\, d_M(u, v)$$

   *is a well-defined nonnegative real number;*
4. *the Wiener index*

   $$WI_M(G_M)$$

   *is a well-defined nonnegative real number;*
5. *for every pair $u, v \in V_M^*$, there exists an uncertain geodesic from $u$ to $v$.*

*Proof.* Since $M$ is an uncertain model, its degree-domain

$$\mathrm{Dom}(M) \subseteq [0,1]^k$$

is fixed. Since $M$ is Wiener-evaluable, the distinguished element

$$0_M \in \mathrm{Dom}(M),$$

the map

$$\Lambda_M : \mathrm{Dom}(M) \setminus \{0_M\} \to (0,\infty),$$

and the map

$$\omega_M : \mathrm{Dom}(M) \to [0,\infty)$$

are fixed as part of the structure.

Because

$$\sigma_M : V \to \mathrm{Dom}(M) \qquad \text{and} \qquad \eta_M : \binom{V}{2} \to \mathrm{Dom}(M)$$

are functions, the pairs

$$(V, \sigma_M) \qquad \text{and} \qquad \left(\binom{V}{2}, \eta_M\right)$$

are well-defined Uncertain Sets of type $M$.

Hence, for each $u \in V$, the statement

$$\sigma_M(u) \neq 0_M$$

has a definite truth value, so

$$V_M^* := \{u \in V : \sigma_M(u) \neq 0_M\}$$

is a well-defined subset of $V$.

Likewise, for each

$$\{u,v\} \in \binom{V_M^*}{2},$$

the statement

$$\eta_M(\{u,v\}) \neq 0_M$$

has a definite truth value, and therefore

$$E_M^*(G_M) := \left\{ \{u,v\} \in \binom{V_M^*}{2} : \eta_M(\{u,v\}) \neq 0_M \right\}$$

is a well-defined subset of $\binom{V_M^*}{2}$.

Hence

$$G_M^* = (V_M^*, E_M^*(G_M))$$

is a well-defined crisp graph. This proves (1).

Now fix $u, v \in V_M^*$. If $u = v$, then by definition

$$d_M(u,u) = 0,$$

which is well-defined.

Assume next that $u \neq v$. Since the support graph $G_M^*$ is connected, there exists at least one path from $u$ to $v$ in $G_M^*$. Because $V_M^*$ is finite, there are only finitely many simple $u$-$v$ paths in $G_M^*$. Hence the set

$$\mathcal{P}_{u,v}(G_M) := \{\, P : P \text{ is a path from } u \text{ to } v \text{ in } G_M \,\}$$

is finite and nonempty.

Let

$$P : u_0, u_1, \ldots, u_n$$

be a path in $\mathcal{P}_{u,v}(G_M)$. For each $i = 1, \ldots, n$, we have

$$\{u_{i-1}, u_i\} \in E^*_M(G_M),$$

and therefore

$$\eta_M(\{u_{i-1}, u_i\}) \neq 0_M.$$

Hence

$$\Lambda_M\big(\eta_M(\{u_{i-1}, u_i\})\big) \in (0, \infty)$$

is well-defined for each $i$. Thus

$$\ell_M(P) = \sum_{i=1}^{n} \Lambda_M\big(\eta_M(\{u_{i-1}, u_i\})\big)$$

is a finite sum of positive real numbers, so $\ell_M(P) \in (0, \infty)$ is well-defined.

Therefore the set

$$\{\ell_M(P) : P \in \mathcal{P}_{u,v}(G_M)\}$$

is a finite nonempty subset of $(0, \infty)$. Every finite nonempty subset of $\mathbb{R}$ has a minimum, so

$$d_M(u, v) = \min\big\{\ell_M(P) : P \text{ is a path from } u \text{ to } v \text{ in } G_M\big\}$$

is a well-defined positive real number. Thus $d_M(u, v)$ is a well-defined nonnegative real number for all $u, v \in V^*_M$. This proves (2).

Now let $u, v \in V^*_M$ with $u \neq v$. Since

$$\sigma_M(u), \sigma_M(v) \in \mathrm{Dom}(M),$$

the values

$$\omega_M(\sigma_M(u)) \qquad \text{and} \qquad \omega_M(\sigma_M(v))$$

are well-defined elements of $[0, \infty)$. By (2), $d_M(u, v) \in [0, \infty)$ is well-defined. Hence the product

$$\omega_M(\sigma_M(u))\, \omega_M(\sigma_M(v))\, d_M(u, v)$$

is a well-defined nonnegative real number. This proves (3).

Since $V$ is finite, the support set $V^*_M \subseteq V$ is finite. Therefore the set

$$\binom{V^*_M}{2}$$

of unordered pairs of distinct support vertices is finite. By (3), each summand

$$\omega_M(\sigma_M(u))\, \omega_M(\sigma_M(v))\, d_M(u, v)$$

is a well-defined nonnegative real number. Hence

$$WI_M(G_M) = \sum_{\{u,v\} \in \binom{V^*_M}{2}} \omega_M(\sigma_M(u))\, \omega_M(\sigma_M(v))\, d_M(u, v)$$

is a finite sum of well-defined nonnegative real numbers. Therefore $WI_M(G_M)$ is a well-defined nonnegative real number. This proves (4).

Finally, fix $u, v \in V^*_M$. If $u = v$, the trivial length-zero path at $u$ is a geodesic. If $u \neq v$, then the finite nonempty set

$$\{\ell_M(P) : P \in \mathcal{P}_{u,v}(G_M)\}$$

has a minimum, and this minimum is attained by at least one path

$$P_0 \in \mathcal{P}_{u,v}(G_M).$$

Hence

$$\ell_M(P_0) = d_M(u, v),$$

so $P_0$ is an uncertain geodesic from $u$ to $v$. This proves (5). □

For convenience, Table 5.5 summarizes representative Wiener-index-related concepts according to the dimension $k$ of the information associated with vertices and/or edges.

Table 5.5: Representative Wiener-index-related concepts under uncertainty-aware graph frameworks, classified by the dimension $k$ of the information attached to vertices and/or edges.

| $k$ | **Wiener-index-related concept** | **Typical coordinate form** | **Canonical information attached to vertices/edges** |
|---|---|---|---|
| 1 | Wiener Index in a Fuzzy Graph | $\mu$ | The Wiener index is studied in a fuzzy graph, where each vertex and edge is associated with a single membership degree in $[0, 1]$. |
| 2 | Wiener Index in an Intuitionistic Fuzzy Graph | $(\mu, \nu)$ | The Wiener index is defined on an intuitionistic fuzzy graph, where each vertex and edge carries a membership degree and a non-membership degree, usually satisfying $\mu + \nu \leq 1$. |
| 3 | Wiener Index in a Neutrosophic Graph [703] | $(T, I, F)$ | The Wiener index is defined on a neutrosophic graph, where each vertex and edge is described by truth, indeterminacy, and falsity degrees. |

Related extensions and generalizations of the Wiener index include weighted Wiener index [704], hyper-Wiener index [705, 706], terminal Wiener index [707, 708], edge Wiener index [709, 710], Steiner Wiener index [711, 712], degree-distance index [713, 714], Gutman index [715, 716], and Schultz index [717, 718].

## 5.12 Sombor index in Uncertain graphs

The Sombor index of a fuzzy graph sums edgewise square-root expressions of endpoint weighted degrees, quantifying overall structural irregularity through membership-sensitive vertex contributions in networks [719–721].

**Definition 5.12.1** (Sombor Index in a Fuzzy Graph)**.** Let

$$\widetilde{G} = (V, \xi, \Omega)$$

be a finite fuzzy graph, where

$$\xi : V \to [0, 1], \qquad \Omega : V \times V \to [0, 1], \qquad \Omega(u, v) \leq \min\{\xi(u), \xi(v)\} \quad (\forall\, u, v \in V),$$

and $\Omega$ is symmetric.

Define the support edge set by

$$E^*(\widetilde{G}) := \big\{\{u, v\} \subseteq V : u \neq v,\ \Omega(u, v) > 0\big\}.$$

For each vertex $u \in V$, the (fuzzy) degree of $u$ is defined by

$$\Gamma_{\widetilde{G}}(u) := \sum_{\substack{v \in V \\ v \neq u}} \Omega(u, v).$$

Then the *Sombor index* of the fuzzy graph $\widetilde{G}$ is defined by

$$SO_F(\widetilde{G}) := \sum_{\{u,v\} \in E^*(\widetilde{G})} \sqrt{\big(\xi(u)\Gamma_{\widetilde{G}}(u)\big)^2 + \big(\xi(v)\Gamma_{\widetilde{G}}(v)\big)^2}.$$

Table 5.6: Representative Sombor-index-related concepts under uncertainty-aware graph frameworks, classified by the dimension $k$ of the information attached to vertices and/or edges.

| $k$ | **Sombor-index-related concept** | **Typical coordinate form** | **Canonical information attached to vertices/edges** |
|---|---|---|---|
| 1 | Sombor Index in a Fuzzy Graph | $\mu$ | The Sombor index is studied in a fuzzy graph, where each vertex and edge is associated with a single membership degree in $[0, 1]$. |
| 2 | Sombor Index in an Intuitionistic Fuzzy Graph | $(\mu, \nu)$ | The Sombor index is defined on an intuitionistic fuzzy graph, where each vertex and edge carries a membership degree and a non-membership degree, usually satisfying $\mu + \nu \leq 1$. |
| 3 | Sombor Index in a Neutrosophic Graph [722] | $(T, I, F)$ | The Sombor index is defined on a neutrosophic graph, where each vertex and edge is described by truth, indeterminacy, and falsity degrees. |

For convenience, Table 5.6 summarizes representative Sombor-index-related concepts according to the dimension $k$ of the information associated with vertices and/or edges.

The Sombor index of an uncertain graph aggregates edgewise Euclidean contributions of uncertainty-weighted endpoint degrees, thereby measuring global structural irregularity under uncertainty.

**Definition 5.12.2** (Sombor-Evaluable Uncertain Model)**.** Let $M$ be an uncertain model with degree-domain

$$\mathrm{Dom}(M) \subseteq [0, 1]^k.$$

We say that $M$ is *Sombor-evaluable* if it is equipped with:

1. a distinguished element
$$0_M \in \mathrm{Dom}(M),$$
called the *zero degree*;
2. an *edge-degree evaluation map*
$$\delta_M : \mathrm{Dom}(M) \to [0, \infty);$$
3. a *vertex-weight evaluation map*
$$\omega_M : \mathrm{Dom}(M) \to [0, \infty);$$

such that

$$\delta_M(d) = 0 \iff d = 0_M, \qquad \omega_M(d) = 0 \iff d = 0_M \qquad (\forall\, d \in \mathrm{Dom}(M)).$$

**Definition 5.12.3** (Uncertain Graph of Type $M$)**.** Let $V$ be a finite set, and let $M$ be a Sombor-evaluable uncertain model. An *uncertain graph* of type $M$ on $V$ is a triple

$$G_M = (V, \sigma_M, \eta_M),$$

where

$$\sigma_M : V \to \mathrm{Dom}(M), \qquad \eta_M : \binom{V}{2} \to \mathrm{Dom}(M)$$

are functions satisfying

$$\delta_M\big(\eta_M(\{u, v\})\big) \leq \min\Big\{\omega_M\big(\sigma_M(u)\big), \omega_M\big(\sigma_M(v)\big)\Big\} \qquad \left(\forall\, \{u, v\} \in \binom{V}{2}\right).$$

Equivalently,

$$(V, \sigma_M)$$

is an Uncertain Set of type $M$ on the vertex set $V$, and

$$\left(\binom{V}{2}, \eta_M\right)$$

is an Uncertain Set of type $M$ on the set of unordered pairs of distinct vertices.

**Definition 5.12.4** (Sombor Index in an Uncertain Graph)**.** Let

$$G_M = (V, \sigma_M, \eta_M)$$

be a finite uncertain graph of type $M$, where $M$ is Sombor-evaluable.

Define the support vertex set by

$$V_M^* := \{u \in V : \sigma_M(u) \neq 0_M\},$$

and the support edge set by

$$E_M^*(G_M) := \left\{ \{u, v\} \in \binom{V_M^*}{2} : \eta_M(\{u, v\}) \neq 0_M \right\}.$$

For each vertex $u \in V$, define the *uncertain degree* of $u$ by

$$\Gamma_{G_M}(u) := \sum_{\substack{v \in V \\ v \neq u}} \delta_M\big(\eta_M(\{u, v\})\big).$$

Then the *Sombor index* of $G_M$ is defined by

$$SO_M(G_M) := \sum_{\{u,v\} \in E_M^*(G_M)} \sqrt{\big(\omega_M(\sigma_M(u))\, \Gamma_{G_M}(u)\big)^2 + \big(\omega_M(\sigma_M(v))\, \Gamma_{G_M}(v)\big)^2}.$$

**Theorem 5.12.5** (Well-definedness of the Sombor Index in an Uncertain Graph)**.** *Let $V$ be a finite set, let $M$ be a Sombor-evaluable uncertain model, and let*

$$G_M = (V, \sigma_M, \eta_M)$$

*be an uncertain graph of type $M$ on $V$. Then:*

1. *the support vertex set*

   $$V_M^* = \{u \in V : \sigma_M(u) \neq 0_M\}$$

   *and the support edge set*

   $$E_M^*(G_M) = \left\{ \{u, v\} \in \binom{V_M^*}{2} : \eta_M(\{u, v\}) \neq 0_M \right\}$$

   *are well-defined finite sets;*

2. *for every vertex $u \in V$, the uncertain degree*

   $$\Gamma_{G_M}(u) = \sum_{\substack{v \in V \\ v \neq u}} \delta_M\big(\eta_M(\{u, v\})\big)$$

   *is a well-defined element of $[0, \infty)$;*

3. *for every support edge*

   $$\{u, v\} \in E_M^*(G_M),$$

   *the quantity*

   $$\sqrt{\big(\omega_M(\sigma_M(u))\, \Gamma_{G_M}(u)\big)^2 + \big(\omega_M(\sigma_M(v))\, \Gamma_{G_M}(v)\big)^2}$$

   *is a well-defined element of $[0, \infty)$;*

4. *the Sombor index*

$$SO_M(G_M)$$

*is a well-defined element of* $[0,\infty)$*.*

*Proof.* Since $M$ is an uncertain model, its degree-domain

$$\mathrm{Dom}(M) \subseteq [0,1]^k$$

is fixed. Since $M$ is Sombor-evaluable, the distinguished element

$$0_M \in \mathrm{Dom}(M)$$

and the maps

$$\delta_M : \mathrm{Dom}(M) \to [0,\infty), \qquad \omega_M : \mathrm{Dom}(M) \to [0,\infty)$$

are fixed as part of the structure.

Because

$$\sigma_M : V \to \mathrm{Dom}(M) \qquad \text{and} \qquad \eta_M : \binom{V}{2} \to \mathrm{Dom}(M)$$

are functions, the pairs

$$(V, \sigma_M) \qquad \text{and} \qquad \left(\binom{V}{2}, \eta_M\right)$$

are well-defined Uncertain Sets of type $M$.

For each $u \in V$, the statement

$$\sigma_M(u) \neq 0_M$$

has a definite truth value, because both $\sigma_M(u)$ and $0_M$ belong to $\mathrm{Dom}(M)$. Hence

$$V_M^* := \{u \in V : \sigma_M(u) \neq 0_M\}$$

is a well-defined subset of $V$.

Now let

$$\{u,v\} \in \binom{V}{2}$$

and assume that

$$\eta_M(\{u,v\}) \neq 0_M.$$

Since $\delta_M(d) = 0 \iff d = 0_M$, it follows that

$$\delta_M\big(\eta_M(\{u,v\})\big) > 0.$$

By the compatibility condition in the definition of uncertain graph,

$$\delta_M\big(\eta_M(\{u,v\})\big) \leq \min\Big\{\omega_M\big(\sigma_M(u)\big), \omega_M\big(\sigma_M(v)\big)\Big\}.$$

Hence

$$\omega_M(\sigma_M(u)) > 0 \qquad \text{and} \qquad \omega_M(\sigma_M(v)) > 0.$$

Using $\omega_M(d) = 0 \iff d = 0_M$, we obtain

$$\sigma_M(u) \neq 0_M \qquad \text{and} \qquad \sigma_M(v) \neq 0_M.$$

Therefore every edge with nonzero degree has both endpoints in $V_M^*$, so

$$E_M^*(G_M) = \left\{\{u,v\} \in \binom{V_M^*}{2} : \eta_M(\{u,v\}) \neq 0_M\right\}$$

is a well-defined subset of $\binom{V_M^*}{2}$.

Since $V$ is finite, $V_M^* \subseteq V$ is finite, and therefore $\binom{V_M^*}{2}$ is finite. Hence $E_M^*(G_M)$ is finite. This proves (1).

Fix a vertex $u \in V$. For every $v \in V$ with $v \neq u$, the value

$$\eta_M(\{u, v\}) \in \mathrm{Dom}(M)$$

is well-defined. Hence

$$\delta_M\big(\eta_M(\{u, v\})\big) \in [0, \infty)$$

is well-defined. Since $V$ is finite, the set

$$\{\, v \in V : v \neq u \,\}$$

is finite, and therefore

$$\Gamma_{G_M}(u) = \sum_{\substack{v \in V \\ v \neq u}} \delta_M\big(\eta_M(\{u, v\})\big)$$

is a finite sum of well-defined nonnegative real numbers. Thus

$$\Gamma_{G_M}(u) \in [0, \infty)$$

is well-defined. This proves (2).

Now let

$$\{u, v\} \in E_M^*(G_M).$$

By definition of support edge set,

$$u, v \in V_M^*,$$

so

$$\sigma_M(u) \neq 0_M \qquad \text{and} \qquad \sigma_M(v) \neq 0_M.$$

Hence

$$\omega_M(\sigma_M(u)) > 0 \qquad \text{and} \qquad \omega_M(\sigma_M(v)) > 0.$$

By part (2),

$$\Gamma_{G_M}(u), \Gamma_{G_M}(v) \in [0, \infty)$$

are well-defined. Therefore the real numbers

$$\omega_M(\sigma_M(u))\, \Gamma_{G_M}(u) \qquad \text{and} \qquad \omega_M(\sigma_M(v))\, \Gamma_{G_M}(v)$$

are well-defined and nonnegative. Consequently,

$$\big(\omega_M(\sigma_M(u))\, \Gamma_{G_M}(u)\big)^2 + \big(\omega_M(\sigma_M(v))\, \Gamma_{G_M}(v)\big)^2$$

is a well-defined nonnegative real number, and hence its square root

$$\sqrt{\big(\omega_M(\sigma_M(u))\, \Gamma_{G_M}(u)\big)^2 + \big(\omega_M(\sigma_M(v))\, \Gamma_{G_M}(v)\big)^2}$$

is well-defined and belongs to $[0, \infty)$. This proves (3).

Finally, since $E_M^*(G_M)$ is finite by part (1), and each summand in

$$SO_M(G_M) = \sum_{\{u,v\} \in E_M^*(G_M)} \sqrt{\big(\omega_M(\sigma_M(u))\, \Gamma_{G_M}(u)\big)^2 + \big(\omega_M(\sigma_M(v))\, \Gamma_{G_M}(v)\big)^2}$$

is a well-defined nonnegative real number by part (3), it follows that $SO_M(G_M)$ is a finite sum of well-defined nonnegative real numbers. Therefore

$$SO_M(G_M) \in [0, \infty)$$

is well-defined. This proves (4). □

## 5.13 Uncertain Graph Energy

Graph energy is the sum of the absolute values of adjacency-matrix eigenvalues, quantifying a graph's overall spectral magnitude and reflecting structural complexity within networks globally [723]. Fuzzy graph energy is the sum of absolute eigenvalues of the fuzzy adjacency matrix, measuring the spectral intensity of uncertain weighted relationships across the graph [236, 724, 725].

**Definition 5.13.1** (Adjacency Matrix of a Fuzzy Graph)**.** Let

$$G = (V, \sigma, \mu)$$

be a finite fuzzy graph, where

$$V = \{v_1, v_2, \ldots, v_n\}, \qquad \sigma : V \to [0,1], \qquad \mu : V \times V \to [0,1],$$

such that

$$\mu(v_i, v_j) = \mu(v_j, v_i) \qquad \text{and} \qquad \mu(v_i, v_j) \le \min\{\sigma(v_i), \sigma(v_j)\}$$

for all $i, j$.

The *adjacency matrix* of $G$ is the $n \times n$ real matrix

$$A(G) = [a_{ij}], \qquad a_{ij} := \mu(v_i, v_j) \qquad (1 \le i, j \le n).$$

**Definition 5.13.2** (Spectrum of a Fuzzy Graph)**.** Let

$$A(G)$$

be the adjacency matrix of a finite fuzzy graph $G$. The multiset of eigenvalues of $A(G)$,

$$\mathrm{Spec}(G) = \{\lambda_1, \lambda_2, \ldots, \lambda_n\},$$

is called the *spectrum* of $G$.

**Definition 5.13.3** (Energy of a Fuzzy Graph)**.** Let

$$G = (V, \sigma, \mu)$$

be a finite fuzzy graph with adjacency matrix

$$A(G).$$

Since $A(G)$ is a real symmetric matrix, all its eigenvalues

$$\lambda_1, \lambda_2, \ldots, \lambda_n$$

are real.

The *energy* of the fuzzy graph $G$, denoted by

$$E(G),$$

is defined by

$$E(G) := \sum_{i=1}^{n} |\lambda_i|,$$

where $\lambda_1, \lambda_2, \ldots, \lambda_n$ are the eigenvalues of $A(G)$.

In the uncertain-set framework, this notion is obtained by evaluating uncertainty degrees on edges through a real-valued adjacency map.

**Definition 5.13.4** (Energy-Evaluable Uncertain Model)**.** Let $M$ be an uncertain model with degree-domain

$$\mathrm{Dom}(M) \subseteq [0,1]^k.$$

We say that $M$ is *energy-evaluable* if it is equipped with:

1. a distinguished element
$$0_M \in \mathrm{Dom}(M),$$
called the *zero degree*;

2. a map
$$\Phi_M : \mathrm{Dom}(M) \to \mathbb{R},$$
called the *adjacency evaluation map*, such that
$$\Phi_M(0_M) = 0.$$

**Definition 5.13.5** (Uncertain Graph of Type $M$)**.** Let

$$V = \{v_1, v_2, \ldots, v_n\}$$

be a finite nonempty set, and let $M$ be an energy-evaluable uncertain model. An *uncertain graph* of type $M$ on $V$ is a triple

$$G_M = (V, \sigma_M, \eta_M),$$

where

$$\sigma_M : V \to \mathrm{Dom}(M), \qquad \eta_M : \binom{V}{2} \to \mathrm{Dom}(M)$$

are functions.

Equivalently,

$$(V, \sigma_M)$$

is an Uncertain Set of type $M$ on the vertex set $V$, and

$$\left(\binom{V}{2}, \eta_M\right)$$

is an Uncertain Set of type $M$ on the set of unordered pairs of distinct vertices.

**Definition 5.13.6** (Adjacency Matrix of an Uncertain Graph)**.** Let

$$G_M = (V, \sigma_M, \eta_M)$$

be a finite uncertain graph of type $M$, where

$$V = \{v_1, v_2, \ldots, v_n\}.$$

The *adjacency matrix* of $G_M$ is the $n \times n$ real matrix

$$A_M(G_M) = [a_{ij}],$$

defined by

$$a_{ii} := 0 \qquad (1 \leq i \leq n),$$

and

$$a_{ij} := \Phi_M\big(\eta_M(\{v_i, v_j\})\big) \qquad (1 \leq i \neq j \leq n).$$

**Definition 5.13.7** (Spectrum of an Uncertain Graph)**.** Let

$$A_M(G_M)$$

be the adjacency matrix of a finite uncertain graph $G_M$. The multiset of eigenvalues of $A_M(G_M)$,

$$\mathrm{Spec}_M(G_M) = \{\lambda_1, \lambda_2, \ldots, \lambda_n\},$$

is called the *spectrum* of $G_M$.

**Definition 5.13.8** (Energy of an Uncertain Graph)**.** Let

$$G_M = (V, \sigma_M, \eta_M)$$

be a finite uncertain graph of type $M$, and let

$$A_M(G_M)$$

be its adjacency matrix. Since $A_M(G_M)$ is a real symmetric matrix, all its eigenvalues

$$\lambda_1, \lambda_2, \ldots, \lambda_n$$

are real.

The *energy* of the uncertain graph $G_M$, denoted by

$$E_M(G_M),$$

is defined by

$$E_M(G_M) := \sum_{i=1}^{n} |\lambda_i|,$$

where

$$\mathrm{Spec}_M(G_M) = \{\lambda_1, \lambda_2, \ldots, \lambda_n\}.$$

**Theorem 5.13.9** (Well-definedness of the Energy of an Uncertain Graph)**.** *Let*

$$V = \{v_1, v_2, \ldots, v_n\}$$

*be a finite nonempty set, let $M$ be an energy-evaluable uncertain model, and let*

$$G_M = (V, \sigma_M, \eta_M)$$

*be an uncertain graph of type $M$ on $V$. Then:*

1. *the adjacency matrix*

   $$A_M(G_M)$$

   *is a well-defined real symmetric $n \times n$ matrix;*
2. *the spectrum*

   $$\mathrm{Spec}_M(G_M)$$

   *is well-defined and consists of $n$ real eigenvalues counted with algebraic multiplicity;*
3. *the energy*

   $$E_M(G_M) = \sum_{i=1}^{n} |\lambda_i|$$

   *is a well-defined nonnegative real number;*
4. *the multiset* $\mathrm{Spec}_M(G_M)$ *and the value* $E_M(G_M)$ *are independent of the chosen ordering of the vertices of $V$.*

*Proof.* Since $M$ is an uncertain model, its degree-domain

$$\mathrm{Dom}(M) \subseteq [0,1]^k$$

is fixed. Since $M$ is energy-evaluable, the distinguished element

$$0_M \in \mathrm{Dom}(M)$$

and the map

$$\Phi_M : \mathrm{Dom}(M) \to \mathbb{R}$$

are fixed as part of the structure.

Because

$$\sigma_M : V \to \mathrm{Dom}(M) \qquad \text{and} \qquad \eta_M : \binom{V}{2} \to \mathrm{Dom}(M)$$

are functions, the pairs

$$(V, \sigma_M) \qquad \text{and} \qquad \left(\binom{V}{2}, \eta_M\right)$$

are well-defined Uncertain Sets of type $M$.

For each pair of distinct indices $i, j$, the unordered pair

$$\{v_i, v_j\} \in \binom{V}{2}$$

is well-defined, hence

$$\eta_M(\{v_i, v_j\}) \in \mathrm{Dom}(M)$$

is well-defined, and therefore

$$\Phi_M\big(\eta_M(\{v_i, v_j\})\big) \in \mathbb{R}$$

is well-defined. Also, for each $i$,

$$a_{ii} = 0$$

is well-defined. Thus every entry $a_{ij}$ of

$$A_M(G_M) = [a_{ij}]$$

is well-defined and real, so $A_M(G_M)$ is a well-defined real $n \times n$ matrix.

To prove symmetry, let $i \neq j$. Since

$$\{v_i, v_j\} = \{v_j, v_i\}$$

as unordered pairs, we have

$$a_{ij} = \Phi_M\big(\eta_M(\{v_i, v_j\})\big) = \Phi_M\big(\eta_M(\{v_j, v_i\})\big) = a_{ji}.$$

Also, $a_{ii} = a_{ii}$ trivially. Hence

$$A_M(G_M)^T = A_M(G_M),$$

so $A_M(G_M)$ is symmetric. This proves (1).

Since $A_M(G_M)$ is a real symmetric matrix, the spectral theorem implies that all of its eigenvalues are real. Therefore the spectrum

$$\mathrm{Spec}_M(G_M) = \{\lambda_1, \lambda_2, \ldots, \lambda_n\}$$

is well-defined as a multiset of $n$ real numbers counted with algebraic multiplicity. This proves (2).

Because each $\lambda_i \in \mathbb{R}$, the absolute value

$$|\lambda_i|$$

is well-defined and belongs to $[0, \infty)$. Since there are finitely many eigenvalues, the sum

$$E_M(G_M) := \sum_{i=1}^{n} |\lambda_i|$$

is a finite sum of well-defined nonnegative real numbers. Hence $E_M(G_M)$ is a well-defined nonnegative real number. This proves (3).

Finally, consider another ordering of the same vertex set $V$, say

$$V = \{w_1, w_2, \ldots, w_n\}.$$

Since both $(v_1, \ldots, v_n)$ and $(w_1, \ldots, w_n)$ are enumerations of the same finite set, there exists a permutation $\pi$ of $\{1, 2, \ldots, n\}$ such that

$$w_i = v_{\pi(i)} \qquad (1 \leq i \leq n).$$

Let $P$ be the permutation matrix corresponding to $\pi$. If

$$A'_M(G_M)$$

denotes the adjacency matrix obtained from the ordering $(w_1, \ldots, w_n)$, then

$$A'_M(G_M) = P^T A_M(G_M) P.$$

Thus $A'_M(G_M)$ is permutation-similar to $A_M(G_M)$, and hence both matrices have the same characteristic polynomial and the same multiset of eigenvalues. Therefore the spectrum $\mathrm{Spec}_M(G_M)$ is independent of the chosen ordering of the vertices.

Since the energy is defined as the sum of the absolute values of the eigenvalues, it also remains unchanged under permutation similarity. Hence

$$E_M(G_M)$$

is independent of the chosen ordering of $V$. This proves (4). □

Representative graph-energy-related concepts under uncertainty-aware graph frameworks are listed in Table 5.7.

Table 5.7: Representative graph-energy-related concepts under uncertainty-aware graph frameworks, classified by the dimension $k$ of the information attached to vertices and/or edges.

| $k$ | **Graph-energy-related concept** | **Typical coordinate form** | **Canonical information attached to vertices/edges** |
|---|---|---|---|
| 1 | Fuzzy Graph Energy | $\mu$ | Graph energy is studied in a fuzzy graph, where each vertex and edge is associated with a single membership degree in $[0, 1]$. |
| 2 | Intuitionistic Fuzzy Graph Energy [725, 726] | $(\mu, \nu)$ | Graph energy is defined on an intuitionistic fuzzy graph, where each vertex and edge carries a membership degree and a non-membership degree, usually satisfying $\mu + \nu \leq 1$. |
| 3 | Neutrosophic Graph Energy [725] | $(T, I, F)$ | Graph energy is defined on a neutrosophic graph, where each vertex and edge is described by truth, indeterminacy, and falsity degrees. |

Related extensions and generalizations of graph energy include Laplacian energy [727], signless Laplacian energy [728, 729], distance energy [730, 731], distance Laplacian energy [732, 733], incidence energy [734, 735], Randić energy [736], skew energy [737], and Seidel energy [738, 739].

# Chapter 6

# Applications

In this chapter, we discuss several applications of fuzzy graphs, neutrosophic graphs, uncertain graphs, and related frameworks.

## 6.1 Uncertain Molecular Graph

Molecular graph is a graph-theoretic representation of a molecule, where vertices denote atoms and edges denote chemical bonds, capturing structural connectivity for mathematical analysis precisely [740, 741]. Fuzzy molecular graph extends a molecular graph by assigning membership degrees to atoms and bonds, enabling uncertain, weighted, or partial molecular structures to be modeled (cf. [742]).

**Definition 6.1.1** (Molecular Fuzzy Graph)**.** (cf. [742]) Let $\Sigma_V$ and $\Sigma_E$ be finite nonempty sets of *atom attributes* and *bond attributes*, respectively.

A *molecular fuzzy graph* is a sextuple

$$MF = (V, E, \lambda_V, \lambda_E, \sigma, \mu),$$

satisfying the following conditions:

1. $V$ is a finite nonempty set whose elements represent atoms;
2. $$E \subseteq \{\{u, v\} \subseteq V : u \neq v\}$$
   is a finite set of undirected edges, whose elements represent chemical bonds;
3. $$\lambda_V : V \to \Sigma_V$$
   is a vertex-labeling map assigning to each atom its chemical attribute (such as element type, charge, or isotope);
4. $$\lambda_E : E \to \Sigma_E$$
   is an edge-labeling map assigning to each bond its chemical attribute (such as bond type or bond order);
5. $$\sigma : V \to [0, 1]$$
   is the vertex-membership function;

6.
$$\mu : E \to [0,1]$$
is the edge-membership function;

7. for every edge
$$e = \{u, v\} \in E,$$
the fuzzy consistency condition
$$\mu(e) \le \min\{\sigma(u), \sigma(v)\}$$
holds.

The quadruple
$$(V, E, \lambda_V, \lambda_E)$$
is called the *underlying molecular graph* of $MF$.

If
$$\sigma(v) = 1 \quad (\forall\, v \in V), \qquad \mu(e) = 1 \quad (\forall\, e \in E),$$
then $MF$ reduces to the corresponding crisp molecular graph.

**Definition 6.1.2** (Molecular-Compatible Uncertain Model)**.** Let $M$ be an uncertain model with degree-domain
$$\mathrm{Dom}(M) \subseteq [0,1]^k.$$
We say that $M$ is *molecular-compatible* if it is equipped with:

1. a partial order
$$\preceq_M \subseteq \mathrm{Dom}(M) \times \mathrm{Dom}(M);$$
2. a symmetric map
$$\Gamma_M : \mathrm{Dom}(M) \times \mathrm{Dom}(M) \to \mathrm{Dom}(M),$$
called the *bond-compatibility map*, such that
$$\Gamma_M(a, b) = \Gamma_M(b, a) \qquad (\forall\, a, b \in \mathrm{Dom}(M)).$$

**Definition 6.1.3** (Molecular Uncertain Graph)**.** Let $\Sigma_V$ and $\Sigma_E$ be finite nonempty sets of *atom attributes* and *bond attributes*, respectively. Let $M$ be a molecular-compatible uncertain model.

A *molecular uncertain graph* of type $M$ is a sextuple
$$MU_M = (V, E, \lambda_V, \lambda_E, \sigma_M, \eta_M),$$
satisfying the following conditions:

1. $V$ is a finite nonempty set whose elements represent atoms;
2.
$$E \subseteq \big\{\{u, v\} \subseteq V : u \ne v\big\}$$
is a finite set of undirected edges, whose elements represent chemical bonds;
3.
$$\lambda_V : V \to \Sigma_V$$
is a vertex-labeling map assigning to each atom its chemical attribute (such as element type, charge, isotope, or valence class);

4. $$\lambda_E : E \to \Sigma_E$$
   is an edge-labeling map assigning to each bond its chemical attribute (such as bond type, bond order, or bond category);
5. $$\sigma_M : V \to \mathrm{Dom}(M)$$
   is the *vertex uncertainty-degree function.* Equivalently,
   $$(V, \sigma_M)$$
   is an Uncertain Set of type $M$ on the atom set $V$;
6. $$\eta_M : E \to \mathrm{Dom}(M)$$
   is the *edge uncertainty-degree function.* Equivalently,
   $$(E, \eta_M)$$
   is an Uncertain Set of type $M$ on the bond set $E$;
7. for every bond
   $$e = \{u, v\} \in E,$$
   the *uncertain molecular consistency condition*
   $$\eta_M(e) \preceq_M \Gamma_M\big(\sigma_M(u), \sigma_M(v)\big)$$
   holds.

The quadruple
$$(V, E, \lambda_V, \lambda_E)$$
is called the *underlying molecular graph* of $MU_M$.

**Theorem 6.1.4** (Well-definedness of Molecular Uncertain Graph)**.** *Let $\Sigma_V$ and $\Sigma_E$ be finite nonempty sets, let $M$ be a molecular-compatible uncertain model, and let*
$$MU_M = (V, E, \lambda_V, \lambda_E, \sigma_M, \eta_M)$$
*satisfy the conditions in the above definition. Then:*

1. *the underlying labeled molecular graph*
   $$(V, E, \lambda_V, \lambda_E)$$
   *is well-defined;*
2. $$(V, \sigma_M)$$
   *and*
   $$(E, \eta_M)$$
   *are well-defined Uncertain Sets of type $M$;*
3. *for every bond*
   $$e = \{u, v\} \in E,$$
   *the compatibility statement*
   $$\eta_M(e) \preceq_M \Gamma_M\big(\sigma_M(u), \sigma_M(v)\big)$$
   *is meaningful and has a definite truth value;*

4. *consequently,*

$$MU_M = (V, E, \lambda_V, \lambda_E, \sigma_M, \eta_M)$$

*determines a unique mathematical object, namely a molecular uncertain graph of type $M$.*

*Proof.* Since $M$ is an uncertain model, its degree-domain

$$\mathrm{Dom}(M) \subseteq [0,1]^k$$

is fixed. Since $M$ is molecular-compatible, the partial order

$$\preceq_M$$

and the symmetric map

$$\Gamma_M : \mathrm{Dom}(M) \times \mathrm{Dom}(M) \to \mathrm{Dom}(M)$$

are fixed as part of the structure.

By assumption, $V$ is a finite nonempty set and

$$E \subseteq \big\{\{u,v\} \subseteq V : u \neq v\big\}$$

is a finite set of unordered pairs of distinct vertices. Hence $(V, E)$ is a well-defined finite simple undirected graph. Moreover,

$$\lambda_V : V \to \Sigma_V \qquad \text{and} \qquad \lambda_E : E \to \Sigma_E$$

are functions. Therefore the quadruple

$$(V, E, \lambda_V, \lambda_E)$$

is a well-defined labeled molecular graph. This proves (1).

Next, since

$$\sigma_M : V \to \mathrm{Dom}(M)$$

is a function, the pair

$$(V, \sigma_M)$$

is a well-defined Uncertain Set of type $M$ on the set $V$. Likewise, since

$$\eta_M : E \to \mathrm{Dom}(M)$$

is a function, the pair

$$(E, \eta_M)$$

is a well-defined Uncertain Set of type $M$ on the set $E$. This proves (2).

Now let

$$e = \{u, v\} \in E.$$

Since $\sigma_M : V \to \mathrm{Dom}(M)$, both

$$\sigma_M(u) \in \mathrm{Dom}(M) \qquad \text{and} \qquad \sigma_M(v) \in \mathrm{Dom}(M)$$

are well-defined. Hence

$$\Gamma_M\big(\sigma_M(u), \sigma_M(v)\big) \in \mathrm{Dom}(M)$$

is well-defined. Also, since $\eta_M : E \to \mathrm{Dom}(M)$,

$$\eta_M(e) \in \mathrm{Dom}(M)$$

is well-defined. Therefore the comparison

$$\eta_M(e) \preceq_M \Gamma_M\big(\sigma_M(u), \sigma_M(v)\big)$$

is meaningful, because $\preceq_M$ is a fixed partial order on $\mathrm{Dom}(M)$. Hence this compatibility statement has a definite truth value for every $e \in E$. This proves (3).

Finally, all six components

$$V, \quad E, \quad \lambda_V, \quad \lambda_E, \quad \sigma_M, \quad \eta_M$$

are uniquely specified, and the compatibility condition in item (7) is meaningful for every bond. Therefore the sextuple

$$MU_M = (V, E, \lambda_V, \lambda_E, \sigma_M, \eta_M)$$

determines a unique mathematical object. Hence the notion of a molecular uncertain graph of type $M$ is well-defined. This proves (4). □

Related extensions of molecular graphs include chemical graphs, weighted molecular graphs, labeled molecular graphs [743, 744], molecular trees (chemical trees) [745, 746], signed molecular graphs [747], molecular hypergraph [748, 748, 749], molecular superhypergraphs [486, 750, 751], and periodic crystal graphs.

## 6.2 Uncertain ANP (Uncertain Decision-Making)

ANP evaluates interdependent criteria and alternatives using pairwise comparisons, builds a supermatrix, and derives global priorities via limit supermatrix [752–754]. Fuzzy ANP extends ANP with fuzzy pairwise judgments, forming fuzzy supermatrices and defuzzified limit priorities to handle uncertainty [755–757].

**Definition 6.2.1** (Fuzzy Analytic Network Process (FANP): supermatrix formulation)**.** [758,759] Let $\mathcal{C} = \{C_1, \dots, C_m\}$ be clusters, where $C_i = \{e_{i1}, \dots, e_{in_i}\}$ with $n_i \geq 1$, and let

$$E := \bigsqcup_{i=1}^{m} C_i$$

be the set of all elements. Let $\mathcal{D} \subseteq \mathcal{C} \times \mathcal{C}$ be a directed dependence relation, where $(C_i, C_j) \in \mathcal{D}$ means that "$C_i$ influences $C_j$" (allowing inner dependence $i = j$ and outer dependence $i \neq j$).

Fix a class $\mathcal{FN}_{>0}$ of positive fuzzy numbers, closed under the operations used below, and fix a ranking (defuzzification) map

$$\mathrm{Score} : \mathcal{FN}_{>0} \longrightarrow \mathbb{R}_{>0}.$$

**(0) Fuzzy pairwise comparisons and local priorities.** A *positive reciprocal fuzzy pairwise comparison matrix* is

$$\widetilde{A} = (\tilde{a}_{rs}) \in \mathcal{FN}_{>0}^{n \times n},$$

satisfying

$$\tilde{a}_{rr} = 1, \qquad \tilde{a}_{sr} = \tilde{a}_{rs}^{-1} \qquad (1 \leq r, s \leq n).$$

Its *local fuzzy priority vector*

$$\widetilde{w} = (\tilde{w}_1, \dots, \tilde{w}_n) \in \mathcal{FN}_{>0}^{n}$$

is obtained, for example, by the fuzzy geometric mean method:

$$\tilde{g}_r := \Big(\prod_{s=1}^{n} \tilde{a}_{rs}\Big)^{1/n}, \qquad \tilde{w}_r := \tilde{g}_r \Big/ \Big(\sum_{t=1}^{n} \tilde{g}_t\Big), \qquad r = 1, \dots, n,$$

where the product, power, sum, and division are understood in the fuzzy-number sense (e.g. via the extension principle or an equivalent admissible fuzzy arithmetic). The associated *crisp normalized priority vector* is defined by

$$w_r^{\sharp} := \frac{\mathrm{Score}(\tilde{w}_r)}{\sum_{t=1}^{n} \mathrm{Score}(\tilde{w}_t)}, \qquad r = 1, \dots, n.$$

Then $w^\sharp = (w_1^\sharp, \dots, w_n^\sharp) \in \mathbb{R}^n_{\geq 0}$ and

$$\sum_{r=1}^{n} w_r^\sharp = 1.$$

**(1) Element-to-element influence vectors and unweighted supermatrix.** Fix $(C_i, C_j) \in \mathcal{D}$ and a target element $e_{jk} \in C_j$. Compare the elements of $C_i$ pairwise with respect to their influence on $e_{jk}$ by a positive reciprocal fuzzy comparison matrix

$$\widetilde{A}^{(i \to j \,|\, k)} \in \mathcal{FN}_{>0}^{n_i \times n_i}.$$

Let

$$w^{\sharp (i \to j \,|\, k)} \in \mathbb{R}^{n_i}_{\geq 0}$$

be the corresponding crisp normalized local priority vector obtained from the above procedure. Define the block $W_{ij} \in \mathbb{R}^{n_i \times n_j}$ by

$$(W_{ij})_{\bullet k} := w^{\sharp (i \to j \,|\, k)} \qquad (k = 1, \dots, n_j),$$

and set $W_{ij} := 0$ if $(C_i, C_j) \notin \mathcal{D}$. The resulting block matrix

$$W := \big(W_{ij}\big)_{1 \leq i,j \leq m} \in \mathbb{R}^{(\sum_i n_i) \times (\sum_j n_j)}$$

is called the *unweighted supermatrix.*

**(2) Cluster weights and weighted (column-stochastic) supermatrix.** For each cluster $C_j$, let

$$\Gamma(j) := \{\, i : (C_i, C_j) \in \mathcal{D} \,\}.$$

Obtain fuzzy cluster-comparison judgments among the influencing clusters $\{C_i : i \in \Gamma(j)\}$, compute the corresponding fuzzy cluster-priority vector, and then defuzzify/normalize it to get

$$\alpha^{(j),\sharp} = (\alpha_i^{(j),\sharp})_{i \in \Gamma(j)} \in \mathbb{R}^{|\Gamma(j)|}_{\geq 0}, \qquad \sum_{i \in \Gamma(j)} \alpha_i^{(j),\sharp} = 1.$$

Define the weighted blocks

$$\bar{W}_{ij} := \begin{cases} \alpha_i^{(j),\sharp}\, W_{ij}, & i \in \Gamma(j), \\ 0, & i \notin \Gamma(j), \end{cases} \qquad \bar{W} := \big(\bar{W}_{ij}\big)_{1 \leq i,j \leq m}.$$

Then $\bar{W}$ is column-stochastic, i.e. each column sums to 1.

**(3) Limit supermatrix and global priorities.** If the limit exists, define the *limit supermatrix* by

$$W^\infty := \lim_{t \to \infty} \bar{W}^{\,t}.$$

If $\bar{W}^{\,t}$ is cyclic with period $N$, use the cycle-average

$$W^\infty := \frac{1}{N} \sum_{t=0}^{N-1} \bar{W}^{\,t}.$$

Let $A \subseteq E$ be the designated set of alternatives (usually a subset of the elements in one cluster). The *global priority* of an alternative is read from the corresponding row of $W^\infty$ (equivalently, from stabilized columns), and the alternatives are ranked in decreasing order of global priority.

We now extend Fuzzy ANP using Uncertain Sets. The related definitions are given below.

**Definition 6.2.2** (Uncertain set and uncertain number)**.** Let $U$ be a nonempty universe. An *uncertain set* on $U$ is a mapping $\mu : U \to [0,1]$.

An *uncertain number* is an uncertain set $\tilde{x}$ on $\mathbb{R}$ whose $\alpha$-cuts

$$[\tilde{x}]_\alpha := \{t \in \mathbb{R} : \mu_{\tilde{x}}(t) \geq \alpha\} \qquad (\alpha \in (0,1])$$

are nonempty compact intervals. Let $\mathsf{UN}$ be a fixed class of uncertain numbers and set

$$\mathsf{UN}_{>0} := \{\tilde{x} \in \mathsf{UN} : \mathrm{supp}(\tilde{x}) \subseteq (0,\infty)\}.$$

Assume $\mathsf{UN}_{>0}$ is closed under the *uncertain inverse*: if $[\tilde{x}]_\alpha = [x_\alpha^-, x_\alpha^+]$ with $0 < x_\alpha^- \leq x_\alpha^+$, define

$$[\tilde{x}^{-1}]_\alpha := \Big[\frac{1}{x_\alpha^+}, \frac{1}{x_\alpha^-}\Big] \qquad (\alpha \in (0,1]).$$

Fix a *score* (crisp representative) map

$$\mathrm{Score} : \mathsf{UN}_{>0} \to (0,\infty),$$

and assume it is *reciprocity-compatible*:

$$\mathrm{Score}(\tilde{x}^{-1}) = \frac{1}{\mathrm{Score}(\tilde{x})} \quad (\forall\, \tilde{x} \in \mathsf{UN}_{>0}), \qquad \mathrm{Score}(\tilde{\mathbf{1}}) = 1,$$

where $\tilde{\mathbf{1}}$ is the uncertain number concentrated at 1.

**Definition 6.2.3** (Uncertain reciprocal judgment matrix)**.** Let $n \geq 2$. An *uncertain reciprocal judgment matrix* is

$$\widetilde{A} = (\tilde{a}_{rs}) \in (\mathsf{UN}_{>0})^{n\times n}$$

such that

$$\tilde{a}_{rr} = \tilde{\mathbf{1}} \quad (\forall r), \qquad \tilde{a}_{sr} = \tilde{a}_{rs}^{-1} \quad (\forall r \neq s).$$

Its induced crisp reciprocal matrix is

$$A := (a_{rs}) \in (0,\infty)^{n\times n}, \qquad a_{rs} := \mathrm{Score}(\tilde{a}_{rs}).$$

**Definition 6.2.4** (Uncertain ANP (UANP): score-induced supermatrix formulation)**.** Let $\mathcal{C} = \{C_1, \dots, C_m\}$ be *clusters*, where

$$C_i = \{e_{i1}, \dots, e_{in_i}\} \quad (n_i \geq 1), \qquad E := \bigsqcup_{i=1}^{m} C_i$$

is the set of all elements. Let $\mathcal{D} \subseteq \mathcal{C} \times \mathcal{C}$ be a directed dependence relation; $(C_i, C_j) \in \mathcal{D}$ means "$C_i$ influences $C_j$" (allowing $i = j$).

**(0) Local priorities from uncertain pairwise judgments.** Fix $(C_i, C_j) \in \mathcal{D}$ and a target element $e_{jk} \in C_j$. Decision makers provide an uncertain reciprocal judgment matrix

$$\widetilde{A}^{(i\to j\,|\,k)} \in (\mathsf{UN}_{>0})^{n_i\times n_i}$$

comparing the elements of $C_i$ with respect to their influence on $e_{jk}$. Let

$$A^{(i\to j\,|\,k)} := \big(\mathrm{Score}(\tilde{a}_{rs}^{(i\to j\,|\,k)})\big) \in (0,\infty)^{n_i\times n_i}$$

be the induced crisp reciprocal matrix. Define the *local priority vector* $w^{(i\to j\,|\,k)} \in \mathbb{R}^{n_i}_{>0}$ as the normalized Perron vector of $A^{(i\to j\,|\,k)}$:

$$A^{(i\to j\,|\,k)}\, w^{(i\to j\,|\,k)} = \lambda_{\max}\, w^{(i\to j\,|\,k)}, \qquad \sum_{r=1}^{n_i} w_r^{(i\to j\,|\,k)} = 1.$$

**(1) Unweighted supermatrix.** For each $(i,j)$, define the block $W_{ij} \in \mathbb{R}^{n_i\times n_j}_{\geq 0}$ by setting its $k$th column as

$$(W_{ij})_{\bullet k} := w^{(i\to j\,|\,k)} \quad (k = 1, \dots, n_j),$$

and set $W_{ij} := 0$ if $(C_i, C_j) \notin \mathcal{D}$. The *unweighted supermatrix* is the block matrix

$$W := \big(W_{ij}\big)_{1\le i,j\le m} \in \mathbb{R}_{\ge 0}^{N\times N}, \qquad N := \sum_{i=1}^{m} n_i.$$

**(2) Cluster weights and weighted supermatrix.** For each target cluster $C_j$, let $\Gamma(j) := \{i : (C_i, C_j) \in \mathcal{D}\}$. Cluster weights are obtained by uncertain pairwise comparisons among clusters in $\Gamma(j)$, yielding (after applying Score and Perron normalization) a vector

$$\alpha^{(j)} = (\alpha_i^{(j)})_{i\in\Gamma(j)} \in \mathbb{R}_{\ge 0}^{|\Gamma(j)|}, \qquad \sum_{i\in\Gamma(j)} \alpha_i^{(j)} = 1.$$

Define the weighted blocks

$$\bar{W}_{ij} := \begin{cases} \alpha_i^{(j)}\, W_{ij}, & i \in \Gamma(j), \\ 0, & i \notin \Gamma(j), \end{cases} \qquad \bar{W} := \big(\bar{W}_{ij}\big)_{1\le i,j\le m} \in \mathbb{R}_{\ge 0}^{N\times N}.$$

**(3) Limit supermatrix and global priorities.** If $\bar{W}$ is *primitive* (some power has strictly positive entries), define

$$W^{\infty} := \lim_{t\to\infty} \bar{W}^{\,t}.$$

In general (even if periodic), define the Cesàro limit (always used in practice when cycling occurs):

$$W^{\infty} := \lim_{T\to\infty} \frac{1}{T} \sum_{t=0}^{T-1} \bar{W}^{\,t}, \quad \text{whenever the limit exists.}$$

Let $A \subseteq E$ be the designated set of alternatives (elements representing alternatives). The *global priority* of $a \in A$ is read from the corresponding row of $W^{\infty}$ (equivalently from stabilized columns), and alternatives are ranked by decreasing global priority.

Related concepts of ANP under uncertainty-aware models are listed in Table 6.1.

Table 6.1: Related concepts of ANP under uncertainty-aware models.

| $k$ | **Related ANP concept(s)** |
|---|---|
| 2 | Intuitionistic Fuzzy ANP [760, 761] |
| 2 | Pythagorean Fuzzy ANP [762, 763] |
| 3 | Hesitant Fuzzy ANP [764] |
| 3 | Spherical Fuzzy ANP [765] |
| 3 | Neutrosophic ANP [766, 767] |

As concepts other than Uncertain ANP, DEMATEL-ANP [768, 769], BOCR-based ANP [770], Group ANP [771, 772], ANP-TOPSIS [771, 773], and Rough ANP [774, 775] are also known.

## 6.3 Uncertain Graph Neural Networks

Fuzzy Graph Neural Network is a graph-learning model that propagates and updates node representations using fuzzy vertex and edge memberships to handle uncertainty and relations [156, 776–780].

**Definition 6.3.1** (Fuzzy Graph Neural Network (F-GNN))**.** Let

$$G = (V, \sigma, \mu)$$

be a finite fuzzy graph, where

$$V \neq \varnothing, \qquad \sigma : V \to [0,1], \qquad \mu : V \times V \to [0,1]$$

is symmetric and satisfies

$$\mu(u,v) \leq \min\{\sigma(u), \sigma(v)\} \qquad (\forall\, u, v \in V).$$

Let

$$X = (x_v)_{v \in V} \in \mathbb{R}^{|V| \times d_0}$$

be the input feature matrix, where each

$$x_v \in \mathbb{R}^{d_0}$$

is the initial feature vector of the vertex $v$.

A *Fuzzy Graph Neural Network* of depth $T \in \mathbb{N}$ on $(G, X)$ is a collection

$$\mathcal{F} = \Big( (\varphi^{(t)}, \psi^{(t)})_{t=0}^{T-1},\, \rho \Big),$$

where, for each layer $t = 0, 1, \ldots, T-1$,

$$\varphi^{(t)} : \mathbb{R}^{d_t} \times \mathbb{R}^{d_t} \times [0,1] \times [0,1] \to \mathbb{R}^{r_t}$$

is a learnable *message function*,

$$\psi^{(t)} : \mathbb{R}^{d_t} \times \mathbb{R}^{r_t} \times [0,1] \to \mathbb{R}^{d_{t+1}}$$

is a learnable *update function*, and

$$\rho : \mathbb{R}^{d_T} \to \mathbb{R}^{q}$$

is an output map.

The hidden states

$$h_v^{(t)} \in \mathbb{R}^{d_t} \qquad (v \in V,\ t = 0, 1, \ldots, T)$$

are defined recursively as follows.

First,

$$h_v^{(0)} := x_v \qquad (\forall\, v \in V).$$

For each $t = 0, 1, \ldots, T-1$, define the fuzzy neighborhood of $v$ by

$$N_\mu(v) := \{u \in V : \mu(u,v) > 0\}.$$

For every $u \in N_\mu(v)$, define the message sent from $u$ to $v$ at layer $t+1$ by

$$m_{u \to v}^{(t+1)} := \mu(u,v)\, \varphi^{(t)}\big(h_v^{(t)},\, h_u^{(t)},\, \sigma(v),\, \sigma(u)\big).$$

Let

$$\mathrm{Agg}$$

be a fixed permutation-invariant aggregation operator (for example, sum, mean, or max). Then the aggregated fuzzy message at $v$ is

$$M_v^{(t+1)} := \mathrm{Agg}\big\{ m_{u \to v}^{(t+1)} :\ u \in N_\mu(v) \big\}.$$

If $N_\mu(v) = \varnothing$, one sets

$$M_v^{(t+1)} := 0.$$

The vertex representation is then updated by

$$h_v^{(t+1)} := \psi^{(t)}\big(h_v^{(t)},\, M_v^{(t+1)},\, \sigma(v)\big).$$

After $T$ layers, the *vertex-level output* is defined by

$$\widehat{y}_v := \rho\big(h_v^{(T)}\big) \qquad (\forall\, v \in V).$$

For graph-level tasks, one may additionally choose a permutation-invariant readout map

$$\text{Readout} : \mathcal{P}_{\text{fin}}(\mathbb{R}^{d_T}) \to \mathbb{R}^q$$

and define

$$\widehat{y} := \text{Readout}\big(\{h_v^{(T)} : v \in V\}\big).$$

**Remark 6.3.2.** The factor

$$\mu(u, v)$$

weights message passing by the fuzzy strength of the edge $(u, v)$, while

$$\sigma(v) \quad \text{and} \quad \sigma(u)$$

allow the update to depend on the fuzzy presence of the incident vertices. Hence an F-GNN is a graph neural network whose propagation rule is modulated by the fuzzy structure of the underlying fuzzy graph.

**Remark 6.3.3.** If

$$\sigma(v) = 1 \qquad (\forall\, v \in V),$$

and

$$\mu(u, v) \in \{0, 1\}$$

coincides with the adjacency indicator of a crisp graph, then the above F-GNN reduces to an ordinary message-passing graph neural network.

Let $M$ be a fixed uncertainty model, and let

$$\text{Dom}(M) \subseteq [0, 1]^k \qquad (k \geq 1)$$

denote its degree domain. For example, $\text{Dom}(M) = [0, 1]$ for a fuzzy model, while $\text{Dom}(M) \subseteq [0, 1]^3$ may be used for a three-component uncertainty model.

**Definition 6.3.4** (Uncertain Graph Neural Network)**.** Let

$$G_M = (V, E, \mu_M)$$

be a finite uncertain graph of type $M$, where

$$V = \{v_1, \ldots, v_n\}$$

is a finite vertex set,

$$E \subseteq \big\{\{u, v\} \subseteq V : u \neq v\big\},$$

and

$$\mu_M : V \cup E \to \text{Dom}(M)$$

assigns to each vertex and each edge its uncertain degree.

For each vertex $v \in V$, define its neighborhood by

$$N(v) := \{u \in V : \{u, v\} \in E\}.$$

An *Uncertain Graph Neural Network* of type $M$ on $G_M$, briefly a *U-GNN*, is a tuple

$$\text{U-GNN}_M = \Big(G_M,\, d_0, \ldots, d_L,\, h^{(0)},\, \big(\text{Msg}^{(\ell)}, \text{Agg}^{(\ell)}, \text{Upd}^{(\ell)}\big)_{\ell=0}^{L-1},\, \text{Readout}\Big),$$

where:

1. $$L \in \mathbb{N}$$
   is the number of layers, and
   $$d_0, d_1, \ldots, d_L \in \mathbb{N}$$
   are feature dimensions;
2. $$h^{(0)} : V \to \mathbb{R}^{d_0}$$
   is the initial vertex-feature map;
3. for each layer $\ell = 0, 1, \ldots, L-1$,
   $$\mathrm{Msg}^{(\ell)} : \mathbb{R}^{d_\ell} \times \mathbb{R}^{d_\ell} \times \mathrm{Dom}(M)^3 \to \mathbb{R}^{d_\ell}$$
   is the *message function*,
   $$\mathrm{Agg}^{(\ell)} : \{\text{finite multisets of } \mathbb{R}^{d_\ell}\} \to \mathbb{R}^{d_\ell}$$
   is a *permutation-invariant aggregation operator*,
   and
   $$\mathrm{Upd}^{(\ell)} : \mathbb{R}^{d_\ell} \times \mathbb{R}^{d_\ell} \to \mathbb{R}^{d_{\ell+1}}$$
   is the *update function*;
4. $$\mathrm{Readout} : \mathbb{R}^{d_L} \to Y$$
   is a task-dependent output map, where $Y$ is the output space.

The forward propagation is defined recursively as follows.

For each $\ell = 0, 1, \ldots, L-1$ and each $v \in V$, define the aggregated uncertain message
$$m_v^{(\ell)} := \mathrm{Agg}^{(\ell)}\Big(\big\{\mathrm{Msg}^{(\ell)}\big(h_v^{(\ell)}, h_u^{(\ell)}, \mu_M(v), \mu_M(u), \mu_M(\{u,v\})\big) :\ u \in N(v)\big\}\Big),$$
where
$$h_v^{(\ell)} := h^{(\ell)}(v).$$

Then define the next-layer representation by
$$h_v^{(\ell+1)} := \mathrm{Upd}^{(\ell)}\big(h_v^{(\ell)}, m_v^{(\ell)}\big).$$

After $L$ layers, the node-level output at $v \in V$ is
$$y_v := \mathrm{Readout}\big(h_v^{(L)}\big).$$

**Remark 6.3.5.** The role of the uncertainty model is entirely encoded in the domain
$$\mathrm{Dom}(M)$$
and in the degree assignment
$$\mu_M : V \cup E \to \mathrm{Dom}(M).$$
Hence the above definition is uniform enough to cover, as special cases, fuzzy graph neural networks, neutrosophic graph neural networks, and other uncertainty-aware graph neural models.

**Remark 6.3.6.** If one takes
$$\mathrm{Dom}(M) = [0,1]$$
and interprets $\mu_M(v)$ and $\mu_M(\{u,v\})$ as scalar memberships, then the above definition reduces to a scalar-valued uncertainty-aware message-passing GNN. If, moreover, the uncertain degrees are ignored in $\mathrm{Msg}^{(\ell)}$, one recovers the usual message-passing graph neural network framework.

Representative graph-neural-network concepts in uncertainty-aware graph frameworks are listed in Table 6.2.

Related concepts such as HyperGraph Neural Networks [6, 783, 784], SuperHyperGraph Neural Networks [8], and Directed Graph Neural Networks [785–787] are also known.

Table 6.2: Representative graph-neural-network concepts under uncertainty-aware graph frameworks, classified by the dimension $k$ of the information attached to vertices and/or edges.

| $k$ | **Graph-neural-network concept** | **Typical coordinate form** | **Canonical information attached to vertices/edges** |
|---|---|---|---|
| 1 | Fuzzy Graph Neural Network | $\mu$ | A graph neural network defined in a fuzzy framework, where each vertex and edge is associated with a single membership degree in $[0,1]$. |
| 2 | Intuitionistic Fuzzy Graph Neural Network | $(\mu,\nu)$ | A graph neural network defined in an intuitionistic fuzzy framework, where each vertex and edge carries a membership degree and a nonmembership degree, usually satisfying $\mu+\nu \le 1$. |
| 3 | Neutrosophic Graph Neural Network [781, 782] | $(T,I,F)$ | A graph neural network defined in a neutrosophic framework, where each vertex and edge is described by truth, indeterminacy, and falsity degrees. |

## 6.4 Uncertain Knowledge Graphs

A fuzzy knowledge graph extends an ordinary knowledge graph by assigning to each factual triple a degree in $[0,1]$, representing the strength, confidence, or plausibility of that fact [788–792].

**Definition 6.4.1** (Knowledge Graph)**.** [789, 793] Let

$$\mathcal{E}$$

be a nonempty set of entities and

$$\mathcal{R}$$

a nonempty set of binary relation symbols. A *knowledge graph* is a triple

$$\mathcal{K} = (\mathcal{E}, \mathcal{R}, \mathcal{T}),$$

where

$$\mathcal{T} \subseteq \mathcal{E} \times \mathcal{R} \times \mathcal{E}.$$

An element

$$(h,r,t) \in \mathcal{T}$$

is called a *fact* or *triple*, where

$$h \in \mathcal{E}$$

is the head entity,

$$r \in \mathcal{R}$$

is the relation, and

$$t \in \mathcal{E}$$

is the tail entity.

**Definition 6.4.2** (Fuzzy Knowledge Graph)**.** [788–790] Let

$$\mathcal{E}$$

be a nonempty set of entities and

$$\mathcal{R}$$

a nonempty set of binary relation symbols. Set

$$\Omega := \mathcal{E} \times \mathcal{R} \times \mathcal{E}.$$

A *fuzzy knowledge graph* is a triple

$$\widetilde{\mathcal{K}} = (\mathcal{E}, \mathcal{R}, \mu),$$

where

$$\mu : \Omega \to [0,1]$$

is a membership function assigning to each possible triple

$$(h, r, t) \in \Omega$$

a degree

$$\mu(h, r, t) \in [0,1].$$

The value

$$\mu(h, r, t)$$

is called the *fuzzy truth degree*, *confidence degree*, or *membership degree* of the fact

$$(h, r, t).$$

The *support* of $\widetilde{\mathcal{K}}$ is the set

$$\operatorname{supp}(\widetilde{\mathcal{K}}) := \{(h, r, t) \in \Omega : \mu(h, r, t) > 0\}.$$

Thus, a fuzzy knowledge graph may be viewed as a fuzzy subset of the set of all possible knowledge triples.

**Definition 6.4.3** (Relation-wise Fuzzy Digraph Induced by a Fuzzy Knowledge Graph)**.** Let

$$\widetilde{\mathcal{K}} = (\mathcal{E}, \mathcal{R}, \mu)$$

be a fuzzy knowledge graph, and fix

$$r \in \mathcal{R}.$$

Define

$$\mu_r : \mathcal{E} \times \mathcal{E} \to [0,1]$$

by

$$\mu_r(h, t) := \mu(h, r, t) \qquad (\forall\, h, t \in \mathcal{E}).$$

Then

$$G_r = (\mathcal{E}, \mu_r)$$

is called the *relation-wise fuzzy digraph* induced by $r$. Hence a fuzzy knowledge graph can be regarded as a family

$$\{G_r : r \in \mathcal{R}\}$$

of labeled fuzzy directed graphs on the common entity set $\mathcal{E}$.

## 6.5 Uncertain Cognitive Map

Cognitive Map is a graphical representation of concepts and their relationships, used to organize knowledge, support reasoning, and model how people perceive systems or problems [794–796]. A Fuzzy Cognitive Map is a weighted directed graph of concepts, where fuzzy causal links model positive or negative influences, supporting reasoning, simulation, and analysis [797–799].

**Definition 6.5.1** (Fuzzy Cognitive Map)**.** [800–802] Let

$$C = \{C_1, C_2, \ldots, C_n\}$$

be a finite set of concepts, and let

$$W = (w_{ij})_{n\times n} \in [-1, 1]^{n\times n}$$

be a real matrix such that

$$w_{ii} = 0 \qquad (i = 1, 2, \ldots, n).$$

A *fuzzy cognitive map* (FCM) is the pair

$$\mathcal{M} = (C, W),$$

where each entry $w_{ij}$ represents the signed causal influence of concept $C_i$ on concept $C_j$, interpreted as follows:

$$w_{ij} > 0 \implies C_i \text{ promotes } C_j,$$

$$w_{ij} < 0 \implies C_i \text{ inhibits } C_j,$$

$$w_{ij} = 0 \implies \text{there is no direct causal influence from } C_i \text{ to } C_j.$$

The associated directed graph

$$G_{\mathcal{M}} = (C, E)$$

is defined by

$$E = \{(C_i, C_j) \in C \times C :\ w_{ij} \neq 0\}.$$

Hence, an FCM is a weighted directed graph whose vertices are concepts and whose arc weights quantify fuzzy causal strengths.

If, in addition, an activation vector

$$A^{(t)} = (a_1^{(t)}, a_2^{(t)}, \ldots, a_n^{(t)}) \in [0, 1]^n$$

is assigned at time $t$, together with an activation function

$$f : \mathbb{R} \to [0, 1],$$

then the induced FCM dynamics is commonly given by

$$a_j^{(t+1)} = f\left(a_j^{(t)} + \sum_{i=1}^{n} a_i^{(t)} w_{ij}\right), \qquad j = 1, 2, \ldots, n.$$

Equivalently,

$$A^{(t+1)} = f\big(A^{(t)} + A^{(t)}W\big),$$

where $f$ is applied componentwise.

In this case, the triple

$$(\mathcal{M}, f, A^{(0)})$$

is called a *dynamical fuzzy cognitive map*.

Representative cognitive-map concepts under uncertainty-aware graph frameworks are listed in Table 6.3.

Table 6.3: Representative cognitive-map concepts under uncertainty-aware graph frameworks, classified by the dimension $k$ of the information attached to concepts and/or causal relations.

| $k$ | **Cognitive-map concept** | **Typical coordinate form** | **Canonical information attached to concepts/causal relations** |
|---|---|---|---|
| 1 | Fuzzy Cognitive Map | $\mu$ | A cognitive map defined in a fuzzy framework, where each concept and causal relation is associated with a single membership or influence degree in $[0, 1]$ (or, in signed settings, a single fuzzy causal strength). |
| 2 | Intuitionistic Fuzzy Cognitive Map [795, 803, 804] | $(\mu, \nu)$ | A cognitive map defined in an intuitionistic fuzzy framework, where each concept and causal relation carries a membership degree and a nonmembership degree, usually satisfying $\mu + \nu \leq 1$. |
| 3 | Neutrosophic Cognitive Map [805–808] | $(T, I, F)$ | A cognitive map defined in a neutrosophic framework, where each concept and causal relation is described by truth, indeterminacy, and falsity degrees. |
| $s + t$ | Plithogenic Cognitive Map [87, 809–811] | $(\mathbf{a}, \mathbf{c}) \in [0, 1]^s \times [0, 1]^t$ | A cognitive map defined in a plithogenic framework, where each concept and causal relation is described by attribute-based information together with an $s$-dimensional appurtenance vector and a $t$-dimensional contradiction vector. |

# Chapter 7

# Conclusions

In this book, we surveyed graph classes that are well known in frameworks such as fuzzy graphs, neutrosophic graphs, and plithogenic graphs. It is hoped that future research will further develop graph algorithms for fundamental tasks such as shortest paths, connectivity analysis, spanning structures, clustering, and optimization in these uncertainty-aware settings. It is also expected that quantitative studies based on computational experiments, benchmark datasets, simulation-based evaluations, and comparative performance analysis will provide a clearer understanding of the practical behavior of these models. In addition, case studies in areas such as molecular networks, decision-making systems, knowledge representation, and uncertain relational data may help clarify the applicability and limitations of these graph frameworks in real-world problems.

# Disclaimer

## Funding

This study was conducted without any financial support from external organizations or grants.

## Acknowledgments

We would like to express our sincere gratitude to everyone who provided valuable insights, support, and encouragement throughout this research. We also extend our thanks to the readers for their interest and to the authors of the referenced works, whose scholarly contributions have greatly influenced this study. Lastly, we are deeply grateful to the publishers and reviewers who facilitated the dissemination of this work.

## Data Availability

Since this research is purely theoretical and mathematical, no empirical data or computational analysis was utilized. Researchers are encouraged to expand upon these findings with data-oriented or experimental approaches in future studies.

## Ethical Statement

As this study does not involve experiments with human participants or animals, no ethical approval was required.

## Conflicts of Interest

The authors declare that they have no conflicts of interest related to the content or publication of this book.

## Code Availability

No code or software was developed for this study.

## Use of Generative AI and AI-Assisted Tools

I use generative AI and AI-assisted tools for tasks such as English grammar checking, and I do not employ them in any way that violates ethical standards.

## Disclaimer (Others)

This work presents theoretical ideas and frameworks that have not yet been empirically validated. Readers are encouraged to explore practical applications and further refine these concepts. Although care has been taken to ensure accuracy and appropriate citations, any errors or oversights are unintentional. The perspectives and interpretations expressed herein are solely those of the authors and do not necessarily reflect the viewpoints of their affiliated institutions.

# Appendix (List of Tables)

*

# Appendix (List of Figures)

*

**This book presents a comprehensive and systematic survey of graph theory under uncertainty, with particular emphasis on the unifying role of the uncertain graph framework. It reviews fundamental concepts, structural properties, graph classes, and graph parameters within fuzzy, neutrosophic, and related models, while also introducing a wide range of extensions such as uncertain digraphs, hypergraphs, superhypergraphs, and dynamic graphs.**

**In addition to theoretical developments, the book explores practical applications, including uncertain molecular graphs, decision-making systems, graph neural networks, knowledge graphs, and cognitive maps. By organizing diverse uncertainty-aware graph models within a common perspective, this work provides a coherent framework for understanding their relationships, capabilities, and applications in complex systems.**

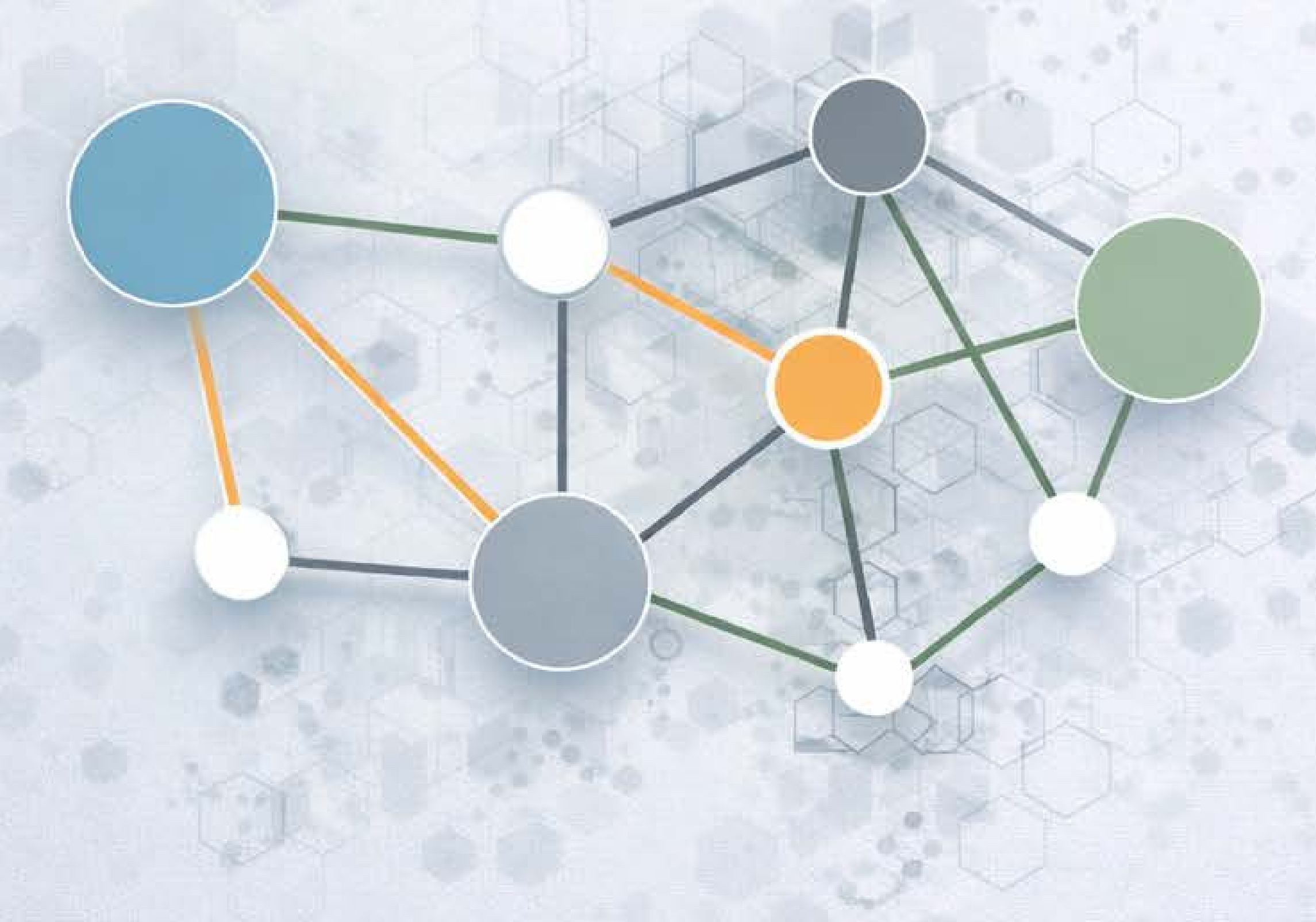

Takaaki Fujita

Florentin Smarandache

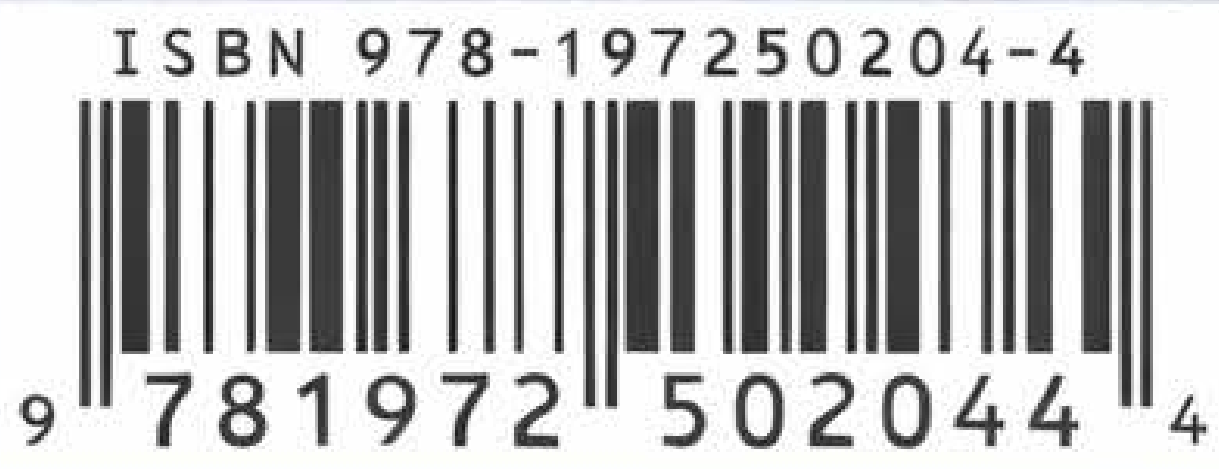